\documentclass[final]{cvpr}

\usepackage{times}
\usepackage{epsfig}
\usepackage{graphicx}
\usepackage{amsmath}
\usepackage{amssymb}
\usepackage[ruled,vlined,linesnumbered]{algorithm2e}

\graphicspath{{./figs/}}
\graphicspath{{./images_supp/}}
\usepackage[title]{appendix}
\usepackage{titling}

\usepackage{soul}
\usepackage{tabularx}
\usepackage{booktabs}       
\usepackage{xcolor}

\newcommand{\set}[1]{\left\{ {#1} \right\}}

\newcommand{\argmax}[2]{\underset{#1}{\text{argmax }} #2} 
\newcommand{\softmax}[2]{\underset{#1}{\text{softmax }} #2} 

\newcommand{\N}{\ensuremath \mathbb{N}}

\newcommand{\R}{\ensuremath \mathbb{R}}

\renewcommand{\epsilon}{\varepsilon}

\def\utilde#1{\mathord{\vtop{\ialign{##\crcr
				$\hfil\displaystyle{#1}\hfil$\crcr\noalign{\kern1.5pt\nointerlineskip}
				$\hfil\tilde{}\hfil$\crcr\noalign{\kern1.5pt}}}}}

\newcommand{\hide}[1]{}

\usepackage[pagebackref=true,breaklinks=true,colorlinks,bookmarks=false]{hyperref}

\begin{document}

\title{VideoClick: Video Object Segmentation with a Single Click}

\author{
	Namdar Homayounfar$^{1,2}$ \quad Justin Liang \quad Wei-Chiu Ma$^{1,3}$ \quad Raquel Urtasun$^{1,2}$\\
	$^{1}$Uber Advanced Technologies Group \quad $^{2}$University of Toronto \quad $^{3}$ MIT\\
	\small\texttt{namdar.homayounfar@mail.utoronto.ca, justin.j.w.liang@gmail.com} \\
		\small\texttt{weichium@mit.edu, urtasun@cs.toronto.edu}
}
\date{}

\maketitle

\begin{abstract}
Annotating videos with object segmentation masks typically involves a two stage procedure of drawing polygons per object instance for all the frames and then linking them through time. 
While simple, this is a very tedious, time consuming and expensive process, making the creation of accurate annotations at scale only possible for well-funded  labs. 
What if we were able to segment an object in the full video with only a single click? This will enable  video segmentation at scale with a very low budget opening the door to many applications. 
Towards this goal, in this paper we propose a bottom up approach where given a  single click for each object in a video, we obtain the segmentation masks of these objects in the full video. In particular, we construct a correlation volume that assigns each pixel in a target frame to either one of the objects in the reference frame or the background. We then refine this correlation volume via a recurrent attention module and decode the final segmentation.  
To evaluate the performance, we label the popular and challenging Cityscapes dataset with video object segmentations. 
Results on this new CityscapesVideo dataset show that our approach outperforms all the baselines in this challenging setting.

\end{abstract}

\section{Introduction}
\label{sec:intro}
Video object segmentation aims to identify all countable objects in the video and produce a \emph{masklet}, i.e. a sequence of masks, for each of them.

By detecting the pixels of unique objects in space and time, we can have a better understanding of the scene, learn a better representation, and even design embodied agents that are capable of interacting with the environment compliantly. 
Most recent approaches, however, rely on deep neural networks that are extremely data-hungry, necessitating  large-scale datasets.

Unfortunately, annotating videos with object segmentation masks is very cumbersome. It requires annotators to painstakingly analyze each video frame, manually delineate the objects from the background, and then associate them across time. 

One potential way to speed up the labeling process is to leverage existing video segmentation approaches to produce initial masklets and then pass it to humans for refinement. In order to track the objects across time, these methods usually assume each object mask is given in the first frame (manually annotated). While this assumption can be further relaxed to bounding boxes \cite{PolyRNN,polygon-rnn++} or extreme points \cite{maninis2018deep,wang2019object}, it still requires the annotators to spent on average 7$-$35 seconds per object \cite{papadopoulos2017extreme}. One thus ponders: can we further reduce the amount of human intervention? 

\begin{figure}
	\begin{centering}
		\includegraphics[width=.97\linewidth]{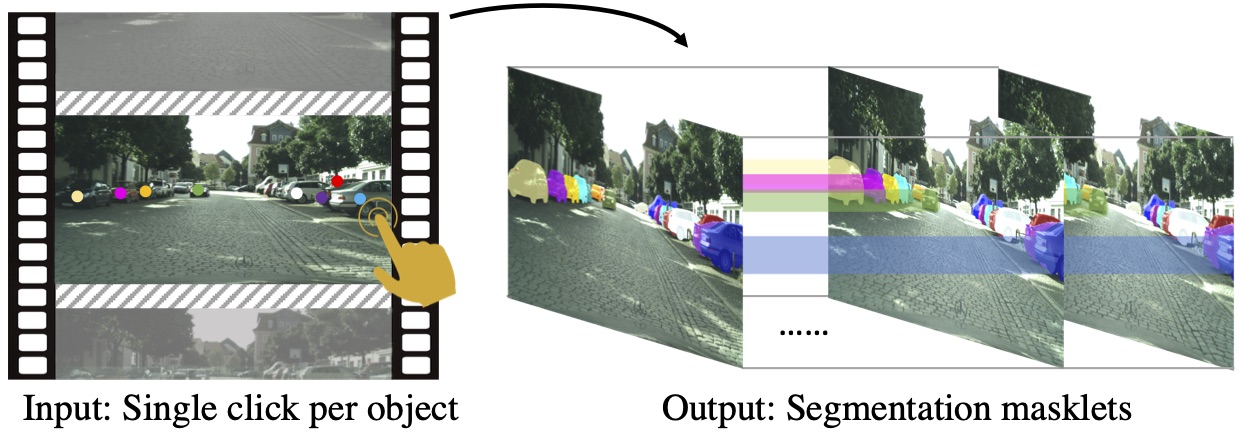}
	\end{centering}
	\caption{\textbf{Single click video object segmentation:} Given a single click anywhere in the object by an annotator, the goal is to obtain the masklet of the object for the entire snippet.} 
\end{figure}

In this paper we look at the problem of segmenting full videos by requiring only a single click for each object that appears in the video. The single click setting has several key advantages: (i) the click does not have to be an extreme point, it can be an arbitrary point so long as it lies within the object, significantly relieving the cognitive load for annotation. (ii) A click is remarkably easy and fast to annotate, allowing for video annotation at scale. 
At the same time, this is a very challenging task since we have to map a single point provided by the annotator in a video to the masklet of an object that can undergo various deformations in shape and appearance with possible occlusions in a potentially crowded scene. 

With these intuitions in mind, we present \emph{VideoClick}, a video object segmentation model that takes as input one single click per object and outputs the corresponding masklets for the full video. Our model is conceptually inspired by the state-of-the-art optical flow network RAFT \cite{RAFT}. Given a pair of consecutive images and a set of keypoints\footnote{We use the term \emph{keypoint} and \emph{click} interchangably.} within the reference frame, we first construct a 3D correlation volume representing the similarities between the pixels within the target frame and the keypoints. We then refine the correlation volume with a novel recurrent attention module based on the extracted visual features. The mask can then be obtained by taking an $\arg\max$ over the score of assigning each pixel in the target frame to one of the keypoints or the background. This approach is akin to clustering point features in the target frame to an instance prototype feature corresponding to annotator clicks.  We apply this procedure iteratively to video frames by propagating the keypoints in an \emph{implicit} fashion. This allows us to handle situations where the objects undergo drastic geometry changes over time (\eg, a vehicle may be split into multiple parts due to occlusion). 

To validate the effectiveness of our approach, we label the popular and challenging Cityscapes dataset \cite{Cordts2016Cityscapes} with video object segmentations.  We named this new dataset CityscapesVideo. 
Our dataset presents several unique challenges and has several key differences comparing to existing datasets: first, while most  benchmarks annotate only a few unique objects per sequence (\eg, DAVIS \cite{DAVIS2017}: $\sim$2.51 objects/sequence, YoutubeVOS \cite{xu2018youtubevos}: $\sim$1.91 objects/sequence), we annotate all visible objects within the scene. This leads to on average 14.5 masklets per sequence, which is an order of magnitude larger than before. Second, previous datasets usually label objects with large size. 
However, in self-driving scenario, the camera can capture not only objects in vicinity but also extremely distant objects, such as those more than a hundred meters away. Furthermore, the sizes of different objects  vary a lot. For instance, objects far from the ego-car are extremely small in the images. Finally, due to the (large) relative motions between the self-driving vehicle and the objects, the objects tend to undergo drastic appearance and geometry changes, which makes the task very difficult.
In comparison to the recently introduced Kitti MOTS \cite{Voigtlaender_2019_CVPR_mots} self driving dataset where only cars and pedestrians were annotated in a semi-automatic fashion, we annotate manually all the 8 categories of cityscapes leading to about 50$\times$ more masklets and 140$\times$ more sequences as described in Table \ref{tab:datacompare}. Morover, while Kitti has captured data in only one city, CityscapesVideo benefits from the regional variation of the original Cityscapes dataset and as such provides a more diverse and challenging set of video sequences.

We evaluate our model and establish a baseline on this challenging dataset. In particular, we obtain a mIOU of 63.5 between our predicted masklets and the ground truth.


\begin{table*}[]
	\centering
	\begin{tabular}{cc|cccccccc}
		& \multicolumn{1}{l|}{} & \multicolumn{8}{c}{Number of Masklets for Each Category}           \\ \hline
		\multicolumn{1}{c|}{Split}      & \# Sequences             & Person & Rider & Car   & Truck & Bus & Train & Motorcycle & Bicycle \\ \hline 
		\multicolumn{1}{c|}{Train}      & 2706                  & 11667  & 1252  & 22415 & 580   & 415 & 129   & 511        & 2251    \\ 
		\multicolumn{1}{c|}{Validation} & 263                   & 1102   & 167   & 1673  & 28    & 27  & 32    & 63         & 192     \\ 
		\multicolumn{1}{c|}{Test}       & 496                   & 2624   & 478   & 3994  & 105   & 101 & 26    & 90         & 841     \\ \hline
	\end{tabular}

	\caption{\textbf{CityscapesVideo Statistics:} Number of annotated masklets (ie. tracked sequence of masks) per each instance category in our new introduced dataset.}
\label{tab:datastats}

\end{table*}

\begin{table*}[t]
	\centering
	\begin{tabular}{ccc|cccccccc}
		& \multicolumn{1}{l}{} & \multicolumn{1}{l|}{} & \multicolumn{7}{c}{Total Number of Masklets for Each Category}           \\ \hline

		\multicolumn{1}{c|}{}& \multicolumn{1}{c|}{\# Sequences} & \# Frames & person& rider & car & truck & bus & train & mcycle & bcycle   \\ 
		 \hline
		\multicolumn{1}{c|}{KITTI MOTS} & \multicolumn{1}{c|}{25} & 10870 & 395 & - & 582 & - & - & - & -\\
		\multicolumn{1}{c|}{CityscapesVideo} & \multicolumn{1}{c|}{3469} & 17345 & 15393 & 1897 & 28082 & 713 & 543 &161& 664 & 3284\\
		\hline

	\end{tabular}
	\caption{\textbf{CityscapesVideo vs KITTI MOTS  Statistics\cite{Voigtlaender_2019_CVPR_mots}:} We compare dataset statistics with the KITTI MOTS dataset. We compare the number of sequences, frames and total number of maskets (ie. tracked sequence of masks) for each category.}
	\label{tab:datacompare}
\end{table*}

\begin{figure*}
	\begin{centering}
		\includegraphics[width=1\linewidth, trim={0.75cm 3.5cm 0.75cm 3.75cm},clip]{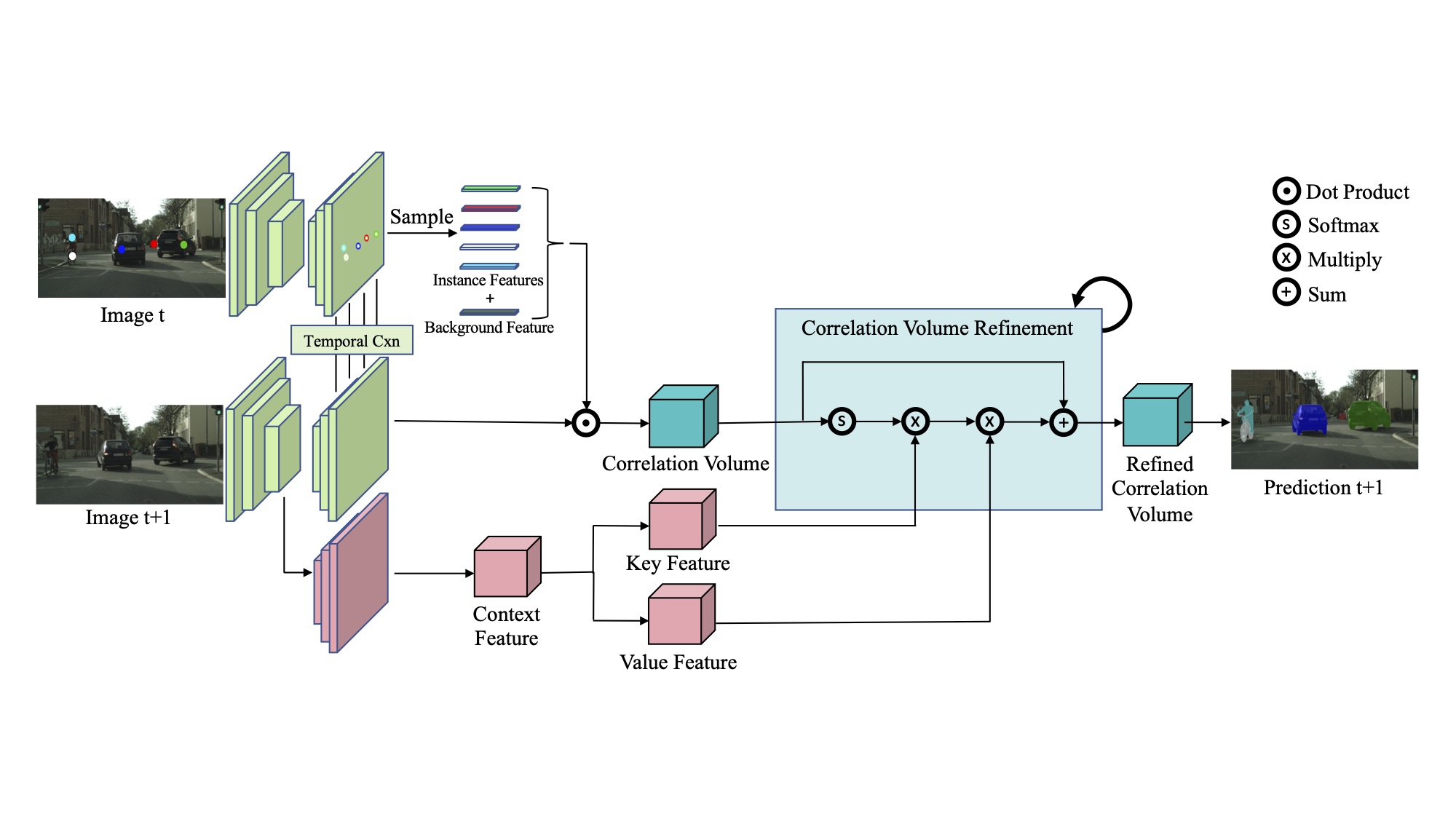}
	\end{centering}
	\caption{\textbf{VideoClick for single click video object segmentation:} Given a single click per object, our model samples instance features that are used to construct a correlation volume. This correlation volume assigns each feature pixel in the target frame to the corresponding instance feature in the reference frame or the background feature vector. Next, the cost volume is iteratively refined with the context feature using a recurrent attention mechanism to obtain the final segmentation.}
	\label{fig:model}
\end{figure*}

\section{Related work}
\paragraph{Instance segmentation:}
Modern instance segmentation methods can be categorized into two paradigms:  top-down or bottom-up.
For top-down approaches \cite{Dai2015InstanceAwareSS, 2016FullyCI, he2017mask, Liu2018PathAN, cai18cascadercnn, Chen2018MaskLab, Kim2018InstanceSA, upsnet, Chen_2019_CVPR, Fu2019RetinaMaskLT, huang2019msrcnn, Xu_2019_ICCV, Liang_2020_CVPR, Peng_2020_CVPR, Cao_2020_CVPR, Zhang_2020_CVPR, Fan_2020_CVPR, Wu_2020_ECCV}, region proposals are predicted and a voting process is used to filter out confident proposals. Most recently, LevelSet R-CNN \cite{Homayounfar_2020_ECCV} combines the Mask R-CNN framework \cite{he2017mask} with the classic variational level set method to segment the instances in a structured fashion. PolarMask \cite{xie2019polarmask} predicts the contour of an instance using polar coordinates. CondInst \cite{Tian_2020_ECCV} employs dynamic instance-aware networks to eliminate the need for ROI cropping and feature alignment. Point-Set Anchors \cite{Wei_2020_ECCV} uses a bounding box polygon as initialization from which they predict an offset for each point to bring it to the object boundary. SeaNet \cite{Chen_2020_ECCV} proposes a supervised edge attention module in the mask head and a new branch to learn IoU scores for the bounding boxes. 

For bottom-up approaches \cite{2015ProposalFreeNF, zhang2016instance, zhang2015monocular, Uhrig2016PixelLevelEA, Brabandere2017SemanticIS, newell2017associative, Fathi2017SemanticIS, kirillov2017instancecut, bai2017deep, liu2017sgn, Kendall_2018_CVPR, Neven_2019_CVPR}, instances in a scene are segmented without an explicit object proposal. For example, PatchPerPix \cite{Mais_2020_ECCV} predicts dense local shape descriptors and assembles them to predict the instance segmentation. \cite{Wolf_2020_ECCV} proposes a greedy algorithm for joint graph partitioning and labeling derived from the efficient Mutex Watershed partitioning algorithm. In \cite{Yuan_2020_NIPS}, the authors adopt a variational loss to handle multiple instances in a permutation-invariant way. 

\paragraph{Video segmentation:}
Video segmentation \cite{Zeng_2019_ICCV, Lin_2019_ICCV, Wang_2019_ICCV, Zhang_2019_ICCV, Duarte_2019_ICCV,  Johnander_2019_CVPR, NIPS2017_6636, Wang_2019_ICCV_RANet, hu2019learning, Yang_2019_ICCV, Oh_2019_ICCV} has become a more and more popular research area. RVOS \cite{Ventura_2019_CVPR} proposes a recurrent model that is recurrent in both spatial and temporal domain. \cite{Voigtlaender_2019_CVPR_mots} introduces a new video instance segmentation dataset and extend Mask R-CNN to video by using 3D convolutions and an association embedding branch to track objects. In \cite{Lin_2020_CVPR}, the authors combine Mask R-CNN with a modified variational autoencoder to output video segmentation and tracking. MaskProp \cite{Bertasius_2020_CVPR} adapts Mask R-CNN to video by adding a mask propagation branch that propagates instance masks from each video frame to one another. In \cite{Huang_2020_CVPR}, the authors propose a temporal aggregation network and a dynamic time evolving template matching mechanism. STEm-Seg \cite{Athar_2020_ECCV} models a video clip as a single 3D spatio-temporal volume, then segments and tracks instances across space and time in a single stage. SipMask \cite{Cao_2020_ECCV} uses a novel light weight spatial preservation module and a tracking branch to conduct single stage  video instance segmentation. 

\paragraph{Semi-automatic annotation:}
There has been significant efforts on speeding up the pixel-level image labeling process, such as incorporating the feedback from humans into the model \cite{Gulshan_2010, 6126343, grabcut, Xu_2016, Li_2018_CVPR, Bai_2014_CVPR, jang2019interactive, Sofiiuk_2020_CVPR, Ding_2020_ECCV, Kont_2020_ECCV, Heo_2020_ECCV,Majumder_2019_CVPR,Zhang_2020_CVPR}, or parameterizing the output in a way that is easy for annotators to adjust and refine \cite{PolyRNN,polygon-rnn++,Liang_2020_CVPR,ling2019fast,Dong2019AutomaticAA,Wang_2019_CVPR}.
In \cite{Oh_2019_CVPR, Miao_2020_CVPR}, annotators provide scribbles drawn on the image to iteratively refine each object mask across video frames. 
ScribbleBox \cite{Chen_2020_ECCV} have annotators interactively correct both box tracks and segmentation masks for video object segmentation. 
\cite{Shen_2020_ECCV} interactively annotate 3D object geometry using 2D scribbles.
In \cite{Lin_2020_CVPR}, users provide an initial click to segment the main body of the object and then iteratively provide more points on mislabeled regions to refine the segmentation. 
Polygon-RNN series \cite{PolyRNN,polygon-rnn++,ling2019fast} take as input the ground truth bounding box and predict the object mask in the form of polygons so that the annotators or other models \cite{Liang_2020_CVPR} can easily refine.  DEXTR \cite{Man+18} adopts extreme points of instances as input to segment objects and show that this is much more efficient than leveraging bounding box. 
In mapping papers, \cite{Liang2018EndtoEndDS, HomayounfarMLU18, LiangHMWU19, Homayounfar_2019_ICCV} have shown representing crosswalks, roads and lanes as polylines can be an efficient way to speed up annotation of such map elements.

\paragraph{Point-based methods:} Extracting image features for a given set of points and directly operating on them has gradually drawn wider attention, due to its strong performance, as well as speed and memory advantage.
PixelNet \cite{PixelNet} extract hypercolumn features from sampled pixel coordinates to perform segmentation. 
PointRend \cite{Kirillov_2020_CVPR} iteratively refines a coarse instance segmentation mask by making predictions on an adaptively selected set of finer points. 
CenterMask \cite{Wang_2020_CVPR} performs instance segmentation by predicting a heat map of point locations from which a feature is extracted and processed to output a mask. 
Similarly, PointINS \cite{qi2020pointins} use an instance aware convolution to perform instance segmentation using single point image features. 
Dense RepPoints \cite{Yang_2020_ECCV} generates a set of initial points, refine them based on the sampled features, and then post-process to obtain the final mask.
In \cite{Xu_2020_ECCV}, the authors tackle the task of multi-object tracking and segmentation by representing them as a set of randomly selected points. 


\section{Single Click Video Object Segmentation}

We tackle the problem of single click video object segmentation. Let $\set{I_t \in \R^{3\times H \times W}}_{t=1}^T$ be a sequence of RGB images taken from a video. 
In our setting we assume the following workflow for the annotator. 
The annotator provides one and only one click per object in the first frame that it appears in the video snippet.
 Note that this click can be at any random point inside the object,  preferably not close to its boundaries.
We denote by $P_t = \set{p_{t,n}}_{n=1}^{N_t}$  the click coordinates at time $t$  and by $N = \sum N_t $ the total number of clicks in the snippet.   If the object disappears, say due to occlusion, the annotator clicks on it once again when it reappears in the video. 
 Given these single clicks for the $N$ objects, our goal is then to obtain their corresponding segmentation masks in the video,  denoted by $\set{M_t \in \N^{ H \times W}}_{t=1}^T$. In particular, at a pixel position $(i,j)$ at time $t$ we have $M_t(i,j) \in \set{0, 1, \dots, N}$ where the label $n  \in \set{1, \dots, N}$ corresponds to one of the $N$ objects specified by the annotator and the label 0 represents the background.

We develop our model to operate on a pair of consecutive RGB images $I_t$ and $I_{t+1} $ and then later extend it to the full snippet $\set{I_t}_{t=1}^T$. Specifically, suppose an annotator highlights with single point clicks $P_t = \set{p_{t, n}}_{n=1}^{N_t}$ all the visible objects in $I_t$. Given these point clicks, our goal is to obtain their corresponding mask $M_{t+1}$ in $I_{t+1}$. At a high level,   our model has three steps as shown in Figure. \ref{fig:model}: First we obtain deep spatio-temporal features on top of $I_t$ and $I_{t+1}$. Next, given the annotator point clicks, we sample instance features and fuse this with target image features to construct a correlation volume. This volume scores the compatibility of each pixel in $I_{t+1}$ to the instances in $I_{t}$ specified by the annotator point clicks or to the background. Finally, the correlation volume is refined for a number of iterations by a recurrent attention network. The correlation volume can be converted into a segmentation mask by taking a $\arg\max$ over the score of these instances and the background assigned to each pixel in $I_{t+1}$. In the following, we describe each component of our model in detail.

\paragraph{Deep Spatio-Temporal Features:}

We input the images $I_{t}$ and $I_{t+1}$ to a siamese encoder-decoder network. The encoder is a 2D residual network \cite{He2015DeepRL}, and the decoder is based on FPN \cite{Lin2016FeaturePN} augmented with temporal connections to aggregate the features across time. We detail the exact decoder architecture in the supplemental material. At the end, we obtain down-sampled features $F_t$ and $F_{t+1} \in \R^{D\times H/4 \times W/4}$ corresponding to $I_t$ and $I_{t+1}$ respectively. These features will contain information about the object masks and also their association in time and will be used to construct a correlation volume between the pixels of $I_{t+1}$ and the keypoints of $N_t$ instances in $I_t$ or the background.

\paragraph{Keypoint and Background Features:}

 Given the keypoints $P_t = \set{p_{t, n}}_{n=1}^{N_t}$ highlighted by the annotator, we extract using bilinear interpolation $N_t$ vectors of dimension $D$  from the feature map $F_t$. Each extracted feature vector captures the spatio-temporal semantic information about its corresponding instance specified by the annotator click. The goal is to match each pixel in $I_{t+1}$ to one of these $N_t$ instances. However, a pixel in $I_{t+1}$ could also correspond to the background which could include a new object appearing for the first time in $I_{t+1}$. As such, we simply define a global learnable $D$-dimensional feature vector corresponding to the background.  Finally, we concatenate the $N_t$ object feature vectors and the background feature vector to create a matrix $E$ of dimensions $  (N_t+1)  \times D$ which will be used in the correlation volume construction.

\paragraph{Correlation Volume:} 
To obtain the segmentation masks in image $I_{t+1}$ corresponding to either the $N_t$ objects in $I_t$ or the background, we construct an initial correlation volume $C^0_{t+1} \in \R^{(N_t+1) \times H / 4 \times W / 4}$ where $C^0_{t+1}(n, i, j)$ scores the compatibility of feature pixel $(i,j)$ in $F_{t+1}$ with row $n$ of $E$ which is the feature vector of either the background or one of the $N_t$ instances in $I_t$.  $C^0_{t+1}$ is constructed by taking the dot product between the rows of the matrix $E$ and each feature pixel in $F_{t+1}$ as follows \footnote{We use a superscript to denote the initial cost volume and its refinement and a subscript for the video timesteps.}:
\begin{equation}
C^0_{t+1}(n, i, j) = \sum_{h} E(n,h) F_{t+1}(h, i, j)
\end{equation}
The correlation volume can then be converted into a segmentation mask by taking an $\arg\max$ along the first dimension of $C^0_{t+1}$:

\begin{align}
	M^0_{t+1}(i,j) = \argmax{n}{C^0_{t+1}(n, i,j)}
\end{align}
Note that if an object disappears in $I_{t+1}$, then its corresponding channel in $C^0_{t+1}$ would not  have the highest score on any of the pixels and as such its label will not appear in the mask. In practice however, we could have spurious pixels in $I_{t+1}$ assigned to an occluded instance.  As such, we consider an instance to be occluded if the area of its mask is less than a threshold obtained from the validation set.

\paragraph{Correlation Volume Refinement:} Next, we proceed to refine the initial estimate $C^0_{t+1}$ through a recurrent attention module that depends on $C^0_{t+1}$ and the  features  extracted solely from $I_{t+1}$. In particular, we input the encoder features of $I_{t+1}$ to a new decoder to obtain features $F_{context}\in \R^{D\times H/4 \times W/4}$ which would have better localized information about the object boundaries in $I_{t+1}$ without being affected by the features of $I_t$.  At a high level, at each refinement step, we find the feature channels from $F_{context}$ that are closest to a predicted instance mask captured in the correlation volume and use them to refine the correlation. Our refinement update rule is inspired by the self attention block of the transformers \cite{NIPS2017_3f5ee243} and proceeds as follows at each refinement timestep $\tau$: 

First, we map the feature map $F_{context}$ to embedding tensors $F_{key}$ and $F_{value}$ using two residual blocks.  Next, we convert the previous correlation volume $C^{\tau-1}_{t+1}$ to a probability tensor $S^{\tau-1}$ by taking a softmax along its first dimension.  Now, each channel of $S^{\tau-1}$ represents the probability mask of the corresponding object or the background.

Then we compute an attention matrix $A^{\tau-1}\in \R^{(N_t+1) \times D}$ between $S^{\tau-1}$ and $F_{key}$:
\begin{align}
A^{\tau-1} (n, d) = \softmax{dim=1}\sum_{i, j} S^{\tau-1}(n, i, j)F_{key}(d, i, j)
\end{align}
Here $A^{\tau-1} (n, d)$ measures the compatibility of the probability mask of the $n$-th instance or the background and the $d$-th feature channel of $F_{key}$.
 Finally, we multiply  $A^{\tau-1}$ by $F_{value}$ to obtain a residual score that will be added to $C^{\tau-1}_{t+1}$. 
\begin{align}
C^{\tau}_{t+1}  = C^{\tau-1}_{t+1} +A^{\tau-1}F_{value} 
\end{align}
We repeat this update rule for a fixed number of timesteps.

\paragraph{Extending to the full Snippet:}
Next, we extend our model from operating on a pair of images to the full snippet.  As before, denote the annotator clicks for new objects appearing at time $t$ by $P_t$. Also in contrast with the two frame case, we have in addition a correlation volume $\hat{C}_t$ from the previous timestep. The next step is to  obtain feature vectors corresponding  to $P_t$ and $\hat{C}_t$. For keypoints $P_t$, we use bilinear interpolation to extract keypoint features from feature map $F_t$. Next, we need to convert $\hat{C}_t$ to a set of of keypoint features. First we obtain the mask of each instance from $\hat{C}_t$.  Then in order to remove potential mask outliers, we find the coordinates of the top 50\% highest scoring mask pixels for each instance. Finally, we average those points on $F_t$ to create a $D$-dimensional feature vector for each instance.

Now, we'll explain the full snippet algorithm \ref{alg:vidclick}:  Given previous correlation volume $\hat{C}_t$ and new keypoints $P_t$, we apply the model on the same frame $I_t$ to obtain an updated correlation volume $C_t$. In particular, whereas $\hat{C}_t$ gives the compatibility of each pixel in $I_t$ to instances appearing in previous frames $ I_{t-1}$, the new correlation $C_t$ takes into account both $\hat{C}_t$ and the new annotator clicks $P_t$. The segmentation mask $M_t$ is obtained from $C_t$. Finally, we supply the consecutive frames $I_t$ and $I_{t+1}$ and the keypoint features corresponding to $C_t$ to our model to obtain $\hat{C}_{t+1}$ and the process continues.

\begin{algorithm}[tb]
	\SetAlgoLined
	\SetKwInOut{Input}{Input}\SetKwInOut{Output}{Output}\SetKwInOut{Init}{Init}
	\Input{Sequence of frames $\set{I_t}_{t=1}^T$ \\ Annotator clicks $\set{P_t}_{t=1}^T$ of new objects appearing at time $t$}
	\Output{Segmentation masks $\set{M_t}_{t=1}^T$}
	\Init{$\hat{C}_{0} \leftarrow None$ }
	\For{$t = 1, \dots, T$}{
	$C_t \leftarrow$ model($I_t, I_t, P_t, \hat{C}_{t}$)\\
	$M_t \leftarrow \softmax{dim=0}{C_t}$\\
	 $P_t \leftarrow None$\\
	 \If{ $t \leq T - 1$}{
	 $\hat{C}_{t+1} \leftarrow$ model($I_t, I_{t+1}, P_t, C_t$)
	 }
	}
	\caption{VideoClick}
	\label{alg:vidclick}
\end{algorithm}

\paragraph{Learning:}
We learn all the parameters of the model in an end-to-end fashion.  In particular, from images $I_{t},I_{t+1}$ and keypoints $P_t$ we obtain a sequence of refined correlation volumes $C^{\tau}_{t+1}$ corresponding to $I_{t+1}$. After applying softmax on the cost volumes, we can directly compare with the ground truth masks $M^{GT}_{t+1}$ using cross entropy. In practice, rather than directly comparing $C^{\tau}_{t+1}$ with $M^{GT}_{t+1}$, we sample using bilinear interpolation a fixed number of random points corresponding to the background, the object interior and around the object boundaries from $C^{\tau}_{t+1}$ and $M^{GT}_{t+1}$ and compute the cross entropy on this points. Our goal is to balance the the effect of background pixels as well as object of varying sizes. We also found that the auxiliary task of binary foreground/background prediction with an extra residual block from the features $F_t$ and $F_{t+1}$ improves the performance. 
Note that we do not use this branch at test time.


\section{Experiments}

\begin{table*}[t!]
	\setlength{\tabcolsep}{2pt}
	\centering

		\begin{tabular}{c|c|c|cccccccc}
			\multicolumn{1}{c|}{} & \multicolumn{1}{c|}{mIOU$_ {\texttt{val}}$} & \multicolumn{1}{c|}{mIOU$_ {\texttt{test}}$} &  \multicolumn{1}{c}{person}& \multicolumn{1}{c}{rider} & \multicolumn{1}{c}{car} & \multicolumn{1}{c}{truck} & \multicolumn{1}{c}{bus} & \multicolumn{1}{c}{train} & \multicolumn{1}{c}{mcycle} & bcycle \\ 
			\hline

			Siamese Bounding Boxes  &$50.2$ &$44.5$ &$34.3$ &$35.6$ &$57.6$ &$58.7$ &$63.2$ &$37.7$ &$33.6$ &$35.6$ \\
			Mask R-CNN + Key Point Matching  &$46.6$  &$44.5$ &$43.1$ &$41.8$ &$59.5$ &$50.2$ &$59.7$ &$33.5$ &$35.3$ &$32.9$ \\
			Mask R-CNN +  Mask Matching    &$55.2$ &$51.6$ &$49.1$ &$47.7$ &$67.6$ &$58.2$ &$65.3$ &$41.3$ &$41.2$ &$42.2$ \\
			
			\hline
			\hline

			Ours  &\textbf{63.5} & \textbf{59.6} 
			& \textbf{53.5}  & \textbf{54.4} & \textbf{75.9} & \textbf{65.0}  & \textbf{73.1}  & \textbf{53.5}   & \textbf{50.8}  & \textbf{50.5} \\

			\hline
			
		\end{tabular}

	\caption{\textbf{Results on CityscapesVideo val and test sets:} This table shows the multi object tracking and segmentation results given a single click. We report mean IOU for both the val and test set and also the mean IOU per class.}
	\label{tab:results}
\end{table*}

\begin{table*}[t!]
	\setlength{\tabcolsep}{2pt}

	\centering

		\begin{tabular}{c|c|c|cccccccc}

			\multicolumn{1}{c|}{} & \multicolumn{1}{c|}{mIOU$_ {\texttt{val}}$} & \multicolumn{1}{c|}{mIOU$_ {\texttt{test}}$} &  \multicolumn{1}{c}{person}& \multicolumn{1}{c}{rider} & \multicolumn{1}{c}{car} & \multicolumn{1}{c}{truck} & \multicolumn{1}{c}{bus} & \multicolumn{1}{c}{train} & \multicolumn{1}{c}{motorcycle} & bicycle \\ 
			\hline

			Ours Evaluate on the Full Snippet    &61.5 & 58.1 & 52.3 & 53.9 & 73.7 & 62.5 & 69.7 & 52.6 & 50.6 & 48.5 \\
			Ours  &\textbf{63.5} & \textbf{59.6} 
			& 53.5  & 54.4 & 75.9 & 65.0  & 73.1  & 53.5   & 50.8  & 50.5 \\
			
			\hline
			
		\end{tabular}

	\caption{\textbf{Results on CityscapesVideo val and test full sequences:} This table shows the multi object tracking and segmentation results given a single click reported on the full 30 frame sequence. We report mean IOU for both the val and test set and also the mean IOU per class. We show that our model can generalize on a different frame rate despite being trained only on 5 frame sequences.}
	\label{tab:fullresults}
\end{table*}

\begin{table}[t]
	\centering
	\begin{tabular}{c|c}
		& mIOU \\ 
		 \hline
		Mask R-CNN + Keypoint Mask Filter & 61.8 \\
		 \hline
		 \hline
		Ours trained on Single Images & 71.4 \\ 
		 \hline
	\end{tabular}
	\caption{We report our model results run on single images and compare with Mask R-CNN. Both models are given the annotator key points and have to predict a segmentation mask. We report the mIOU result across the entire dataset.}
	\label{table:singleimage}
\end{table}

\paragraph{CityscapesVideo:} In this paper, we introduce a new video instance segmentation dataset for autonomous driving built on top of the popular Cityscapes dataset. In particular, we annotate the 8 object categories (bicycle, bus, person, train, truck, motorcycle, car and rider) of Cityscapes instance segmentation task with instance segmentation tracklets for 3475 training and validation sequences of the dataset. In the original dataset, only the 20th frame of a 30 frame sequence is annotated. We re-annotate the 20th frame and annotate 4 more at regular intervals so that we obtain a sequence of 5 frames at 360 ms, specifically we annotate frames 1, 7, 13, 19 and 25 of each sequence. For this paper, we set aside two cities (weimar and zurich) in the training set as validation and consider the actual validation set to be the test set.  This results in a total of 13530/1315/2500 frames for the train/val/test sets.

\paragraph{Implementation Details} We train our model on the Cityscapes video dataset on pairs of images. Each pair could either be the same image or two consecutive images with the original $1024 \times 2048$ resolution. We also augment the training dataset by random horizontal flipping with equal probability. We train the model on 16 RTX5000 GPUs for 25 epochs with the Adam \cite{adam_opt} optimizer with learning rate of 0.0001 and weigh decay of 0.0001. We take the best model from evaluating on the validation set. For the backbone, we employ \texttt{ResNet-101} \cite{He2015DeepRL} pretrained on Imagenet \cite{alexnet2012} and use our modified version of FPN \cite{Lin2016FeaturePN} with random initialization. In order to fit our model operating on original image resolutions, we employ gradient check-pointing \cite{gradckpt} during training.

\paragraph{Metrics} For our metric, we compute the mIOU between the prediction mask volume $\set{M_t}_{t=1}^T$ and the corresponding ground truth mask volume. The volume has a size of $N \times T \times H \times W$, where $N$ is the total number of instances in the sequence and $T$ is the number of time steps in the sequence. That is, each mask has its own 2D canvas that we plot it on. The benefit of such a metric is that it is a measure of both the segmentation and tracking quality of the model. Such a metric will also penalize both false negatives and false positives.

\begin{figure*} 
	\centering
	\setlength\tabcolsep{0.5pt}
	\begin{tabular}{cccccc}
		
		\raisebox{2px}{{t=1}} &	
		\raisebox{2px}{{t=2}} &
		\raisebox{2px}{{t=3}} &
		\raisebox{2px}{{t=4}} &
		\raisebox{2px}{{t=5}}  \\

        \raisebox{21px}{\rotatebox{90}{\small GT}}
        \includegraphics[width=.195\textwidth]{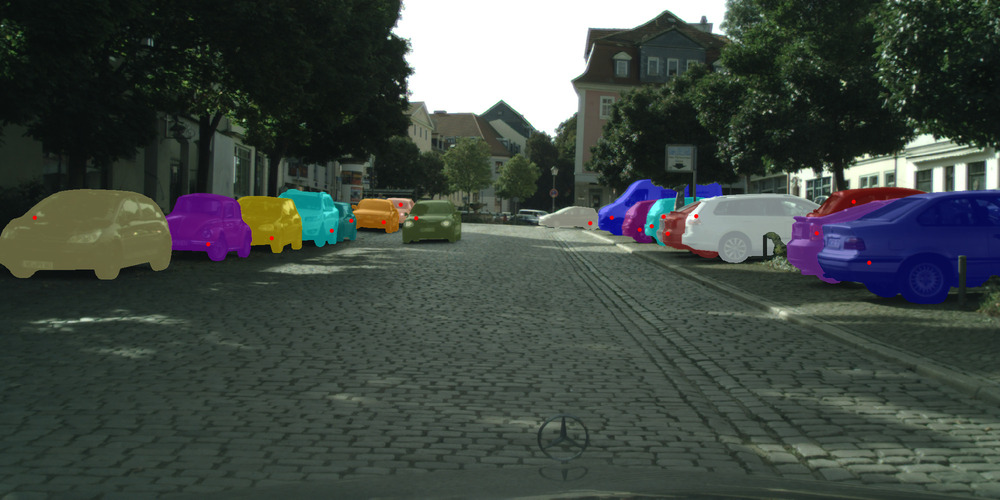} &
        \includegraphics[width=.195\textwidth]{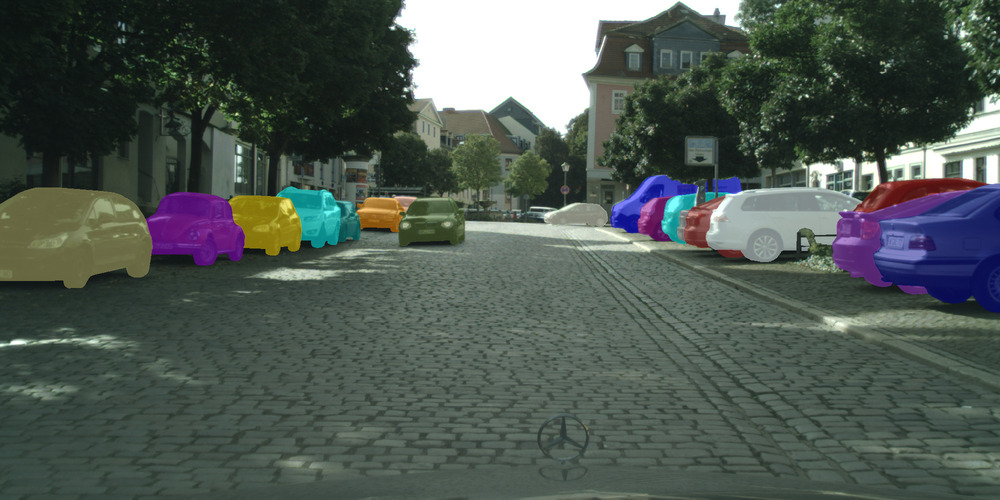} &
        \includegraphics[width=.195\textwidth]{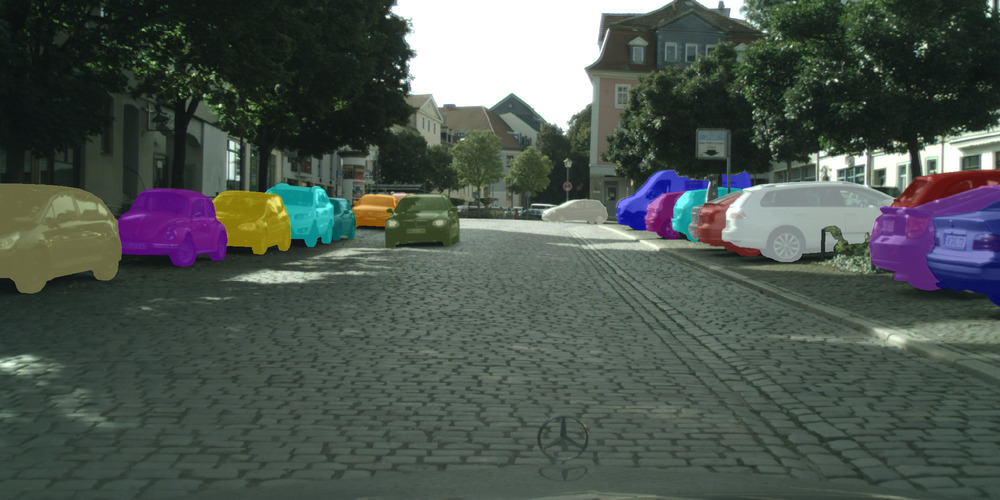} &
        \includegraphics[width=.195\textwidth]{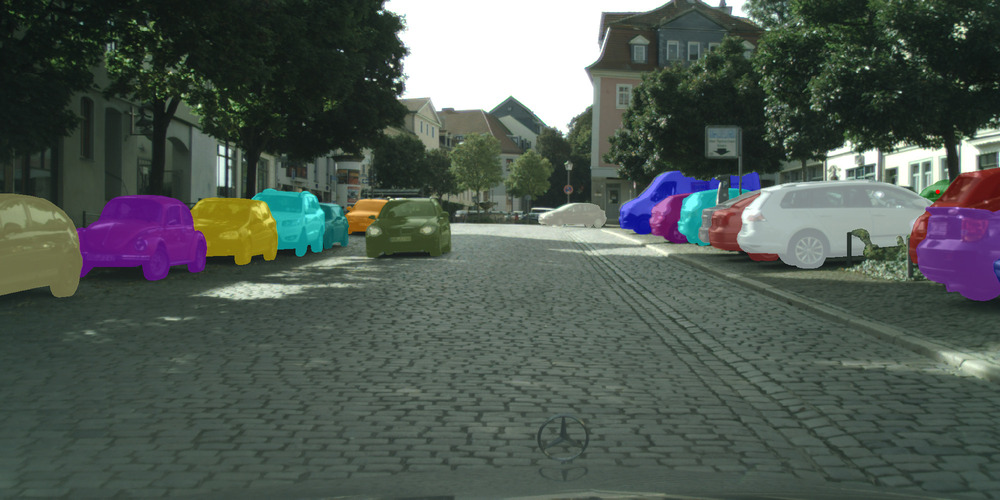} &
        \includegraphics[width=.195\textwidth]{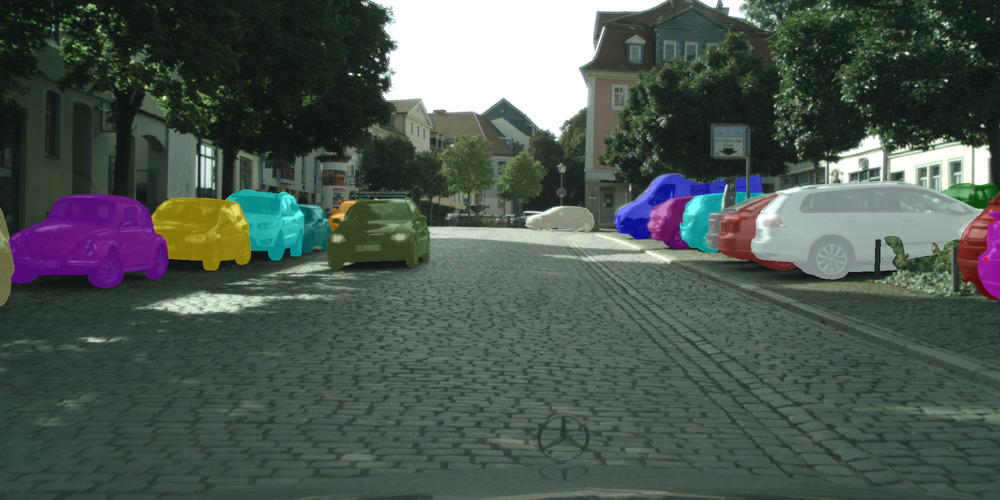} \\     
        
        \raisebox{19px}{\rotatebox{90}{\small Ours}}
        \includegraphics[width=.195\textwidth]{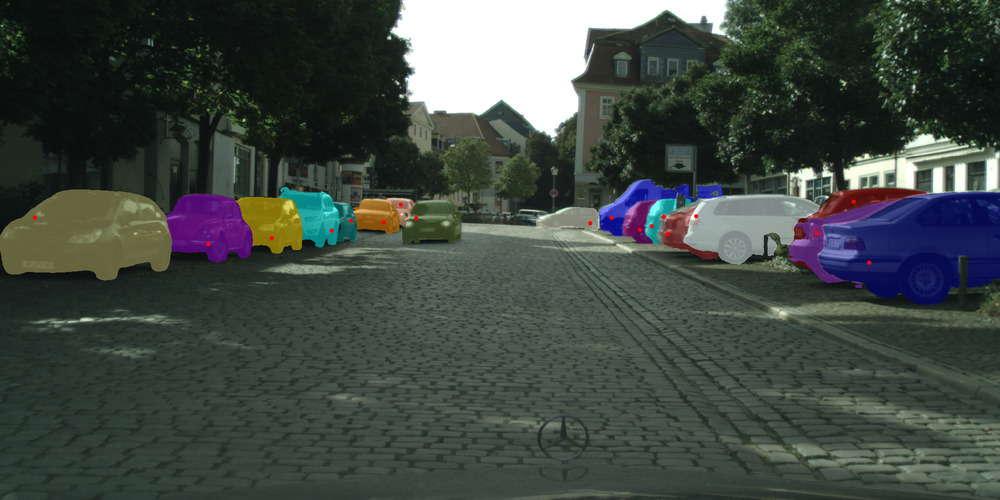} &
        \includegraphics[width=.195\textwidth]{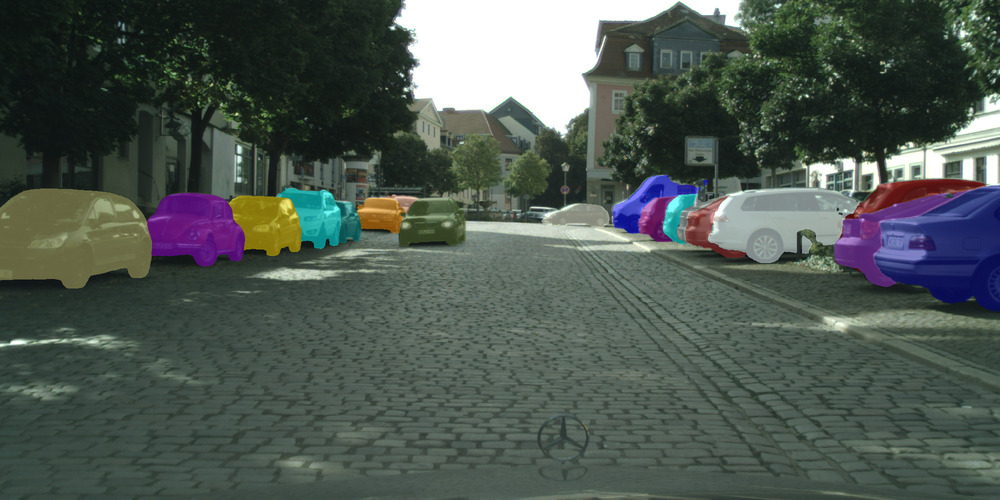} &
        \includegraphics[width=.195\textwidth]{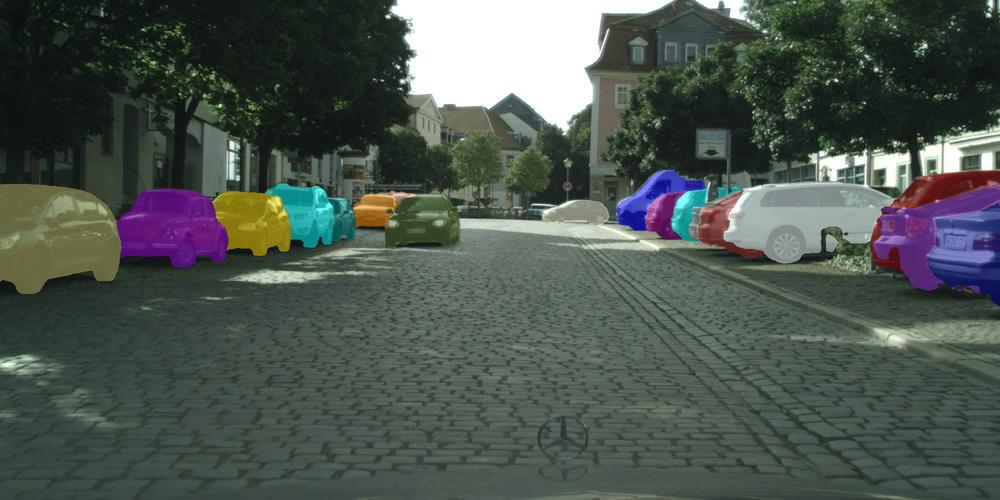} &
        \includegraphics[width=.195\textwidth]{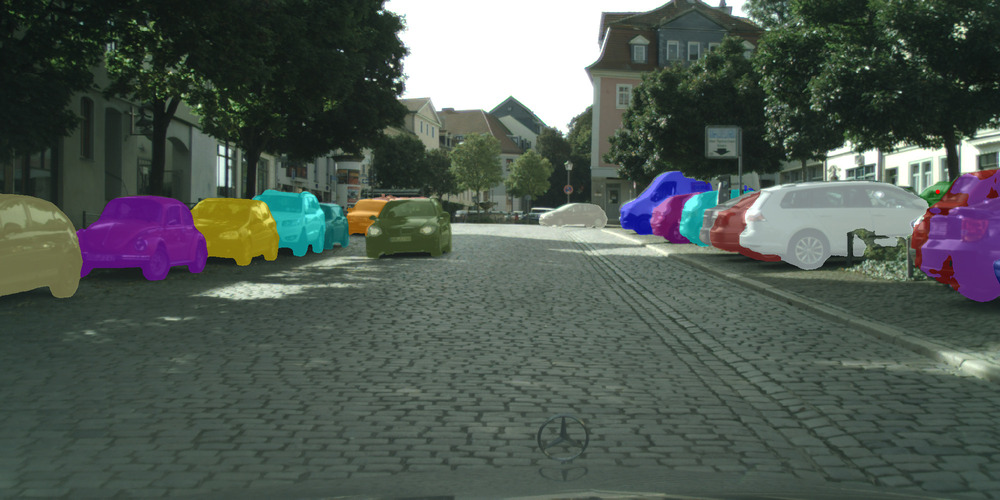} &
        \includegraphics[width=.195\textwidth]{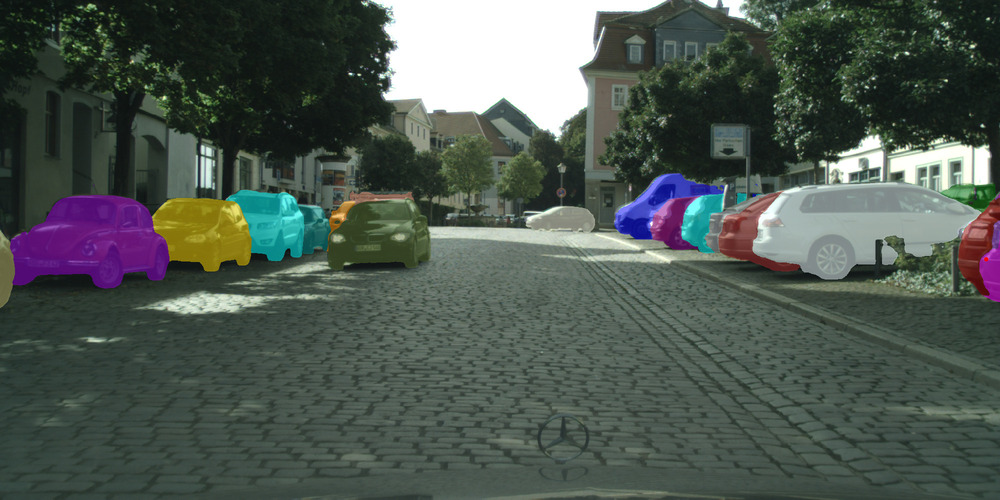} \\   
        
        \raisebox{3px}{\rotatebox{90}{\small MRCNN+M}}
        \includegraphics[width=.195\textwidth]{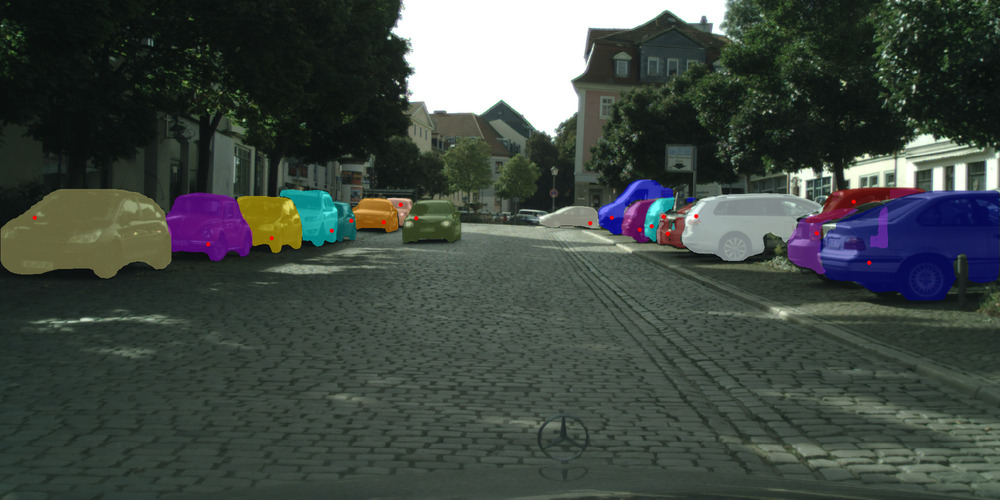} &
        \includegraphics[width=.195\textwidth]{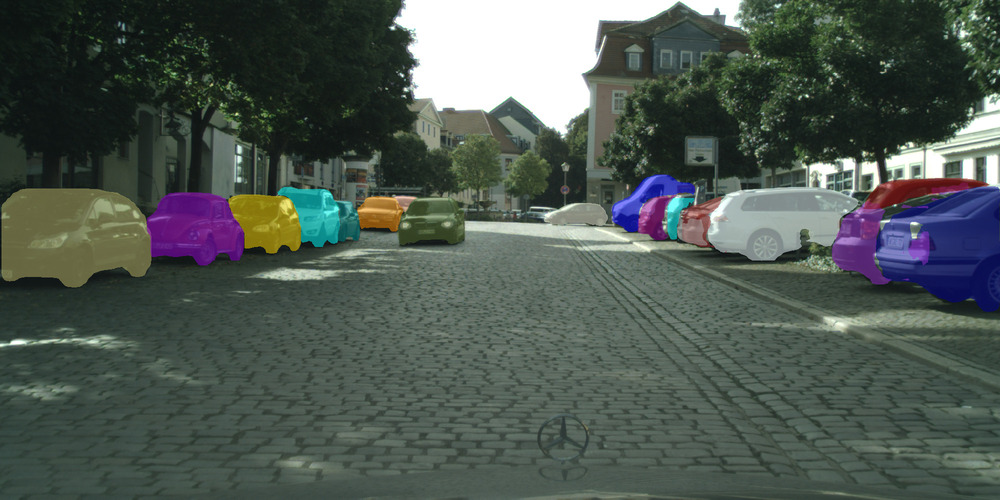} &
        \includegraphics[width=.195\textwidth]{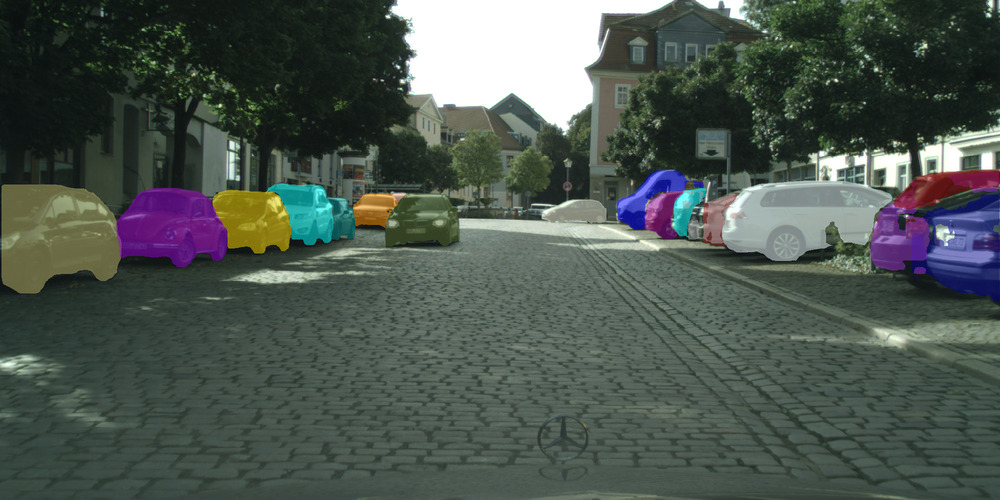} &
        \includegraphics[width=.195\textwidth]{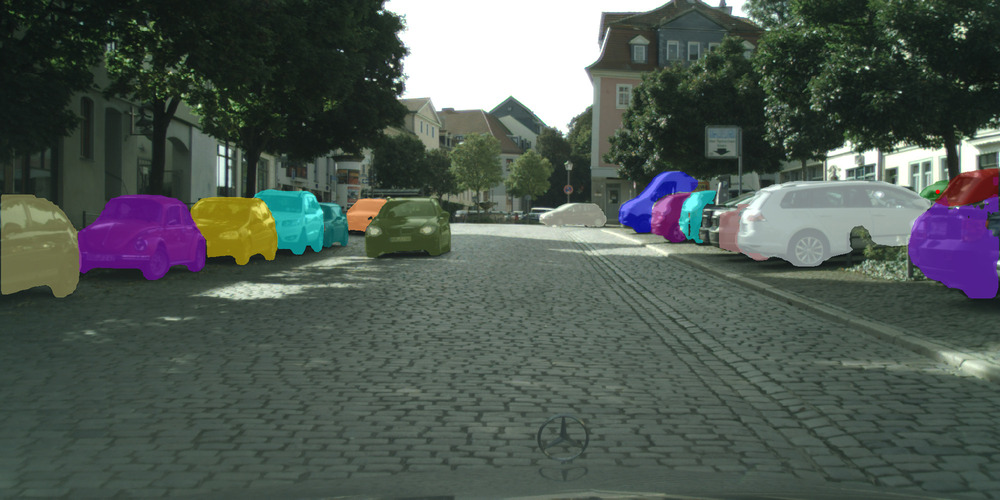} &
        \includegraphics[width=.195\textwidth]{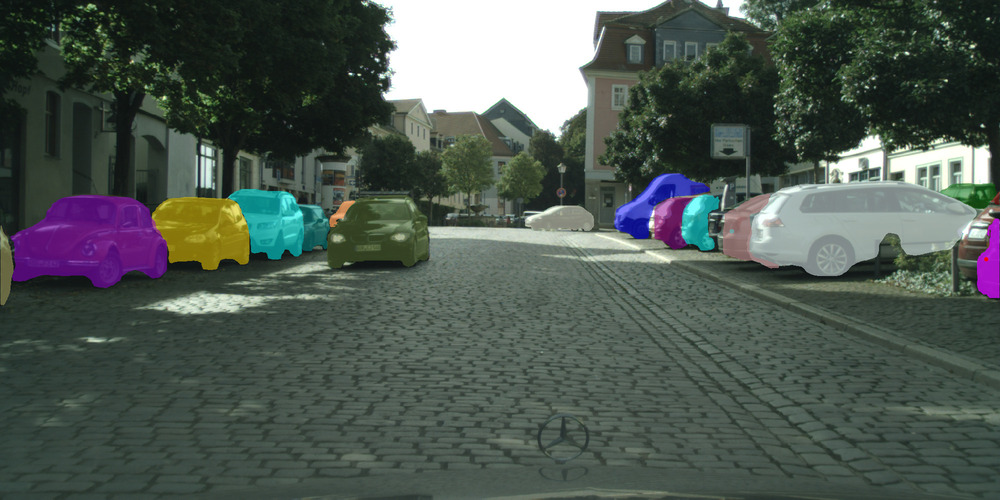} \\
        
        \raisebox{21px}{\rotatebox{90}{\small GT}}
        \includegraphics[width=.195\textwidth]{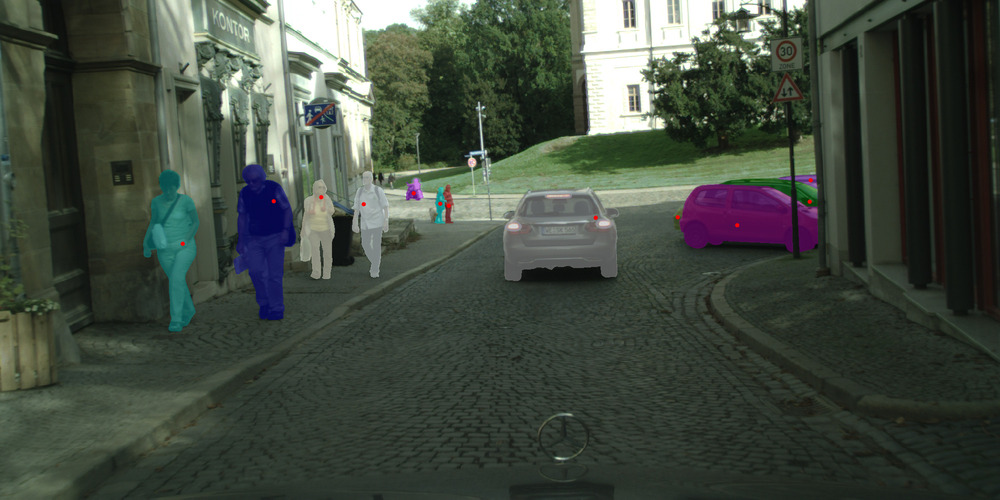} &
        \includegraphics[width=.195\textwidth]{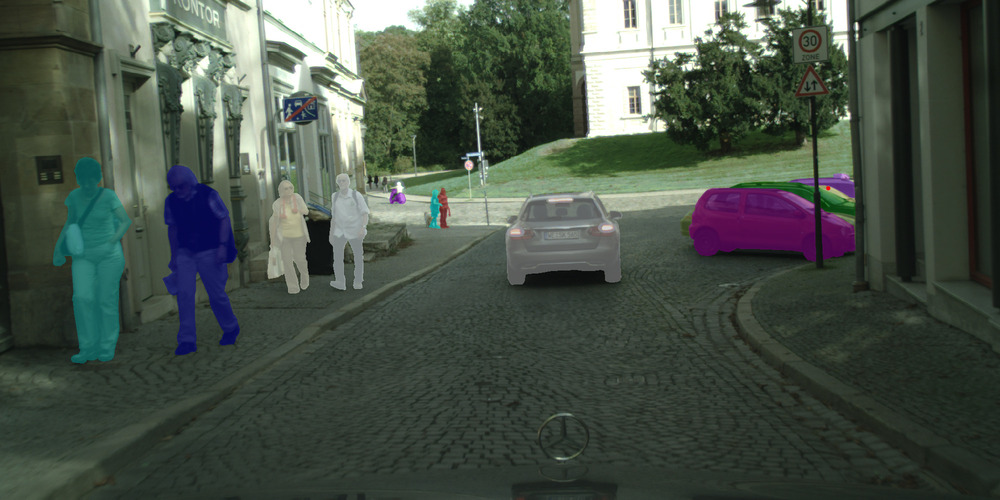} &
        \includegraphics[width=.195\textwidth]{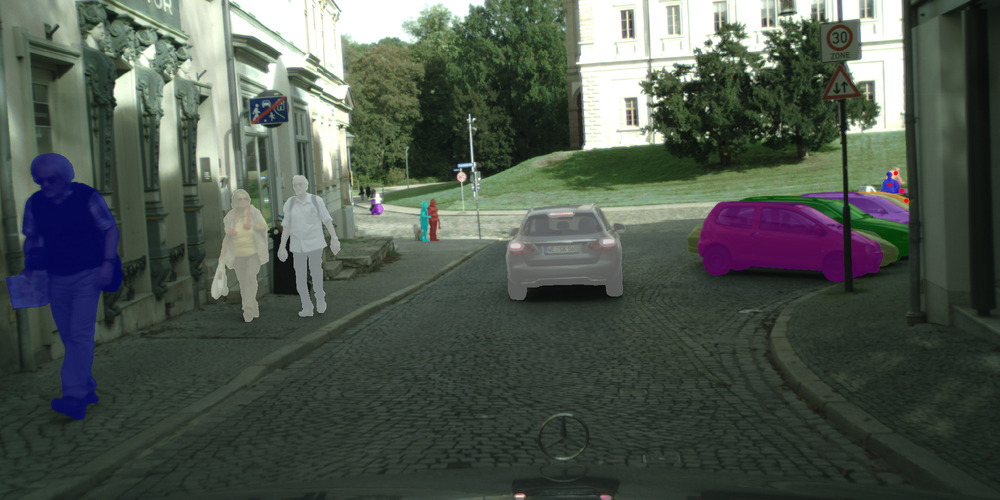} &
        \includegraphics[width=.195\textwidth]{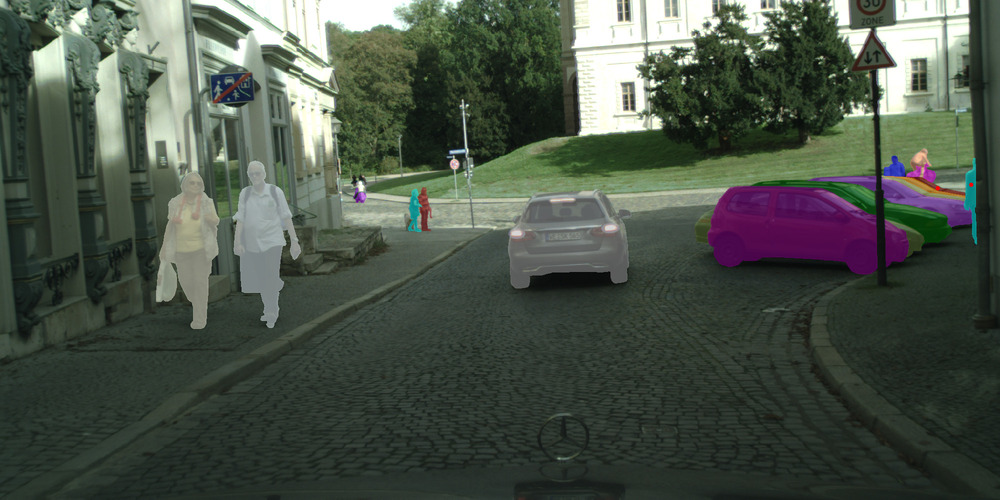} &
        \includegraphics[width=.195\textwidth]{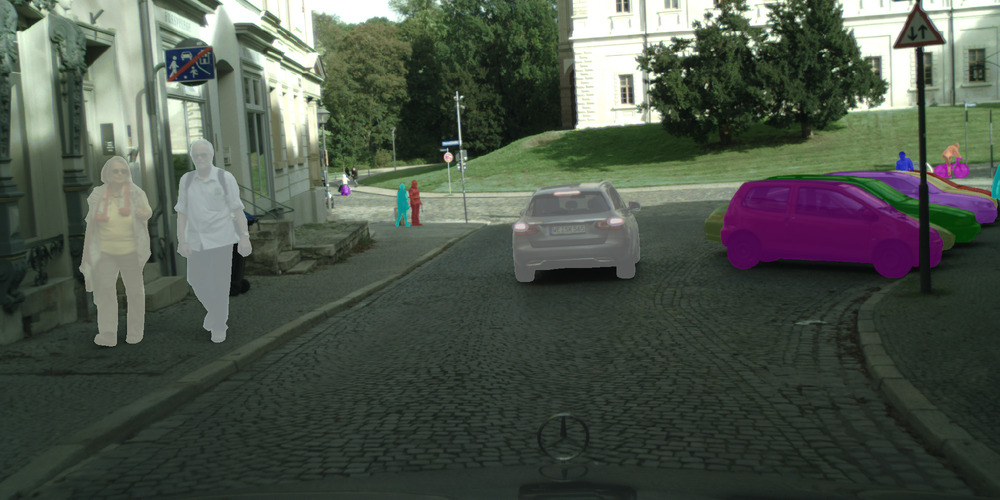} \\     
        
        \raisebox{19px}{\rotatebox{90}{\small Ours}}
        \includegraphics[width=.195\textwidth]{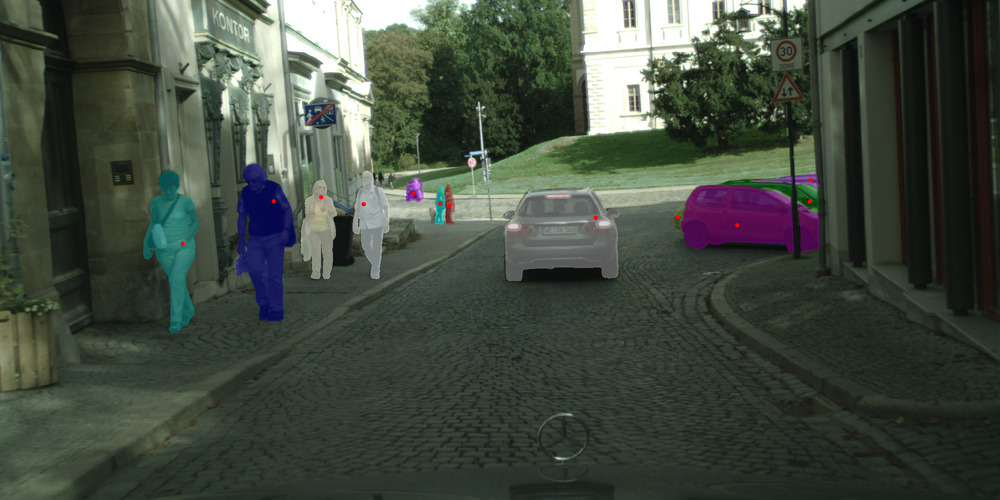} &
        \includegraphics[width=.195\textwidth]{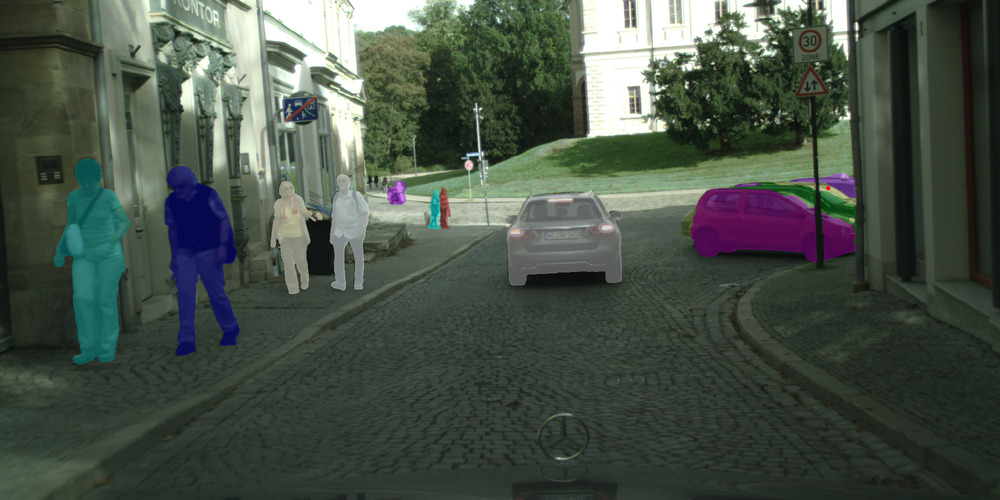} &
        \includegraphics[width=.195\textwidth]{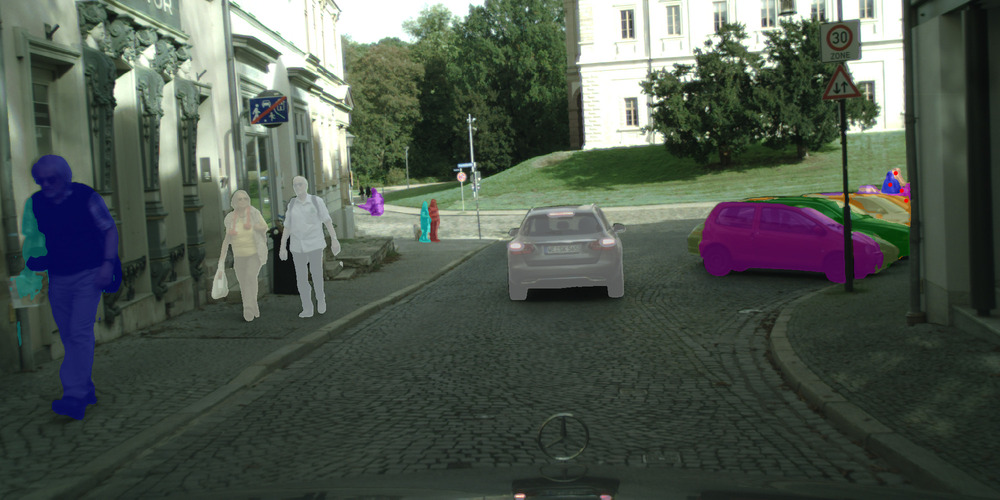} &
        \includegraphics[width=.195\textwidth]{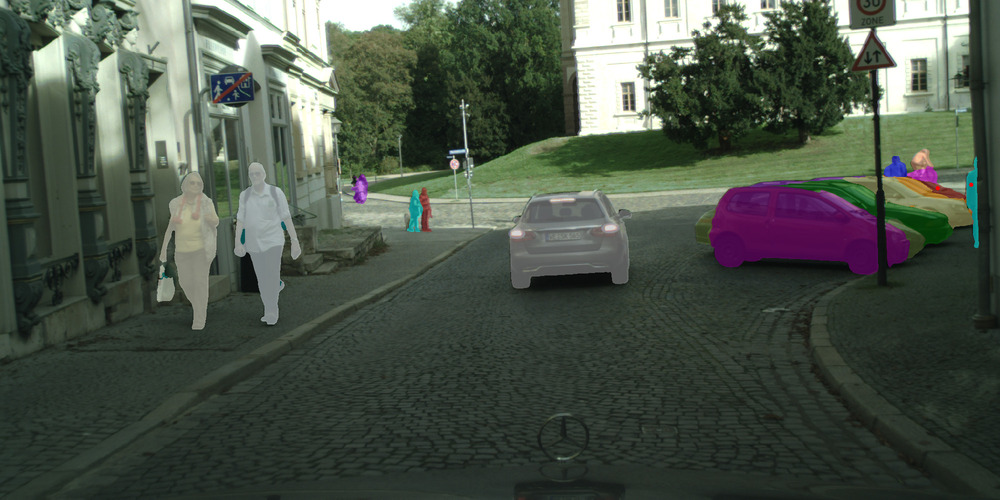} &
        \includegraphics[width=.195\textwidth]{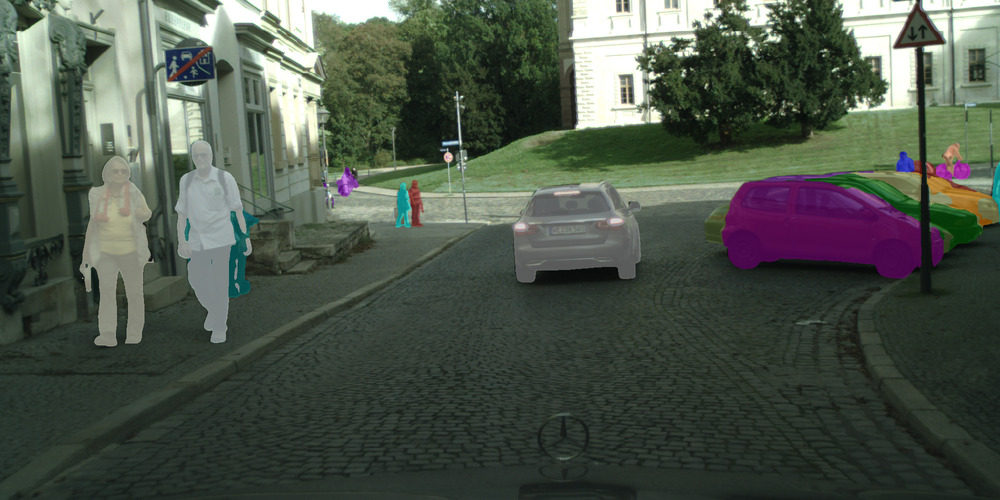} \\   
        
        \raisebox{3px}{\rotatebox{90}{\small MRCNN+M}}
        \includegraphics[width=.195\textwidth]{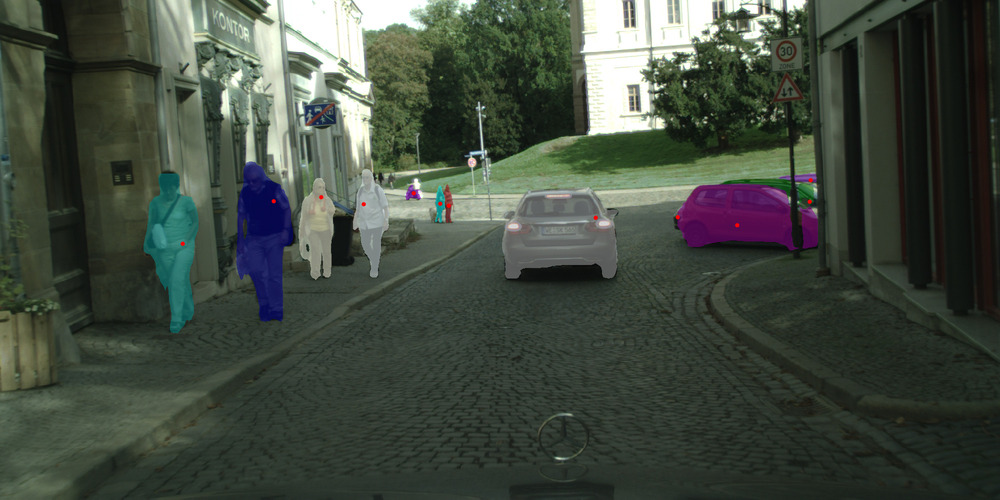} &
        \includegraphics[width=.195\textwidth]{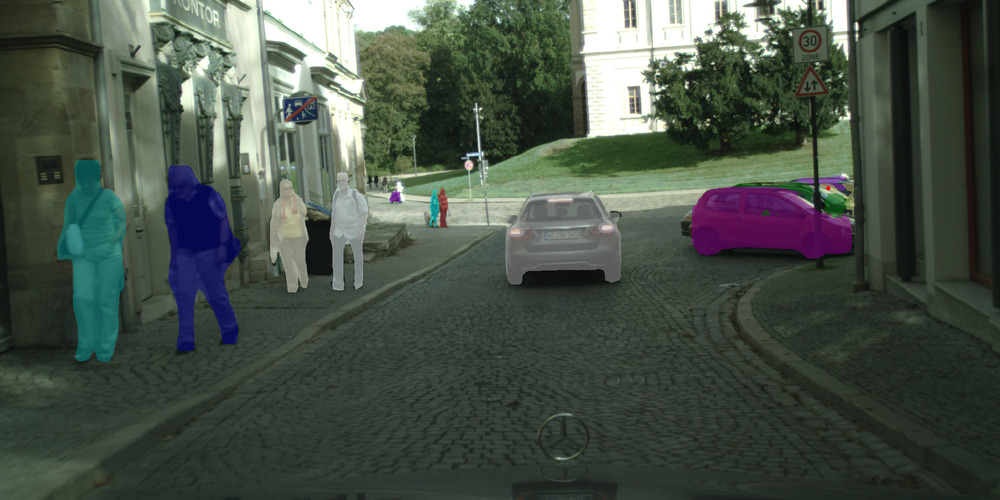} &
        \includegraphics[width=.195\textwidth]{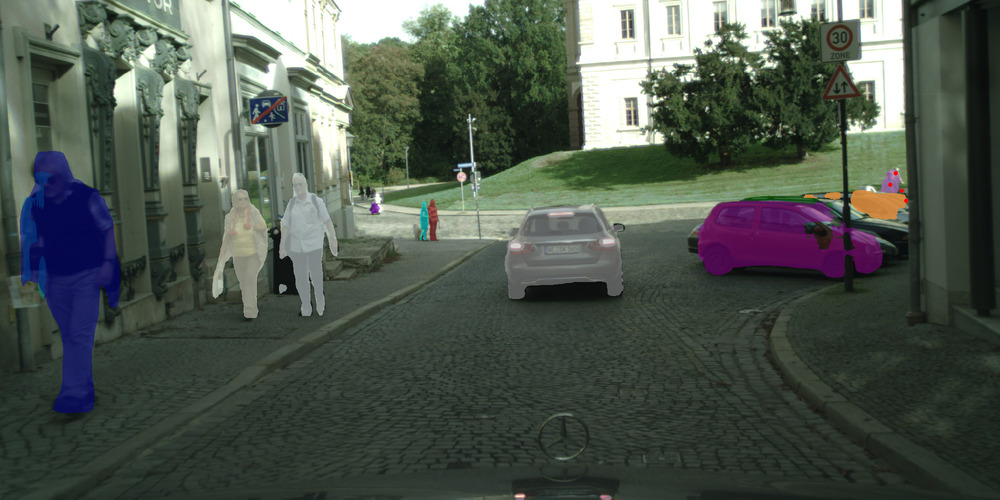} &
        \includegraphics[width=.195\textwidth]{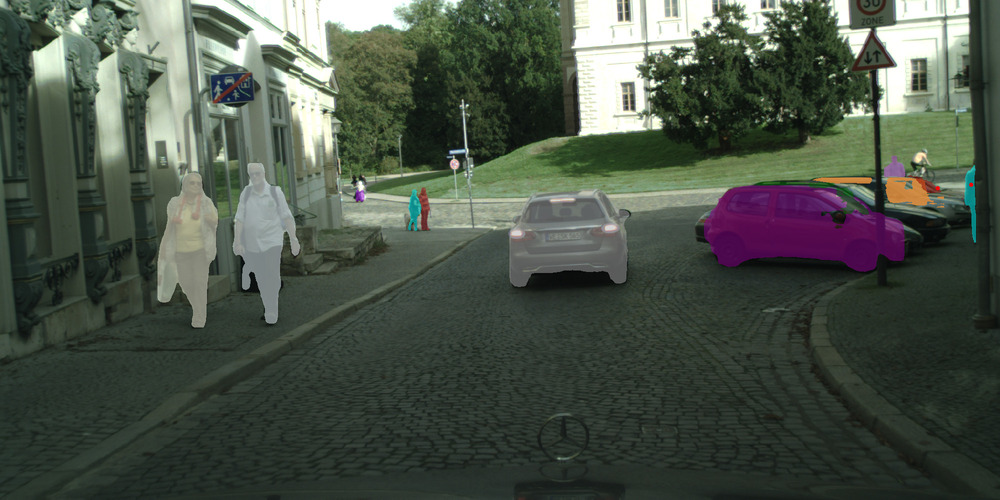} &
        \includegraphics[width=.195\textwidth]{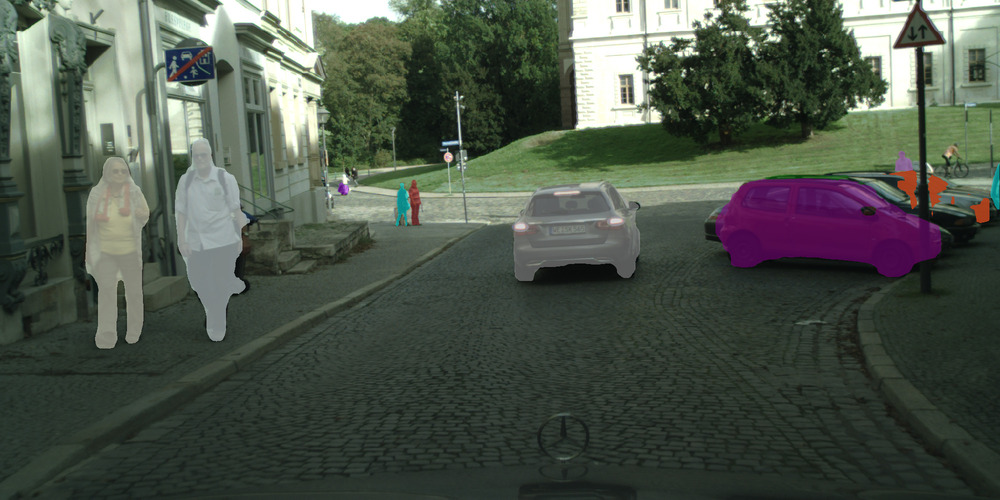} \\
                
        \raisebox{21px}{\rotatebox{90}{\small GT}}
        \includegraphics[width=.195\textwidth]{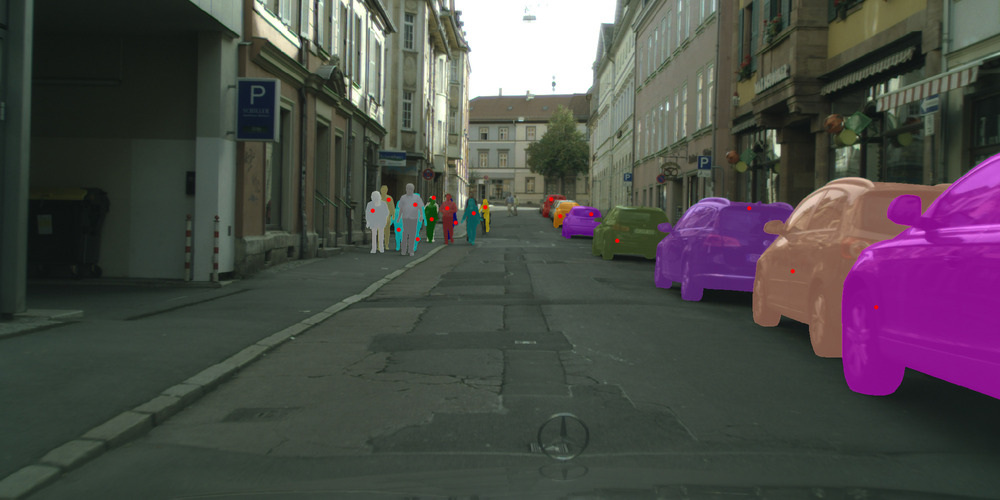} &
        \includegraphics[width=.195\textwidth]{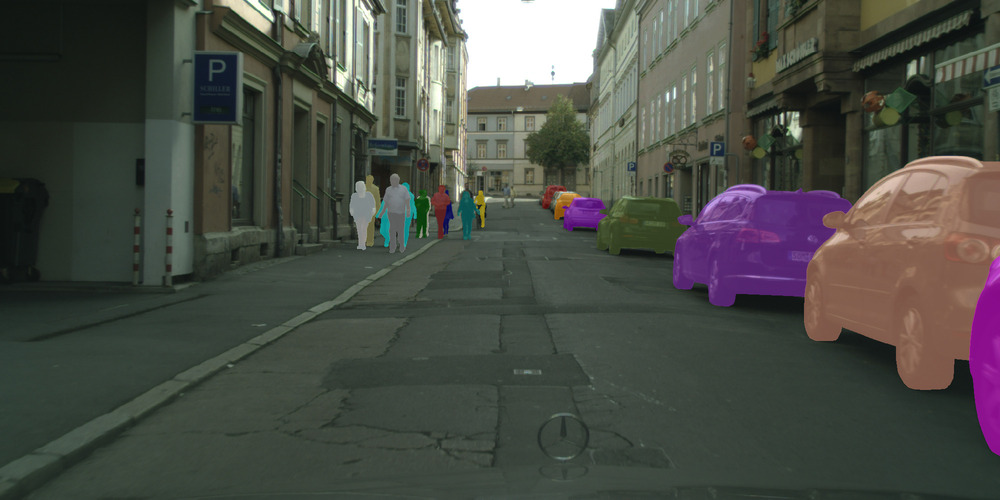} &
        \includegraphics[width=.195\textwidth]{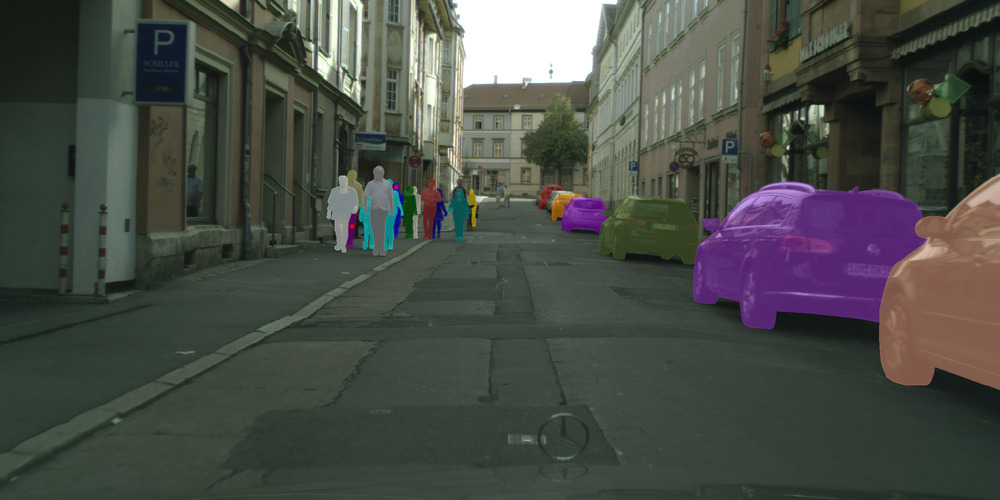} &
        \includegraphics[width=.195\textwidth]{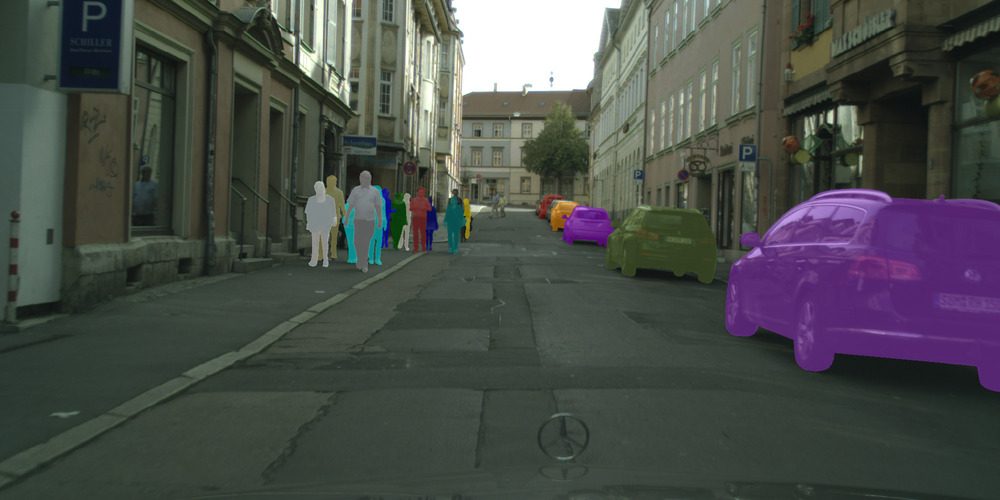} &
        \includegraphics[width=.195\textwidth]{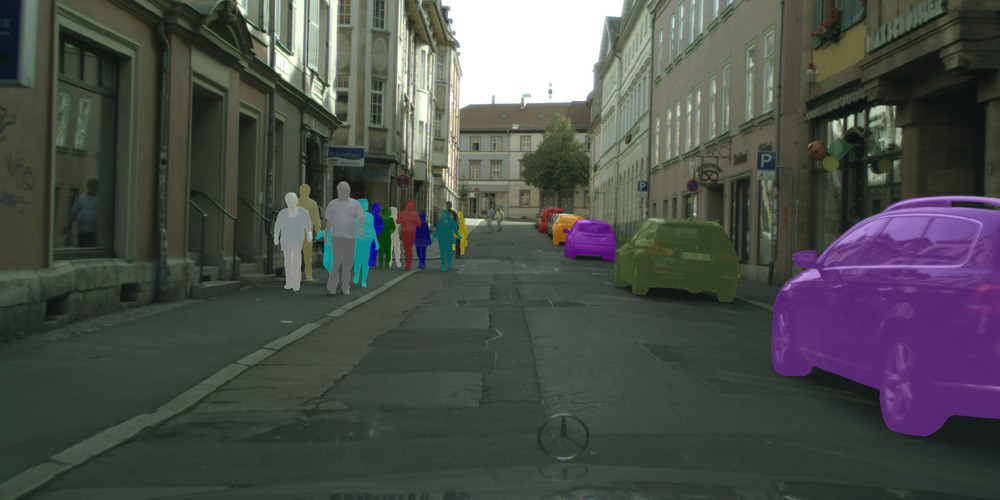} \\     
        
        \raisebox{19px}{\rotatebox{90}{\small Ours}}
        \includegraphics[width=.195\textwidth]{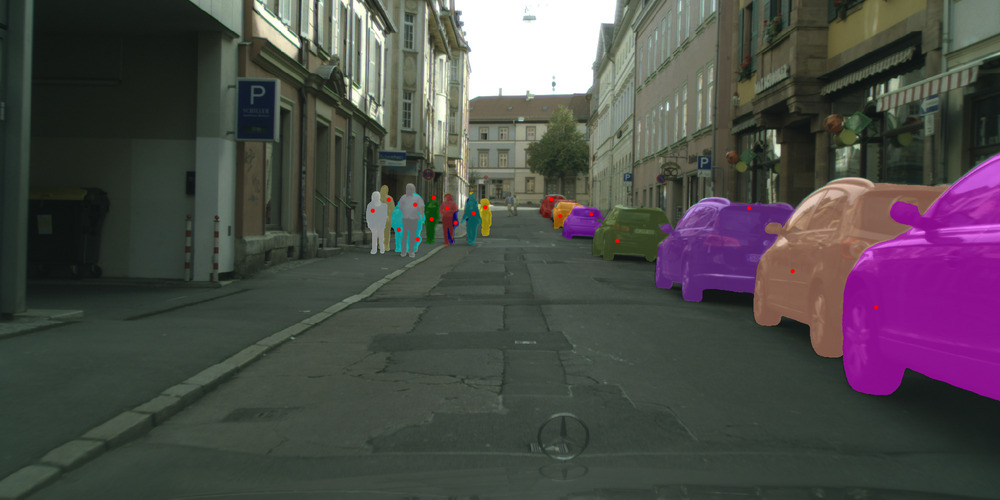} &
        \includegraphics[width=.195\textwidth]{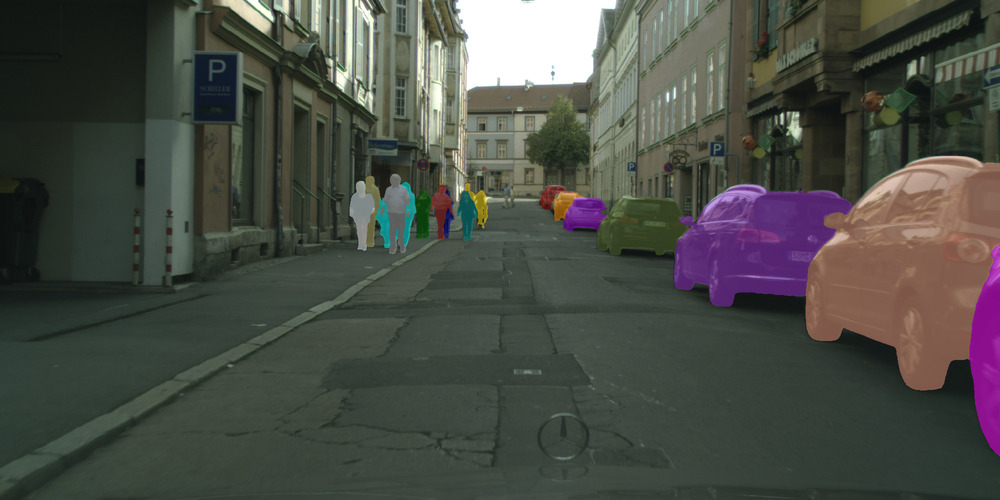} &
        \includegraphics[width=.195\textwidth]{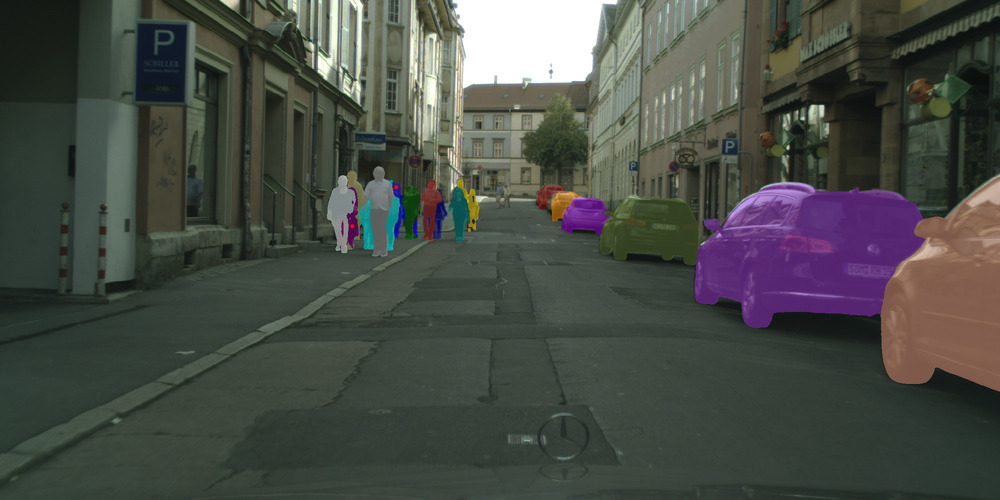} &
        \includegraphics[width=.195\textwidth]{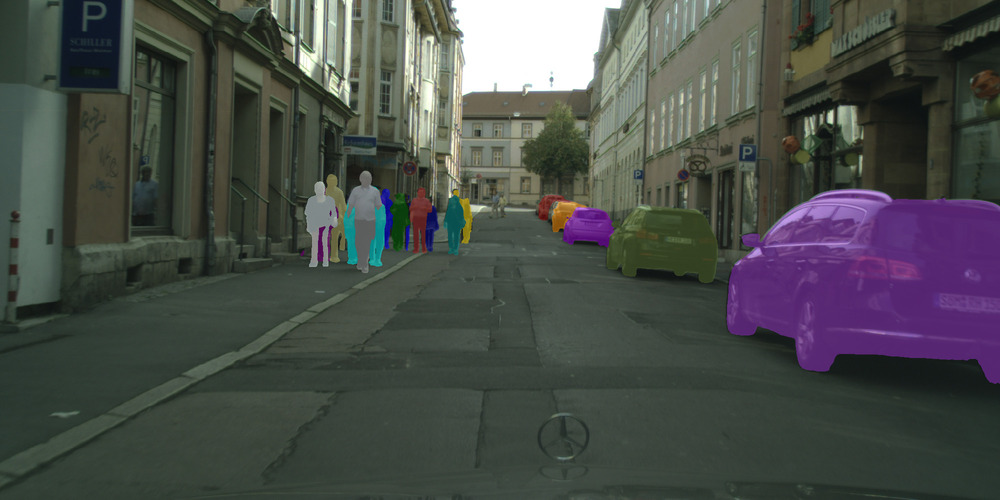} &
        \includegraphics[width=.195\textwidth]{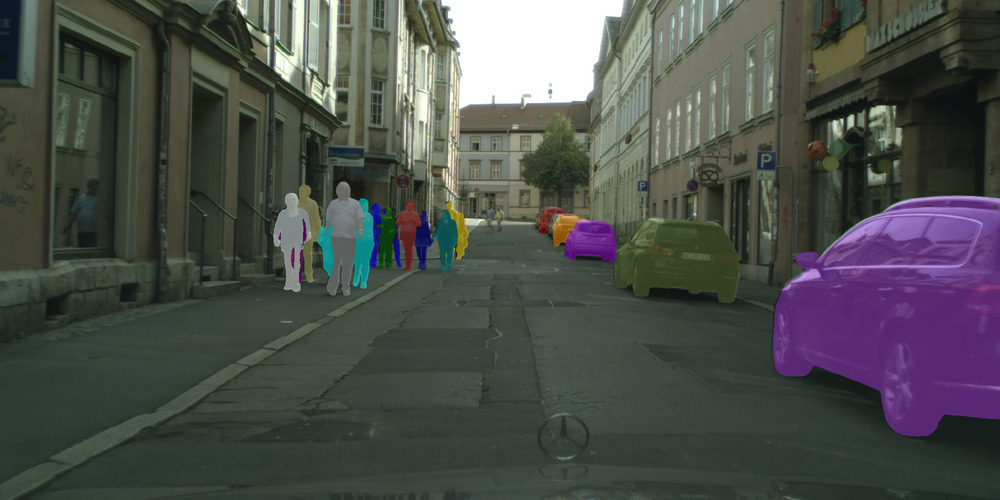} \\   
        
        \raisebox{3px}{\rotatebox{90}{\small MRCNN+M}}
        \includegraphics[width=.195\textwidth]{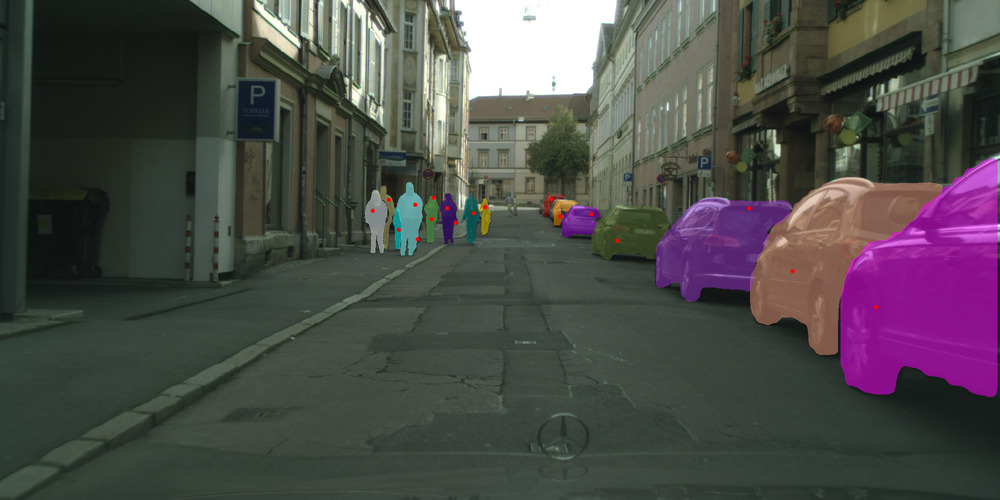} &
        \includegraphics[width=.195\textwidth]{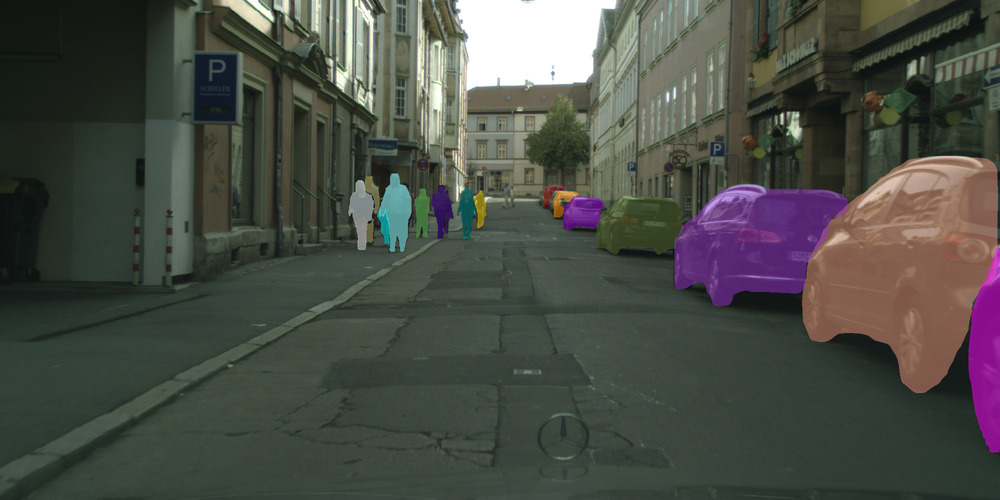} &
        \includegraphics[width=.195\textwidth]{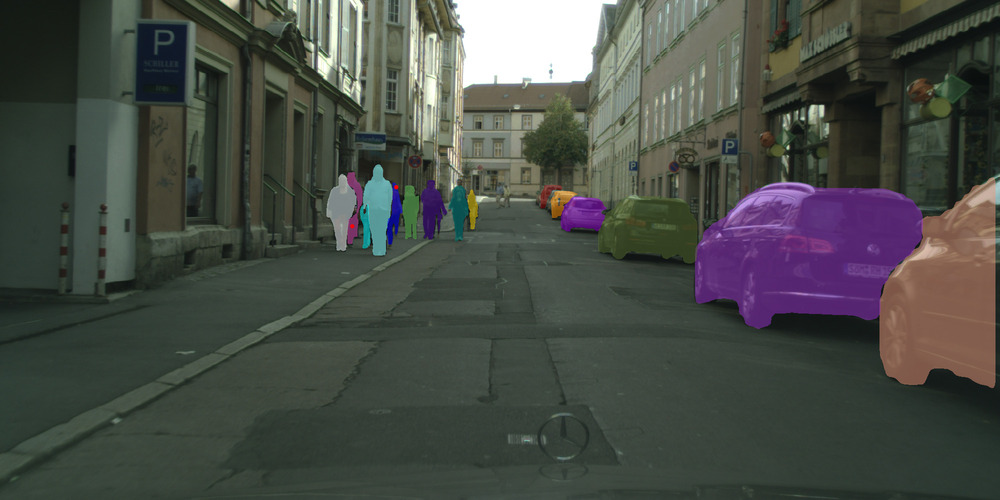} &
        \includegraphics[width=.195\textwidth]{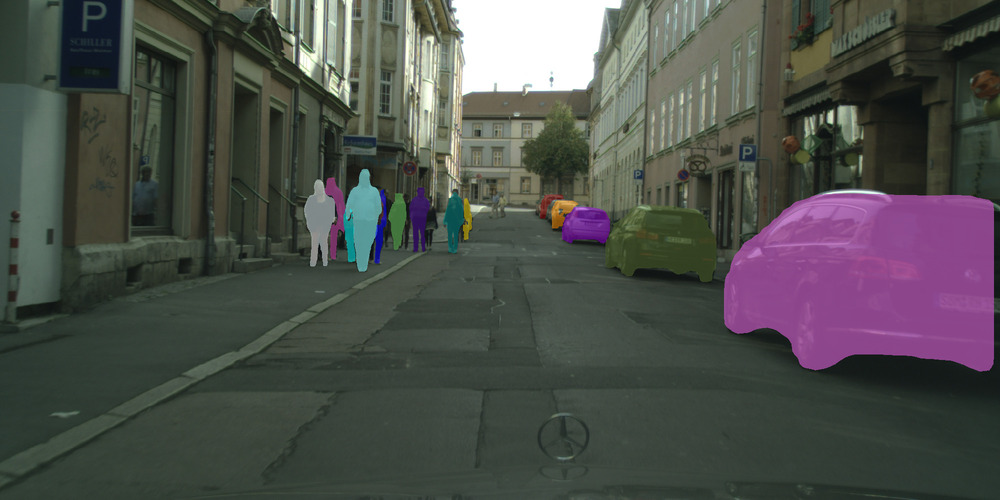} &
        \includegraphics[width=.195\textwidth]{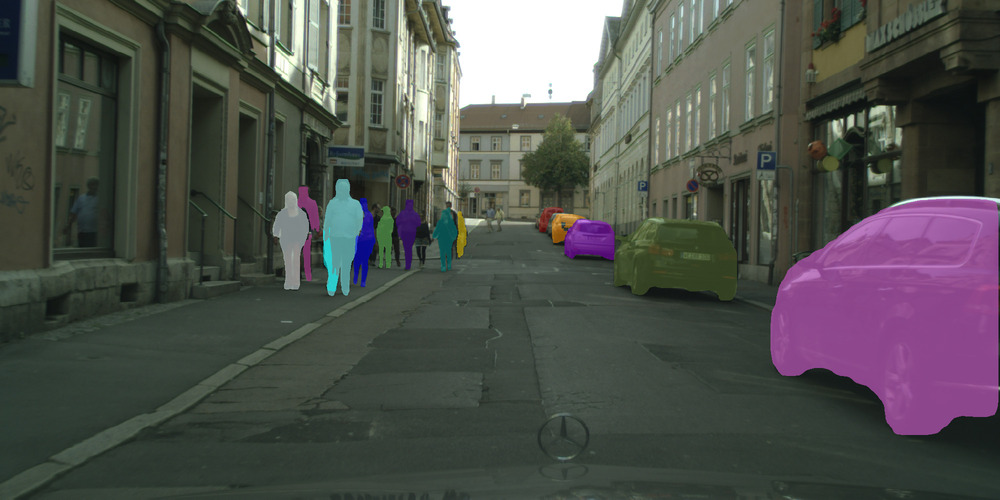} \\                
                
	\end{tabular}
	
	\caption{We showcase qualitative segmentation results of our model on the CityscapesVideo validation set and compare it with the ground truth and the Mask R-CNN + Mask Matching (MRCNN+M) baseline. Red point is the GT key point given by the annotator. }
	\label{fig:results}
\end{figure*}

\paragraph{Baselines:} We establish multiple competitive baselines for the task of single click video object segmentation. In particular, we consider the following baselines:

\begin{enumerate}
	\item Inspired by \cite{Duan_2019_ICCV, Feichtenhofer_2017_ICCV}, we develop a top down baseline called Siamese Bounding Boxes that trained end to end for the leaderboard. We use the same siamese network to input two frames at a time. Given the annotator keypoint, the model extracts a point based feature and predicts a bounding box in the next frame from which features are extracted to predict a mask for each object. This center point of this mask is then used as key point for the next frame. We have a scoring head similar to the one in \cite{he2017mask} to score whether the instance exists in the next image. That is, given the key point feature, we feed it to a mlp to predict a probability of existance in frame 2.
	\item We train Mask R-CNN with the same backbone as ours on the new dataset. Note that since our model is class agnostic, we train Mask R-CNN in a class agnostic fashion. Next, we use the annotator provided instance keypoints in frame $I_t$ to keep only the masks in bounding boxes with the highest confidence. Next, we use the optical flow framework RAFT \cite{RAFT} to warp each instance mask from $I_t$ to $I_{t+1}$. We then average the warped mask coordinates to get new key points and repeat the above process to get the new masks. We call this baseline Mask R-CNN + Key Point Matching. 
	\item As another baseline with Mask R-CNN, after we warp the masks to the next frame, we greedily match the warped masks to the masks in frame $I_{t+1}$. We call this baseline Mask R-CNN + Mask Matching.
	
\end{enumerate}

\paragraph{Comparison with the Baselines:} 
Our results are given in Table. \ref{tab:results}. We find that  Mask R-CNN + Mask Matching to be a strong baseline as the RAFT flow is a strong estimation of the object movement across video frames.  However, we observe that our model significantly outperforms the baselines; we outperform the Mask R-CNN + Mask Matching baseline by 8.3\% and 8\% in the validation and test sets.

\begin{figure*} 
	\centering
	\setlength\tabcolsep{0.5pt}
	\begin{tabular}{cccccc} 
        
        \raisebox{2px}{{t=1}} &	
		\raisebox{2px}{{t=2}} &
		\raisebox{2px}{{t=3}} &
		\raisebox{2px}{{t=4}} &
		\raisebox{2px}{{t=5}}  \\		
		
        \includegraphics[width=.195\textwidth]{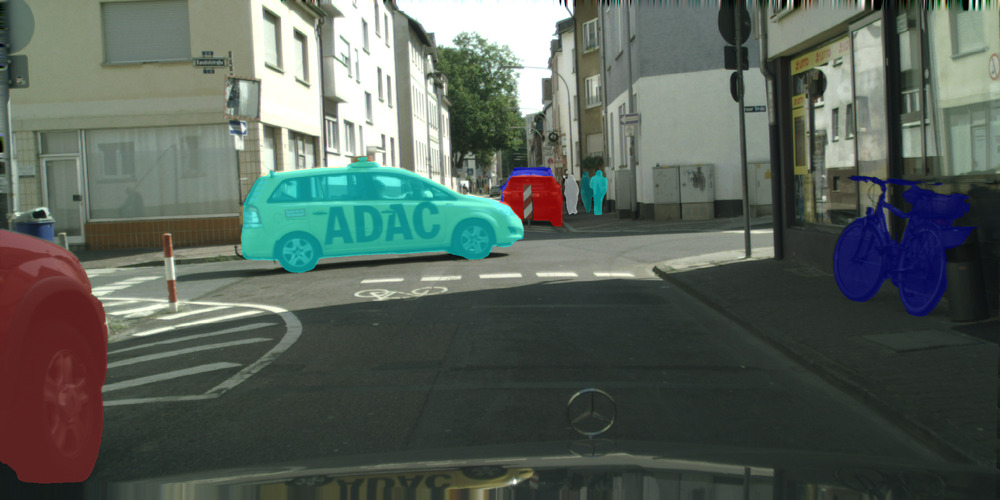} &
        \includegraphics[width=.195\textwidth]{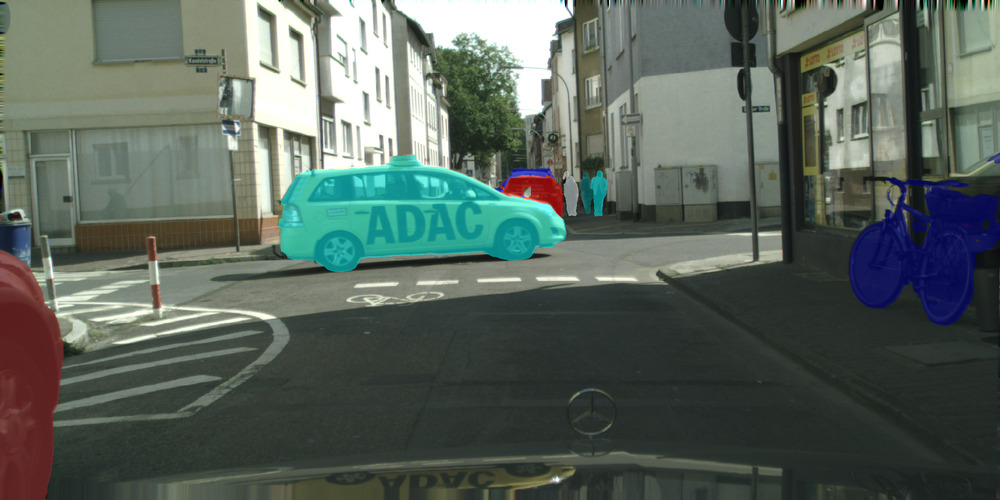} &
        \includegraphics[width=.195\textwidth]{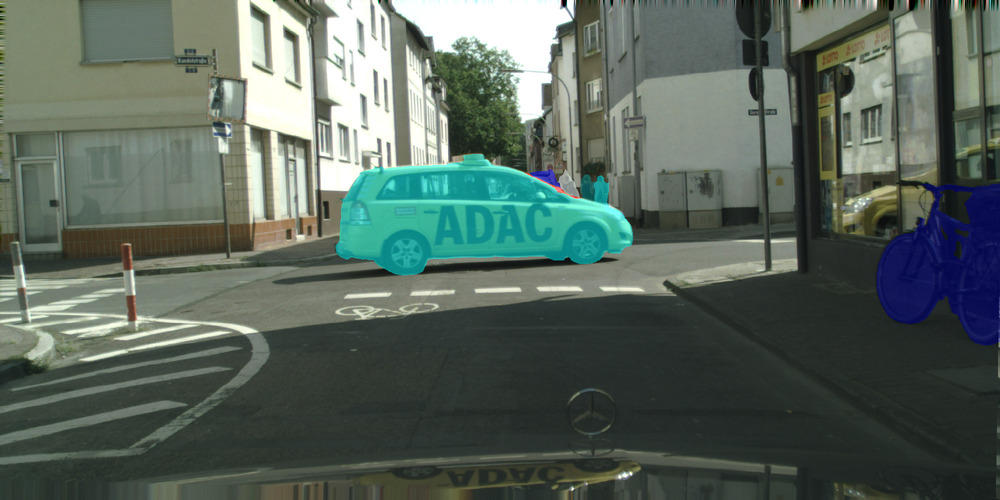} &
        \includegraphics[width=.195\textwidth]{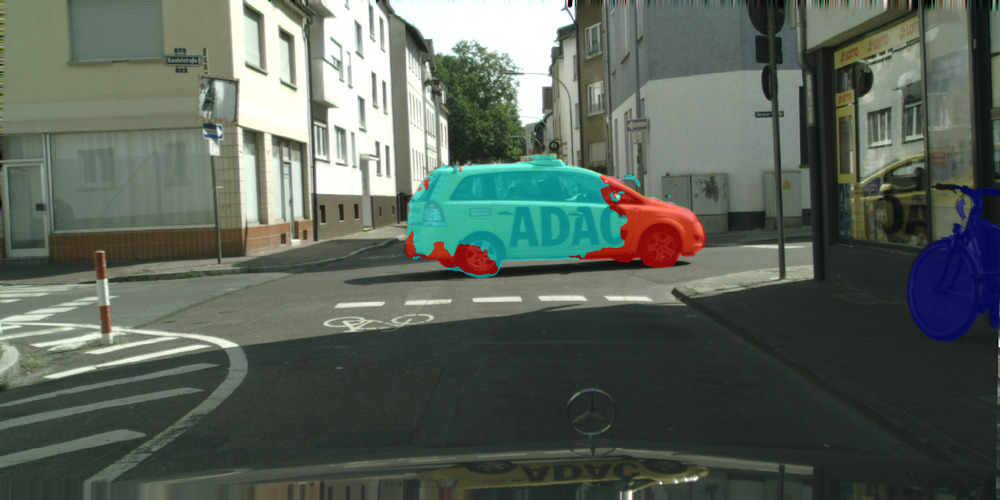} &
        \includegraphics[width=.195\textwidth]{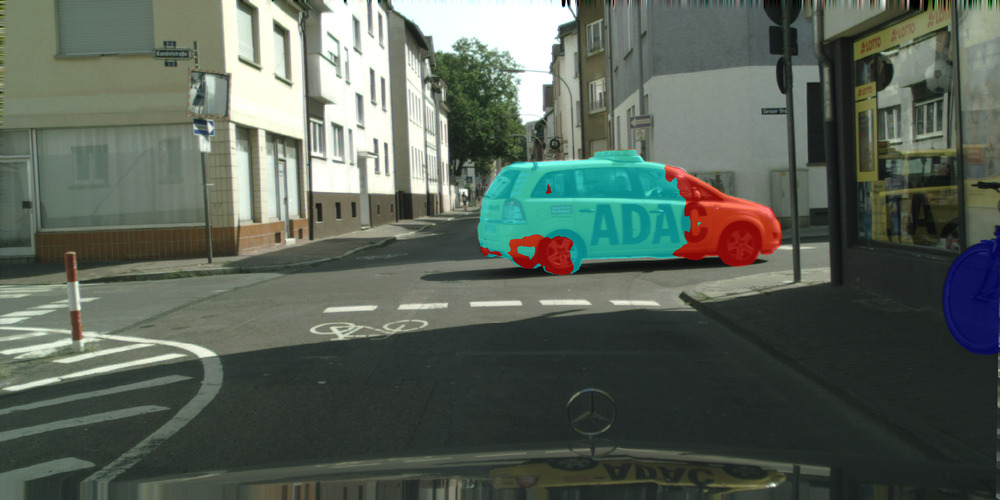} \\   

	\end{tabular}
	
	\caption{We showcase a failure mode of our model. }
	\label{fig:failure}
\end{figure*}

\paragraph{Effect of the Refinement Branch:} To understand the efficacy of the recurrent attention module for refining the initial correlation volume, we train a model without this branch. We obtain a mIOU of 61.6 on the validation set which is about 2 mIOU lower compared to the 63.5 mIOU of the model with the refinemnt branch. 

\paragraph{Extra Decoder for the Refinement Branch:} To obtain the context features that are passed to the refinement module, we have an extra decoder that is applied only on the second frame $I_{t+1}$ without any spatio-temporal context from $I_t$. Intuitively, these features would have better localized information about the object  masks and boundaries in $I_{t+1}$. To test this, we train a model without the extra decoder where the recurrent refinement module takes features form the spatio-temporal features $F_{t+1}$ instead. We obtain a mIOU of 61.1 compared to the 63.5 of the full model on the validation set confirming the importance of this branch.

\paragraph{Annotator Point Selection} In the single click video object segmentation setting, we require that the annotator click on any point inside the object. In this section, we evaluate our model when the annotator point is required to be the center of the object, more specifically, when the point is the one with the highest distance transform value to the boundaries. We observe the performance to be very similar at 63.4 mioU compared to 63.5 our model. In other words, our model is robust to where the annotator clicks.

\paragraph{Frame Rate Generalization}
We trained and evaluated our model on the 5 frames of each Cityscapes sequence that have ground truth. These 5 frames are at 360 ms apart. Here we run the model on the full sequence of 30 frames which is at 60 ms apart to see how our model performs on longer sequences and at a different frame rate. Although we do not have ground truth for all the 30 frames, we can still apply the model to the full snippet and evaluate on the 5 frames. In Table \ref{tab:fullresults} we see in fact the drop in performance to be minimal for both the validation and the test sets. This indicates that our model can generalize to a higher frame rate and also keep track of masklet for longer periods. 

\paragraph{Our Model in Single Image Setting:}
Our model can also be used for single click image object segmentation. In particular, we train and evaluate our model on single images only. Here we have removed the complications of handling motion of the objects between two frames. We compare with Mask R-CNN with the same backbone where the automatically generated masks are filtered out based on the annotator provided keypoints. In Table \ref{table:singleimage} we observe that our model outperforms the baseline by about 10 mIOU points.

\paragraph{Qualitative Examples:} In Figure \ref{fig:results} we showcase qualitative results of our model run on the validation set and compare it with the ground truth and the Mask R-CNN + Mask Flow + Matching baseline. Results are shown on the 5 frames which are 360ms apart. Compared to the baseline we see that our model results in higher resolution segmentation and better tracking.

\paragraph{Failure Mode:} Our model sometimes showcases difficulty when an object occludes another which can cause the occluded object's key point to switch onto the object causing the occlusion.  We showcase an example of such a failure in Figure \ref{fig:failure}: In time step 2, the cyan car is about to pass over the green car in the background. In time step 3, it passes over but there is still a small area (top right) of the green car that is not fully occluded and we predict a green segmentation for that area. This prediction results in a new key point right on the border of both cars.  At the next timestep, the model thinks that this keypoint on the border belongs to the the cyan car as well causing its segmentation mask to be divided into two regions.

\section{Conclusion}
\label{sec:conclusion}

Annotating object instance masks in videos is an expensive process as it requires not only the object masks to be labeled but also instances must be tracked across time. To tackle this problem, we propose VideoClick, a bottom up multi object video segmentation method that takes in a single click for each object from an annotator and outputs the segmentation masks for an entire video. This is done by sampling point based features using the annotator key points and constructing a correlation volume that assigns each pixel in a target frame to one of these keypoints. This correlation volume is refined using a recurrent attention mechanism to output a more precise final segmentation. We demonstrate the effectiveness of our method on a new benchmark, CityscapesVideo, that we introduce in this paper. We show competitive initial results with strong segmentation and tracking on this dataset. 


{\small
	\bibliographystyle{ieee_fullname}
	\bibliography{top}
}

\onecolumn
\title{Appendix - VideoClick: Video Object Segmentation with a Single Click}
\author{\vspace{-5ex}}
\date{\vspace{-5ex}}
\maketitle

	In Section \ref{sec:arch} of this appendix, we present the architectural details of the siamese network. In Section \ref{sec:figs}, we showcase more qualitative examples. 

\begin{appendices}


\section{Siamese Network Architecture}
\label{sec:arch}
For the encoder of our siamese newtork, we use \texttt{ResNet-101} \cite{He2015DeepRL}, where we replace all the batch normalization layers with group normalization \cite{wu2018group} with 32 groups each. For the decoder of the siamese network, we use a modified temporal version of FPN \cite{Lin2016FeaturePN}. In particular, whereas the inner block of each FPN layer is just a simple $3\times 3$ convolution, we use the architecture presented in Table. \ref{tab:decoder_head} instead before upsampling. Note that for temporal aggregation, we simply concatenate the features of two frames along the channel dimension and apply a $1 \times 1$ convolution. Also we use bilinear interpolation for upsampling instead of nearest neighbour. For the decoder of the refinement module, we only use the blocks until row 4 of Table. \ref{tab:decoder_head}.

\begin{table}[h]
	\centering
	\scalebox{0.9}{
	\begin{tabular}{c|c|c|c|c|c|c}
		Row & Type                  & \# of Input Tensors & \# Input Channels       & \# of Output Tensors & \# Output Channels & Kernel Size \\ \hline \hline
		1   & Conv2D + ReLU         & 2                   & Depends on resnet layer & 2                    & 256                & 3           \\ \hline
		2   & Conv2D + ReLU         & 2                   & 256                     & 2                    & 256                & 3           \\ \hline
		3   & Conv2D + GN + ReLU    & 2                   & 256                     & 2                    & 256                & 1           \\ \hline
		4   & Concat along channels & 2                   & 256                     & 1                    & 512                & -           \\ \hline
		5   & Conv2D + ReLU         & 1                   & 512                     & 1                    & 1024               & 1           \\ \hline
		6   & Conv2D + ReLU         & 1                   & 1024                    & 1                    & 512                & 1           \\ \hline
		7   & Split along channles  & 1                   & 512                     & 2                    & 256                & -           \\ \hline
		8   & Conv2D + GN + ReLU    & 2                   & 256                     & 2                    & 256                & 1           \\ \hline
		9   & Add to row 2          & 2                   & 256                     & 2                    & 256                & -          
	\end{tabular}}
	\caption{Architecture of each decoder block. GN corresponds to group normalization with 32 groups. }
	\label{tab:decoder_head}

\end{table}


\section{Qualitative Examples}
\label{sec:figs}
In the following figures, we showcase the inference results of our VideoClick model and the corresponding ground truth for various sequences in the validation set of the new CityscapesVideo dataset. Each sequence of Cityscapes is 1.8 s with 30 frames. We annotated frames 1, 7, 13, 19 and 25 with instance masklets.  Here we are showing the results of our model being applied to this 5 frames only. In the accompanying video, we showcase the results of our model applied to the whole sequence. 

For each sequence, the annotator point clicks are given by red points for new objects appearing in the frame. We observe that our model is able to infer accurately the masklets in the sequence from just a single click of the objects provided by the annotator. 

\begin{figure*} 
	\centering
	\setlength\tabcolsep{0.5pt}
	\begin{tabular}{cc}
		

		
			\raisebox{2px}{{Ours}} &	\raisebox{2px}{{Ground Truth}}  \\

		\raisebox{40px}{\rotatebox{90}{t =  0 s}}
		\includegraphics[width=.45\textwidth]{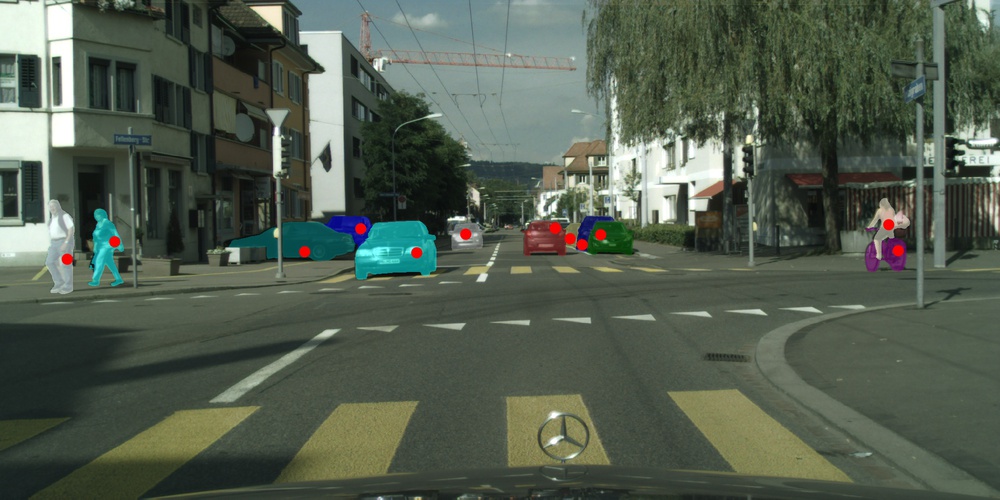} &
		\includegraphics[width=.45\textwidth]{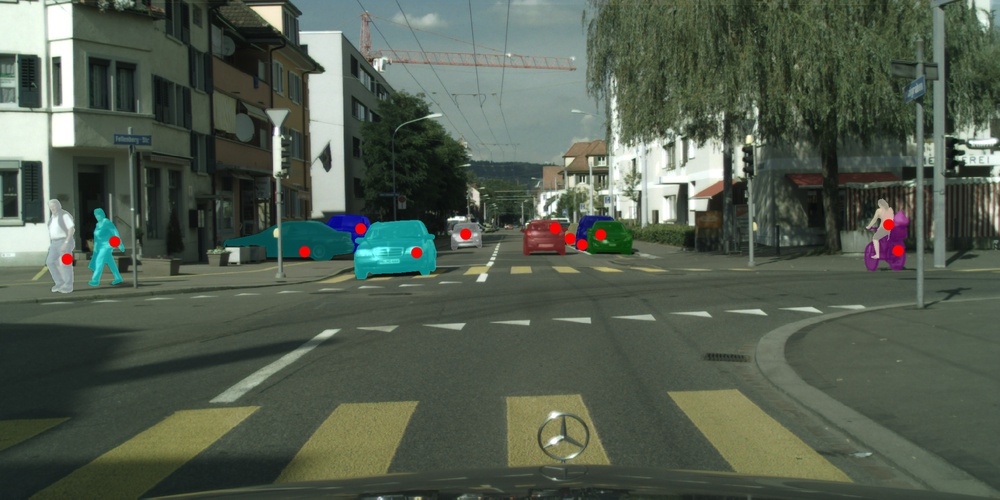} \\

		\raisebox{40px}{\rotatebox{90}{t =  0.43 s}}
		\includegraphics[width=.45\textwidth]{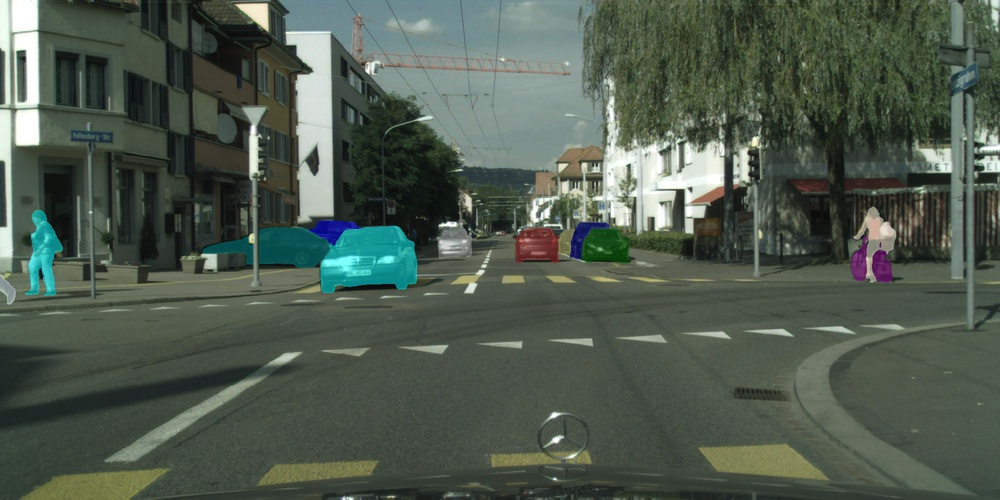} &
		\includegraphics[width=.45\textwidth]{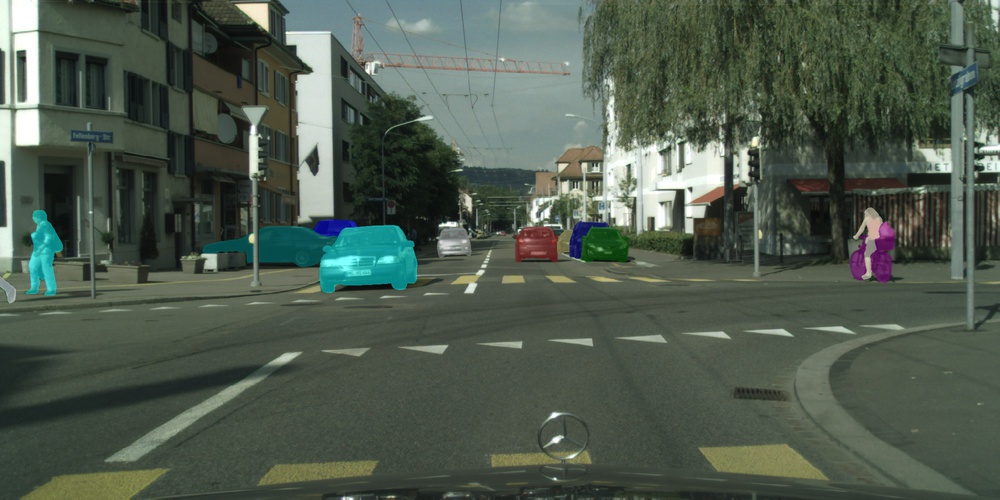} \\
		
		\raisebox{40px}{\rotatebox{90}{t =  0.81 s}}
		\includegraphics[width=.45\textwidth]{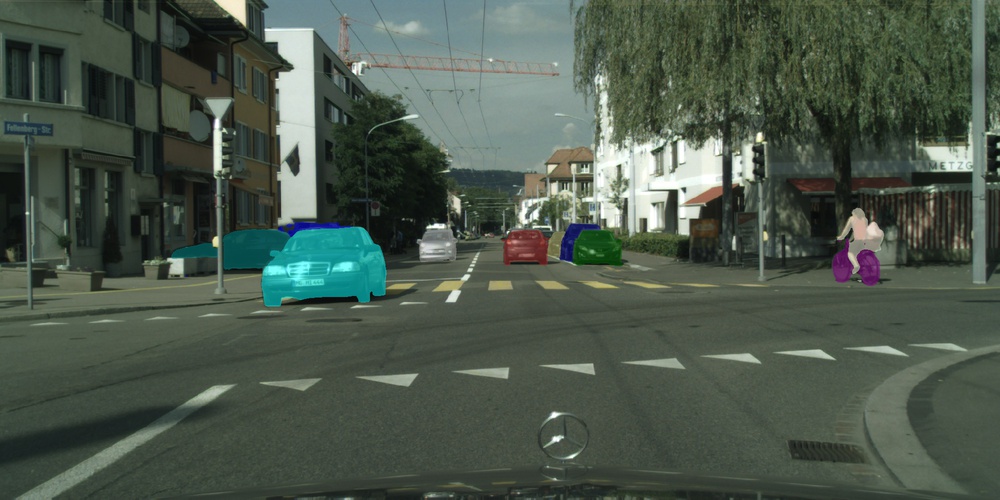} &
		\includegraphics[width=.45\textwidth]{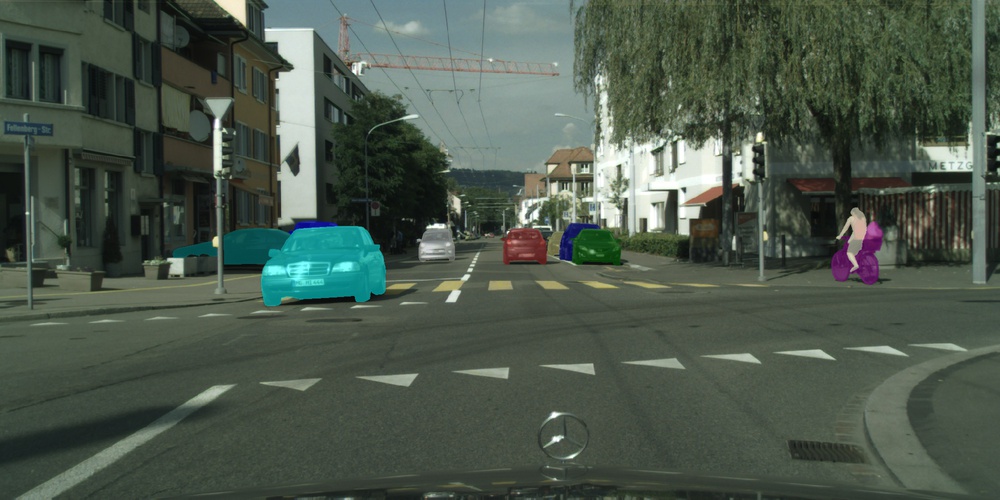} \\
		
		\raisebox{40px}{\rotatebox{90}{t =  1.18 s}}
		\includegraphics[width=.45\textwidth]{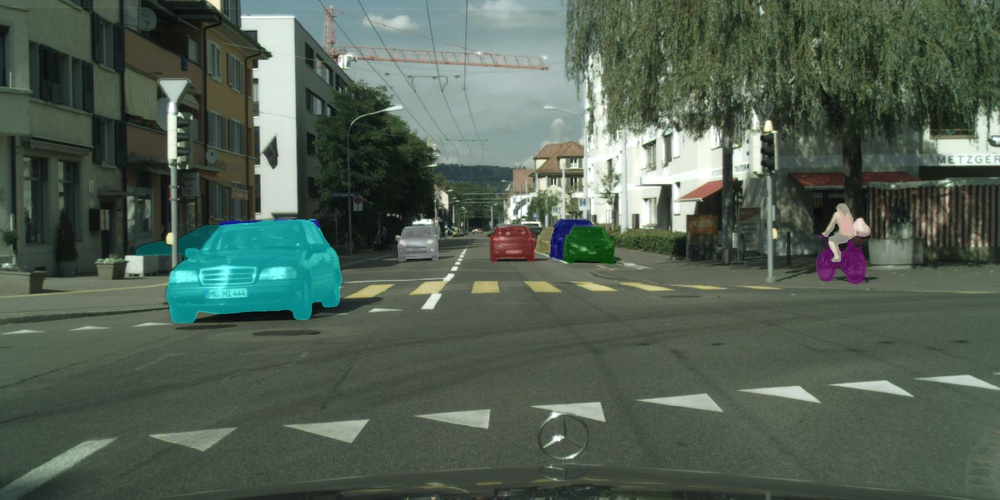} &
		\includegraphics[width=.45\textwidth]{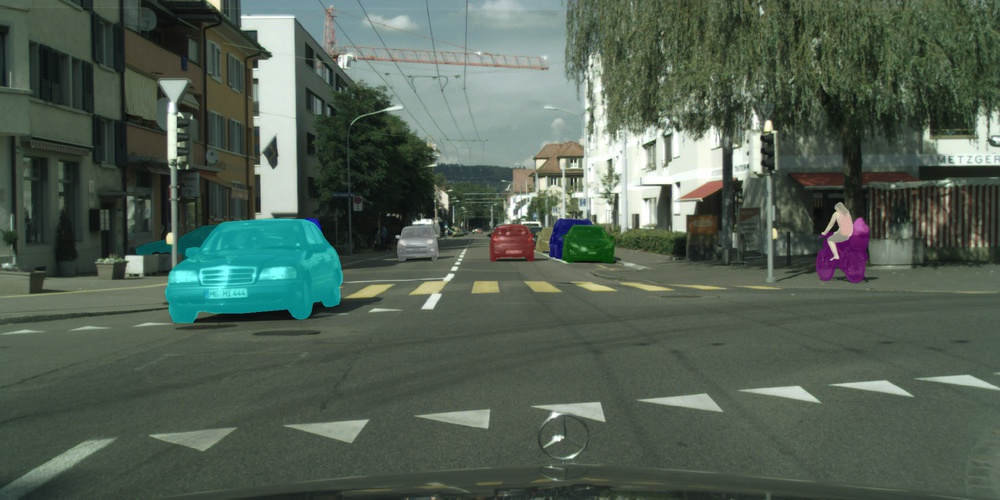} \\
		
		\raisebox{40px}{\rotatebox{90}{t =  1.55 s}}
		\includegraphics[width=.45\textwidth]{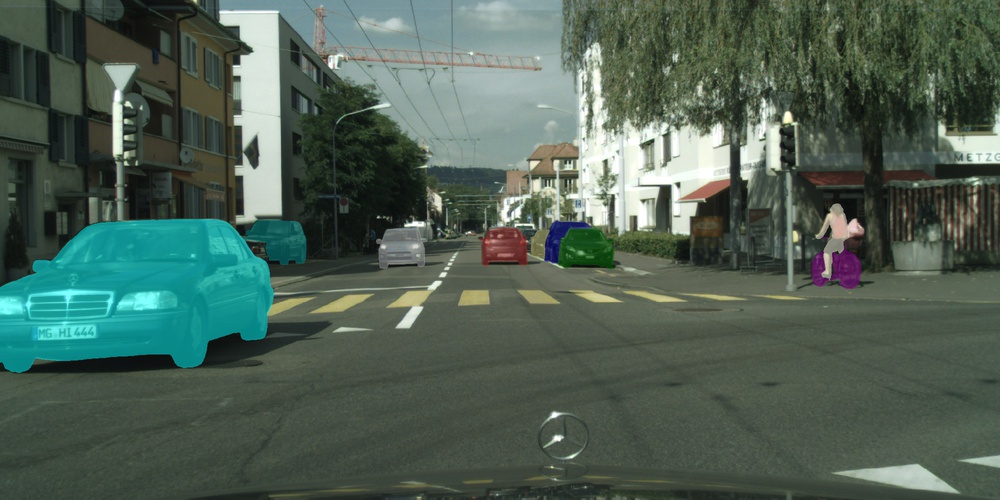} &
		\includegraphics[width=.45\textwidth]{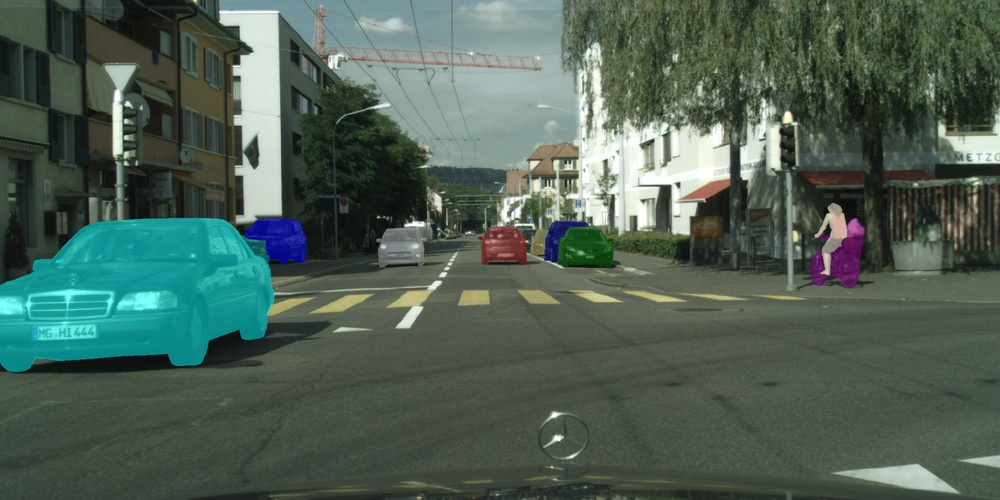} \\

	\end{tabular}
	
	\caption{We showcase qualitative segmentation results of our model on the CityscapesVideo validation set and compare it with the ground truth. Red points are the ground truth key point given by the annotator for the new objects. }
	\label{fig:results1}
\end{figure*}

\begin{figure*} 
	\centering
	\setlength\tabcolsep{0.5pt}
	\begin{tabular}{cc}
		

		
		\raisebox{2px}{{Ours}} &	\raisebox{2px}{{Ground Truth}}  \\
		
		\raisebox{40px}{\rotatebox{90}{t =  0 s}}
		\includegraphics[width=.45\textwidth]{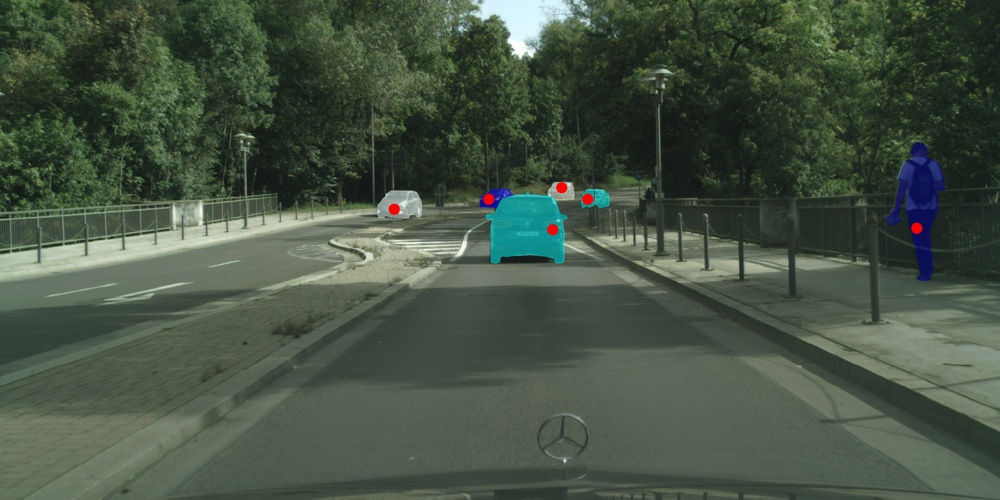} &
		\includegraphics[width=.45\textwidth]{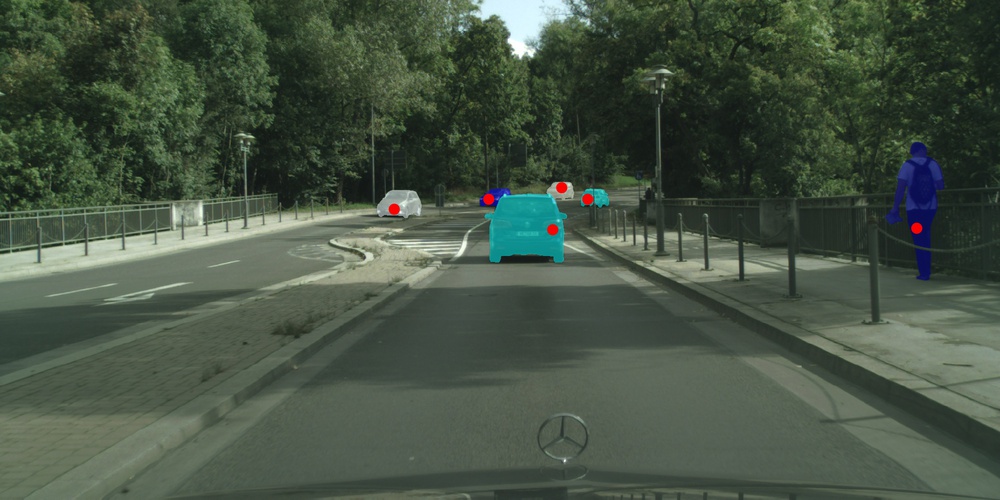} \\
		
		\raisebox{40px}{\rotatebox{90}{t =  0.43 s}}
		\includegraphics[width=.45\textwidth]{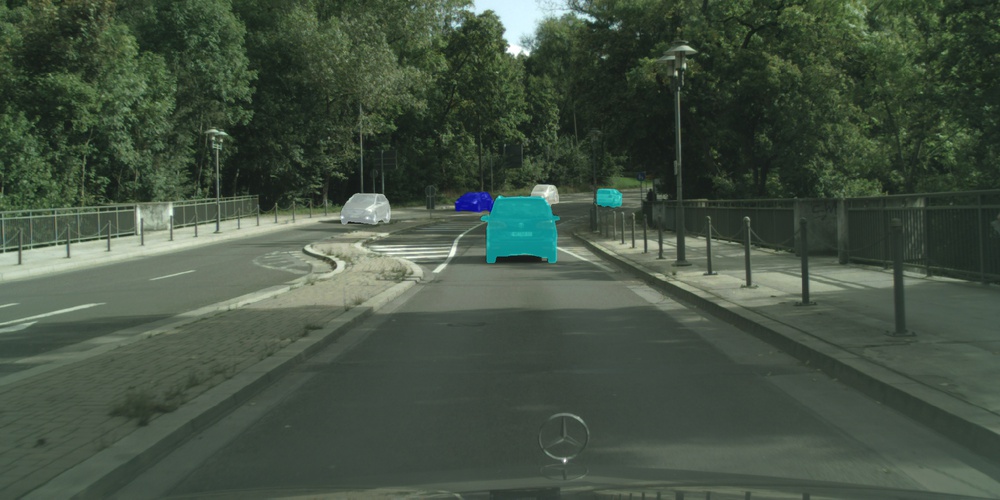} &
		\includegraphics[width=.45\textwidth]{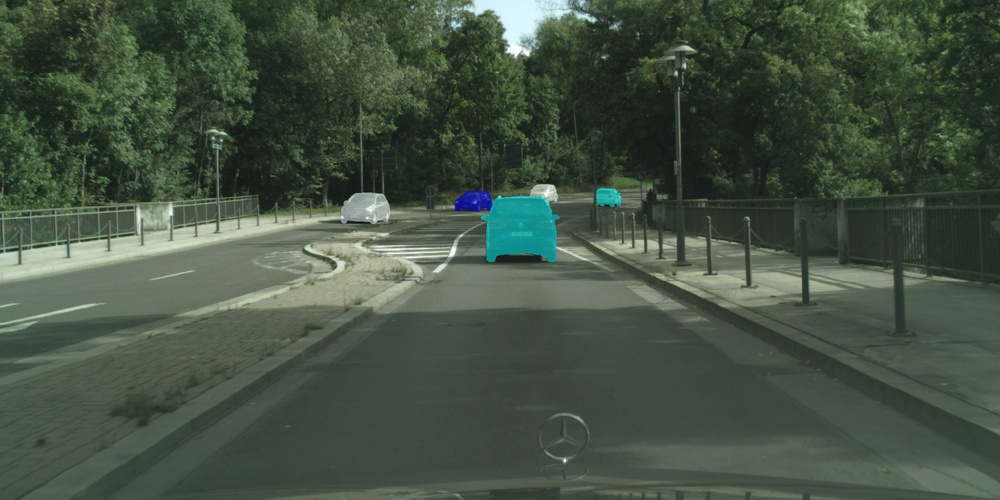} \\
		
		\raisebox{40px}{\rotatebox{90}{t =  0.81 s}}
		\includegraphics[width=.45\textwidth]{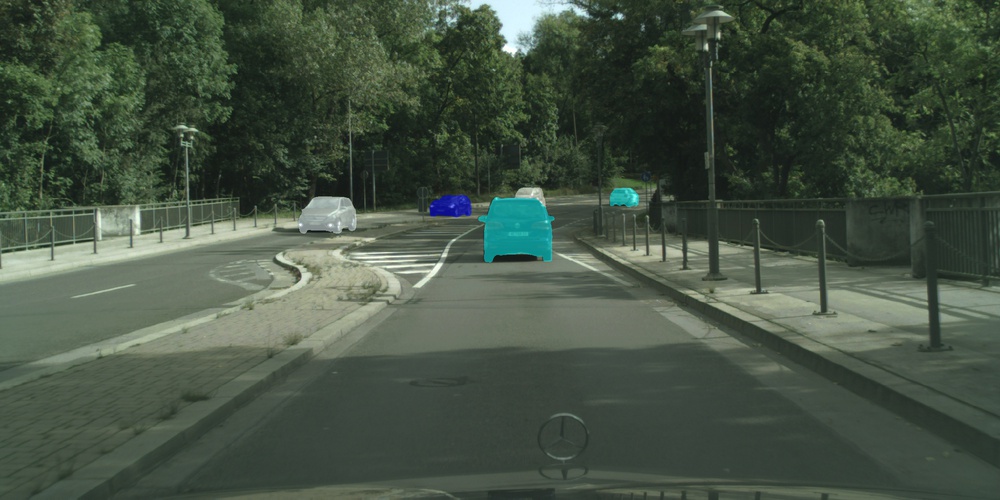} &
		\includegraphics[width=.45\textwidth]{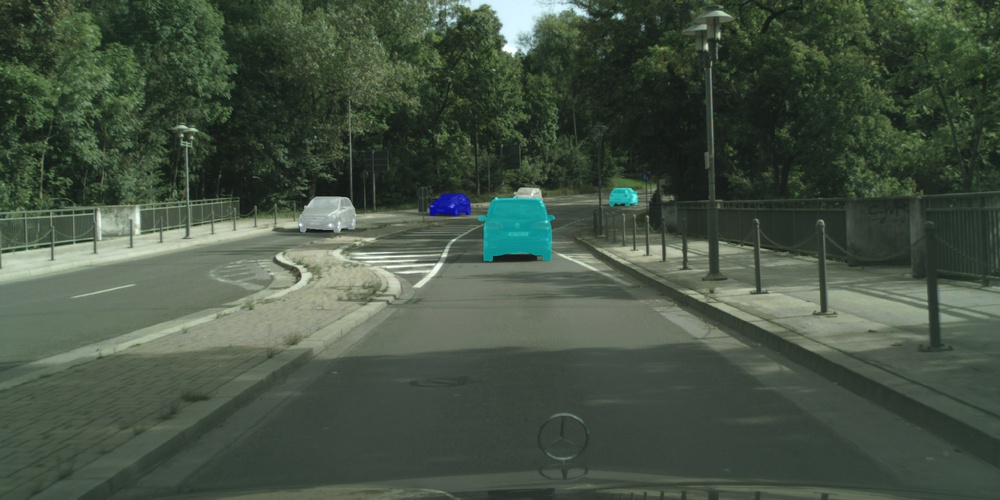} \\
		
		\raisebox{40px}{\rotatebox{90}{t =  1.18 s}}
		\includegraphics[width=.45\textwidth]{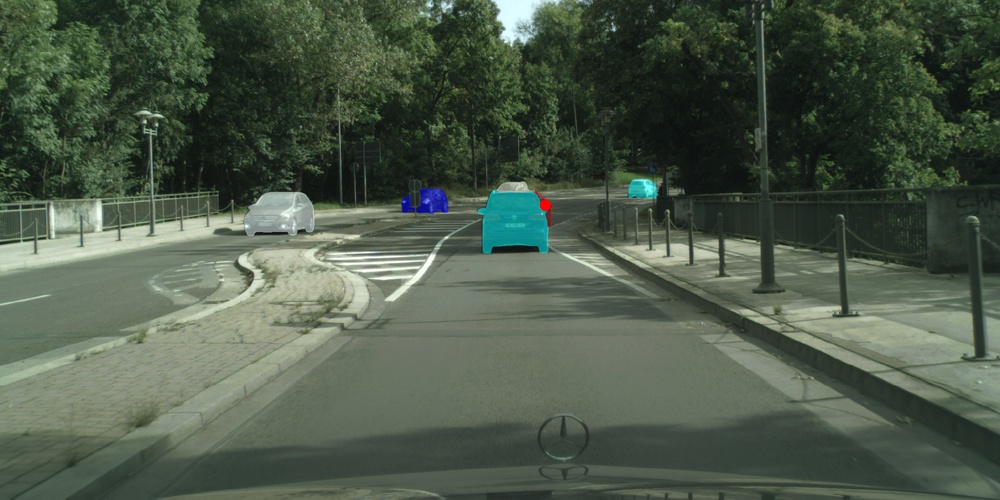} &
		\includegraphics[width=.45\textwidth]{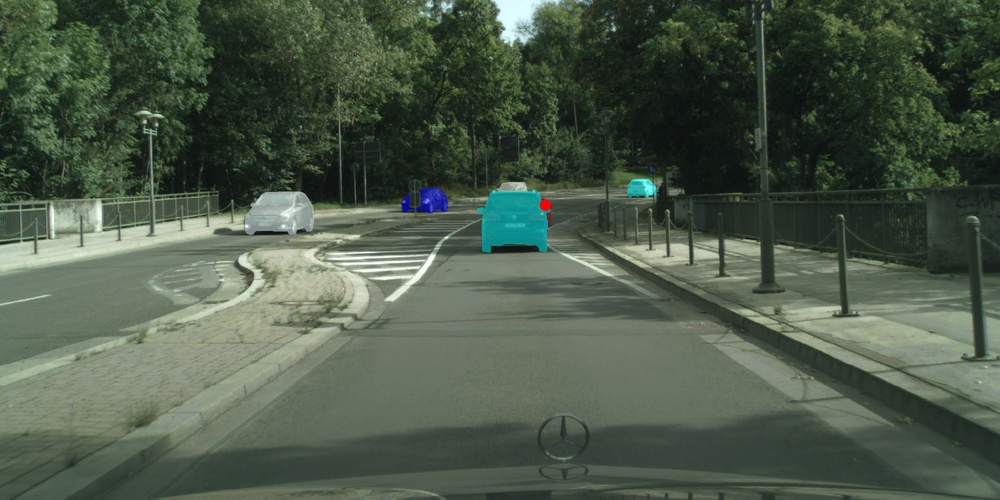} \\
		
		\raisebox{40px}{\rotatebox{90}{t =  1.55 s}}
		\includegraphics[width=.45\textwidth]{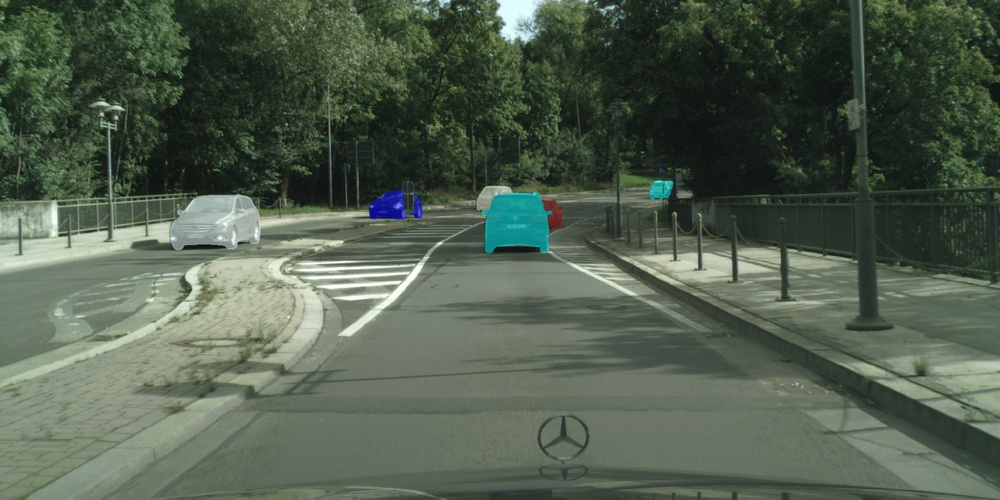} &
		\includegraphics[width=.45\textwidth]{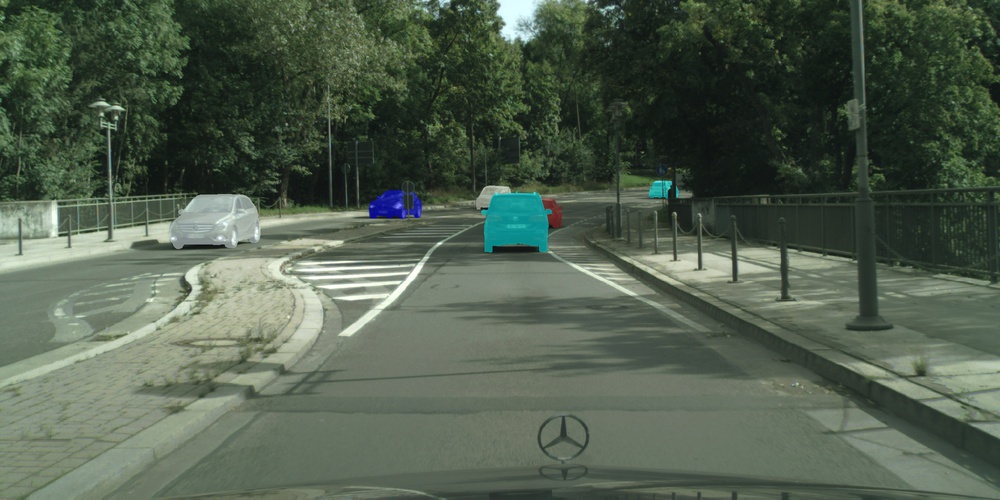} \\
		
	\end{tabular}
	
	\caption{We showcase qualitative segmentation results of our model on the CityscapesVideo validation set and compare it with the ground truth. Red points are the ground truth key point given by the annotator for the new objects. }
	\label{fig:results2}
\end{figure*}


\begin{figure*} 
	\centering
	\setlength\tabcolsep{0.5pt}
	\begin{tabular}{cc}
		

		
		\raisebox{2px}{{Ours}} &	\raisebox{2px}{{Ground Truth}}  \\
		
		\raisebox{40px}{\rotatebox{90}{t =  0 s}}
		\includegraphics[width=.45\textwidth]{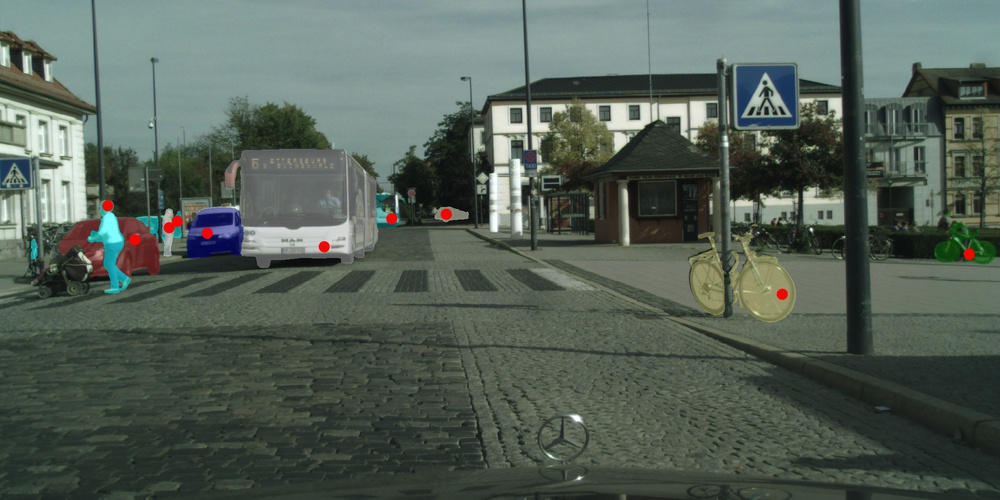} &
		\includegraphics[width=.45\textwidth]{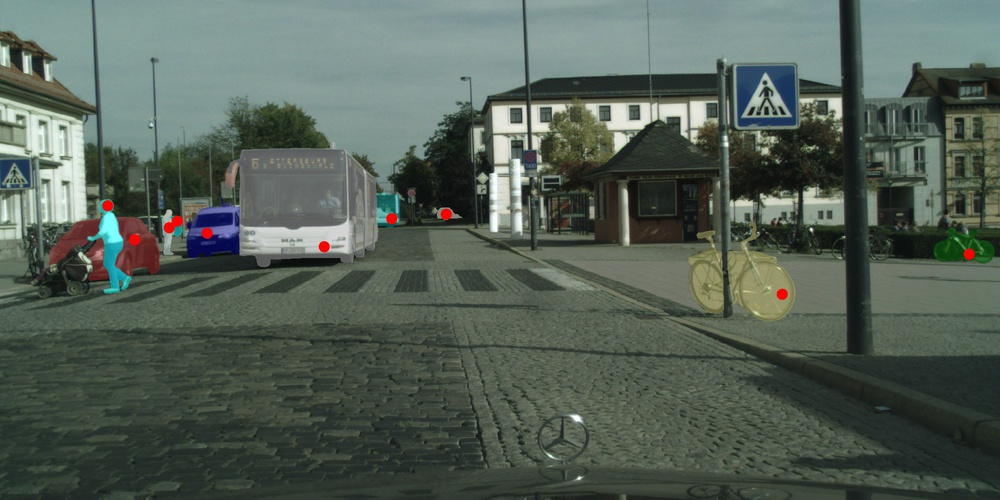} \\
		
		\raisebox{40px}{\rotatebox{90}{t =  0.43 s}}
		\includegraphics[width=.45\textwidth]{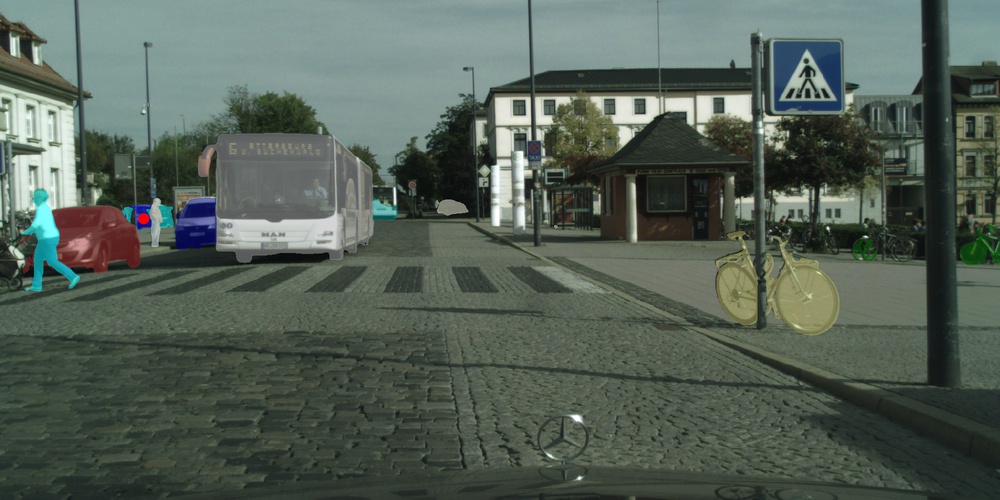} &
		\includegraphics[width=.45\textwidth]{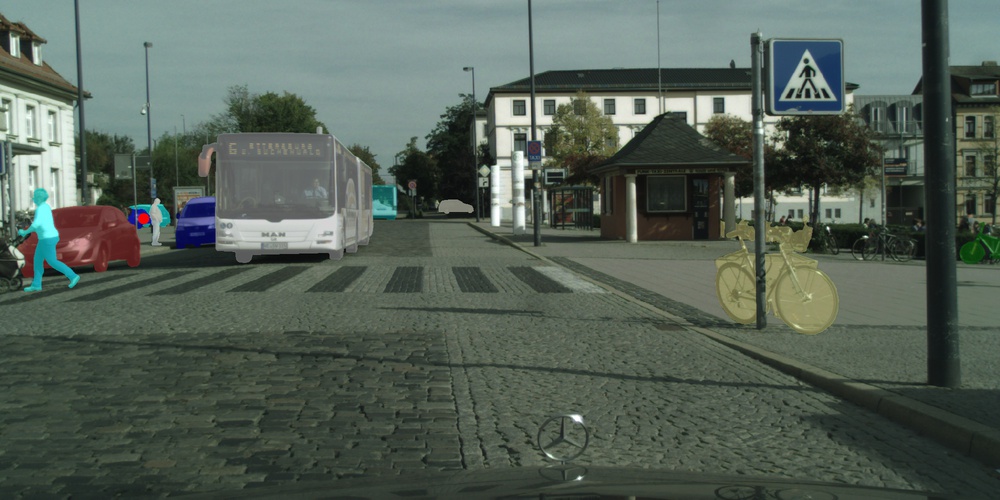} \\
		
		\raisebox{40px}{\rotatebox{90}{t =  0.81 s}}
		\includegraphics[width=.45\textwidth]{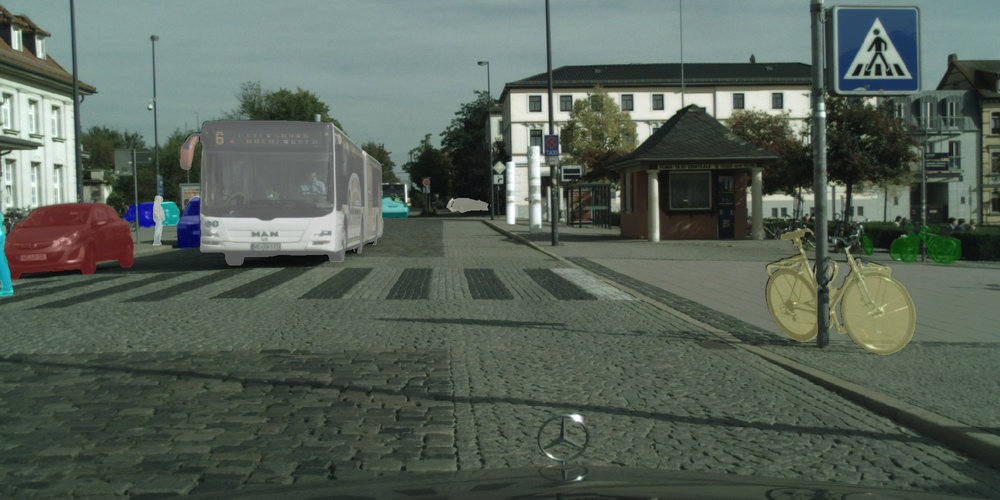} &
		\includegraphics[width=.45\textwidth]{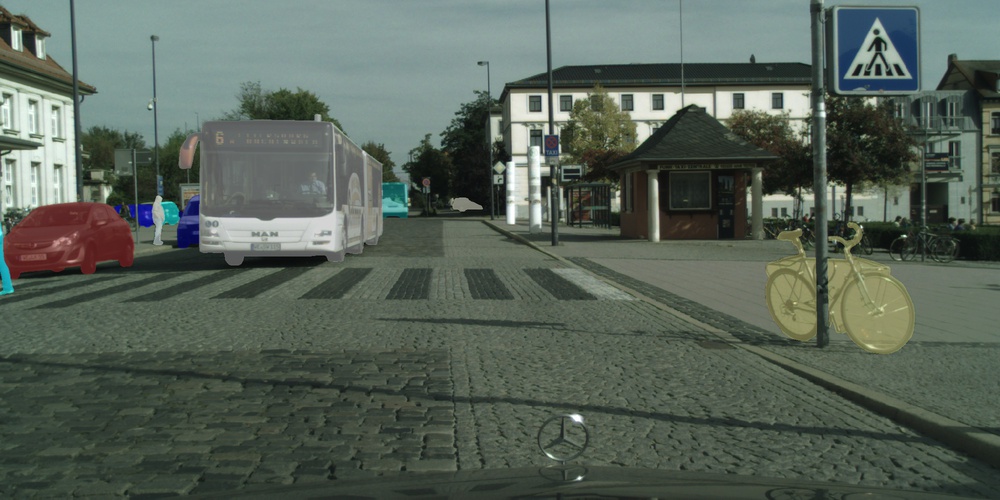} \\
		
		\raisebox{40px}{\rotatebox{90}{t =  1.18 s}}
		\includegraphics[width=.45\textwidth]{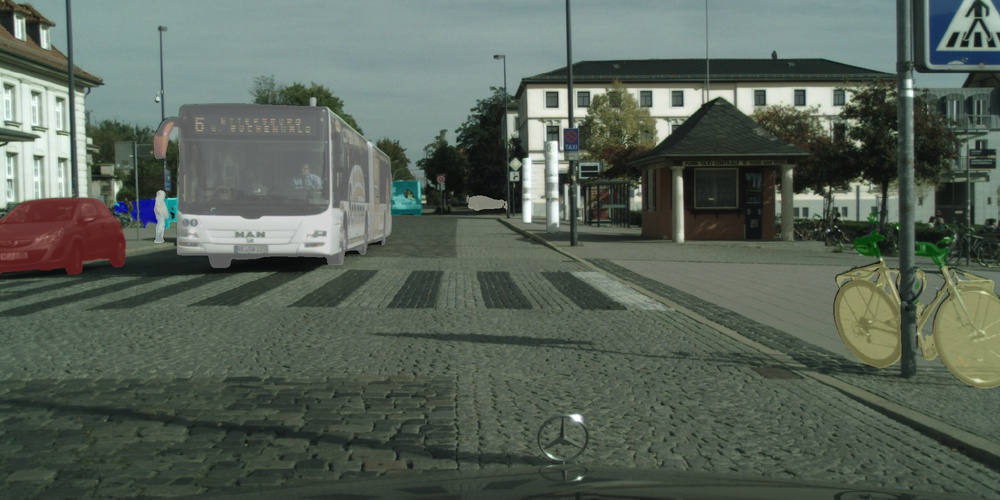} &
		\includegraphics[width=.45\textwidth]{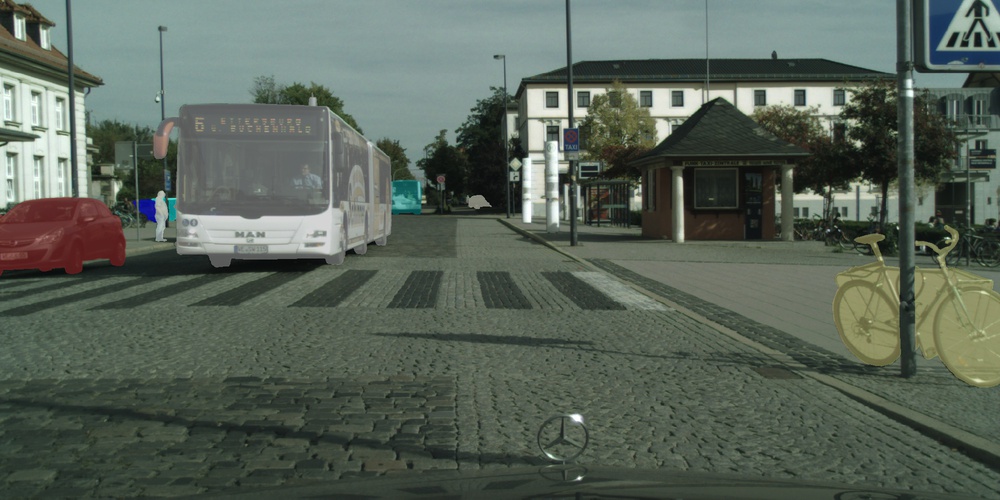} \\
		
		\raisebox{40px}{\rotatebox{90}{t =  1.55 s}}
		\includegraphics[width=.45\textwidth]{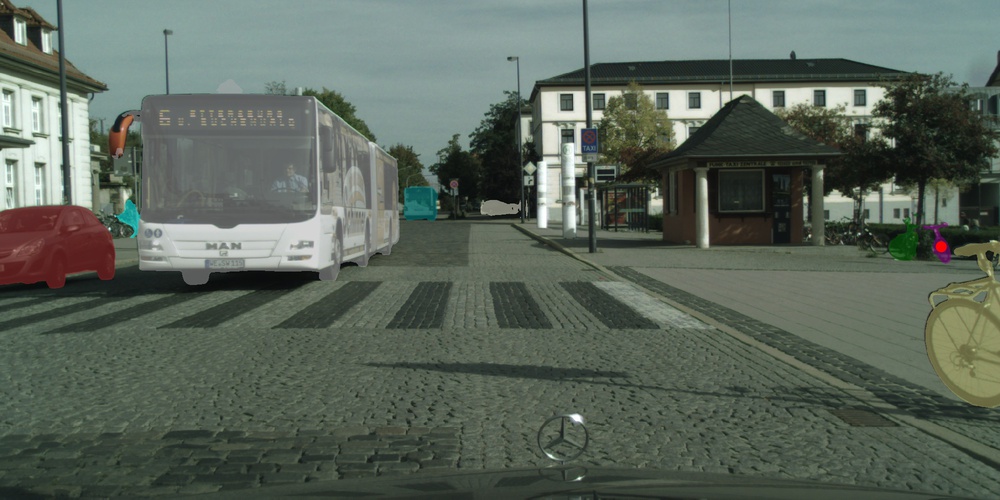} &
		\includegraphics[width=.45\textwidth]{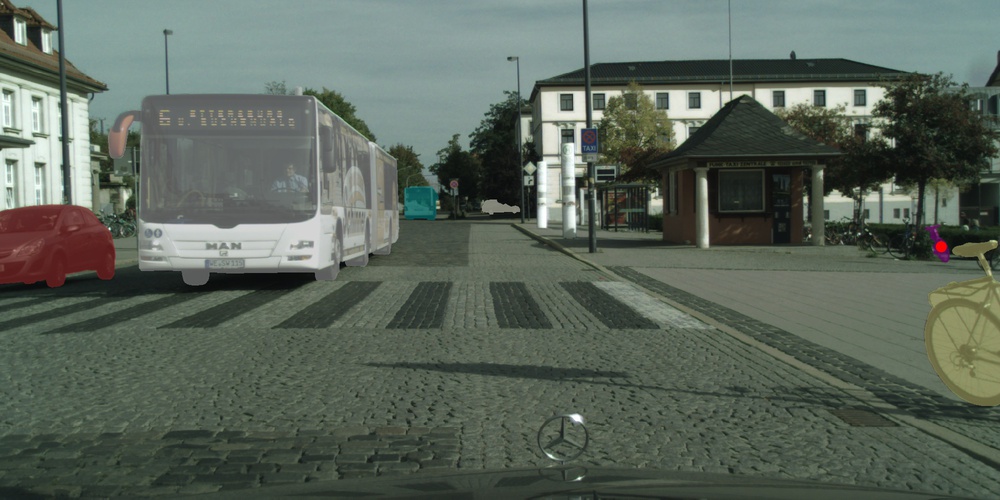} \\
		
	\end{tabular}
	
	\caption{We showcase qualitative segmentation results of our model on the CityscapesVideo validation set and compare it with the ground truth. Red points are the ground truth key point given by the annotator for the new objects. }
	\label{fig:results3}
\end{figure*}

\begin{figure*} 
	\centering
	\setlength\tabcolsep{0.5pt}
	\begin{tabular}{cc}
		

		
		\raisebox{2px}{{Ours}} &	\raisebox{2px}{{Ground Truth}}  \\
		
		\raisebox{40px}{\rotatebox{90}{t =  0 s}}
		\includegraphics[width=.45\textwidth]{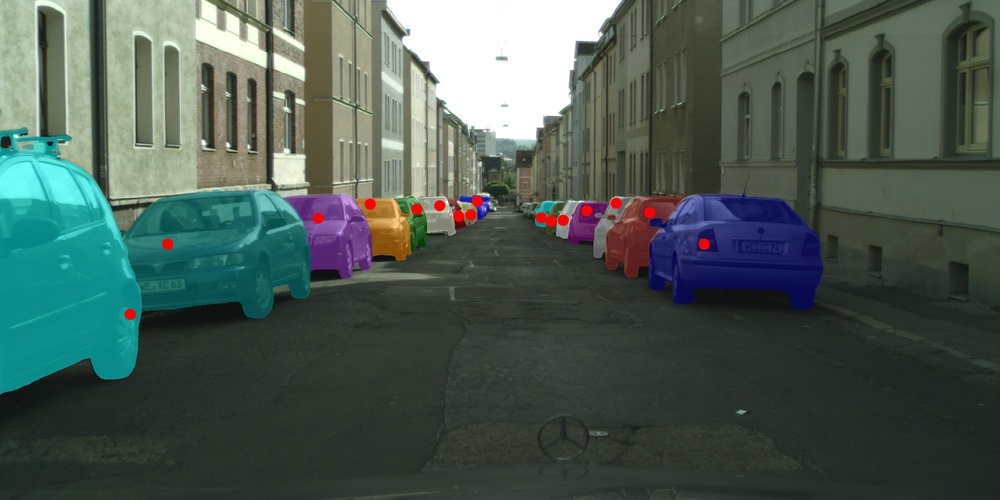} &
		\includegraphics[width=.45\textwidth]{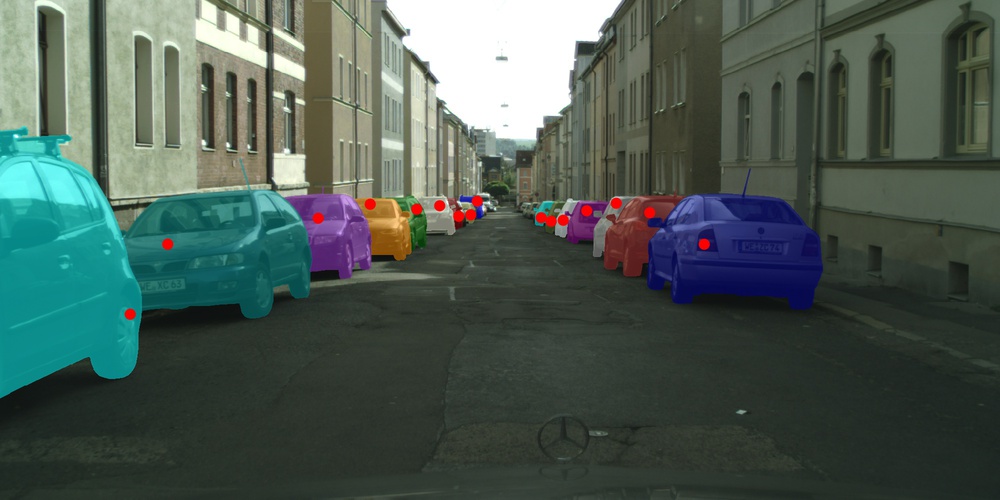} \\
		
		\raisebox{40px}{\rotatebox{90}{t =  0.43 s}}
		\includegraphics[width=.45\textwidth]{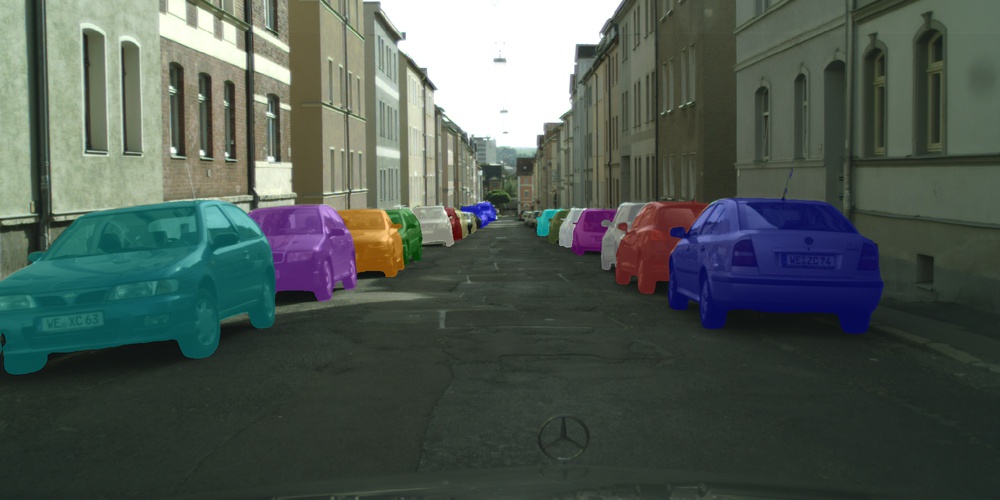} &
		\includegraphics[width=.45\textwidth]{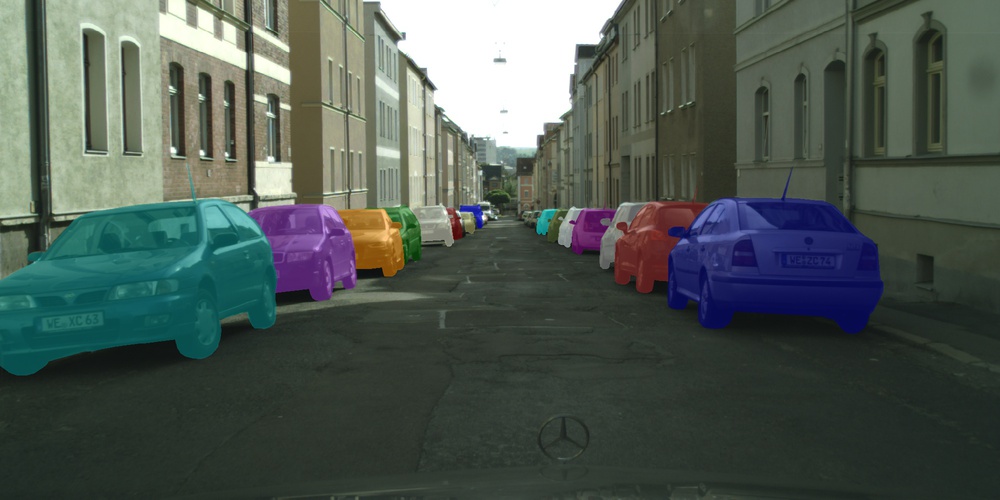} \\
		
		\raisebox{40px}{\rotatebox{90}{t =  0.81 s}}
		\includegraphics[width=.45\textwidth]{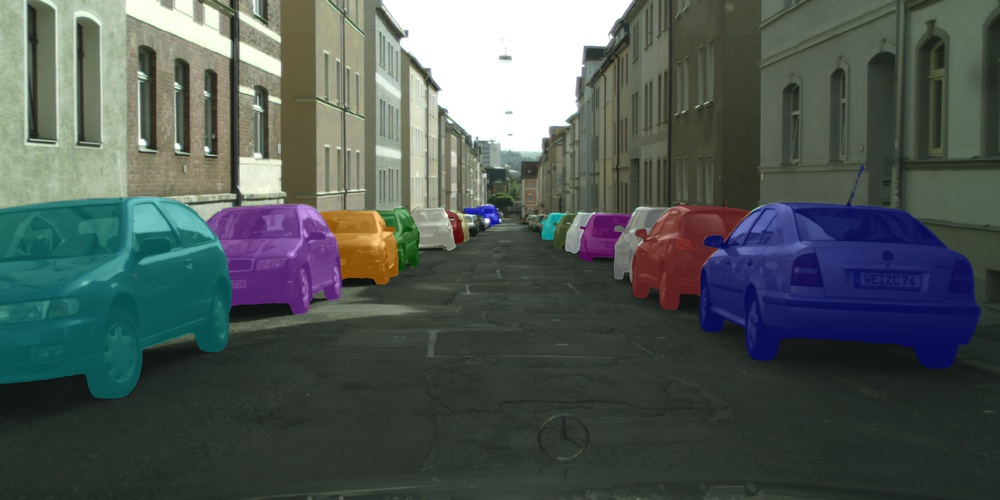} &
		\includegraphics[width=.45\textwidth]{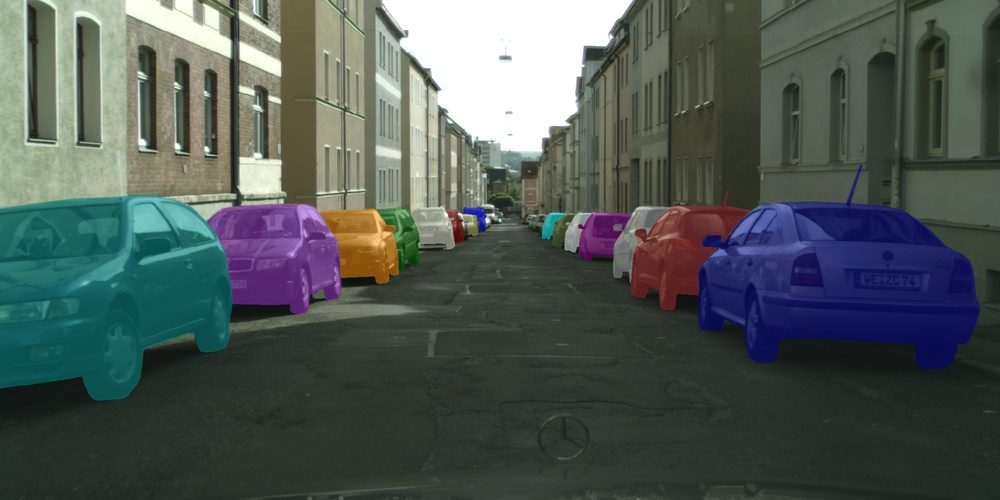} \\
		
		\raisebox{40px}{\rotatebox{90}{t =  1.18 s}}
		\includegraphics[width=.45\textwidth]{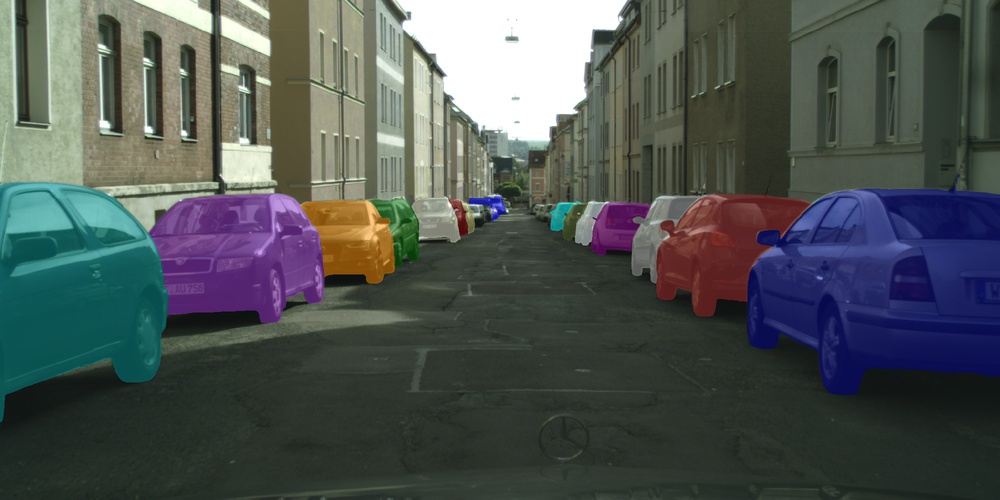} &
		\includegraphics[width=.45\textwidth]{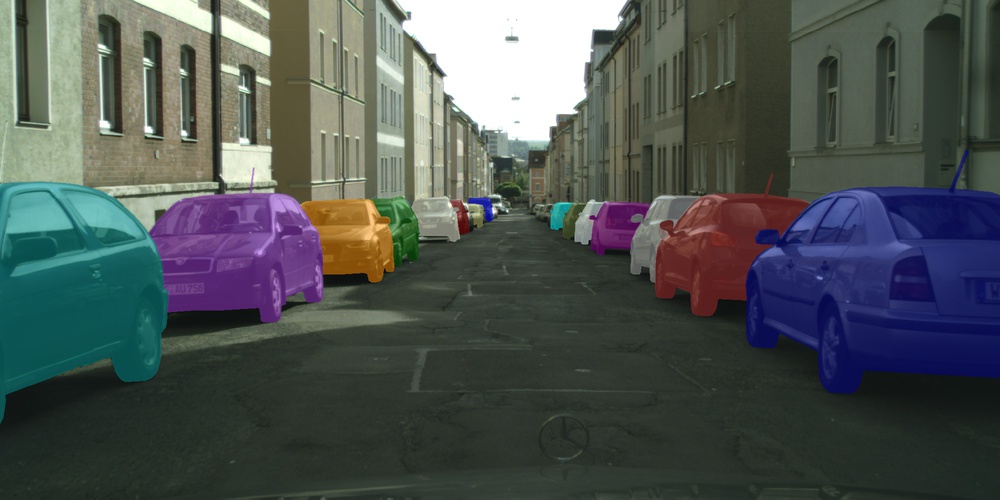} \\
		
		\raisebox{40px}{\rotatebox{90}{t =  1.55 s}}
		\includegraphics[width=.45\textwidth]{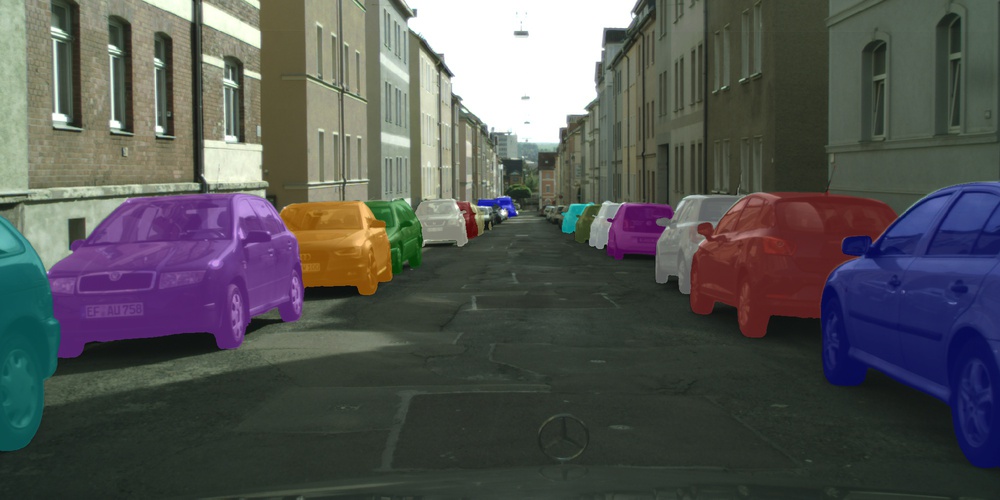} &
		\includegraphics[width=.45\textwidth]{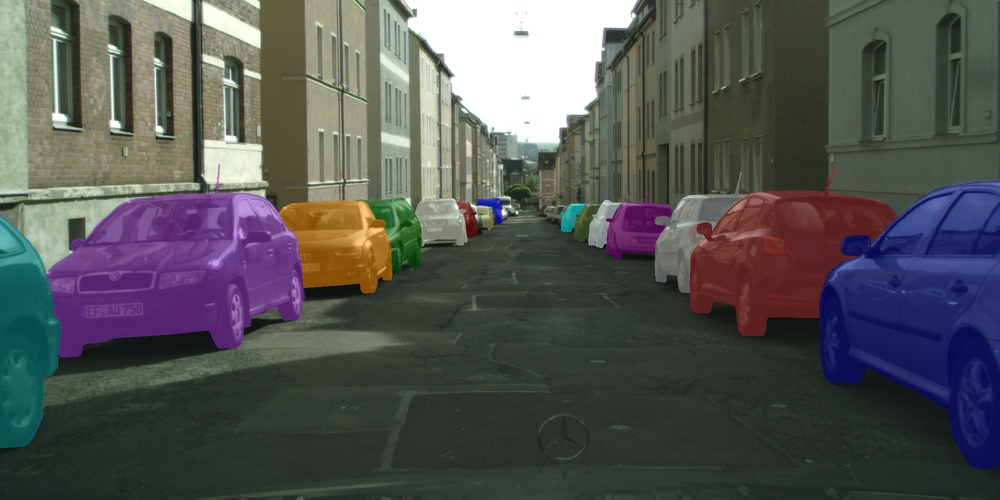} \\
		
	\end{tabular}
	
	\caption{We showcase qualitative segmentation results of our model on the CityscapesVideo validation set and compare it with the ground truth. Red points are the ground truth key point given by the annotator for the new objects. }
	\label{fig:results4}
\end{figure*}

\begin{figure*} 
	\centering
	\setlength\tabcolsep{0.5pt}
	\begin{tabular}{cc}
		

		
		\raisebox{2px}{{Ours}} &	\raisebox{2px}{{Ground Truth}}  \\
		
		\raisebox{40px}{\rotatebox{90}{t =  0 s}}
		\includegraphics[width=.45\textwidth]{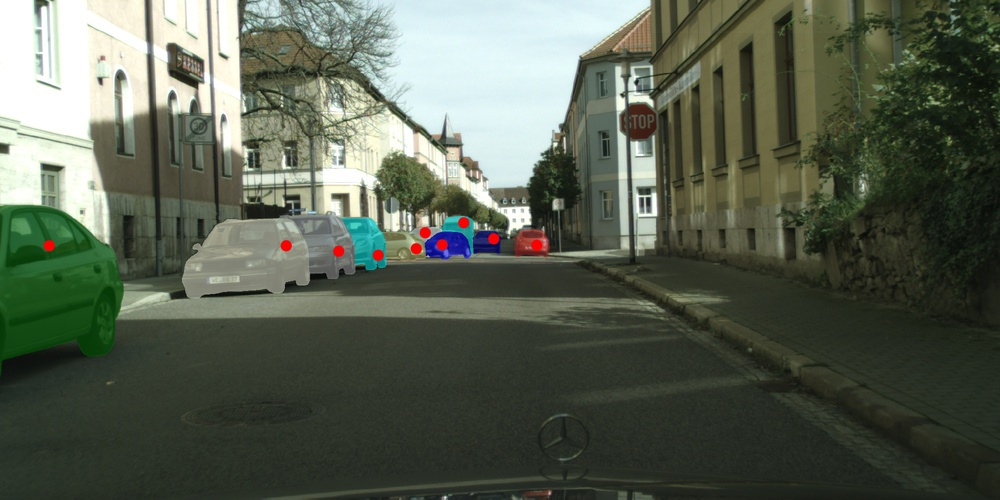} &
		\includegraphics[width=.45\textwidth]{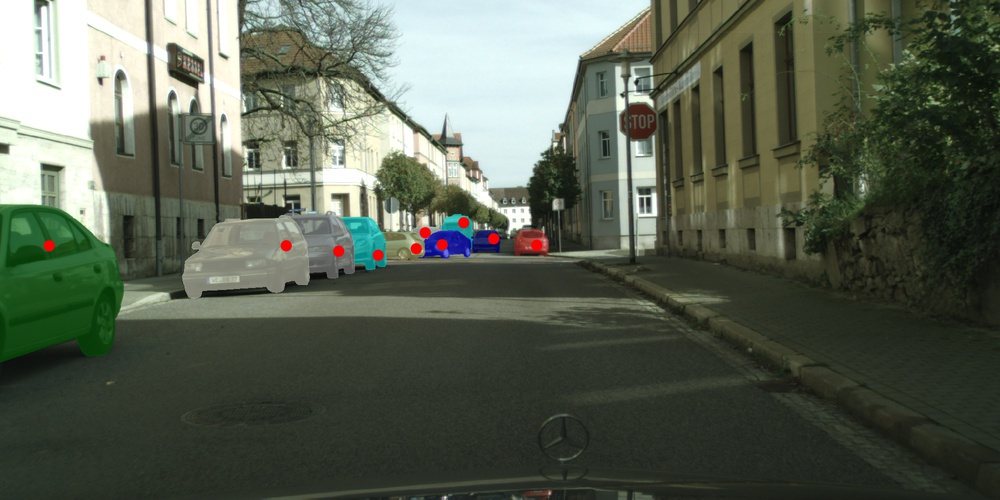} \\
		
		\raisebox{40px}{\rotatebox{90}{t =  0.43 s}}
		\includegraphics[width=.45\textwidth]{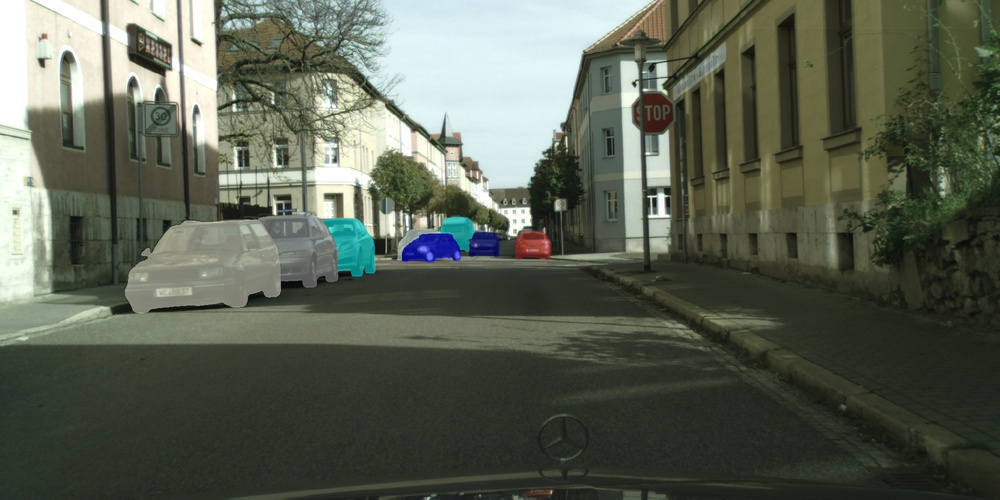} &
		\includegraphics[width=.45\textwidth]{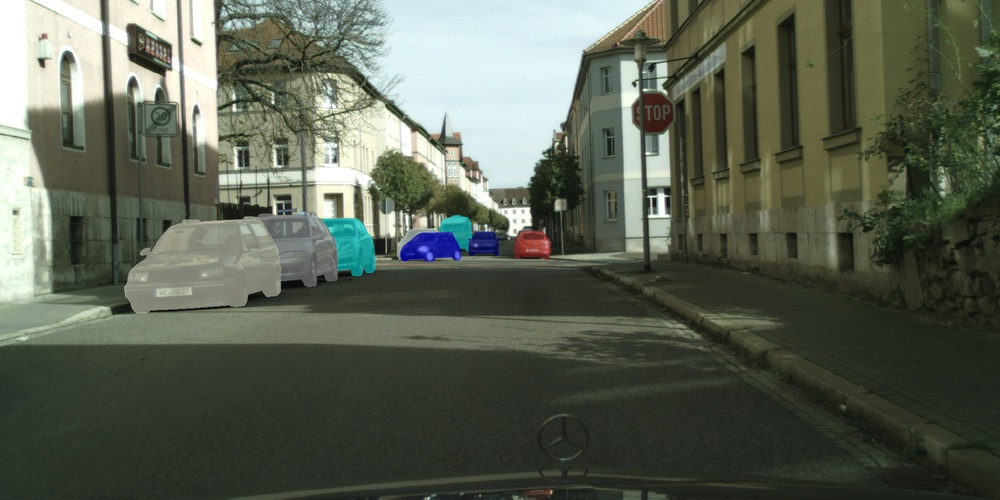} \\
		
		\raisebox{40px}{\rotatebox{90}{t =  0.81 s}}
		\includegraphics[width=.45\textwidth]{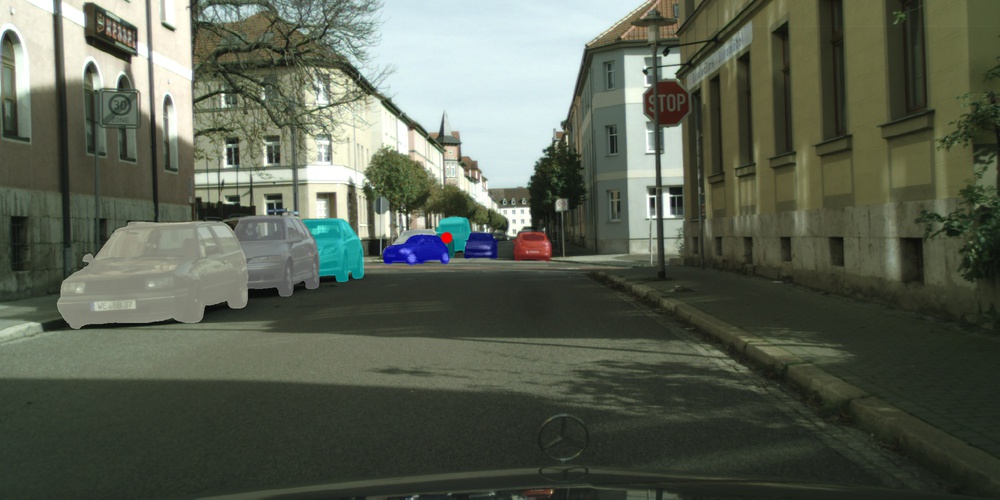} &
		\includegraphics[width=.45\textwidth]{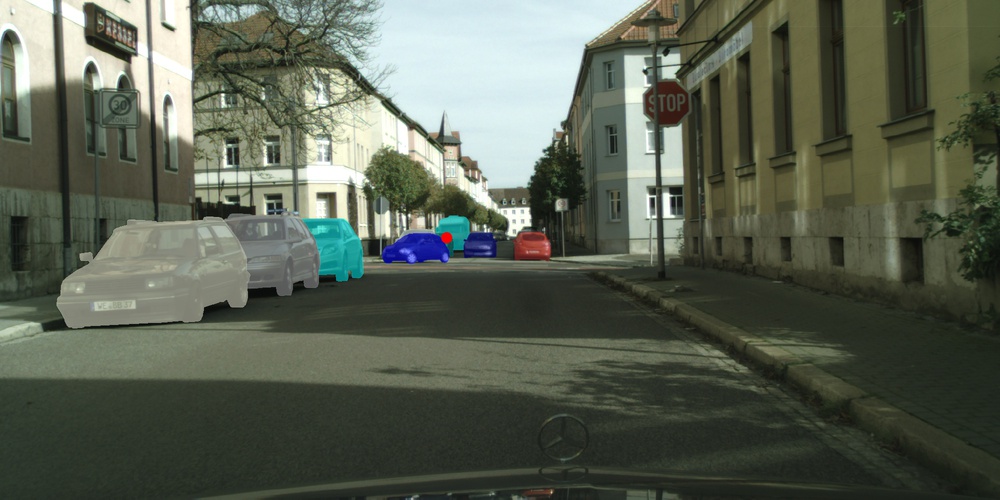} \\
		
		\raisebox{40px}{\rotatebox{90}{t =  1.18 s}}
		\includegraphics[width=.45\textwidth]{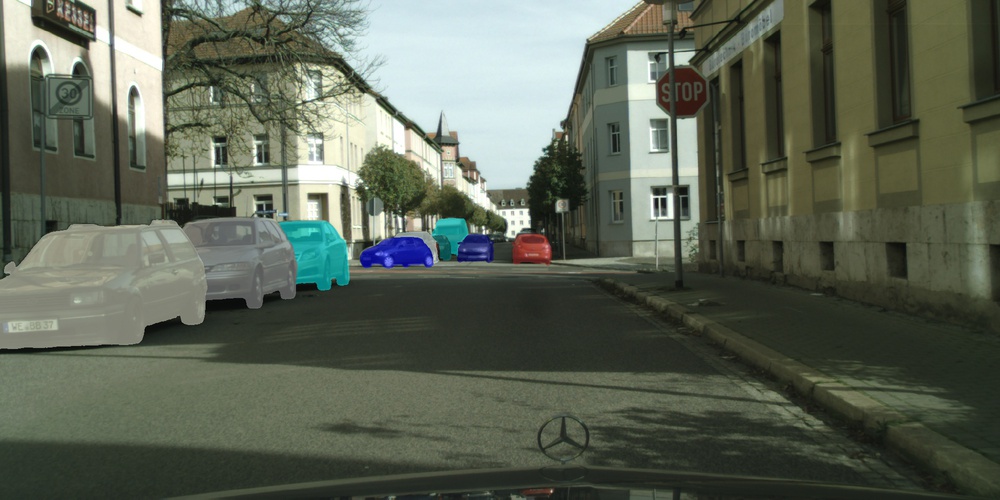} &
		\includegraphics[width=.45\textwidth]{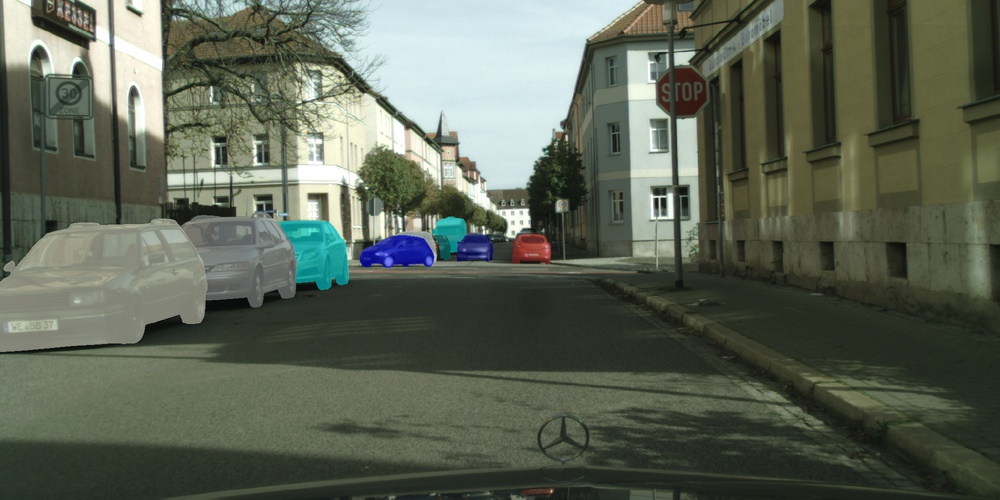} \\
		
		\raisebox{40px}{\rotatebox{90}{t =  1.55 s}}
		\includegraphics[width=.45\textwidth]{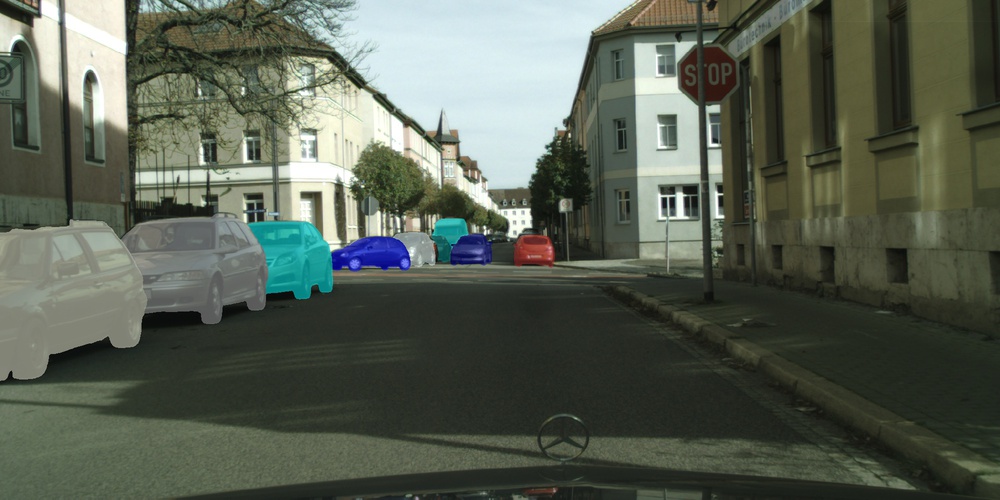} &
		\includegraphics[width=.45\textwidth]{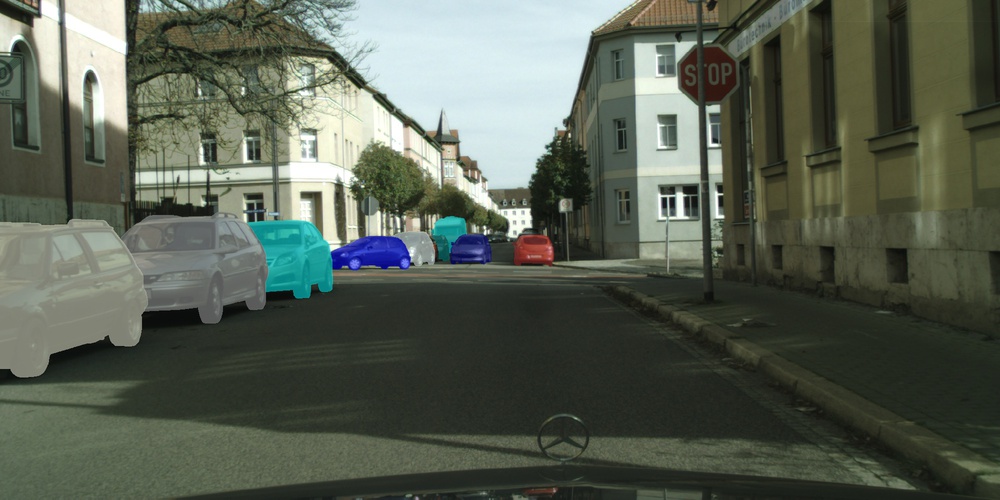} \\
		
	\end{tabular}
	
	\caption{We showcase qualitative segmentation results of our model on the CityscapesVideo validation set and compare it with the ground truth. Red points are the ground truth key point given by the annotator for the new objects. }
	\label{fig:results5}
\end{figure*}

\begin{figure*} 
	\centering
	\setlength\tabcolsep{0.5pt}
	\begin{tabular}{cc}
		

		
		\raisebox{2px}{{Ours}} &	\raisebox{2px}{{Ground Truth}}  \\
		
		\raisebox{40px}{\rotatebox{90}{t =  0 s}}
		\includegraphics[width=.45\textwidth]{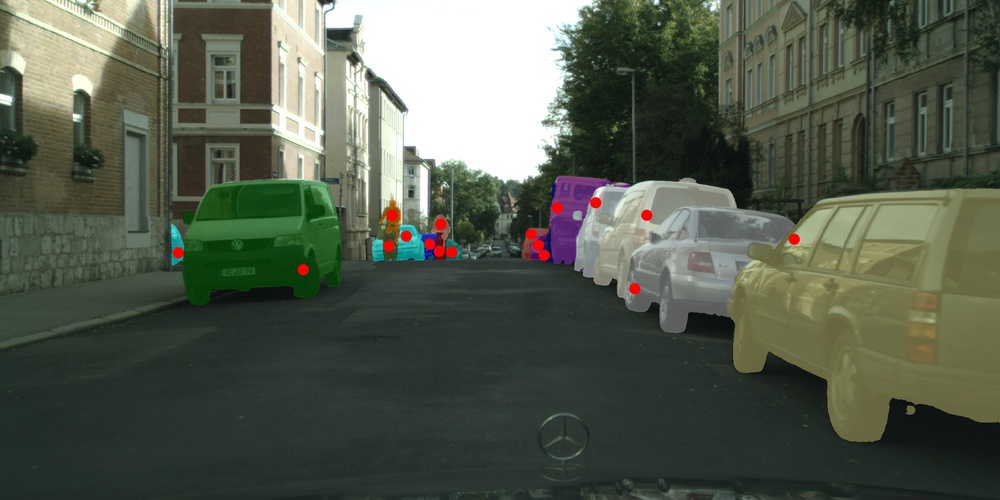} &
		\includegraphics[width=.45\textwidth]{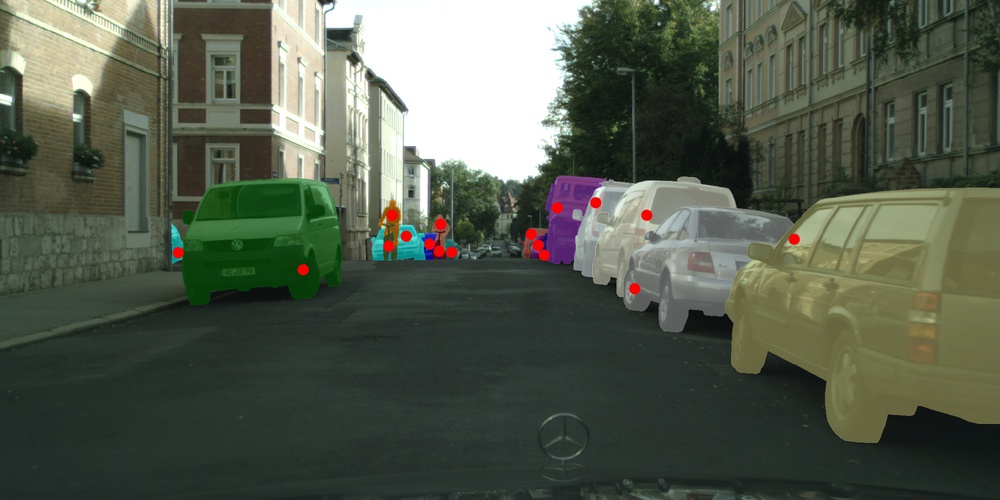} \\
		
		\raisebox{40px}{\rotatebox{90}{t =  0.43 s}}
		\includegraphics[width=.45\textwidth]{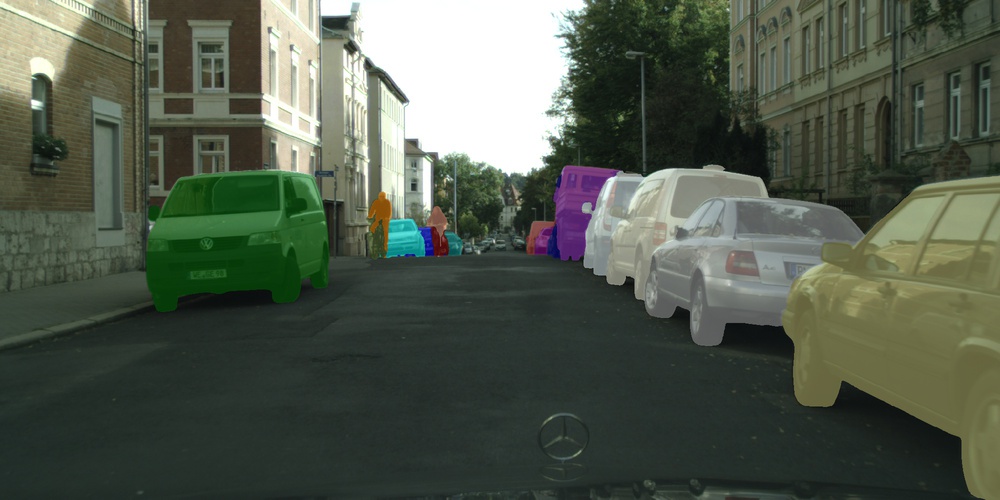} &
		\includegraphics[width=.45\textwidth]{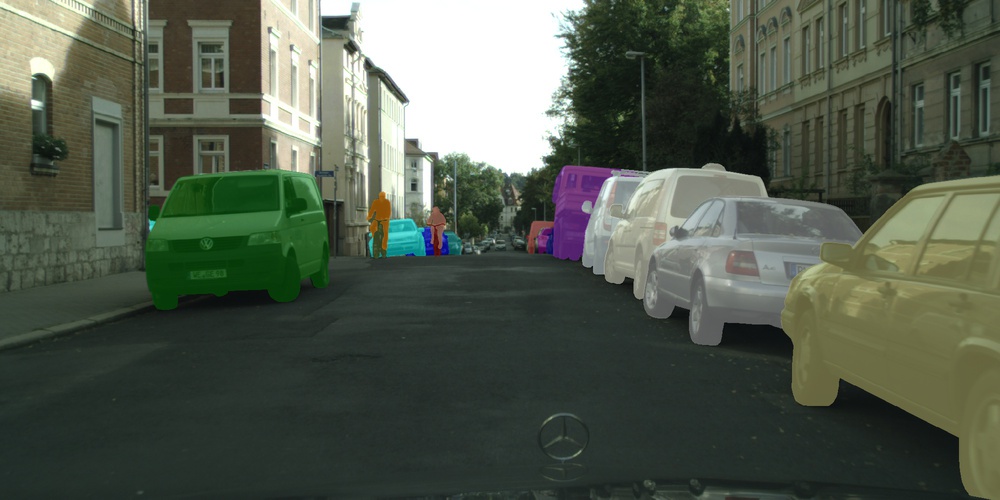} \\
		
		\raisebox{40px}{\rotatebox{90}{t =  0.81 s}}
		\includegraphics[width=.45\textwidth]{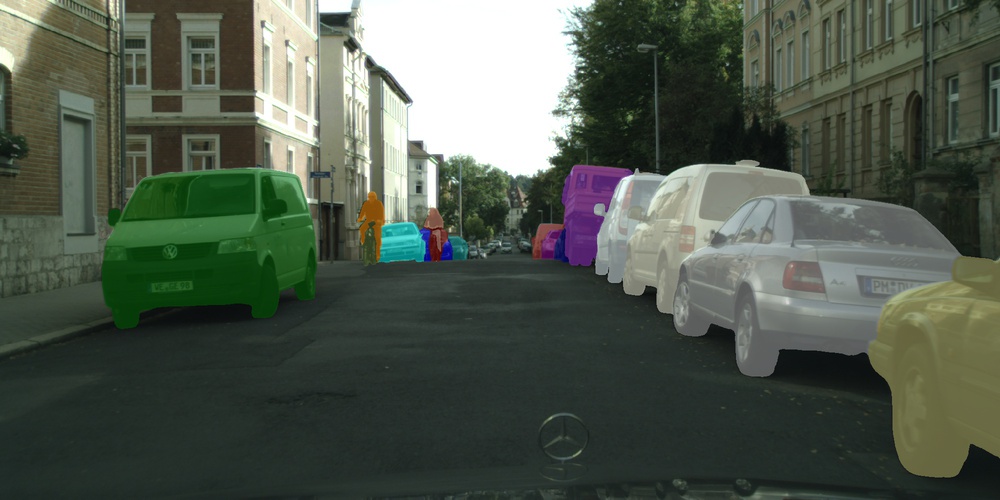} &
		\includegraphics[width=.45\textwidth]{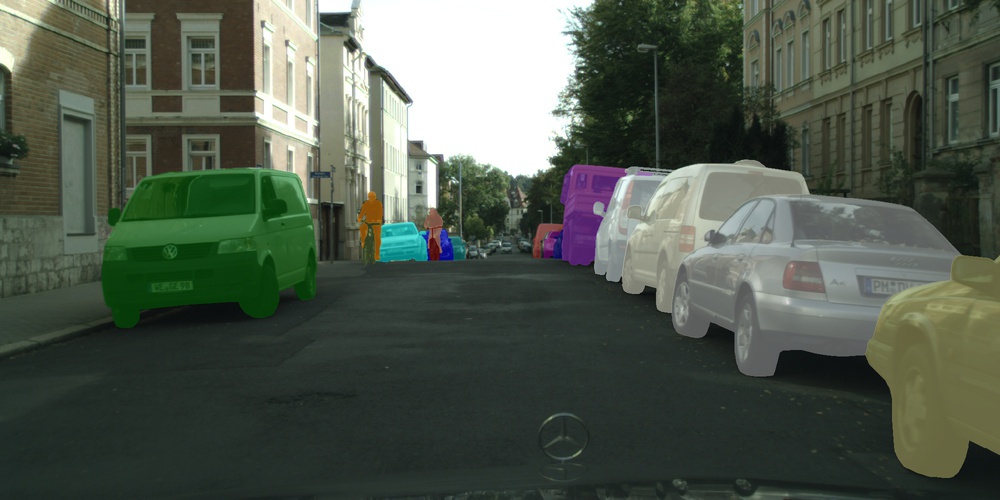} \\
		
		\raisebox{40px}{\rotatebox{90}{t =  1.18 s}}
		\includegraphics[width=.45\textwidth]{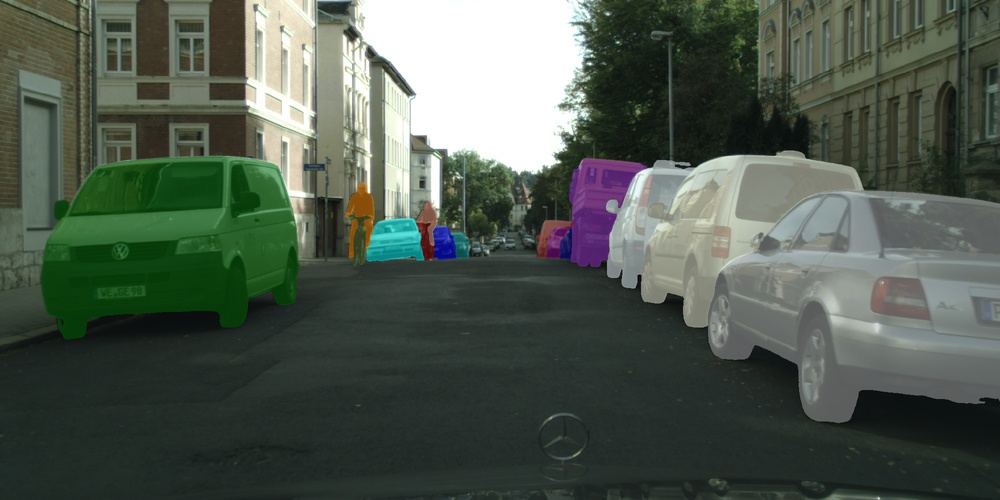} &
		\includegraphics[width=.45\textwidth]{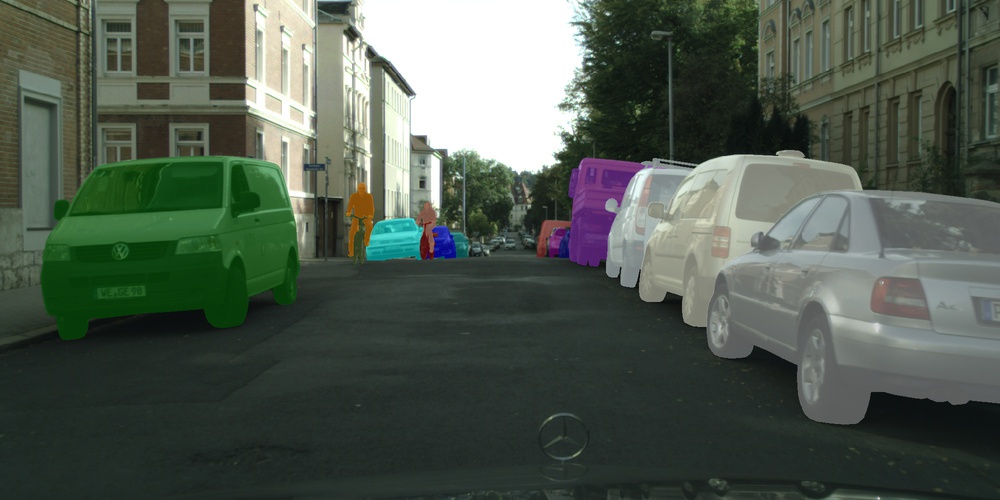} \\
		
		\raisebox{40px}{\rotatebox{90}{t =  1.55 s}}
		\includegraphics[width=.45\textwidth]{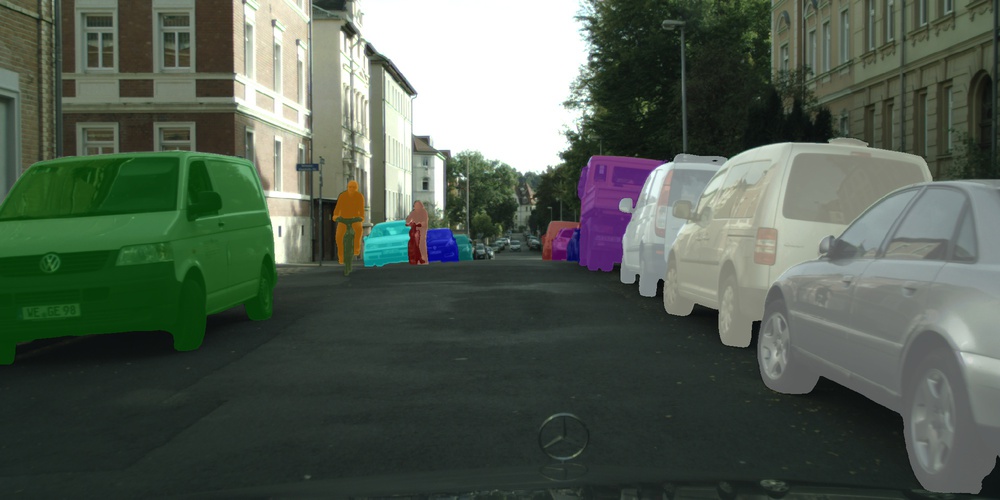} &
		\includegraphics[width=.45\textwidth]{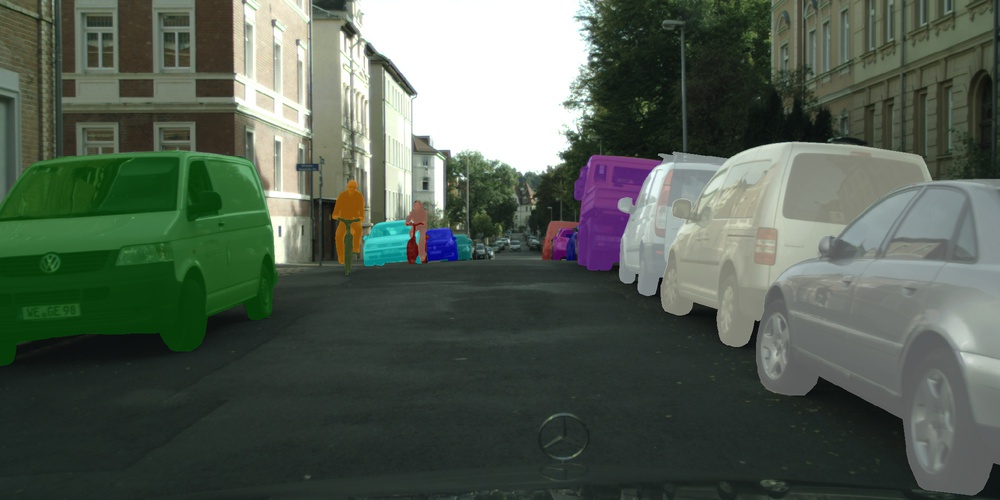} \\
		
	\end{tabular}
	
	\caption{We showcase qualitative segmentation results of our model on the CityscapesVideo validation set and compare it with the ground truth. Red points are the ground truth key point given by the annotator for the new objects. }
	\label{fig:results6}
\end{figure*}

\begin{figure*} 
	\centering
	\setlength\tabcolsep{0.5pt}
	\begin{tabular}{cc}
		

		
		\raisebox{2px}{{Ours}} &	\raisebox{2px}{{Ground Truth}}  \\
		
		\raisebox{40px}{\rotatebox{90}{t =  0 s}}
		\includegraphics[width=.45\textwidth]{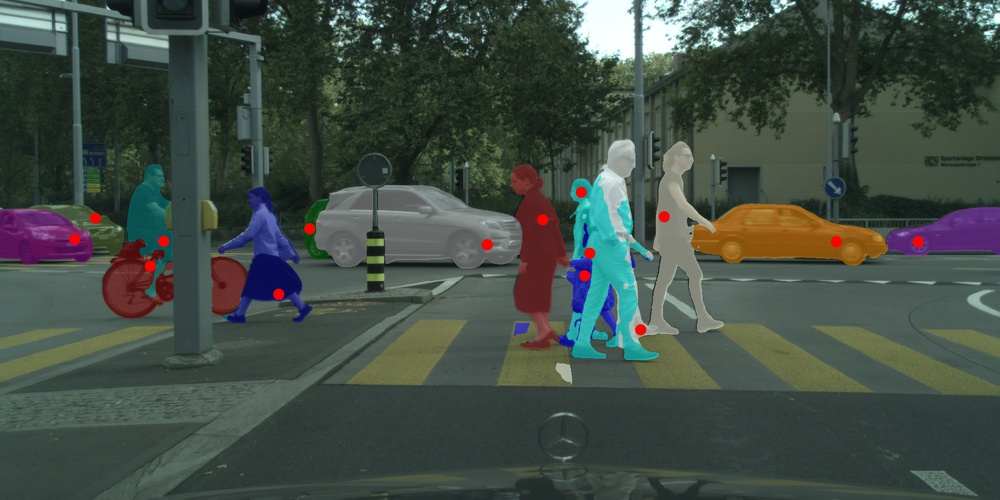} &
		\includegraphics[width=.45\textwidth]{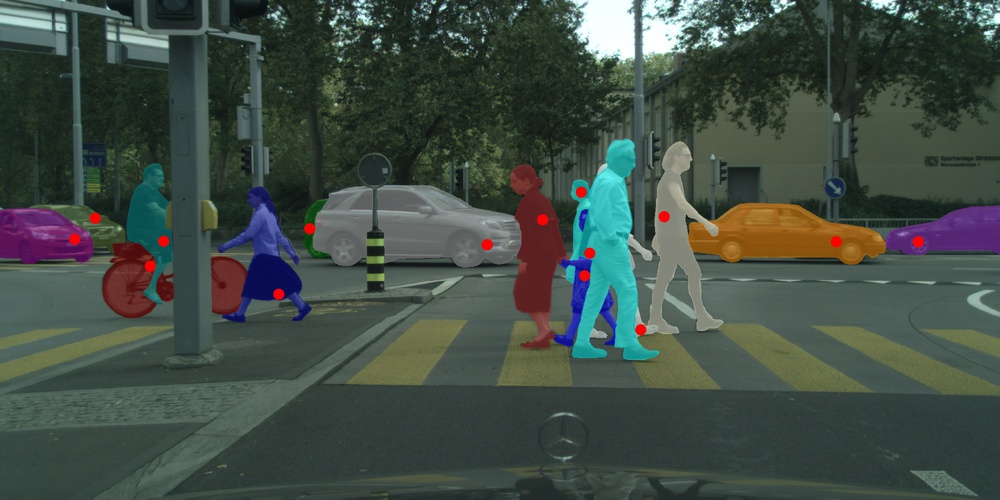} \\
		
		\raisebox{40px}{\rotatebox{90}{t =  0.43 s}}
		\includegraphics[width=.45\textwidth]{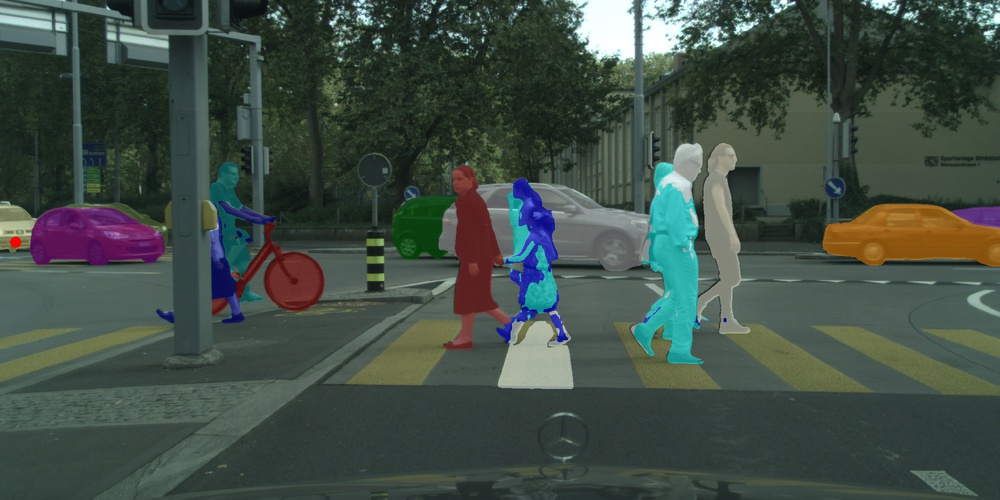} &
		\includegraphics[width=.45\textwidth]{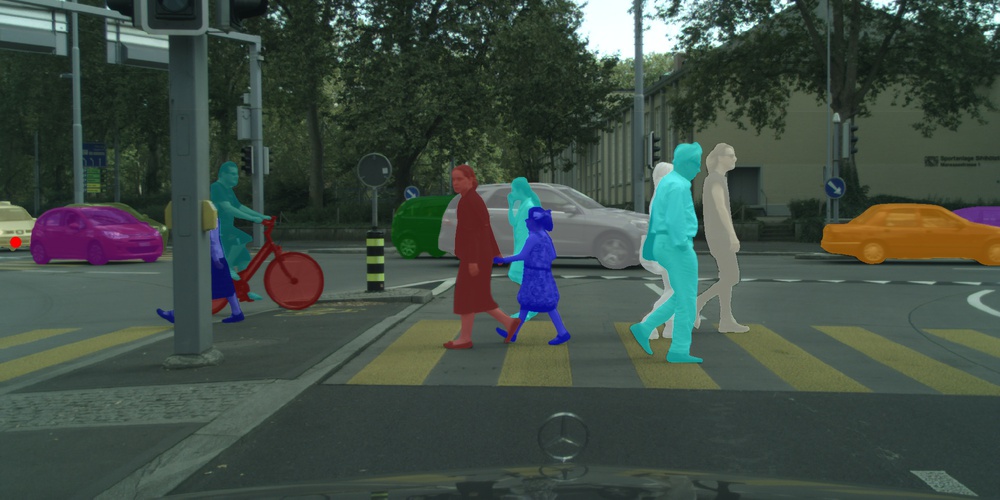} \\
		
		\raisebox{40px}{\rotatebox{90}{t =  0.81 s}}
		\includegraphics[width=.45\textwidth]{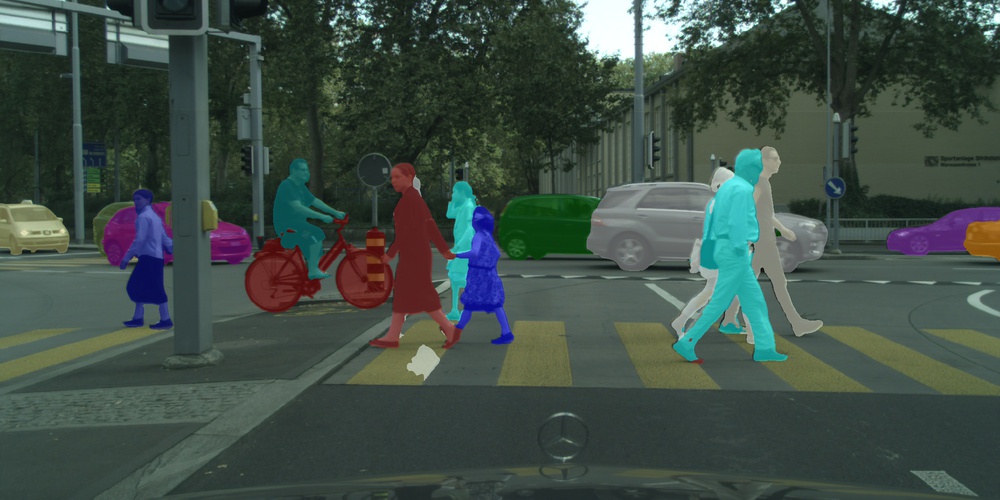} &
		\includegraphics[width=.45\textwidth]{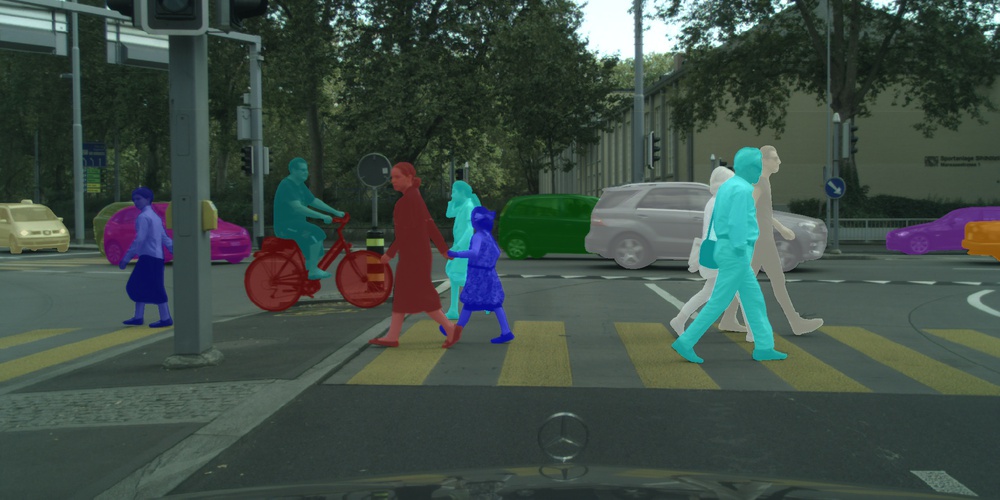} \\
		
		\raisebox{40px}{\rotatebox{90}{t =  1.18 s}}
		\includegraphics[width=.45\textwidth]{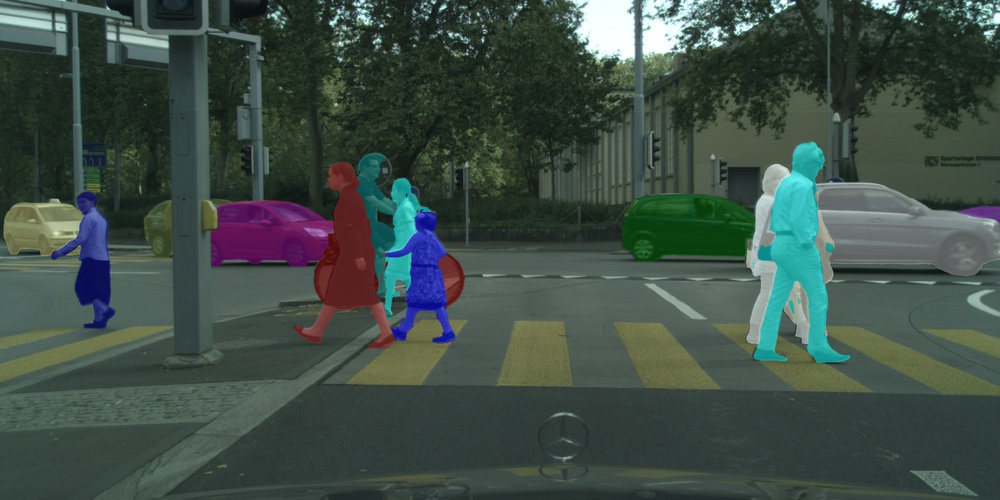} &
		\includegraphics[width=.45\textwidth]{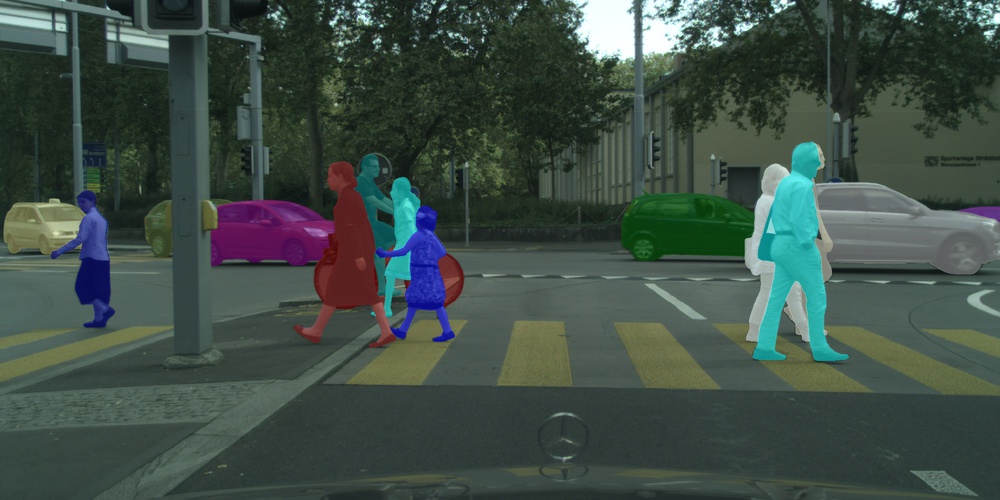} \\
		
		\raisebox{40px}{\rotatebox{90}{t =  1.55 s}}
		\includegraphics[width=.45\textwidth]{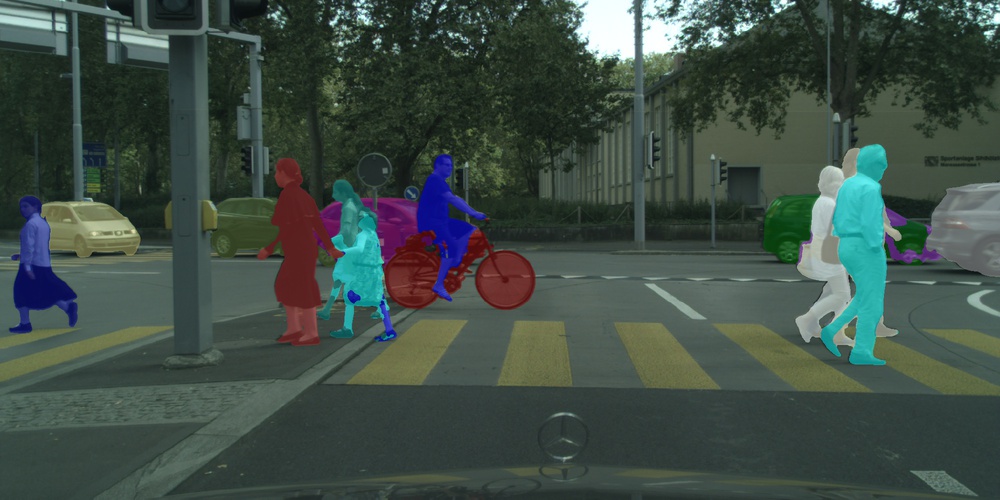} &
		\includegraphics[width=.45\textwidth]{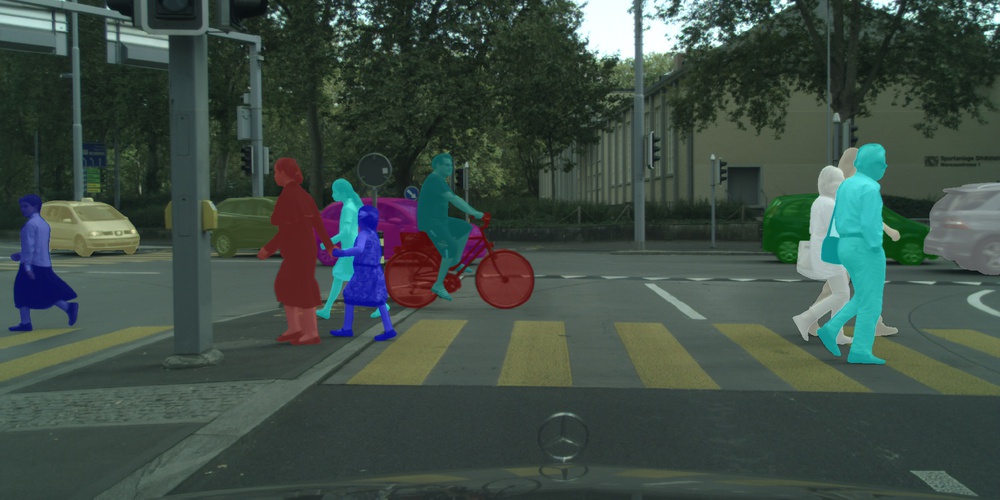} \\
		
	\end{tabular}
	
	\caption{We showcase qualitative segmentation results of our model on the CityscapesVideo validation set and compare it with the ground truth. Red points are the ground truth key point given by the annotator for the new objects. }
	\label{fig:results7}
\end{figure*}

\begin{figure*} 
	\centering
	\setlength\tabcolsep{0.5pt}
	\begin{tabular}{cc}
		

		
		\raisebox{2px}{{Ours}} &	\raisebox{2px}{{Ground Truth}}  \\
		
		\raisebox{40px}{\rotatebox{90}{t =  0 s}}
		\includegraphics[width=.45\textwidth]{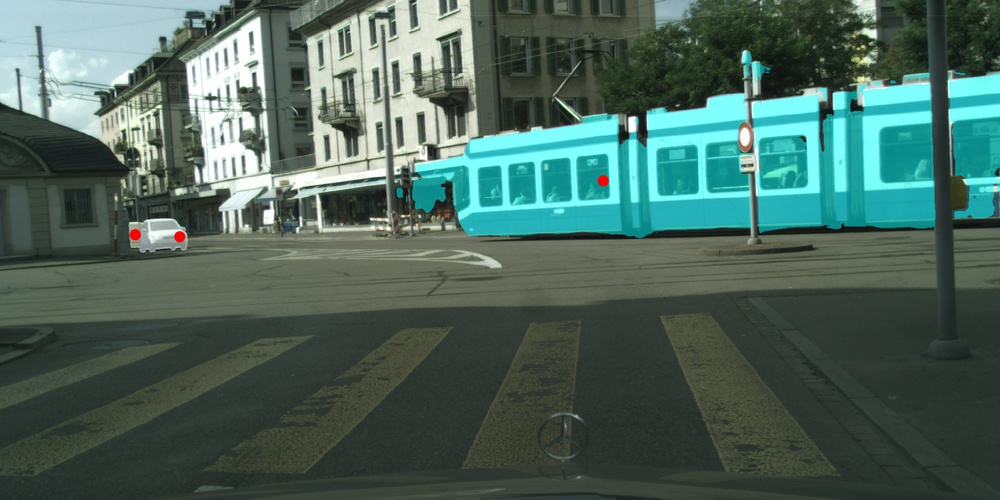} &
		\includegraphics[width=.45\textwidth]{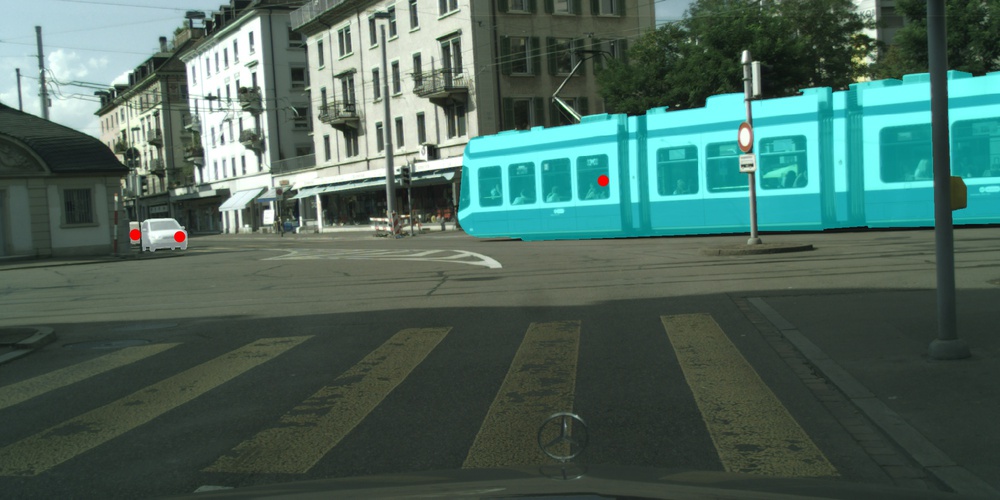} \\
		
		\raisebox{40px}{\rotatebox{90}{t =  0.43 s}}
		\includegraphics[width=.45\textwidth]{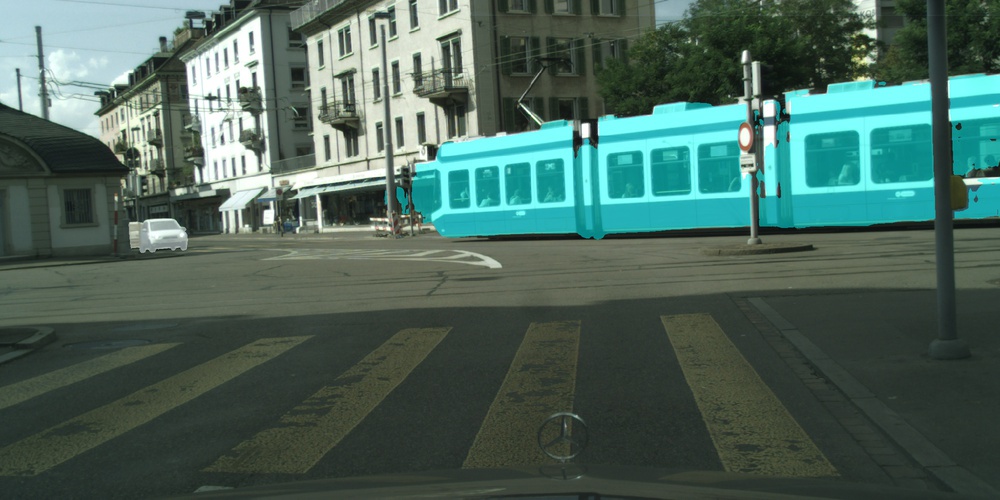} &
		\includegraphics[width=.45\textwidth]{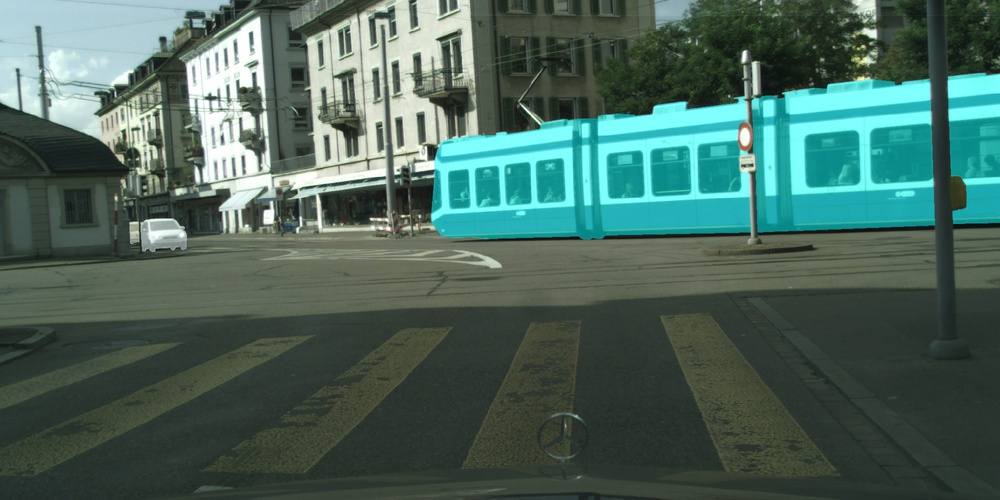} \\
		
		\raisebox{40px}{\rotatebox{90}{t =  0.81 s}}
		\includegraphics[width=.45\textwidth]{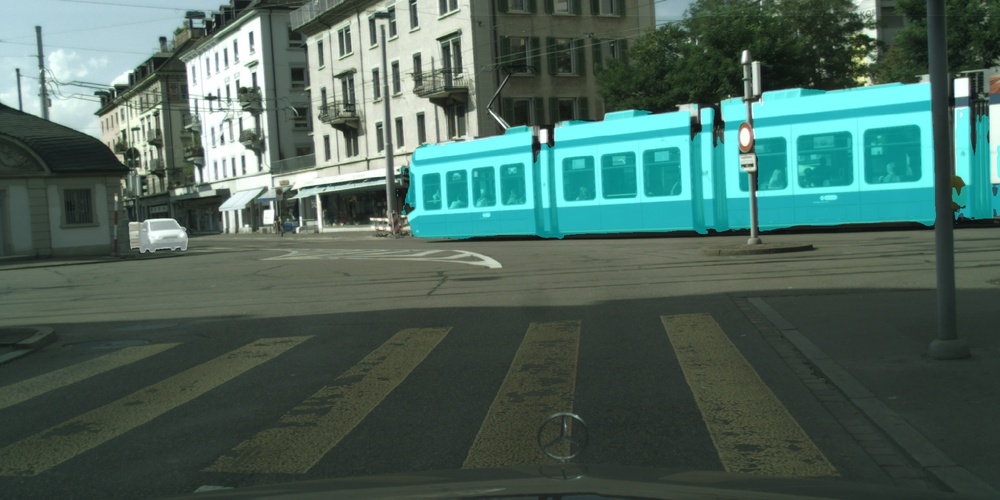} &
		\includegraphics[width=.45\textwidth]{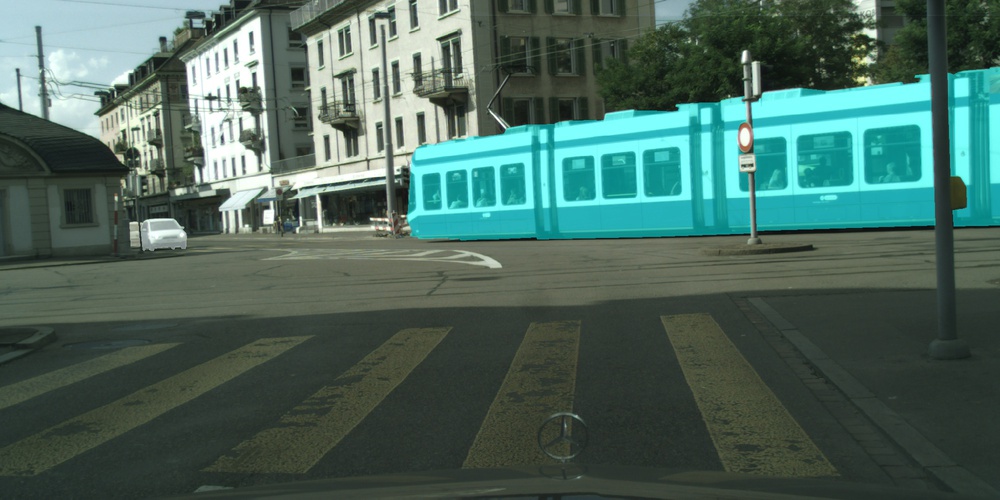} \\
		
		\raisebox{40px}{\rotatebox{90}{t =  1.18 s}}
		\includegraphics[width=.45\textwidth]{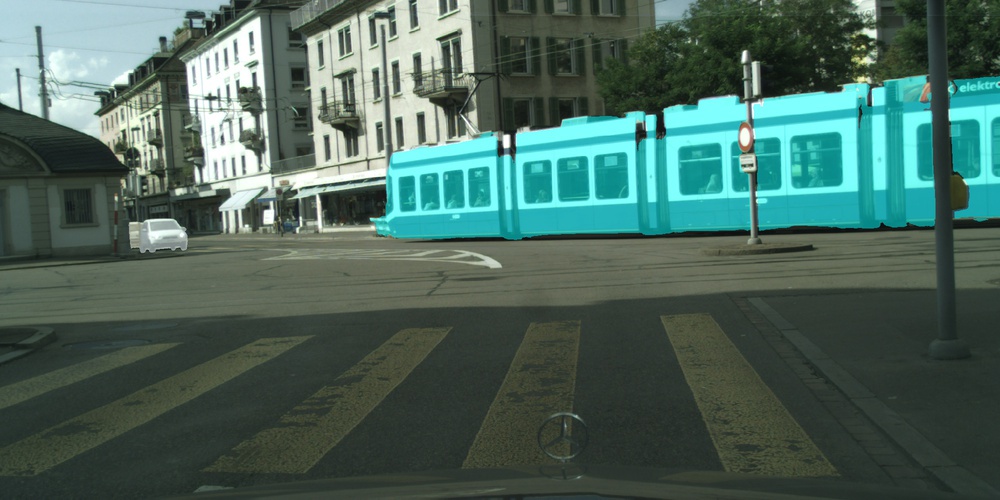} &
		\includegraphics[width=.45\textwidth]{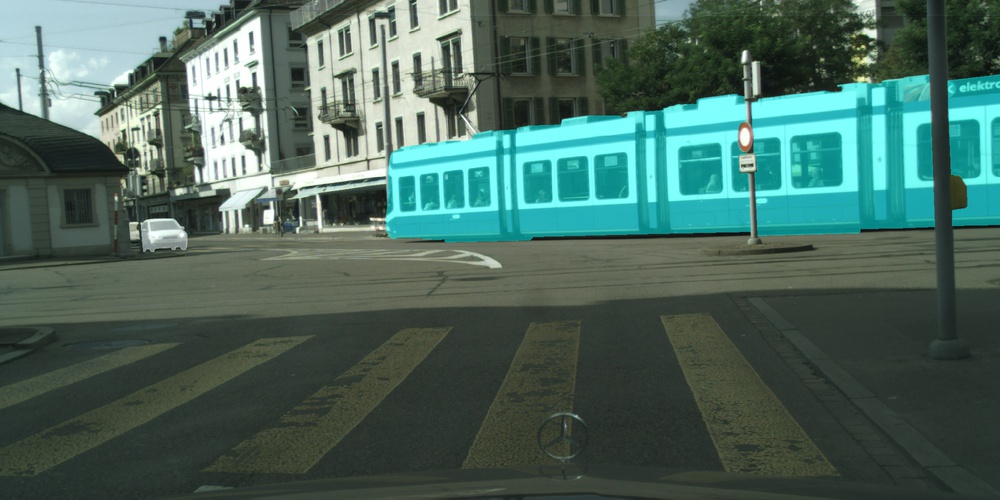} \\
		
		\raisebox{40px}{\rotatebox{90}{t =  1.55 s}}
		\includegraphics[width=.45\textwidth]{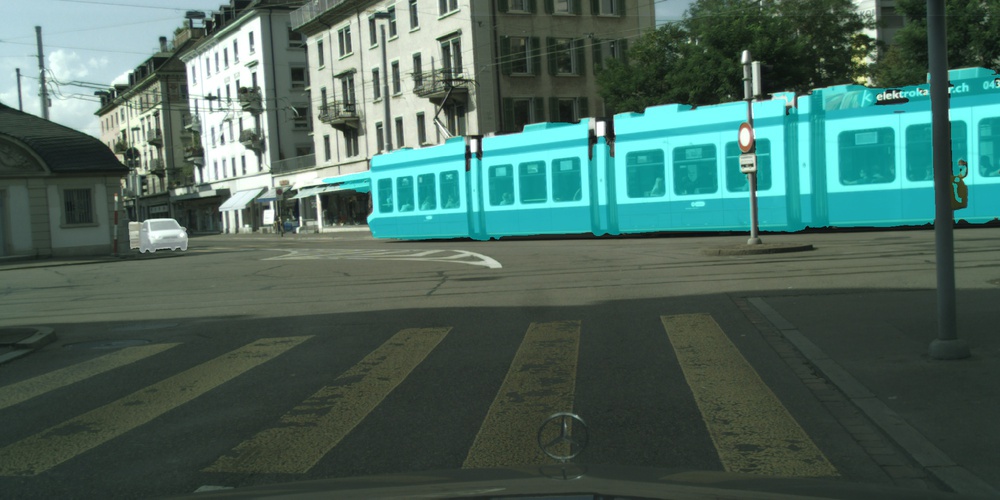} &
		\includegraphics[width=.45\textwidth]{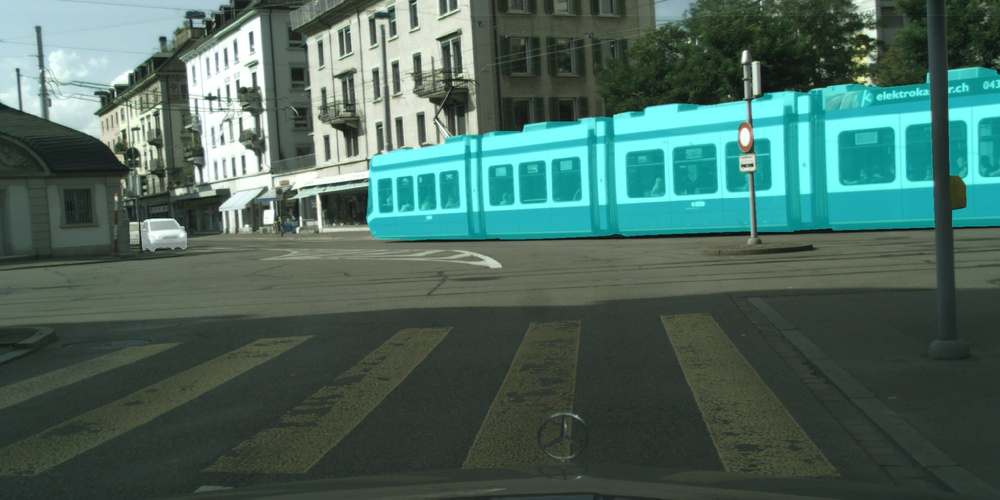} \\
		
	\end{tabular}
	
	\caption{We showcase qualitative segmentation results of our model on the CityscapesVideo validation set and compare it with the ground truth. Red points are the ground truth key point given by the annotator for the new objects. }
	\label{fig:results8}
\end{figure*}

\begin{figure*} 
	\centering
	\setlength\tabcolsep{0.5pt}
	\begin{tabular}{cc}
		

		
		\raisebox{2px}{{Ours}} &	\raisebox{2px}{{Ground Truth}}  \\
		
		\raisebox{40px}{\rotatebox{90}{t =  0 s}}
		\includegraphics[width=.45\textwidth]{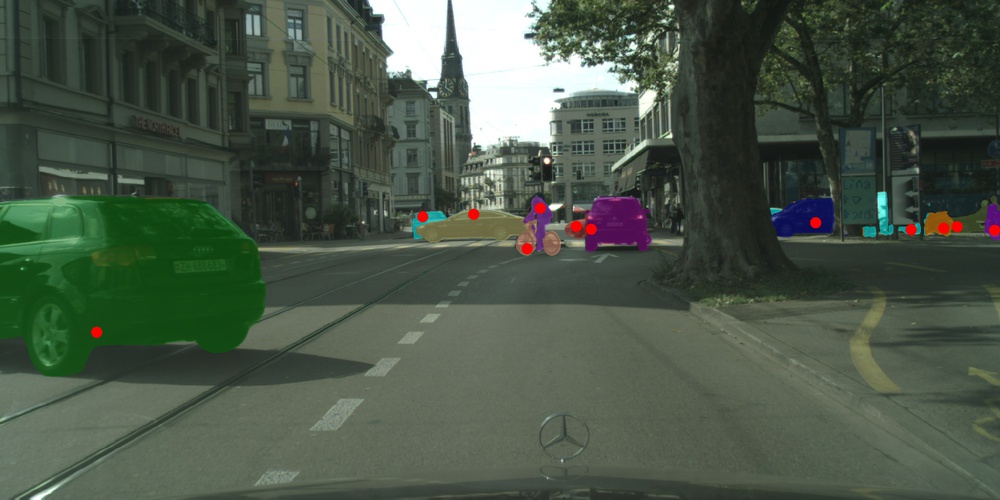} &
		\includegraphics[width=.45\textwidth]{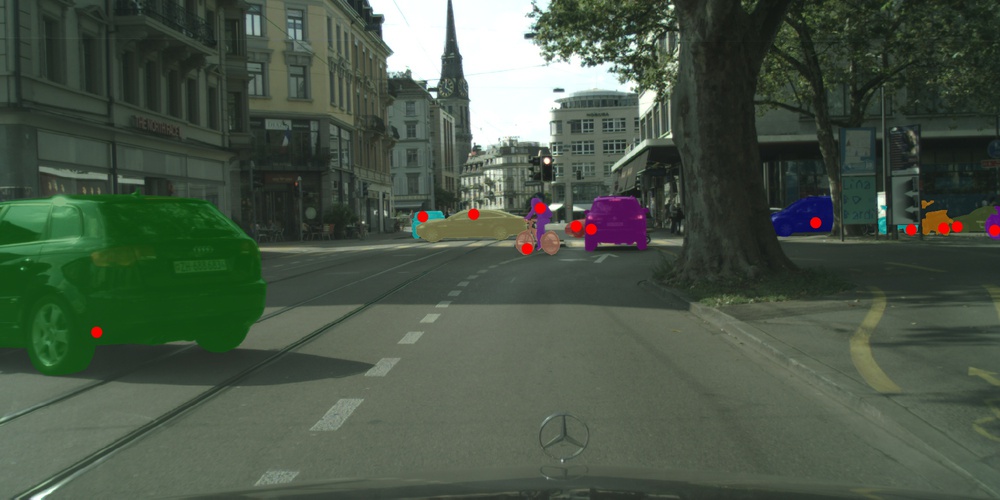} \\
		
		\raisebox{40px}{\rotatebox{90}{t =  0.43 s}}
		\includegraphics[width=.45\textwidth]{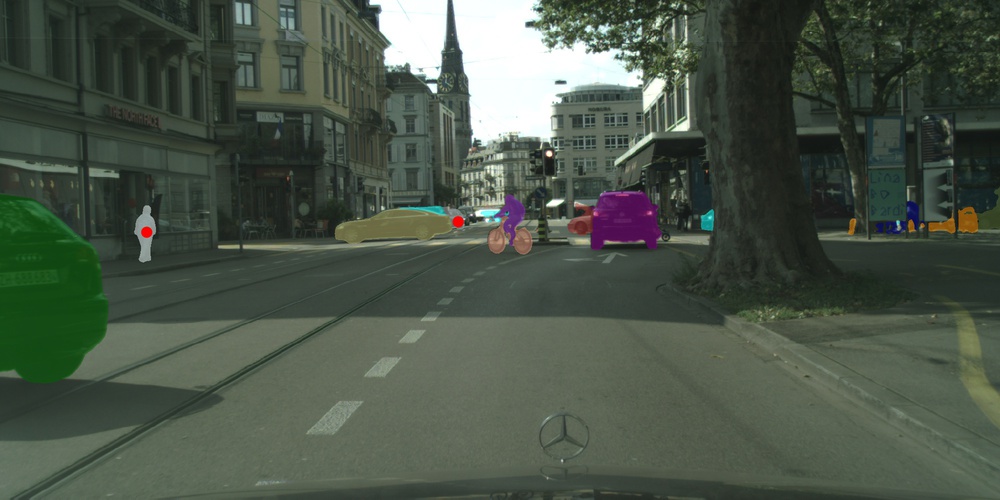} &
		\includegraphics[width=.45\textwidth]{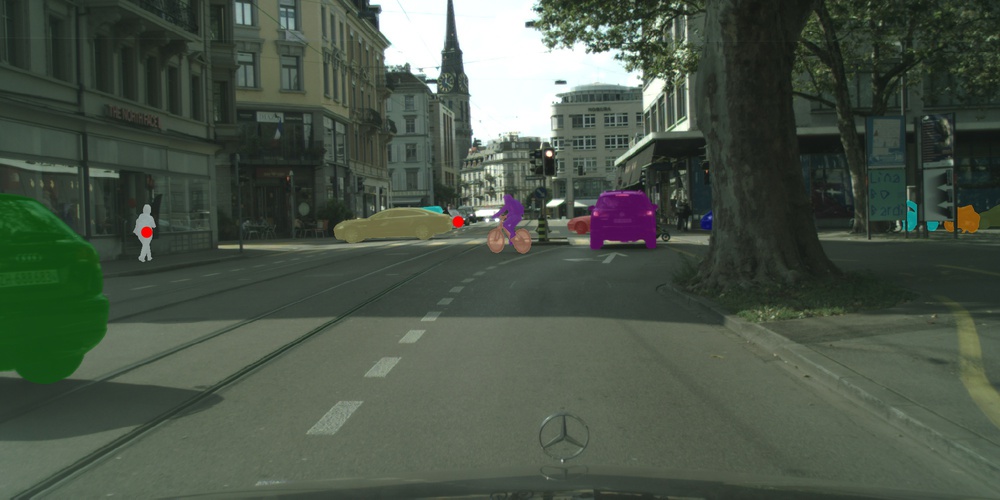} \\
		
		\raisebox{40px}{\rotatebox{90}{t =  0.81 s}}
		\includegraphics[width=.45\textwidth]{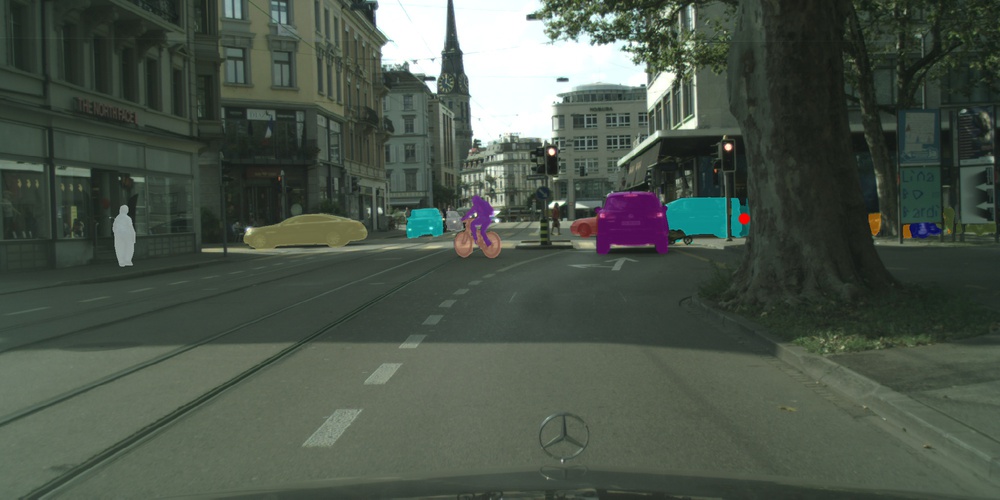} &
		\includegraphics[width=.45\textwidth]{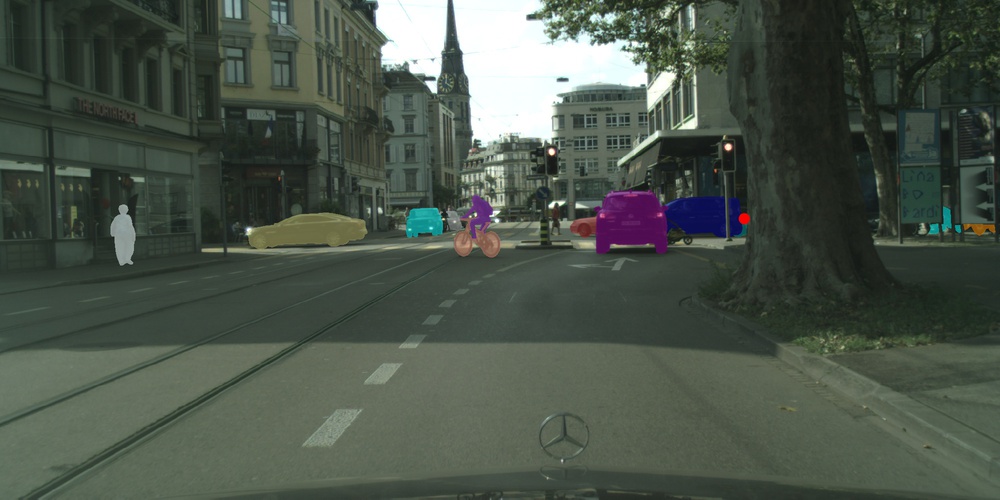} \\
		
		\raisebox{40px}{\rotatebox{90}{t =  1.18 s}}
		\includegraphics[width=.45\textwidth]{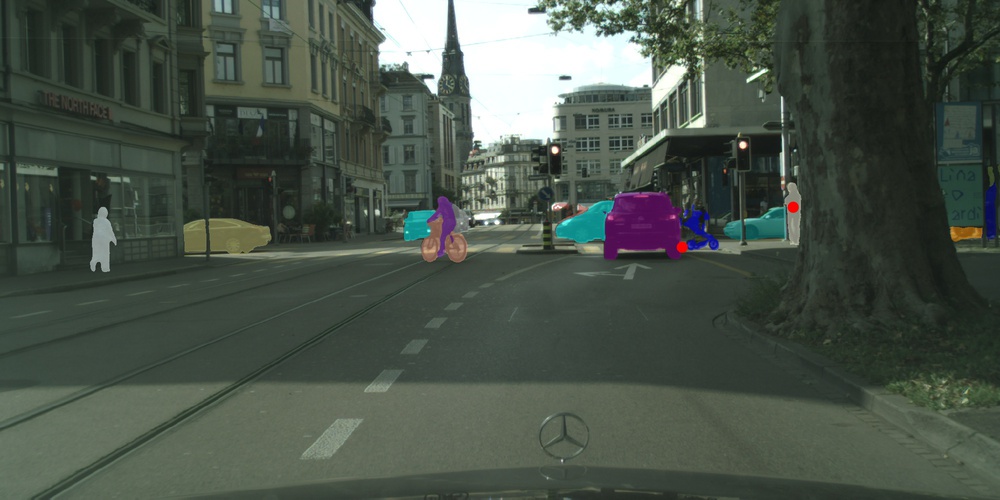} &
		\includegraphics[width=.45\textwidth]{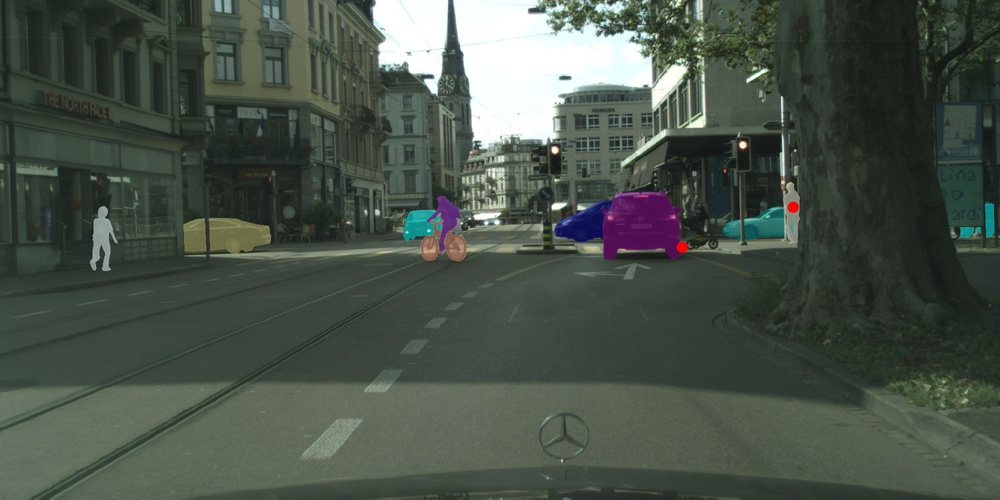} \\
		
		\raisebox{40px}{\rotatebox{90}{t =  1.55 s}}
		\includegraphics[width=.45\textwidth]{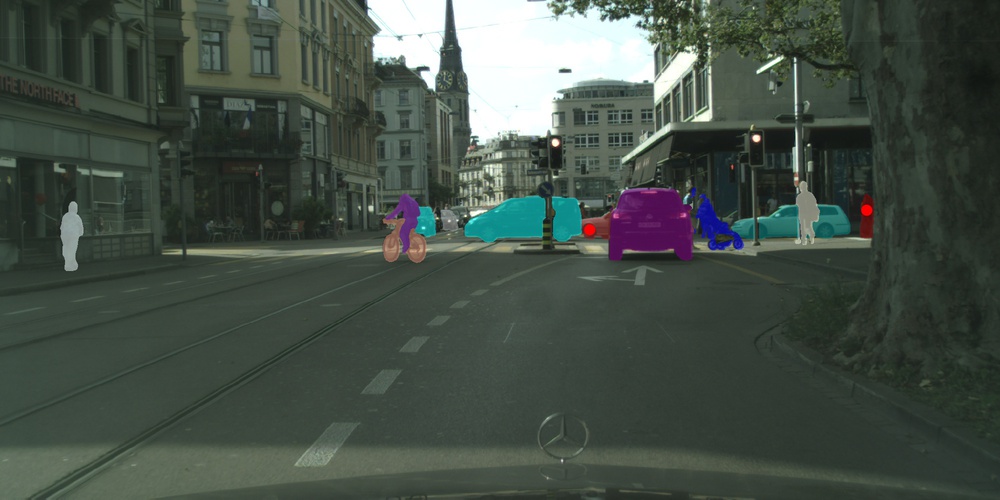} &
		\includegraphics[width=.45\textwidth]{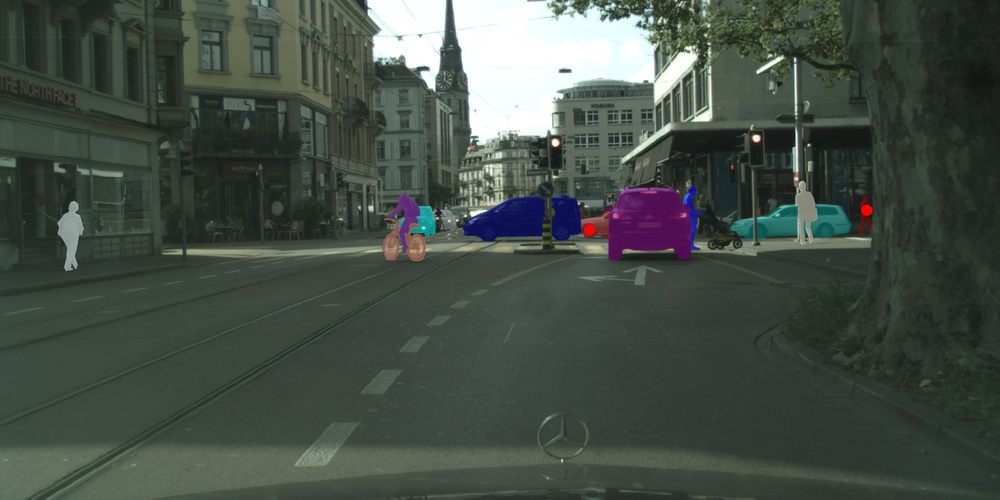} \\
		
	\end{tabular}
	
	\caption{We showcase qualitative segmentation results of our model on the CityscapesVideo validation set and compare it with the ground truth. Red points are the ground truth key point given by the annotator for the new objects. }
	\label{fig:results9}
\end{figure*}

\begin{figure*} 
	\centering
	\setlength\tabcolsep{0.5pt}
	\begin{tabular}{cc}

		\raisebox{2px}{{Ours}} &	\raisebox{2px}{{Ground Truth}}  \\
		
		\raisebox{40px}{\rotatebox{90}{t =  0 s}}
		\includegraphics[width=.45\textwidth]{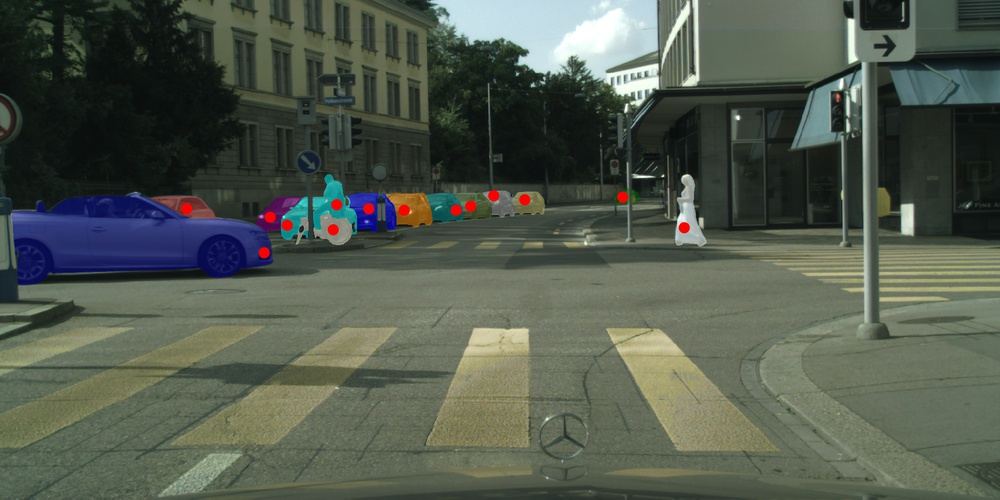} &
		\includegraphics[width=.45\textwidth]{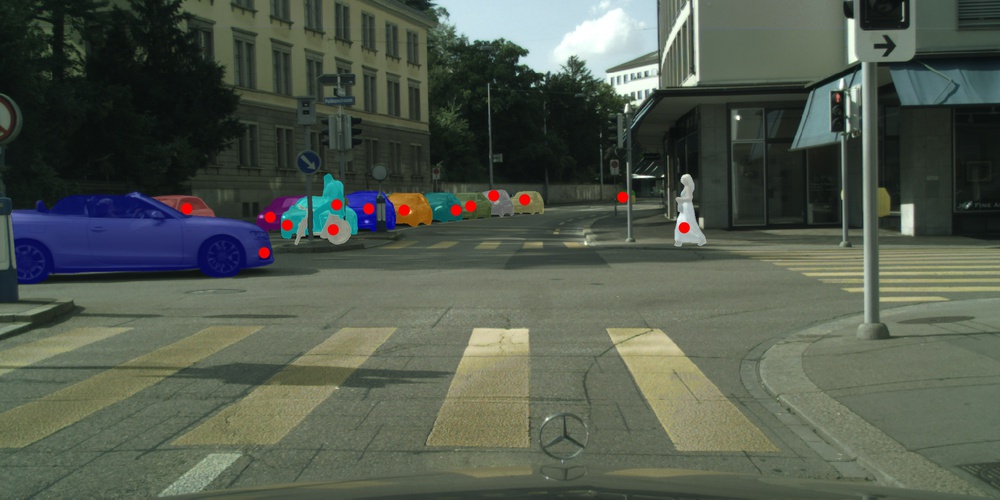} \\
		
		\raisebox{40px}{\rotatebox{90}{t =  0.43 s}}
		\includegraphics[width=.45\textwidth]{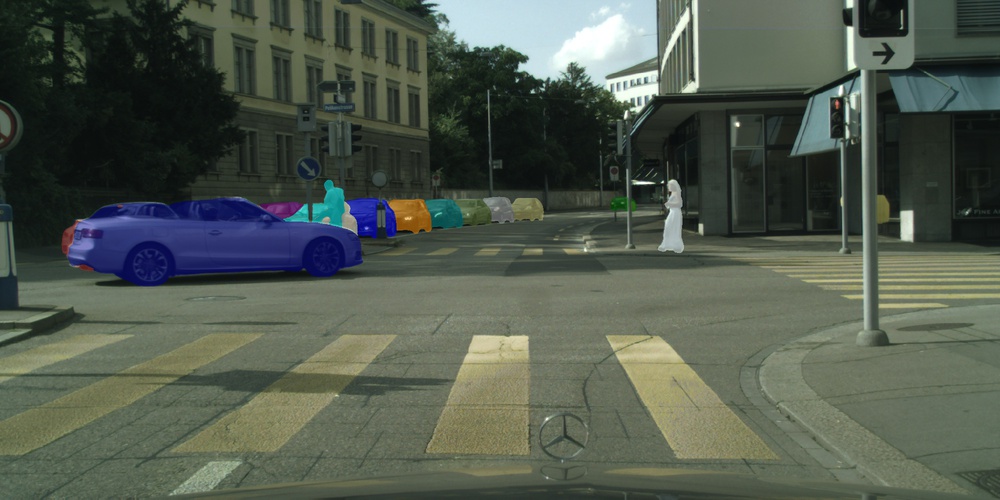} &
		\includegraphics[width=.45\textwidth]{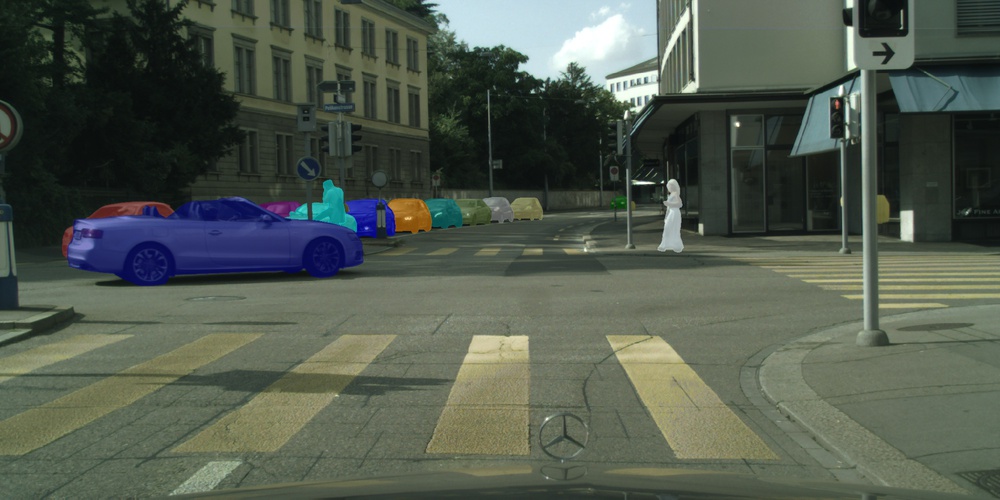} \\
		
		\raisebox{40px}{\rotatebox{90}{t =  0.81 s}}
		\includegraphics[width=.45\textwidth]{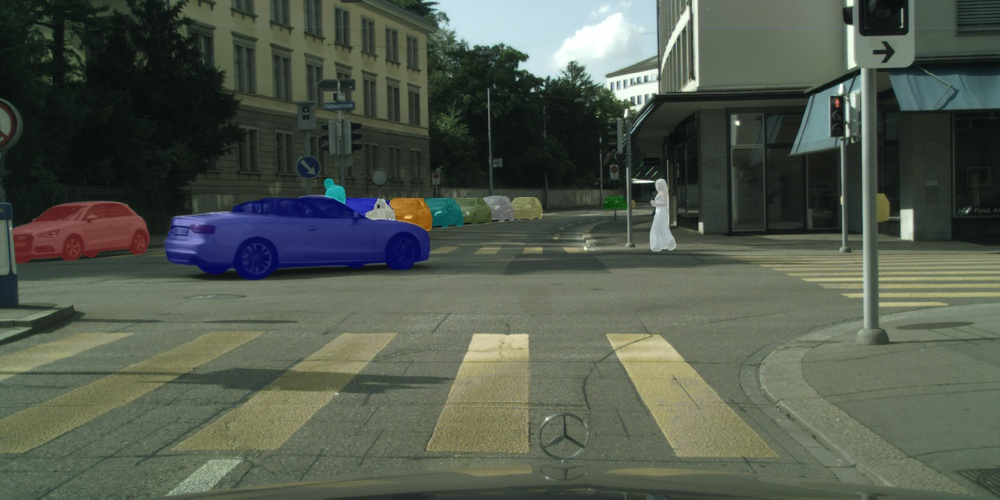} &
		\includegraphics[width=.45\textwidth]{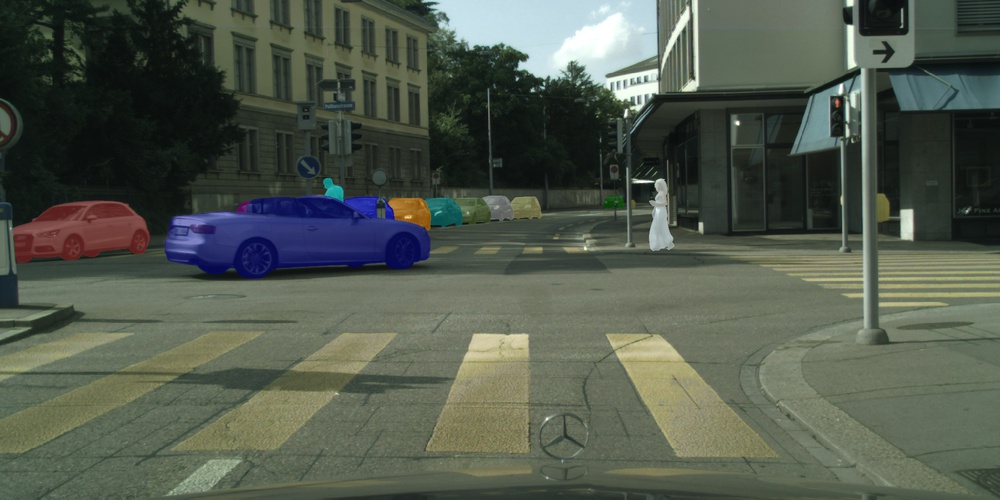} \\
		
		\raisebox{40px}{\rotatebox{90}{t =  1.18 s}}
		\includegraphics[width=.45\textwidth]{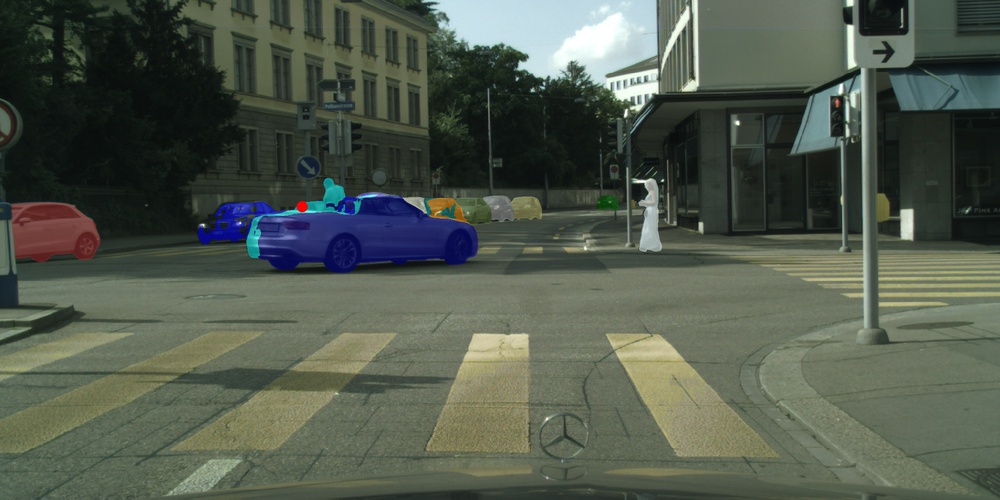} &
		\includegraphics[width=.45\textwidth]{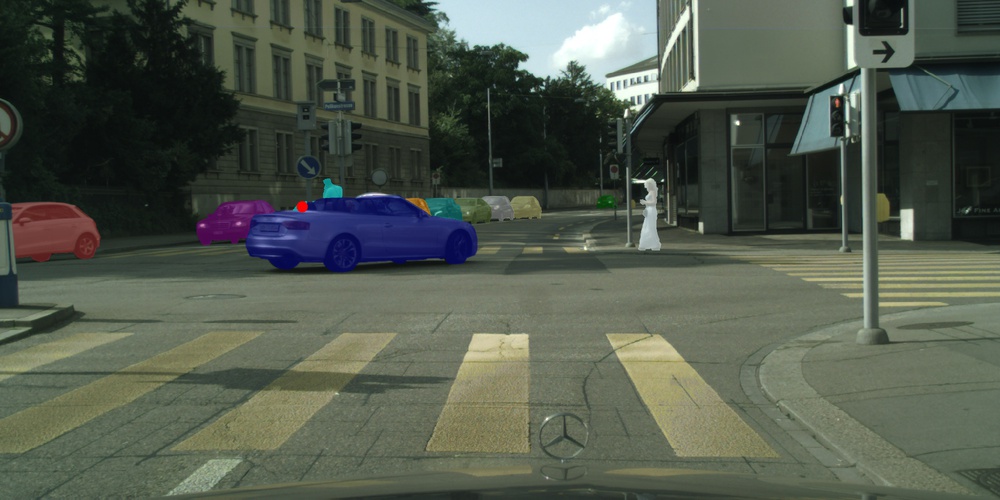} \\
		
		\raisebox{40px}{\rotatebox{90}{t =  1.55 s}}
		\includegraphics[width=.45\textwidth]{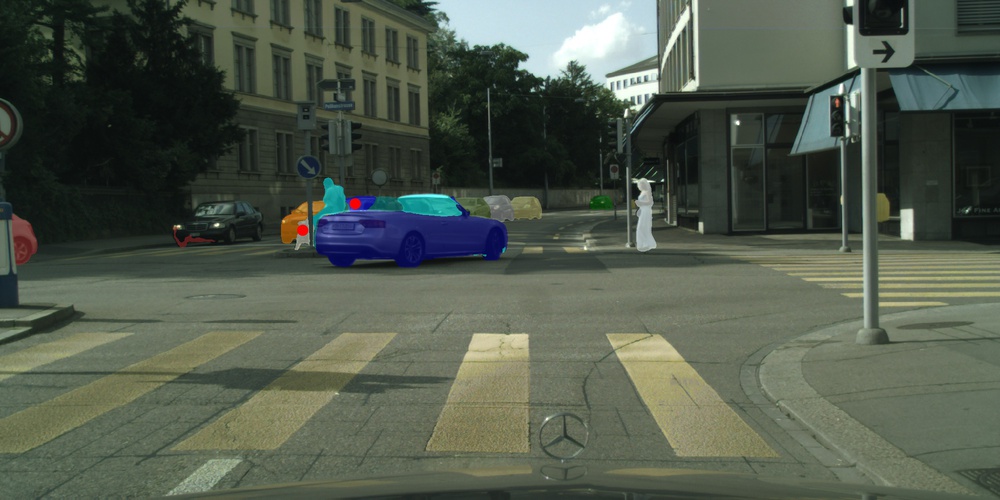} &
		\includegraphics[width=.45\textwidth]{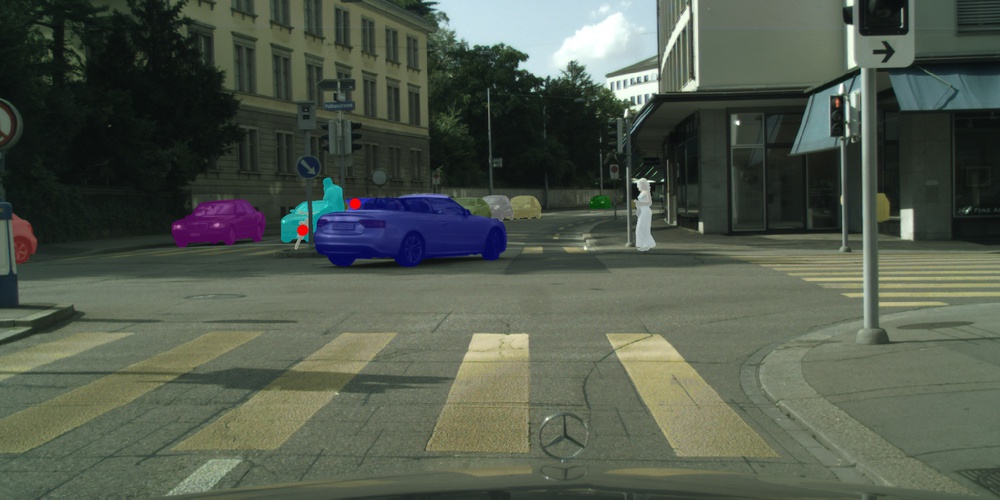} \\
		
	\end{tabular}
	
	\caption{We showcase qualitative segmentation results of our model on the CityscapesVideo validation set and compare it with the ground truth. Red points are the ground truth key point given by the annotator for the new objects. }
	\label{fig:results10}
\end{figure*}

\begin{figure*} 
	\centering
	\setlength\tabcolsep{0.5pt}
	\begin{tabular}{cc}

		\raisebox{2px}{{Ours}} &	\raisebox{2px}{{Ground Truth}}  \\
		
		\raisebox{40px}{\rotatebox{90}{t =  0 s}}
		\includegraphics[width=.45\textwidth]{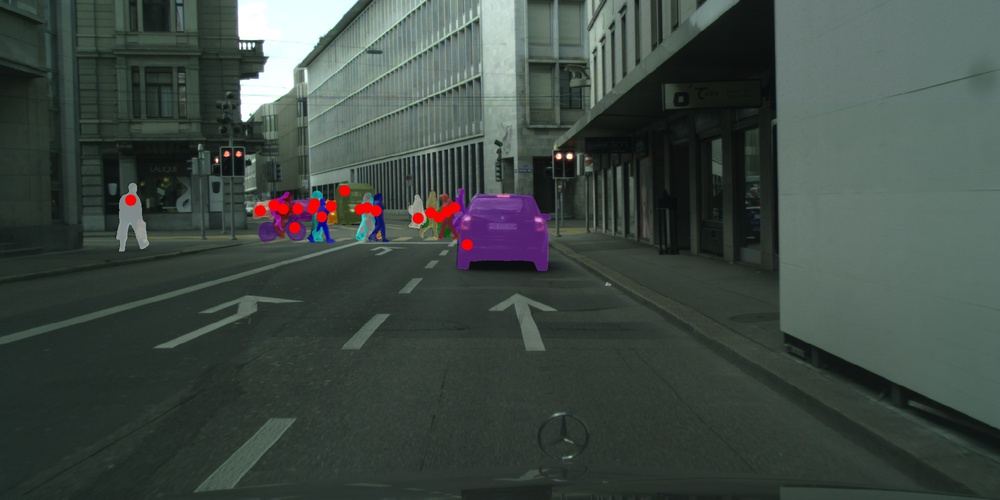} &
		\includegraphics[width=.45\textwidth]{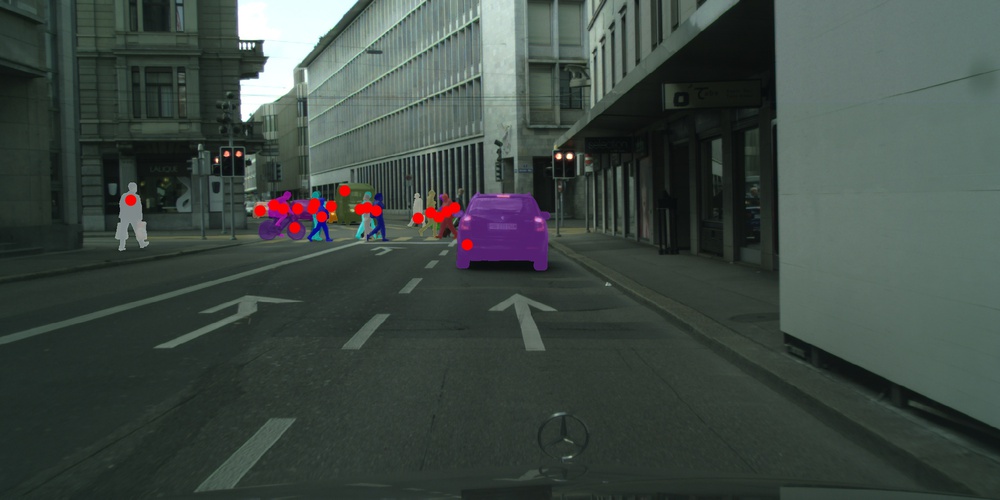} \\
		
		\raisebox{40px}{\rotatebox{90}{t =  0.43 s}}
		\includegraphics[width=.45\textwidth]{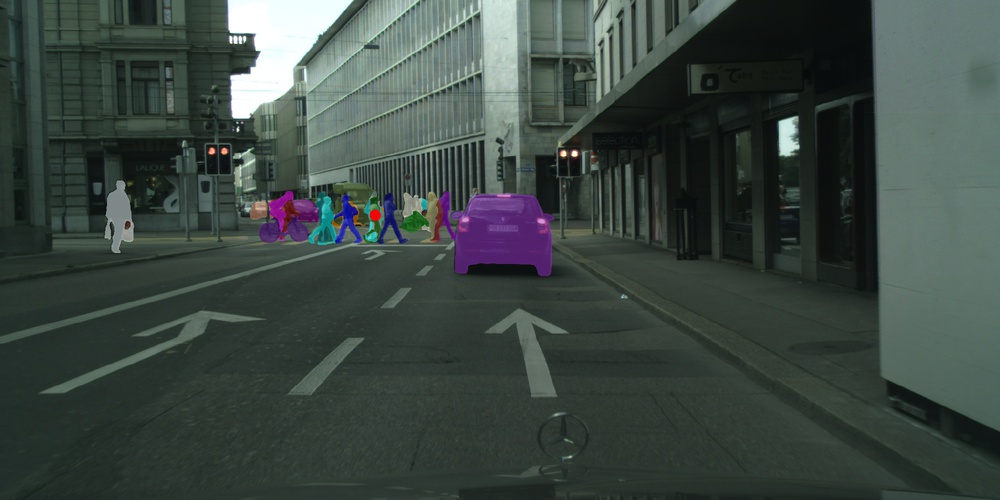} &
		\includegraphics[width=.45\textwidth]{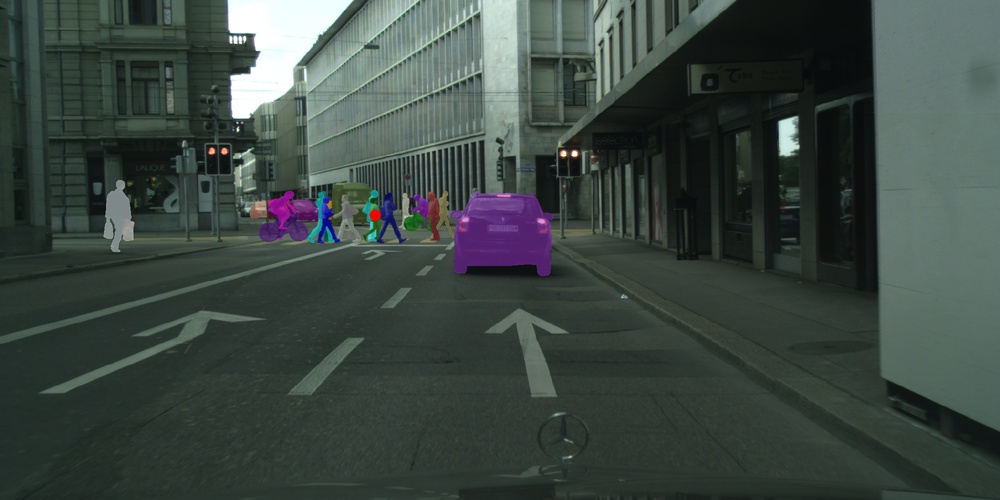} \\
		
		\raisebox{40px}{\rotatebox{90}{t =  0.81 s}}
		\includegraphics[width=.45\textwidth]{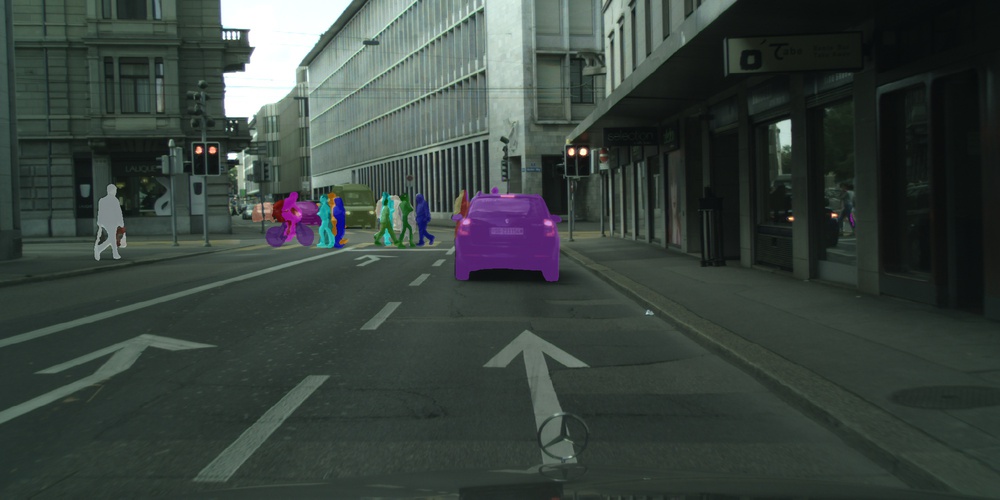} &
		\includegraphics[width=.45\textwidth]{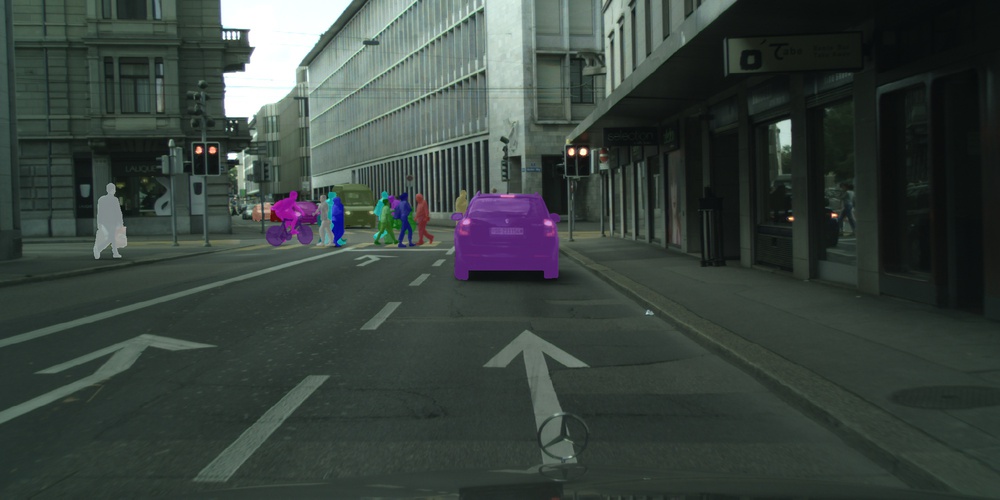} \\
		
		\raisebox{40px}{\rotatebox{90}{t =  1.18 s}}
		\includegraphics[width=.45\textwidth]{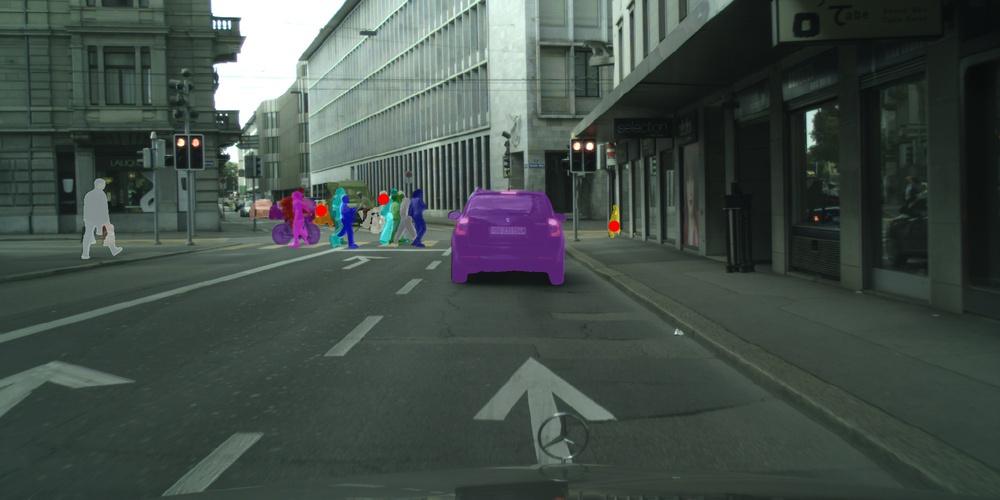} &
		\includegraphics[width=.45\textwidth]{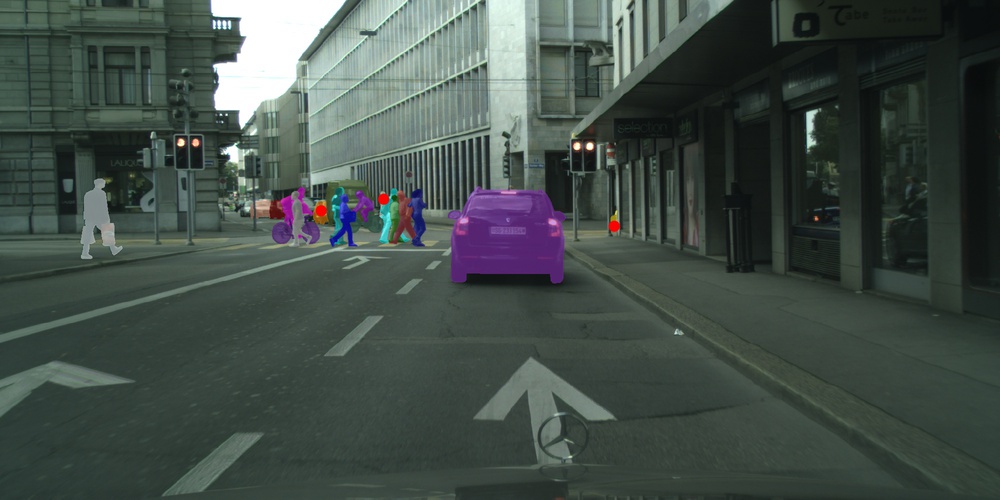} \\
		
		\raisebox{40px}{\rotatebox{90}{t =  1.55 s}}
		\includegraphics[width=.45\textwidth]{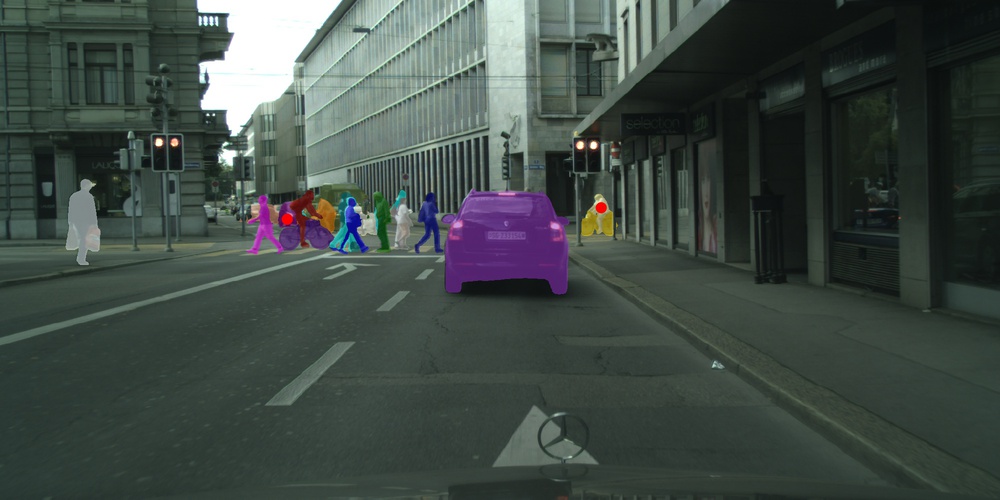} &
		\includegraphics[width=.45\textwidth]{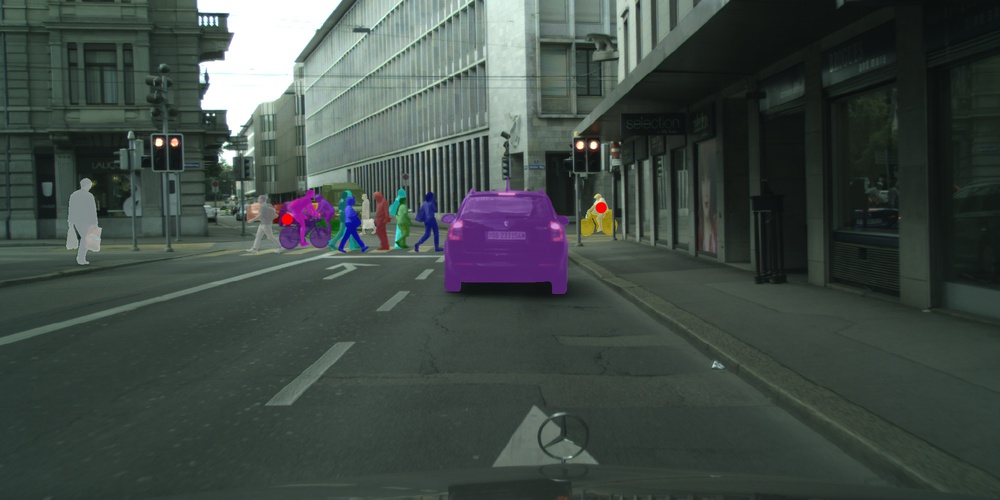} \\
		
	\end{tabular}
	
	\caption{We showcase qualitative segmentation results of our model on the CityscapesVideo validation set and compare it with the ground truth. Red points are the ground truth key point given by the annotator for the new objects. }
	\label{fig:results11}
\end{figure*}

\begin{figure*} 
	\centering
	\setlength\tabcolsep{0.5pt}
	\begin{tabular}{cc}

		\raisebox{2px}{{Ours}} &	\raisebox{2px}{{Ground Truth}}  \\
		
		\raisebox{40px}{\rotatebox{90}{t =  0 s}}
		\includegraphics[width=.45\textwidth]{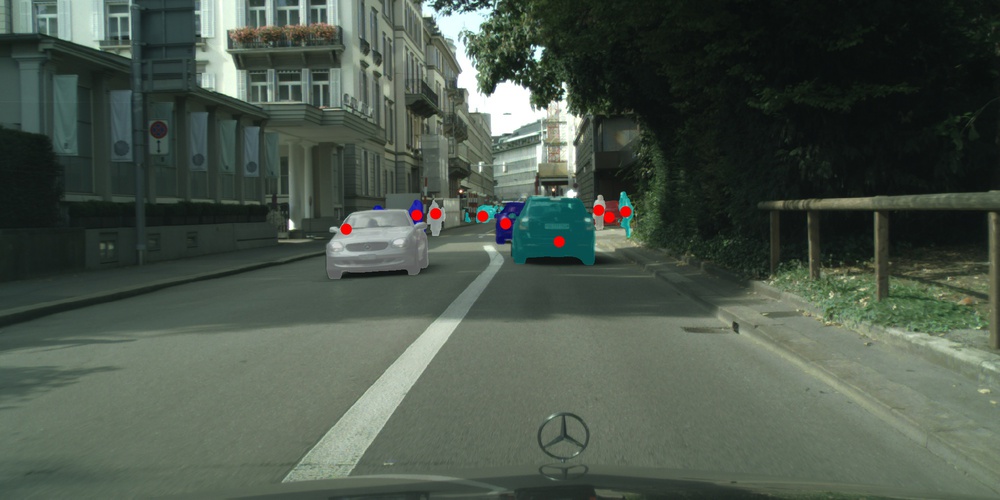} &
		\includegraphics[width=.45\textwidth]{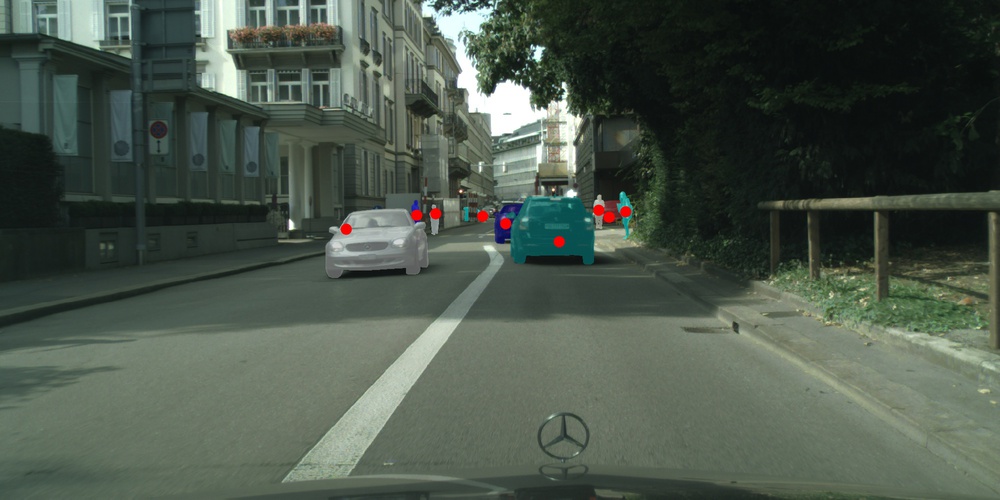} \\
		
		\raisebox{40px}{\rotatebox{90}{t =  0.43 s}}
		\includegraphics[width=.45\textwidth]{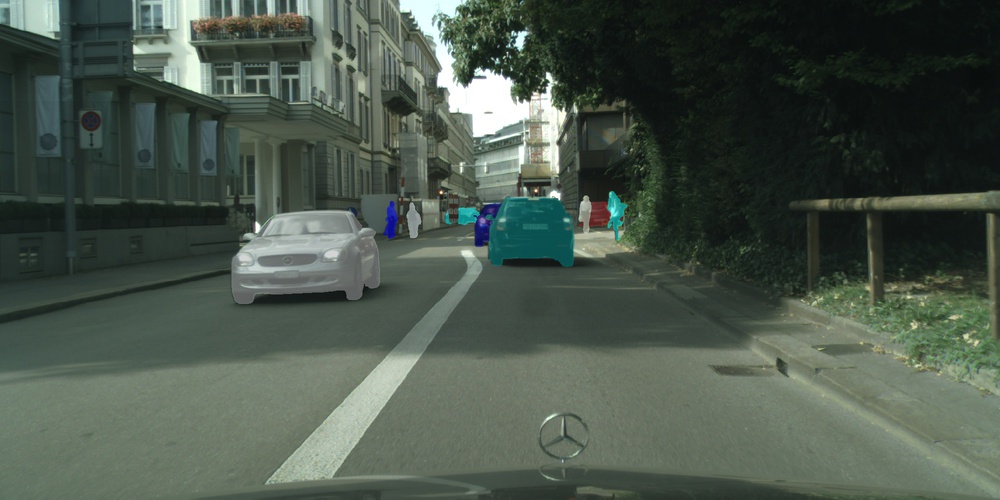} &
		\includegraphics[width=.45\textwidth]{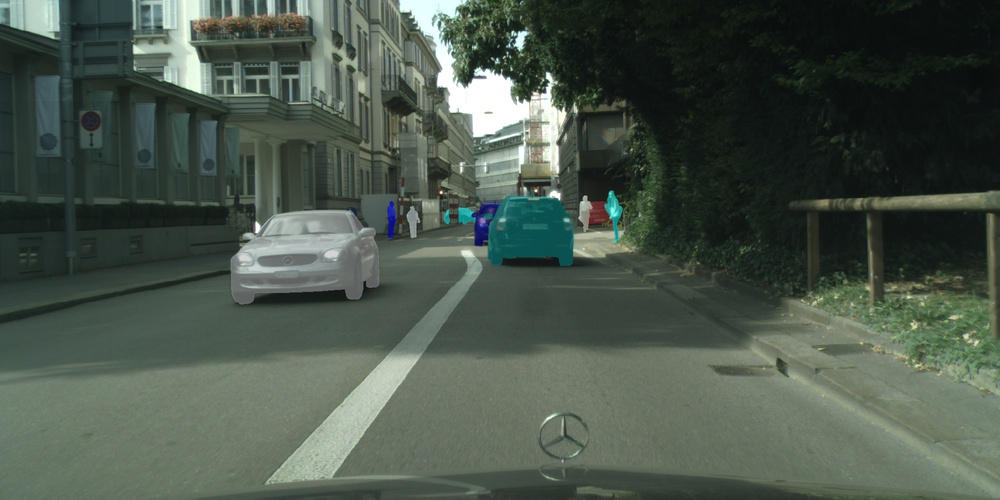} \\
		
		\raisebox{40px}{\rotatebox{90}{t =  0.81 s}}
		\includegraphics[width=.45\textwidth]{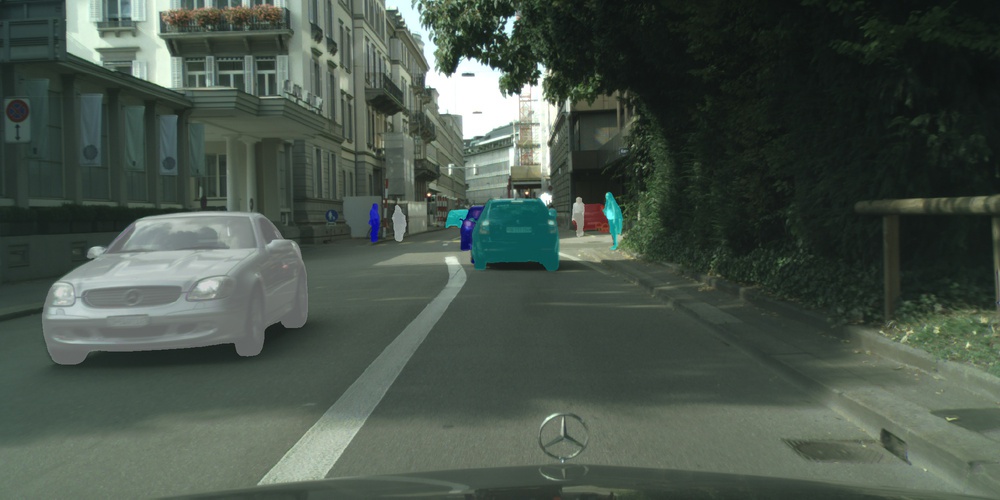} &
		\includegraphics[width=.45\textwidth]{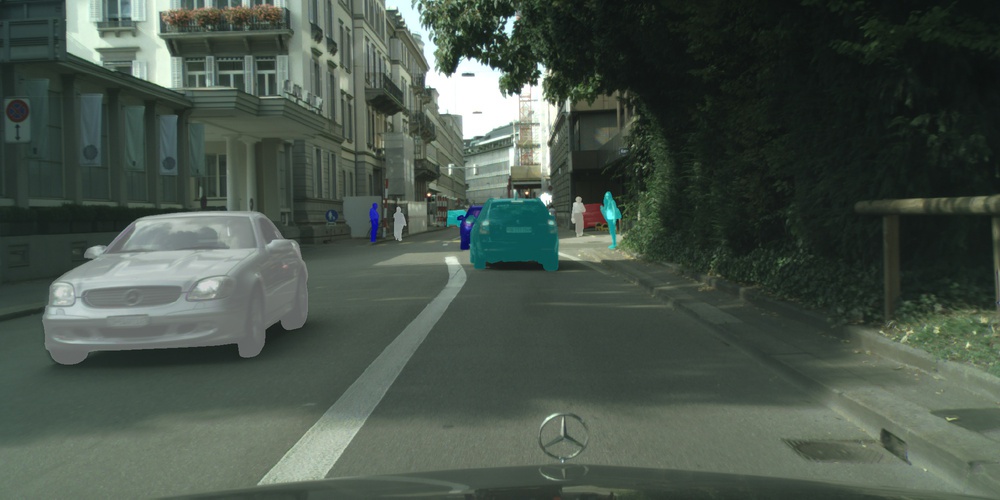} \\
		
		\raisebox{40px}{\rotatebox{90}{t =  1.18 s}}
		\includegraphics[width=.45\textwidth]{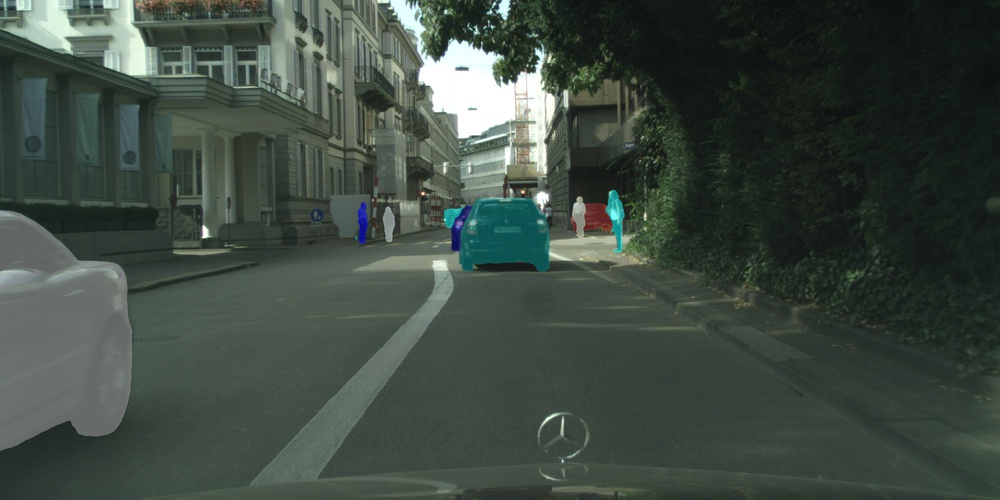} &
		\includegraphics[width=.45\textwidth]{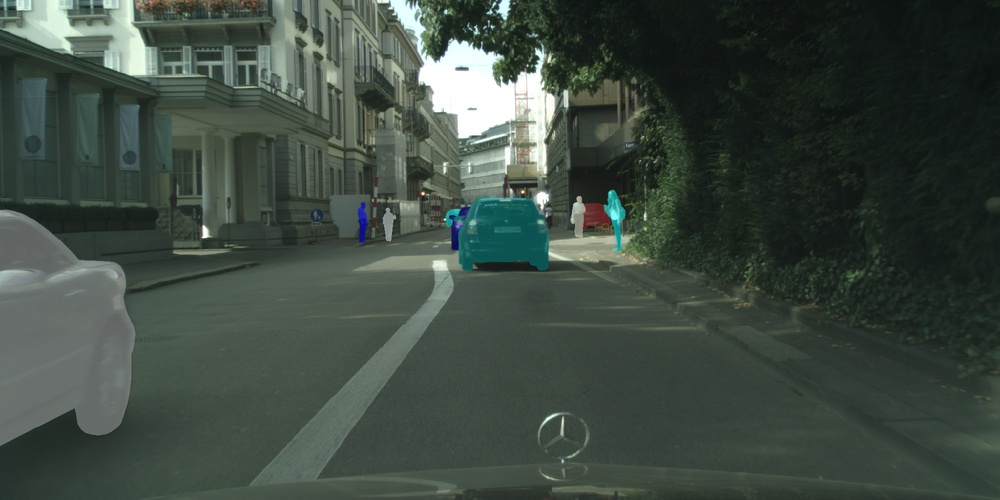} \\
		
		\raisebox{40px}{\rotatebox{90}{t =  1.55 s}}
		\includegraphics[width=.45\textwidth]{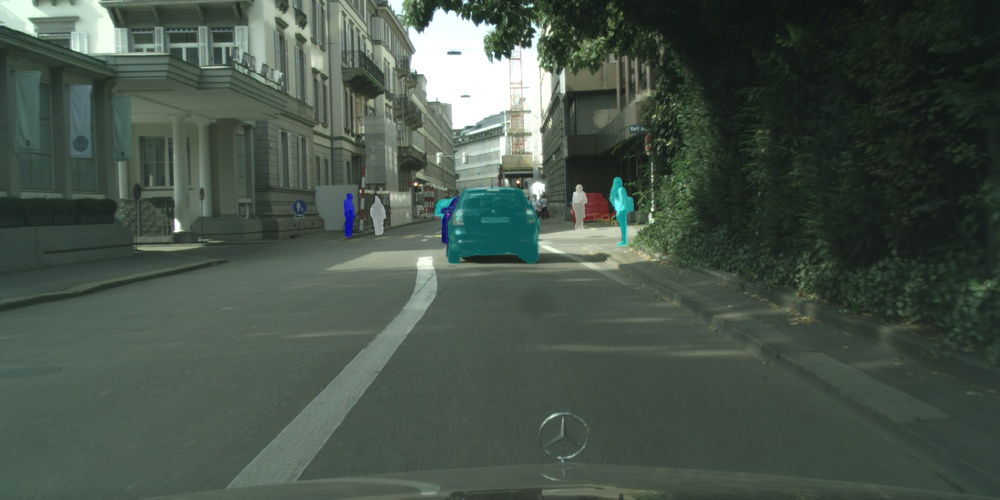} &
		\includegraphics[width=.45\textwidth]{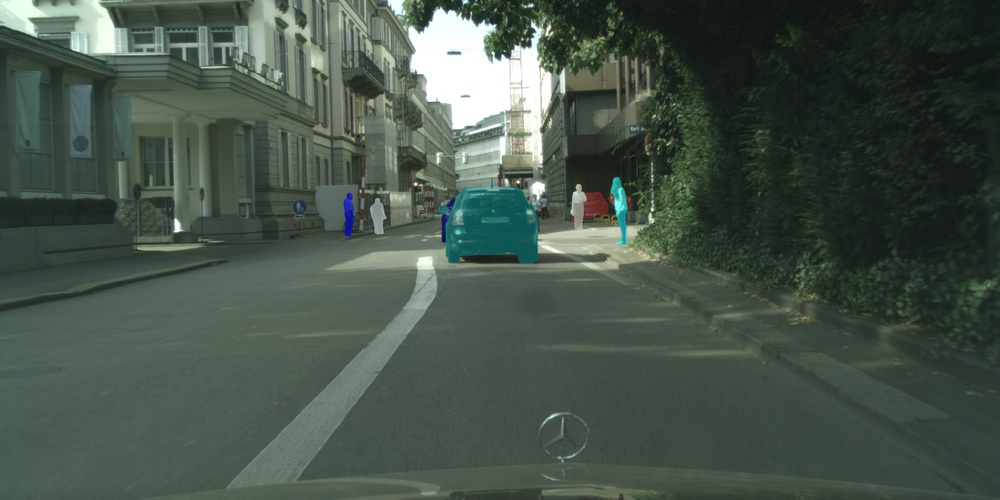} \\
		
	\end{tabular}
	
	\caption{We showcase qualitative segmentation results of our model on the CityscapesVideo validation set and compare it with the ground truth. Red points are the ground truth key point given by the annotator for the new objects. }
	\label{fig:results12}
\end{figure*}

\begin{figure*} 
	\centering
	\setlength\tabcolsep{0.5pt}
	\begin{tabular}{cc}

		\raisebox{2px}{{Ours}} &	\raisebox{2px}{{Ground Truth}}  \\
		
		\raisebox{40px}{\rotatebox{90}{t =  0 s}}
		\includegraphics[width=.45\textwidth]{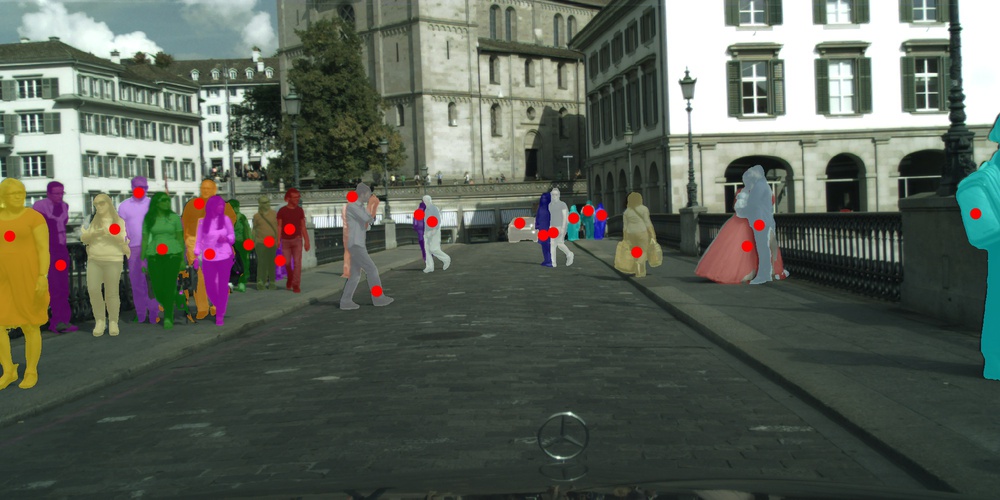} &
		\includegraphics[width=.45\textwidth]{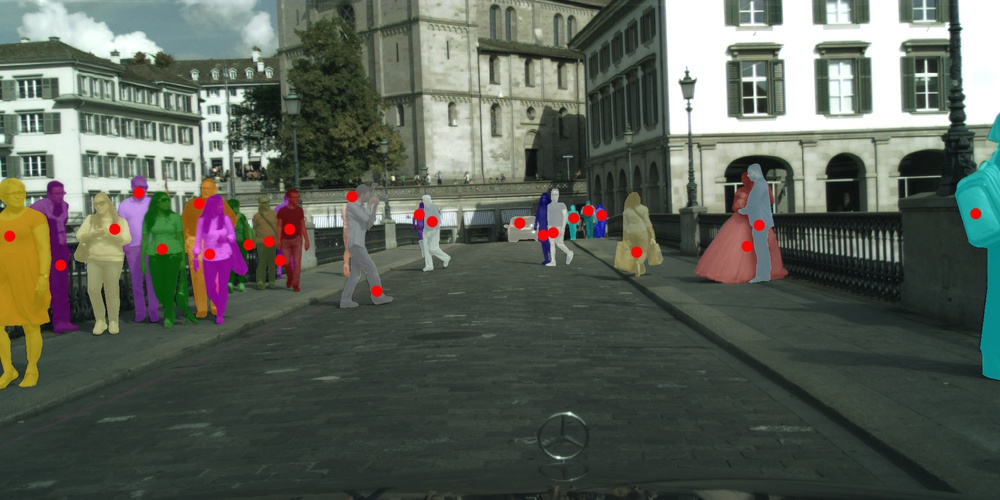} \\
		
		\raisebox{40px}{\rotatebox{90}{t =  0.43 s}}
		\includegraphics[width=.45\textwidth]{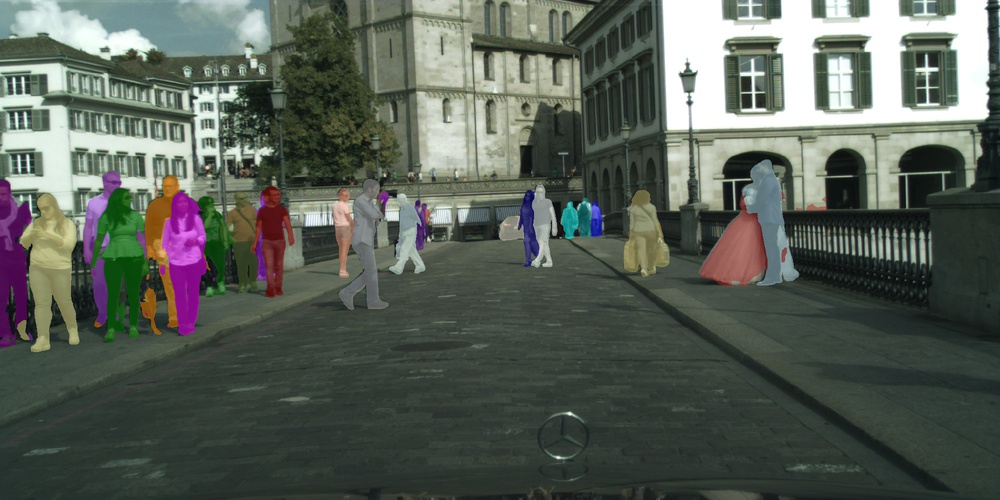} &
		\includegraphics[width=.45\textwidth]{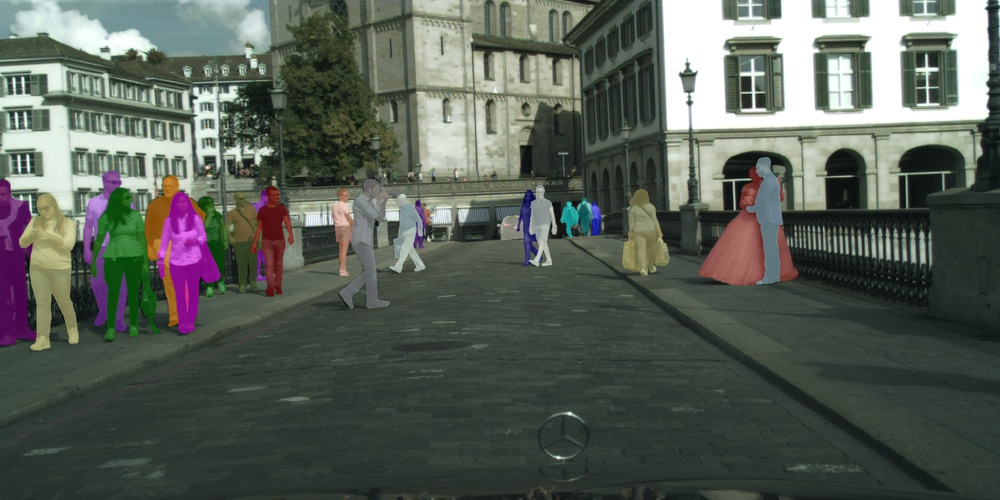} \\
		
		\raisebox{40px}{\rotatebox{90}{t =  0.81 s}}
		\includegraphics[width=.45\textwidth]{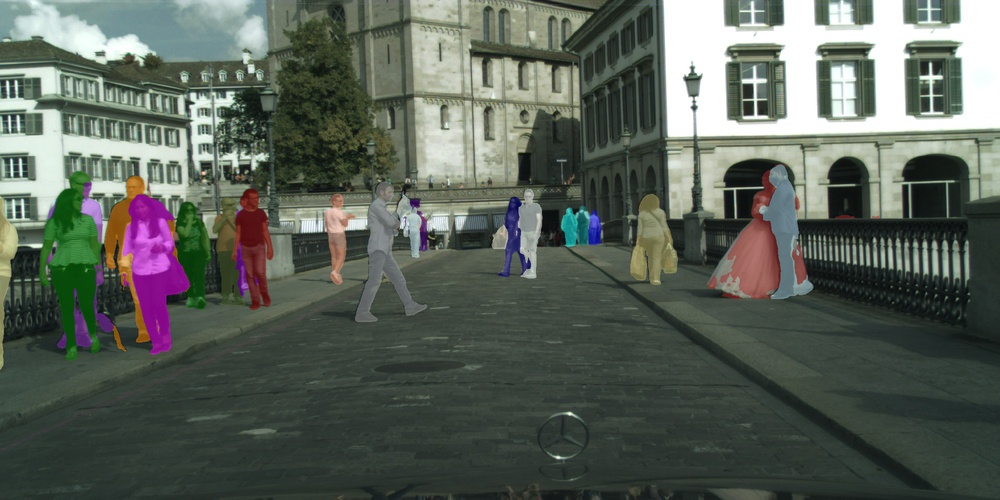} &
		\includegraphics[width=.45\textwidth]{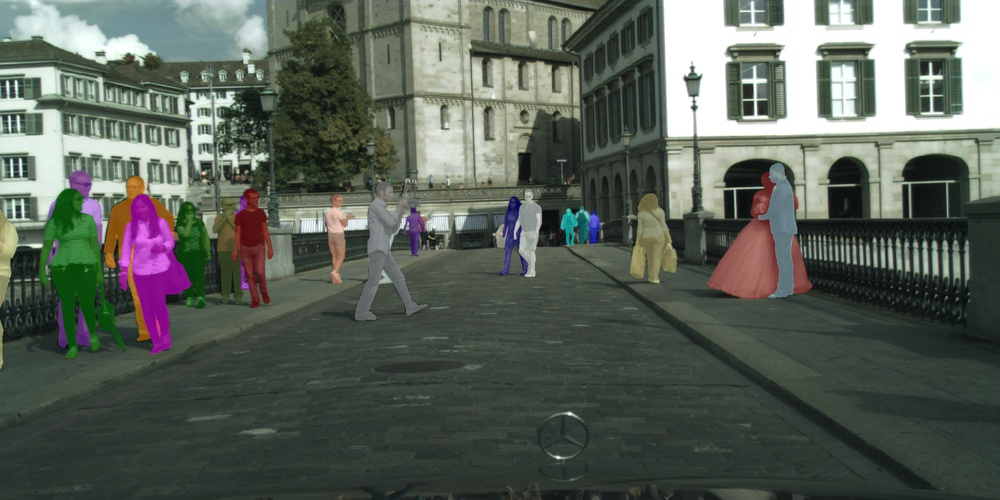} \\
		
		\raisebox{40px}{\rotatebox{90}{t =  1.18 s}}
		\includegraphics[width=.45\textwidth]{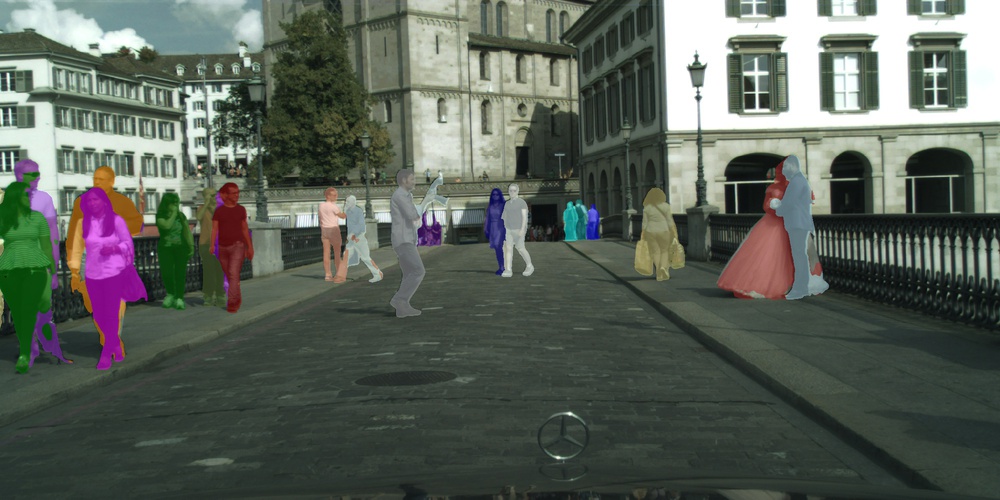} &
		\includegraphics[width=.45\textwidth]{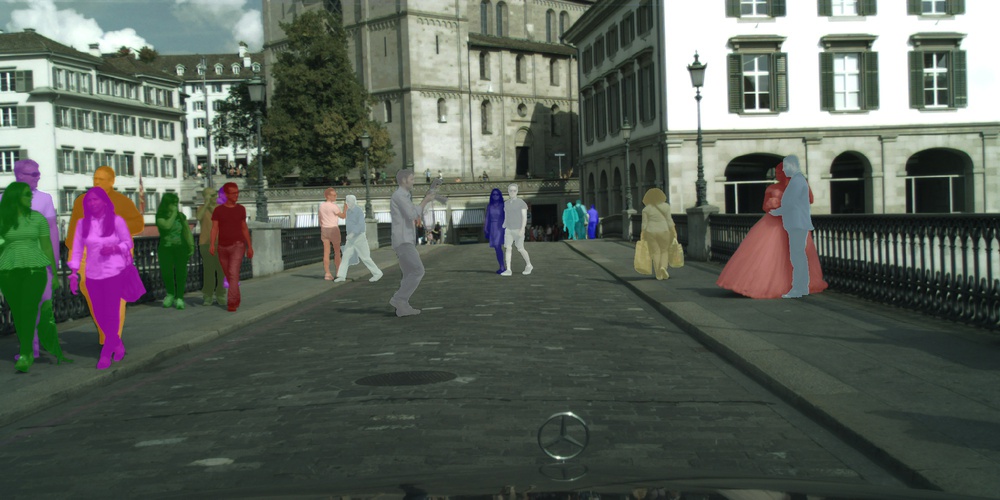} \\
		
		\raisebox{40px}{\rotatebox{90}{t =  1.55 s}}
		\includegraphics[width=.45\textwidth]{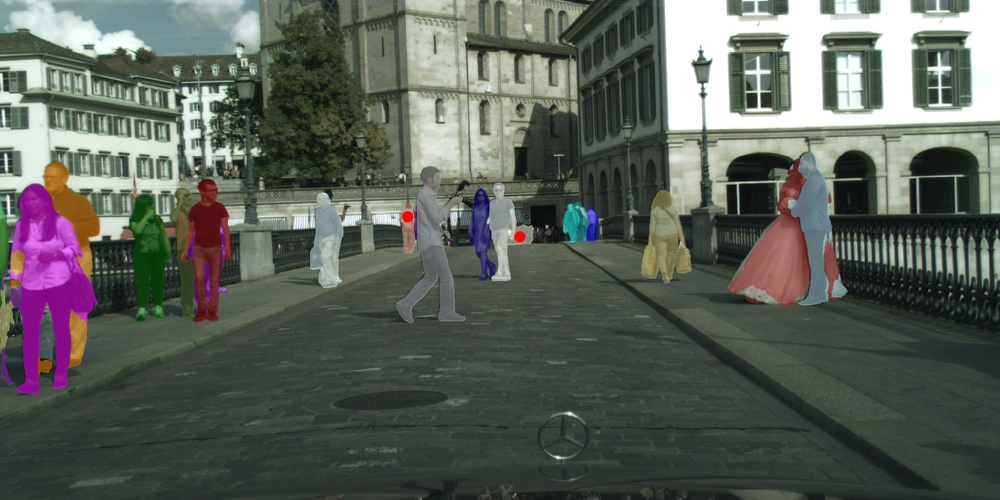} &
		\includegraphics[width=.45\textwidth]{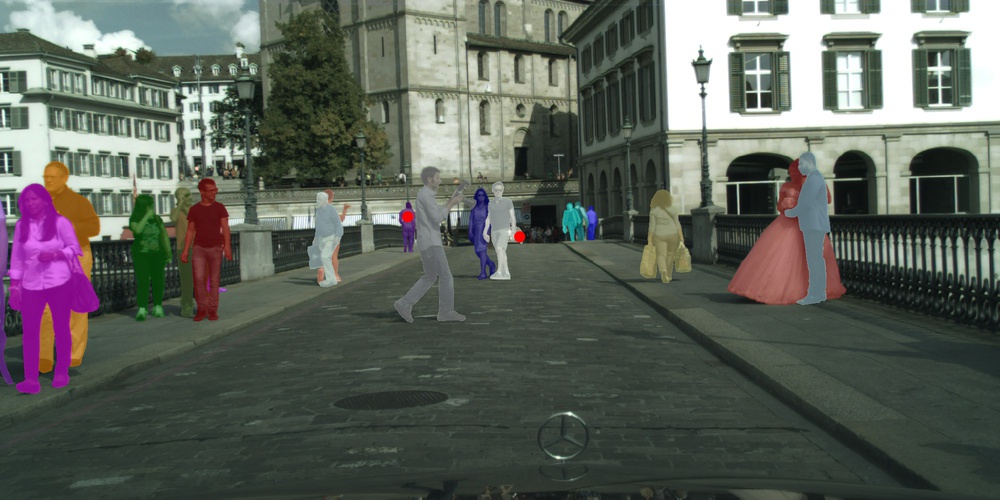} \\
		
	\end{tabular}
	
	\caption{We showcase qualitative segmentation results of our model on the CityscapesVideo validation set and compare it with the ground truth. Red points are the ground truth key point given by the annotator for the new objects. }
	\label{fig:results13}
\end{figure*}

\begin{figure*} 
	\centering
	\setlength\tabcolsep{0.5pt}
	\begin{tabular}{cc}

		\raisebox{2px}{{Ours}} &	\raisebox{2px}{{Ground Truth}}  \\
		
		\raisebox{40px}{\rotatebox{90}{t =  0 s}}
		\includegraphics[width=.45\textwidth]{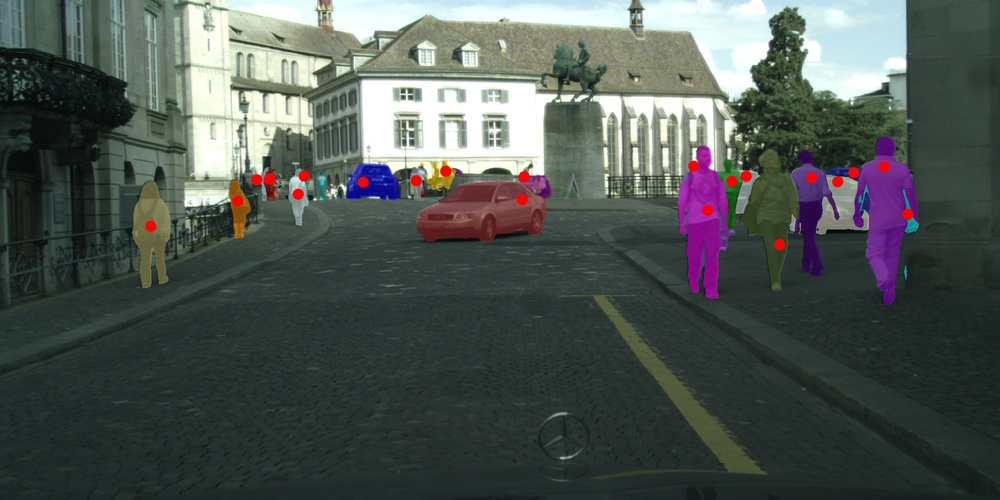} &
		\includegraphics[width=.45\textwidth]{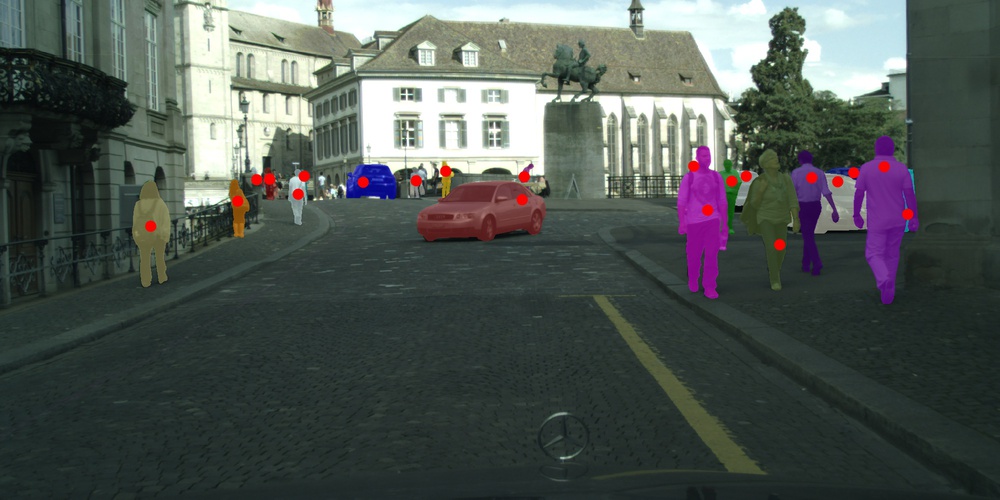} \\
		
		\raisebox{40px}{\rotatebox{90}{t =  0.43 s}}
		\includegraphics[width=.45\textwidth]{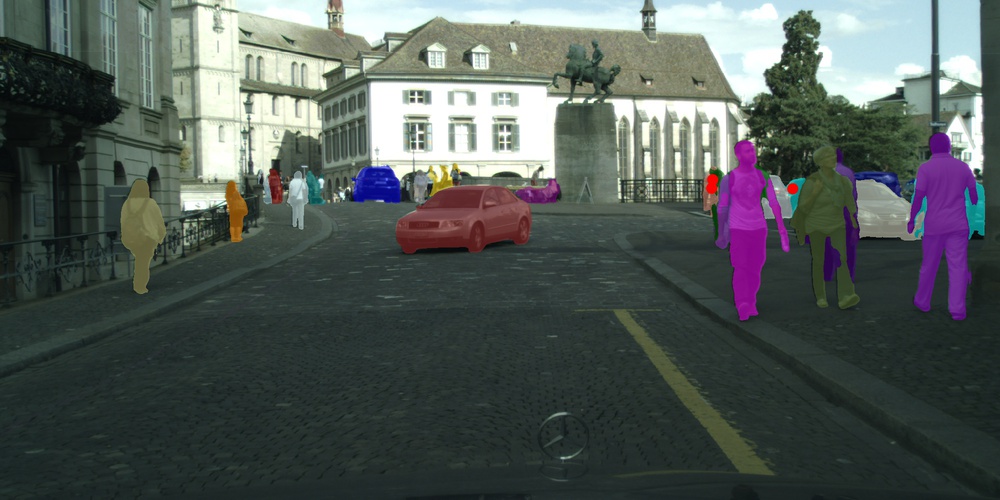} &
		\includegraphics[width=.45\textwidth]{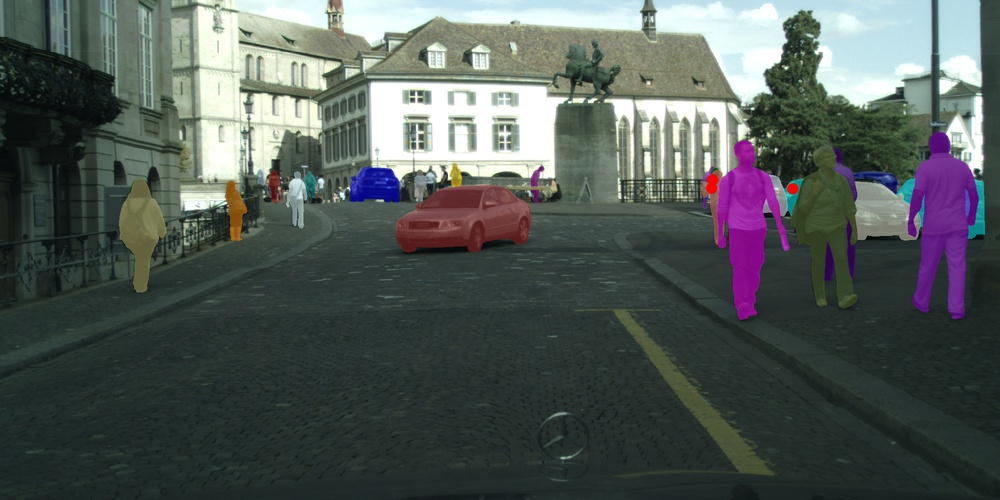} \\
		
		\raisebox{40px}{\rotatebox{90}{t =  0.81 s}}
		\includegraphics[width=.45\textwidth]{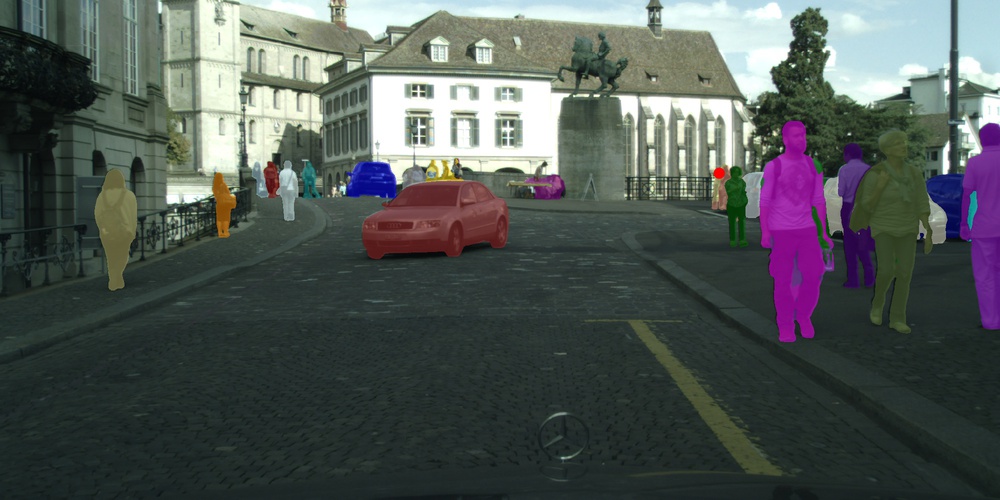} &
		\includegraphics[width=.45\textwidth]{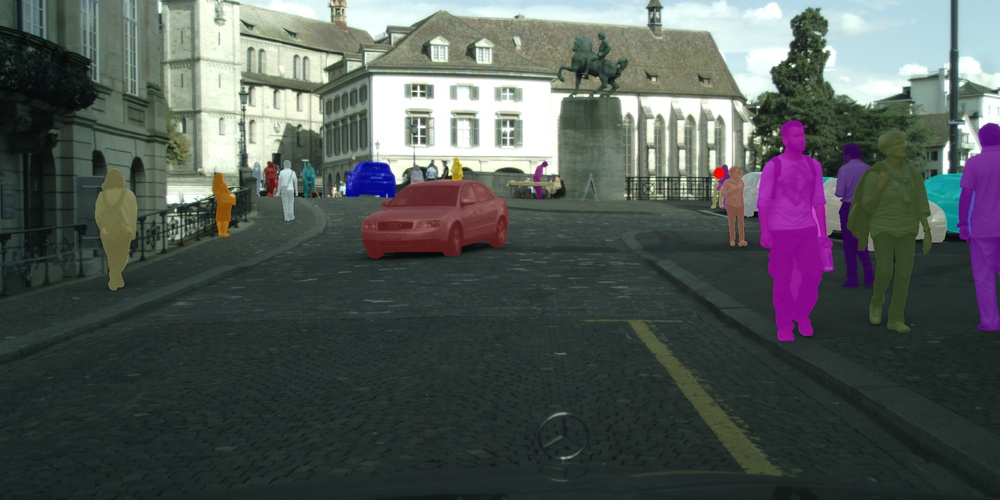} \\
		
		\raisebox{40px}{\rotatebox{90}{t =  1.18 s}}
		\includegraphics[width=.45\textwidth]{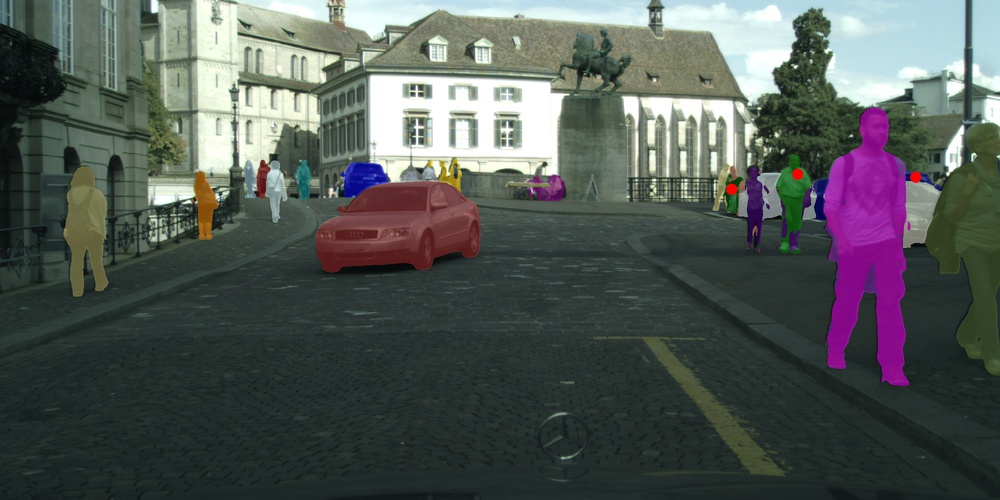} &
		\includegraphics[width=.45\textwidth]{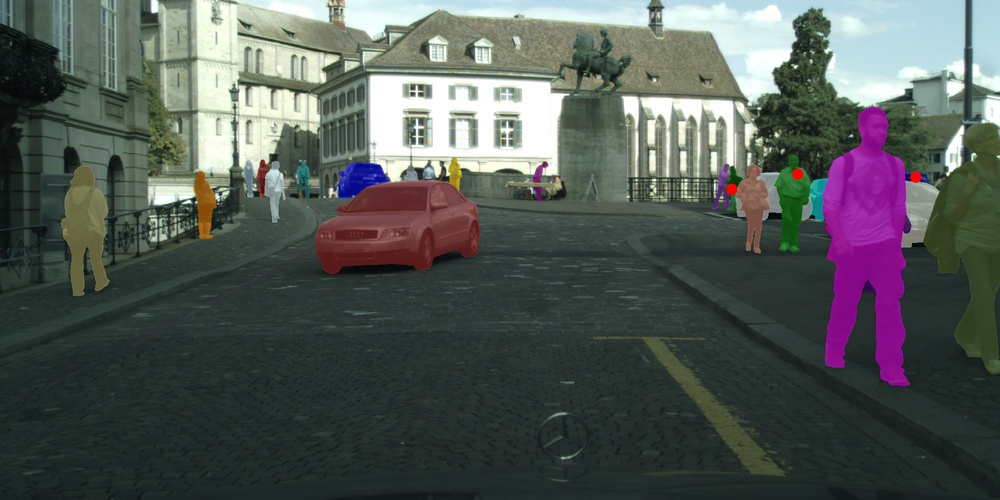} \\
		
		\raisebox{40px}{\rotatebox{90}{t =  1.55 s}}
		\includegraphics[width=.45\textwidth]{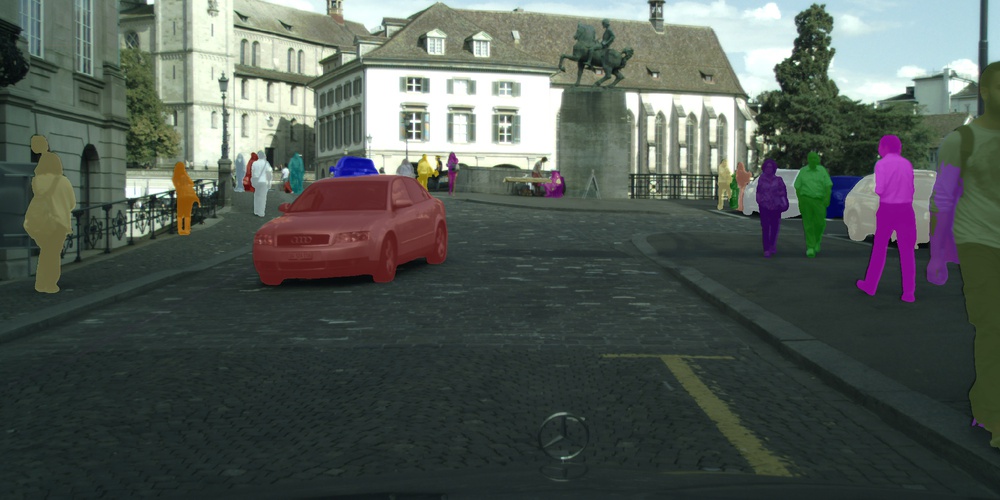} &
		\includegraphics[width=.45\textwidth]{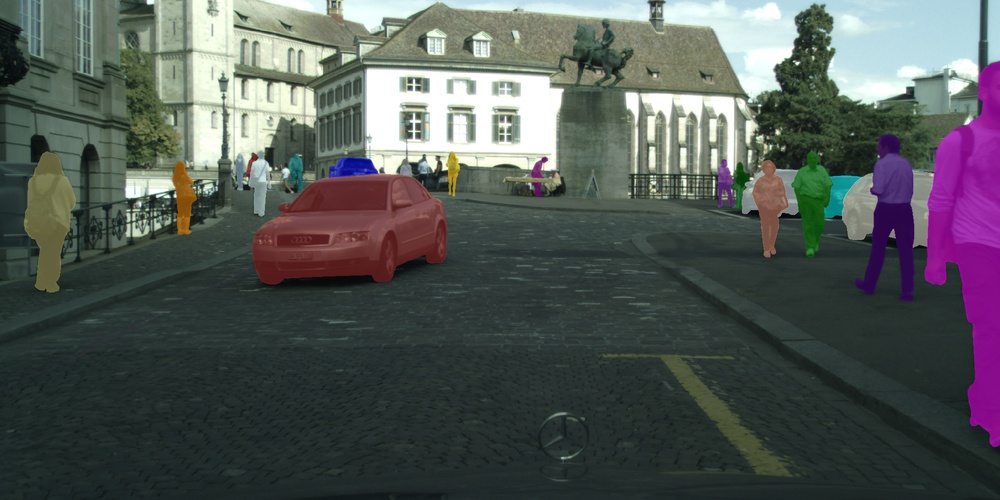} \\
		
	\end{tabular}
	
	\caption{We showcase qualitative segmentation results of our model on the CityscapesVideo validation set and compare it with the ground truth. Red points are the ground truth key point given by the annotator for the new objects. }
	\label{fig:results14}
\end{figure*}

\begin{figure*} 
	\centering
	\setlength\tabcolsep{0.5pt}
	\begin{tabular}{cc}

		\raisebox{2px}{{Ours}} &	\raisebox{2px}{{Ground Truth}}  \\
		
		\raisebox{40px}{\rotatebox{90}{t =  0 s}}
		\includegraphics[width=.45\textwidth]{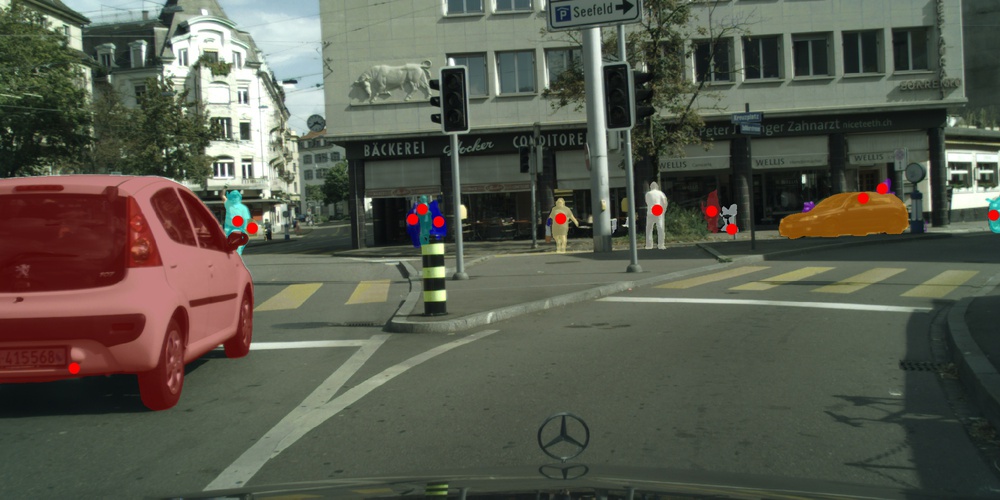} &
		\includegraphics[width=.45\textwidth]{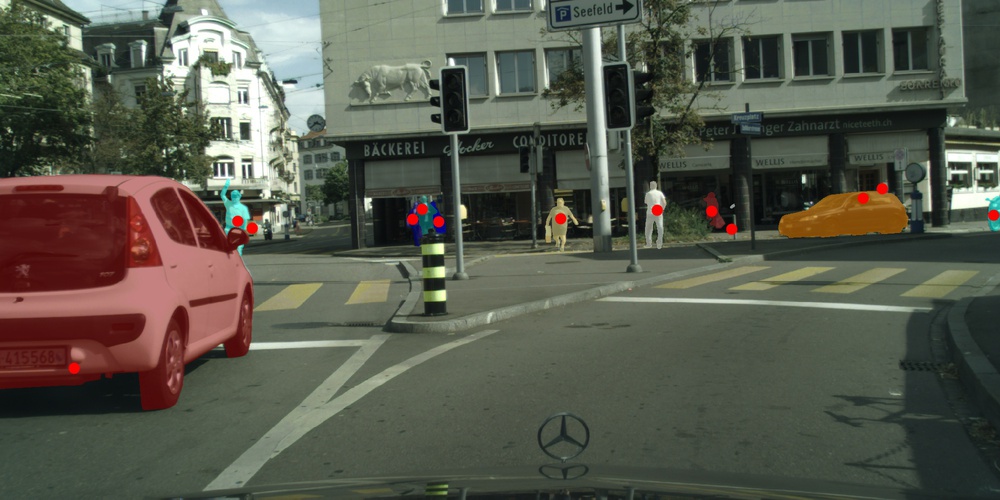} \\
		
		\raisebox{40px}{\rotatebox{90}{t =  0.43 s}}
		\includegraphics[width=.45\textwidth]{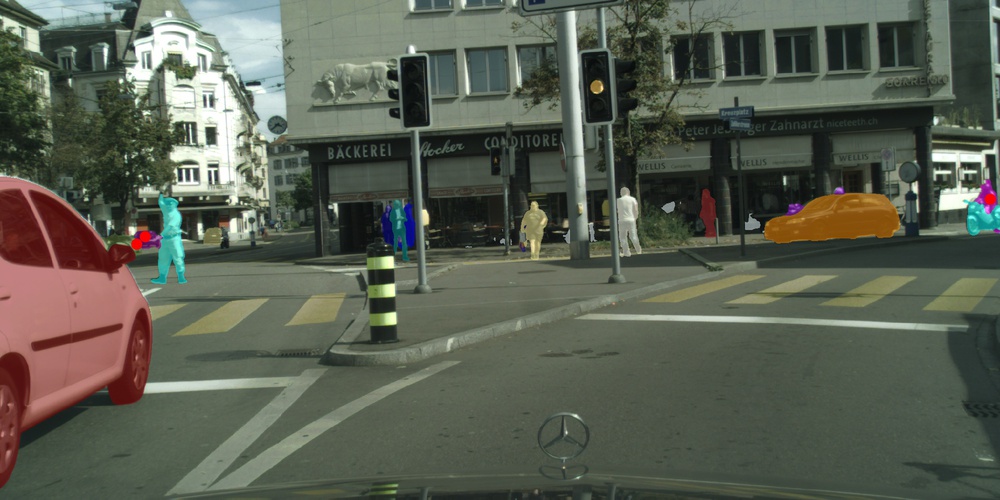} &
		\includegraphics[width=.45\textwidth]{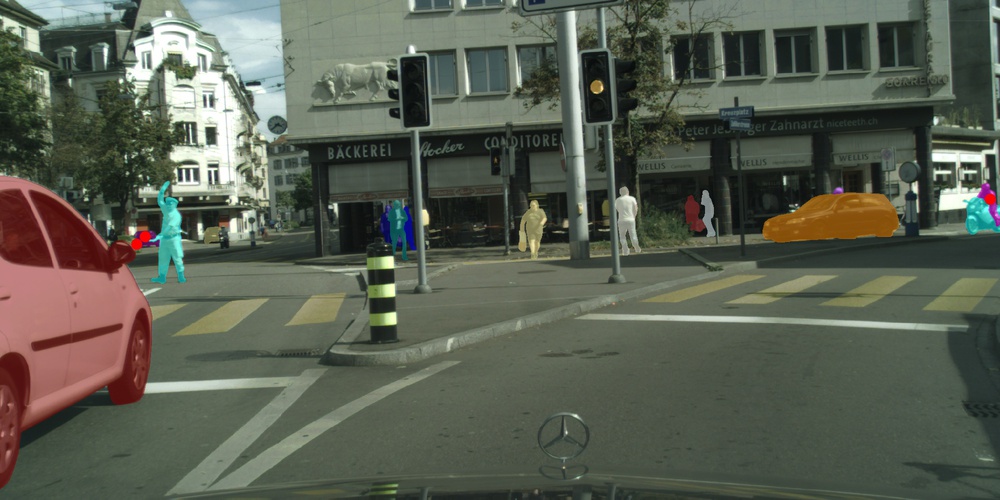} \\
		
		\raisebox{40px}{\rotatebox{90}{t =  0.81 s}}
		\includegraphics[width=.45\textwidth]{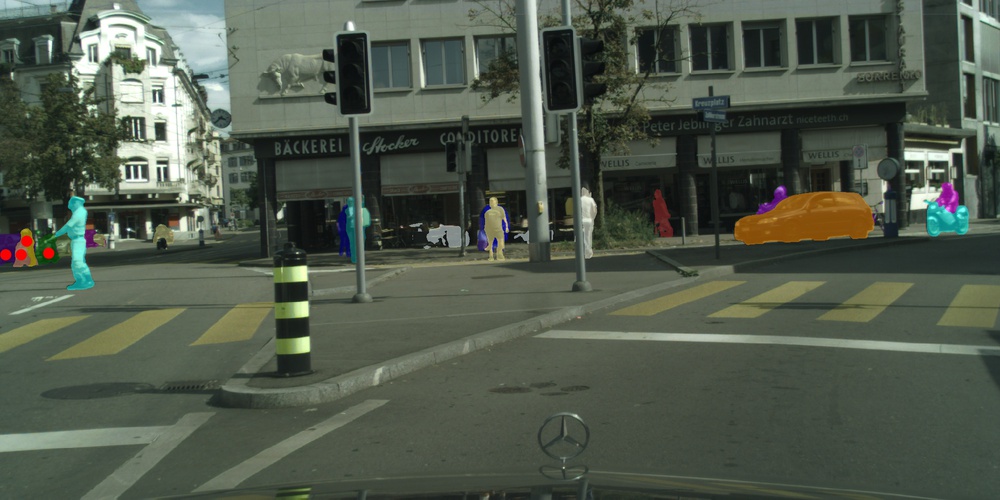} &
		\includegraphics[width=.45\textwidth]{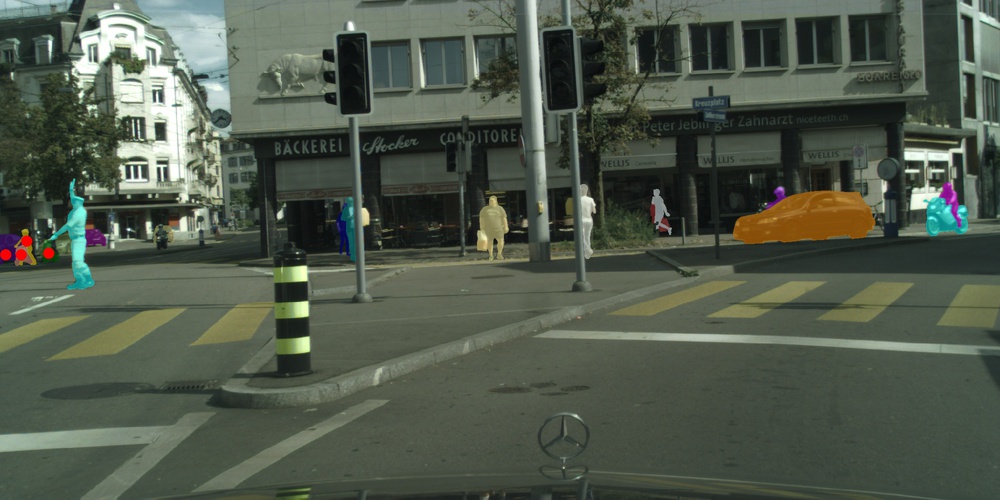} \\
		
		\raisebox{40px}{\rotatebox{90}{t =  1.18 s}}
		\includegraphics[width=.45\textwidth]{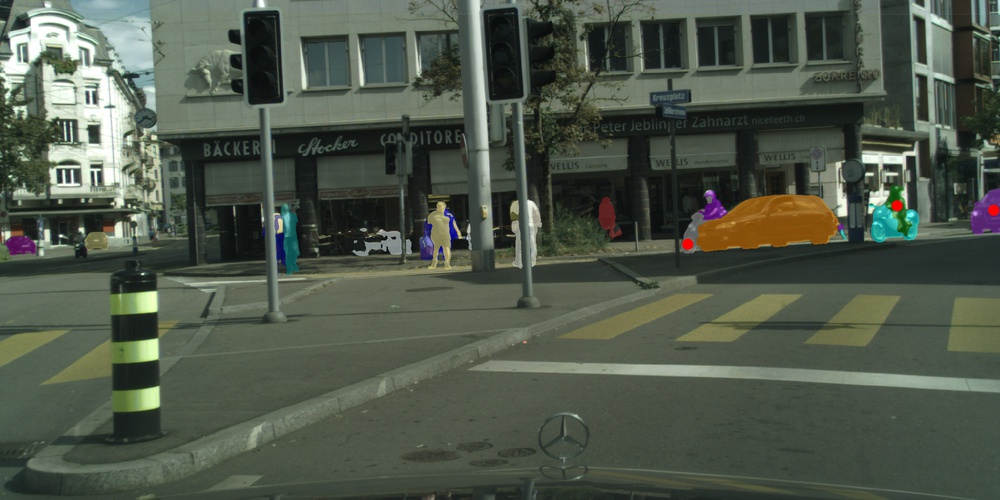} &
		\includegraphics[width=.45\textwidth]{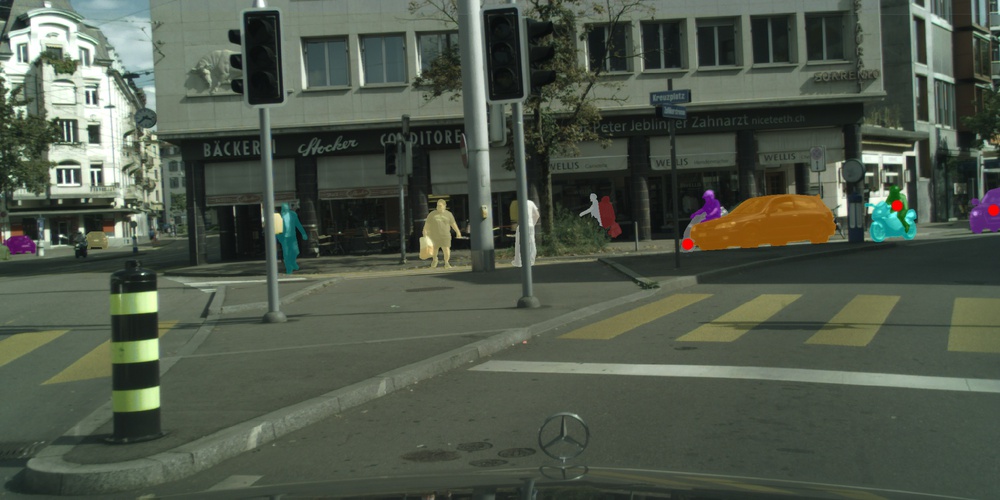} \\
		
		\raisebox{40px}{\rotatebox{90}{t =  1.55 s}}
		\includegraphics[width=.45\textwidth]{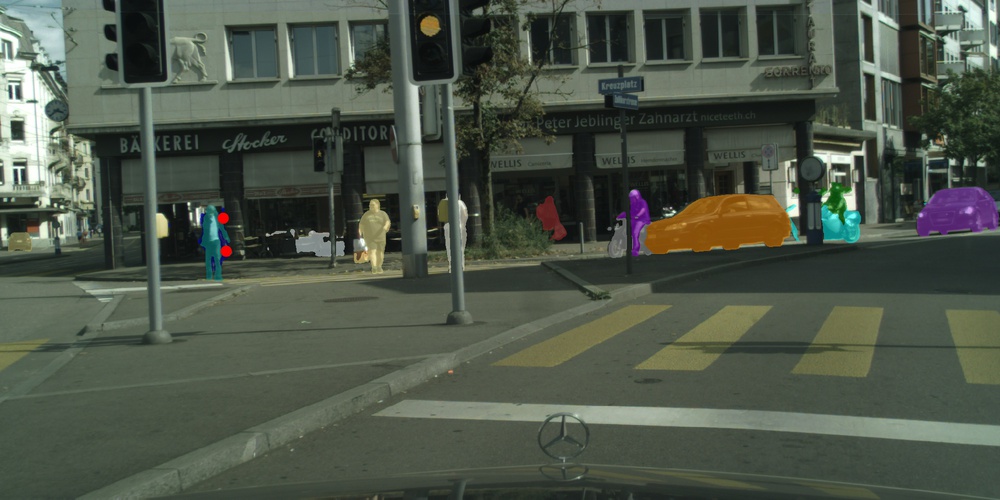} &
		\includegraphics[width=.45\textwidth]{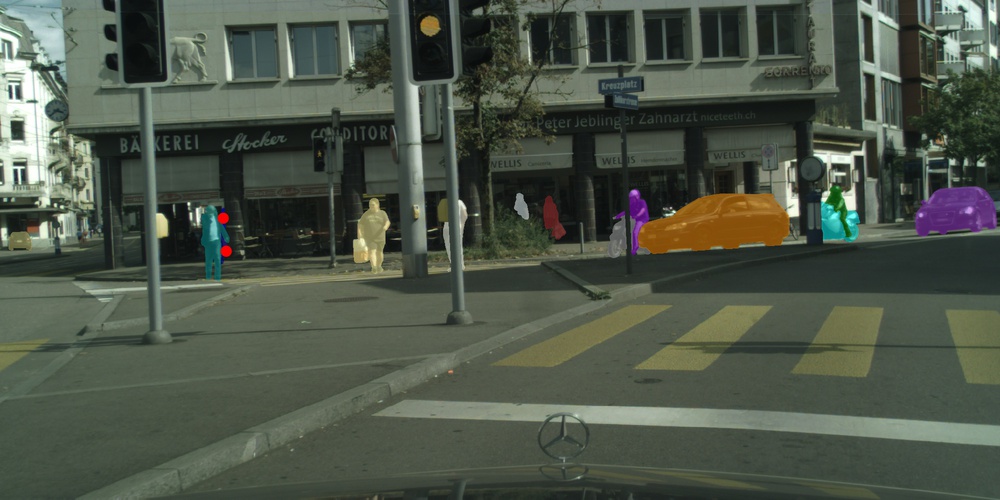} \\
		
	\end{tabular}
	
	\caption{We showcase qualitative segmentation results of our model on the CityscapesVideo validation set and compare it with the ground truth. Red points are the ground truth key point given by the annotator for the new objects. }
	\label{fig:results15}
\end{figure*}

\begin{figure*} 
	\centering
	\setlength\tabcolsep{0.5pt}
	\begin{tabular}{cc}
		

		
		\raisebox{2px}{{Ours}} &	\raisebox{2px}{{Ground Truth}}  \\
		
		\raisebox{40px}{\rotatebox{90}{t =  0 s}}
		\includegraphics[width=.45\textwidth]{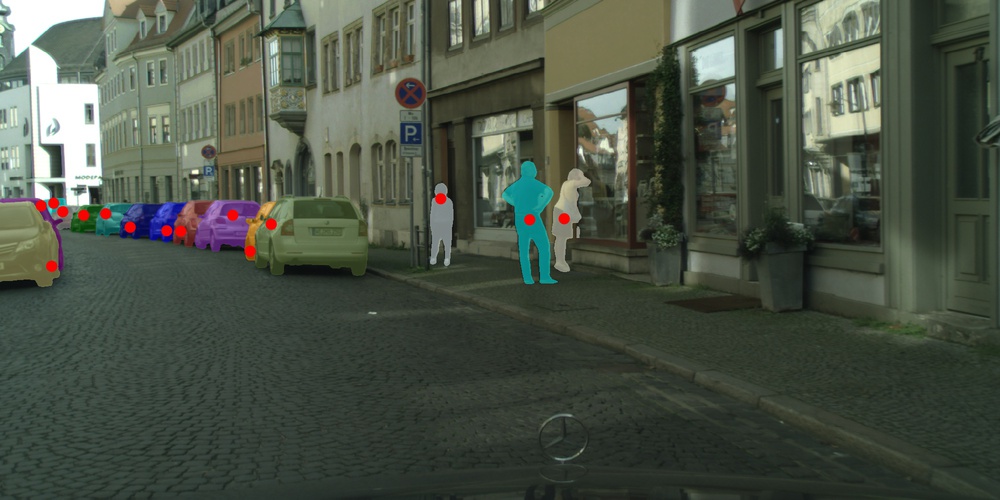} &
		\includegraphics[width=.45\textwidth]{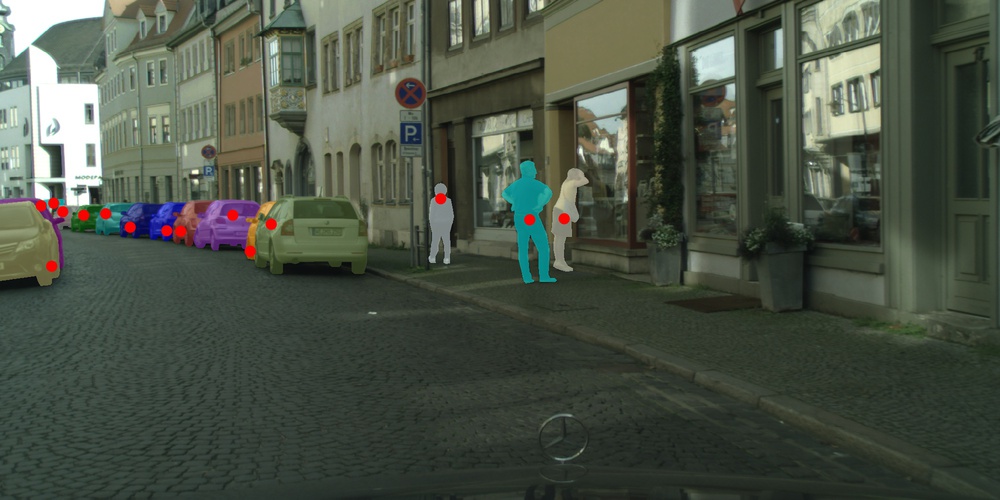} \\
		
		\raisebox{40px}{\rotatebox{90}{t =  0.43 s}}
		\includegraphics[width=.45\textwidth]{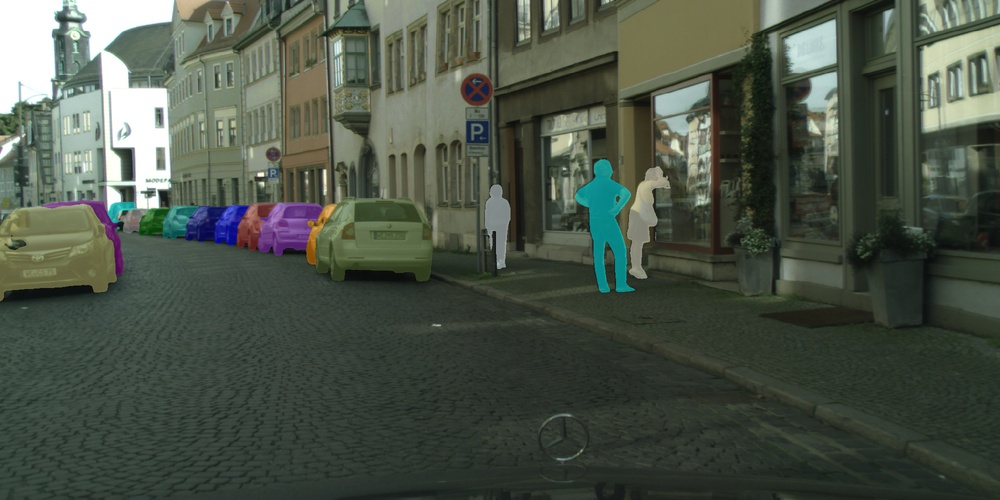} &
		\includegraphics[width=.45\textwidth]{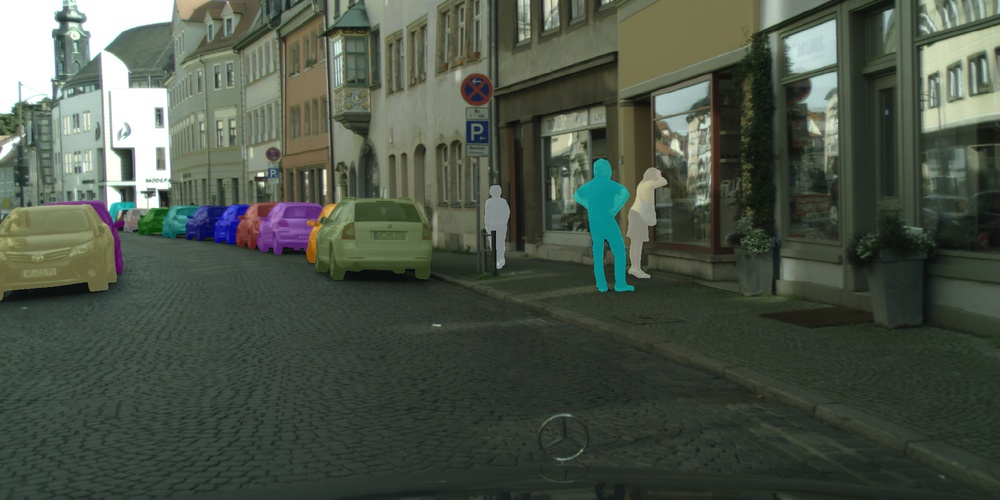} \\
		
		\raisebox{40px}{\rotatebox{90}{t =  0.81 s}}
		\includegraphics[width=.45\textwidth]{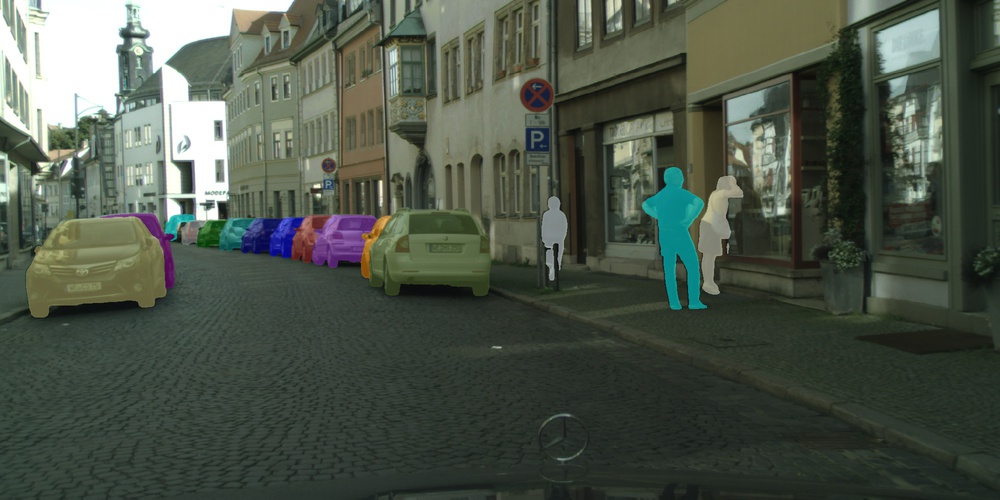} &
		\includegraphics[width=.45\textwidth]{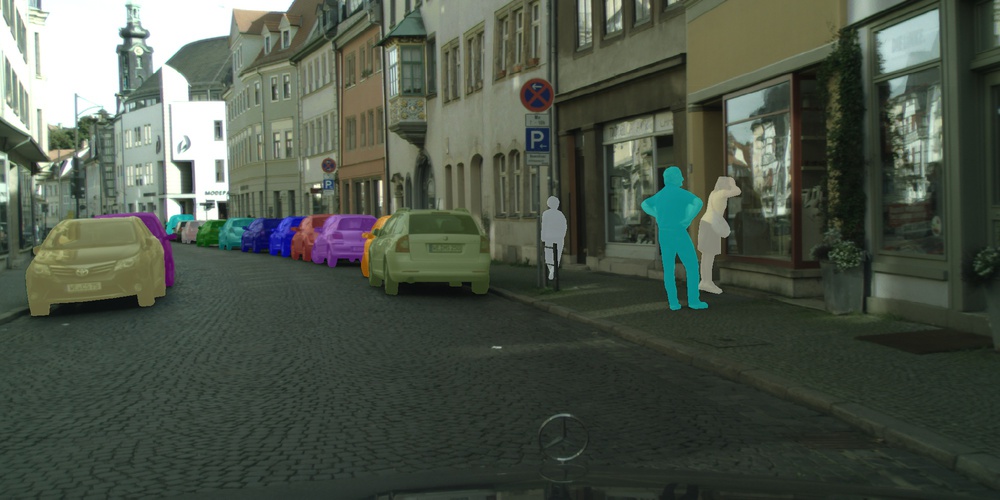} \\
		
		\raisebox{40px}{\rotatebox{90}{t =  1.18 s}}
		\includegraphics[width=.45\textwidth]{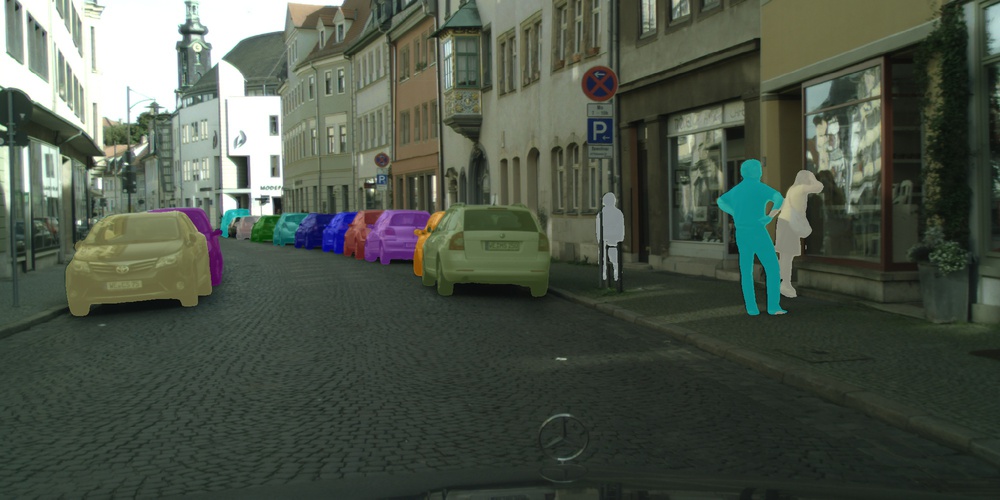} &
		\includegraphics[width=.45\textwidth]{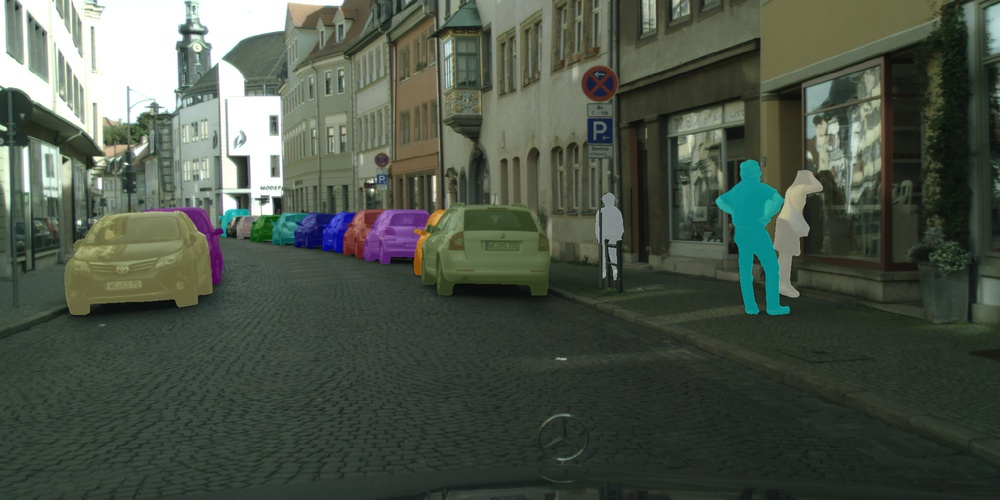} \\
		
		\raisebox{40px}{\rotatebox{90}{t =  1.55 s}}
		\includegraphics[width=.45\textwidth]{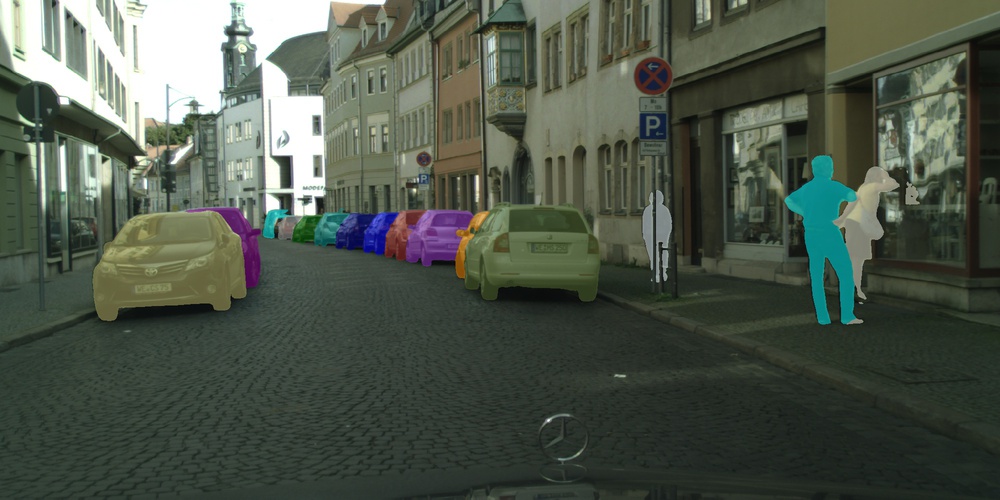} &
		\includegraphics[width=.45\textwidth]{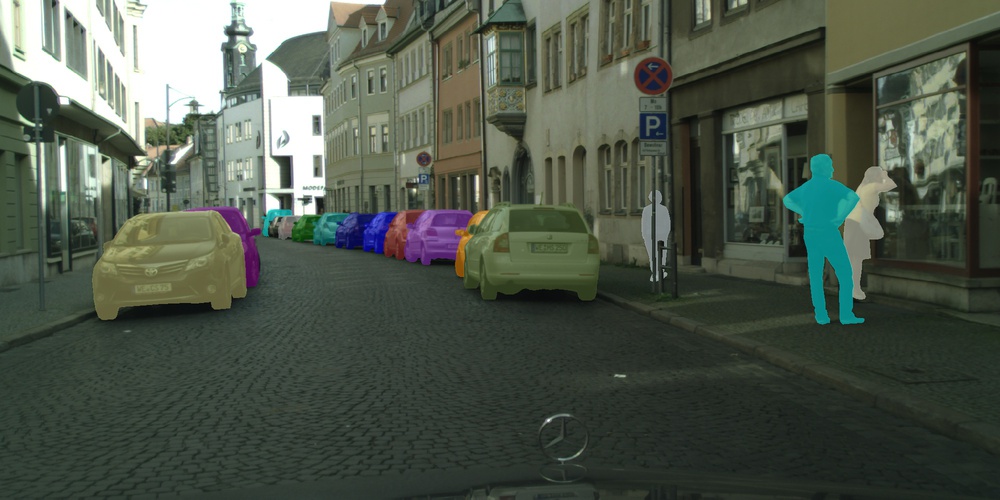} \\
		
	\end{tabular}
	
	\caption{We showcase qualitative segmentation results of our model on the CityscapesVideo validation set and compare it with the ground truth. Red points are the ground truth key point given by the annotator for the new objects. }
	\label{fig:results16}
\end{figure*}

\begin{figure*} 
	\centering
	\setlength\tabcolsep{0.5pt}
	\begin{tabular}{cc}
		

		
		\raisebox{2px}{{Ours}} &	\raisebox{2px}{{Ground Truth}}  \\
		
		\raisebox{40px}{\rotatebox{90}{t =  0 s}}
		\includegraphics[width=.45\textwidth]{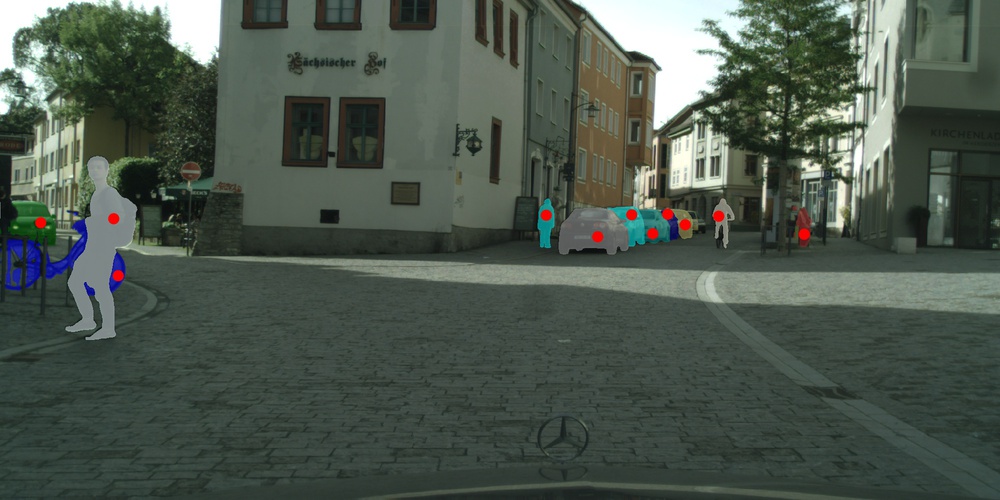} &
		\includegraphics[width=.45\textwidth]{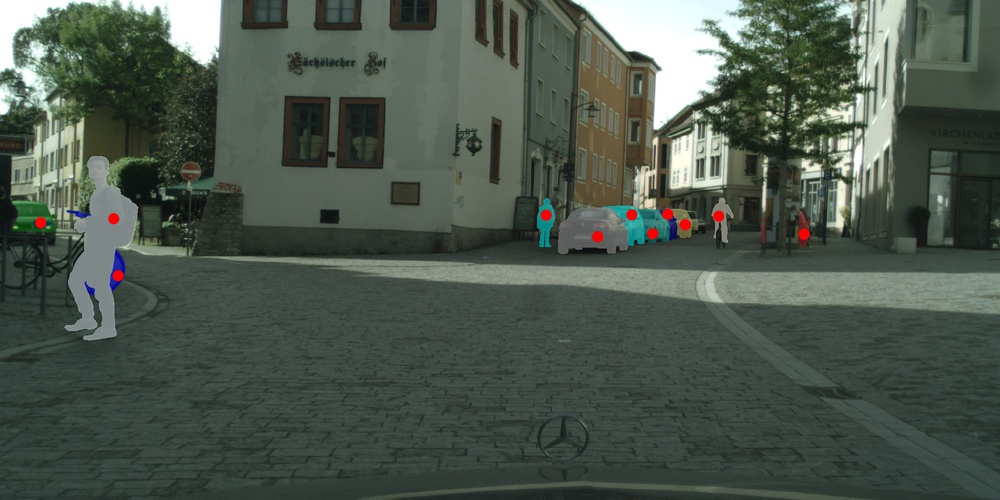} \\
		
		\raisebox{40px}{\rotatebox{90}{t =  0.43 s}}
		\includegraphics[width=.45\textwidth]{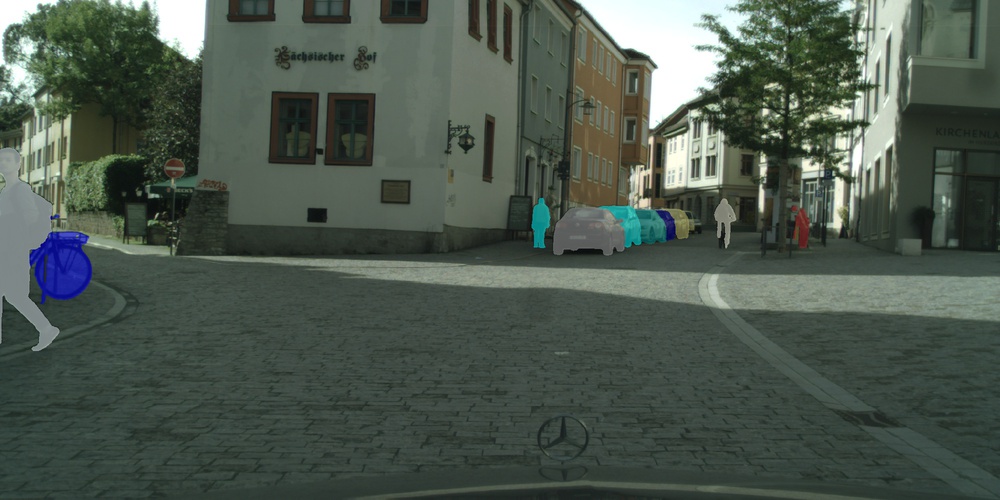} &
		\includegraphics[width=.45\textwidth]{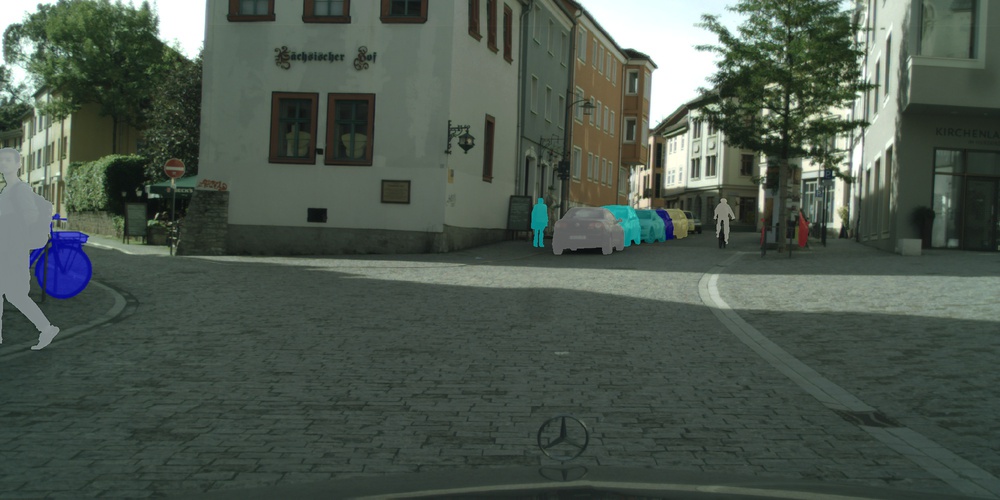} \\
		
		\raisebox{40px}{\rotatebox{90}{t =  0.81 s}}
		\includegraphics[width=.45\textwidth]{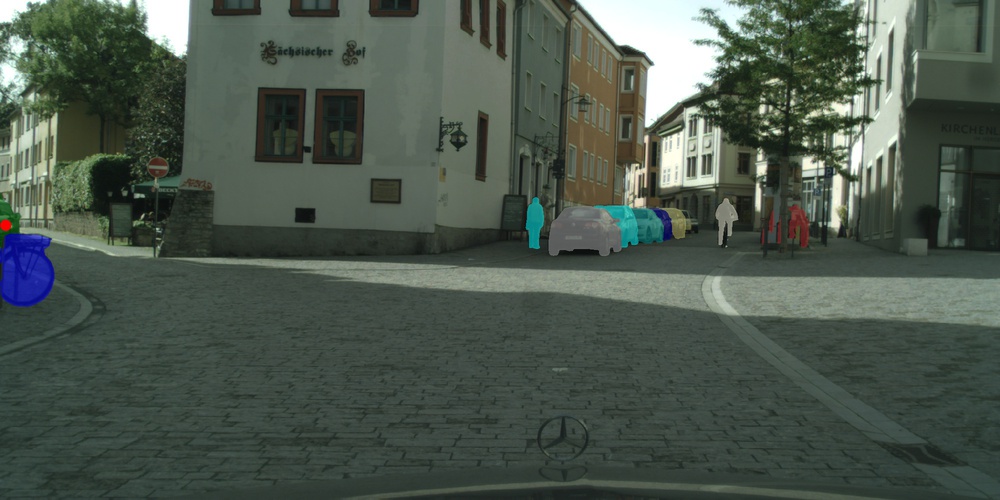} &
		\includegraphics[width=.45\textwidth]{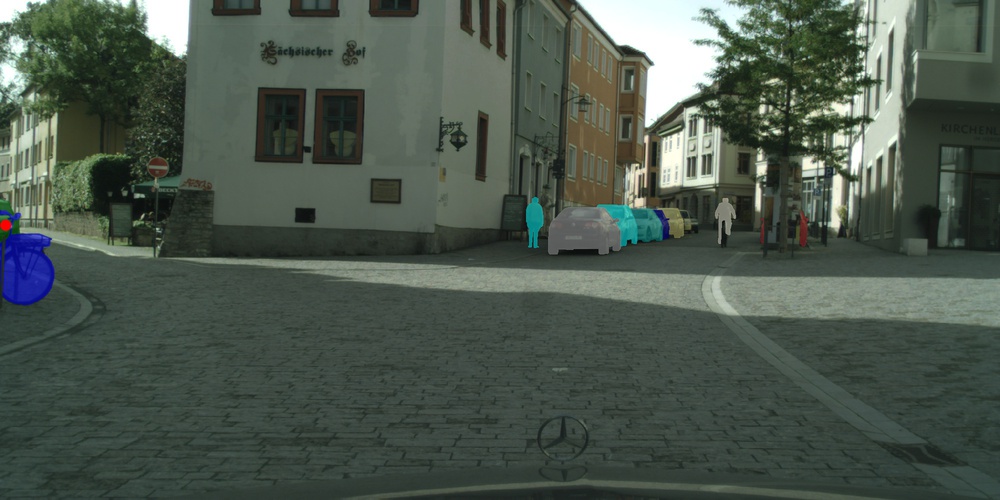} \\
		
		\raisebox{40px}{\rotatebox{90}{t =  1.18 s}}
		\includegraphics[width=.45\textwidth]{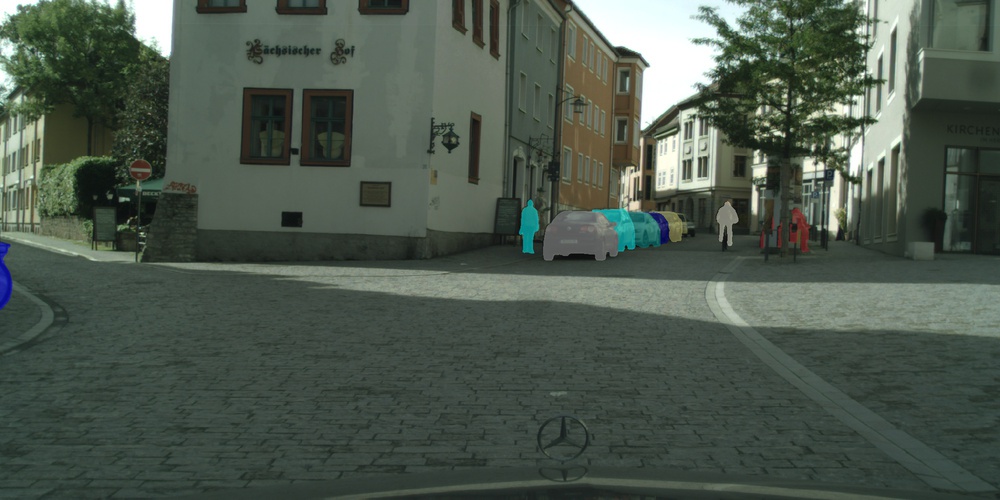} &
		\includegraphics[width=.45\textwidth]{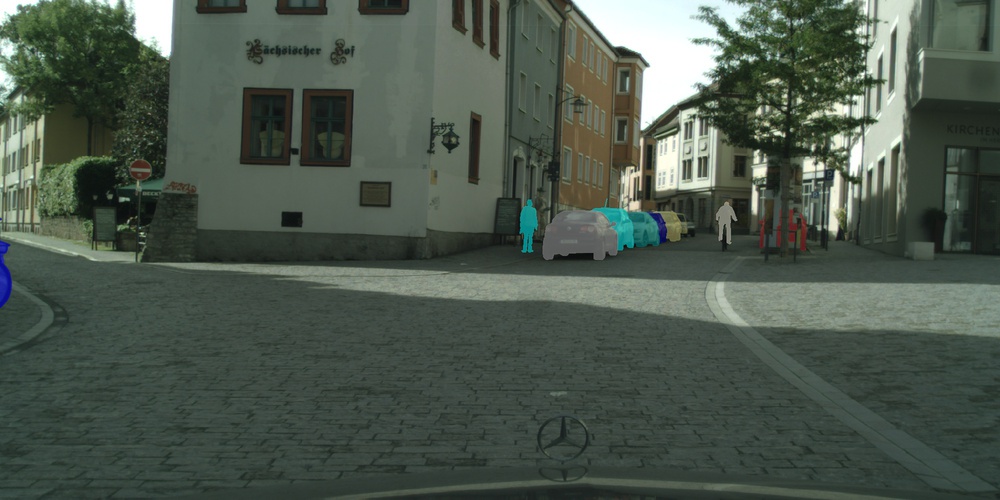} \\
		
		\raisebox{40px}{\rotatebox{90}{t =  1.55 s}}
		\includegraphics[width=.45\textwidth]{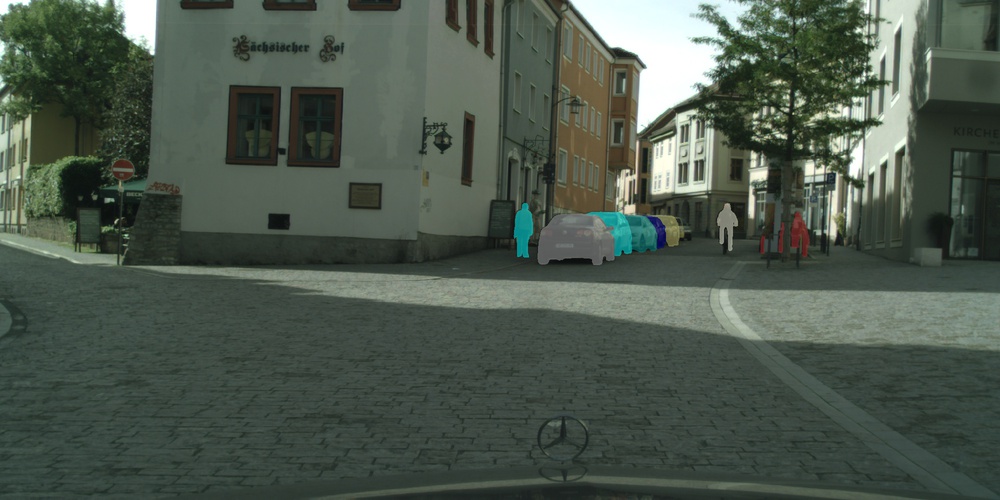} &
		\includegraphics[width=.45\textwidth]{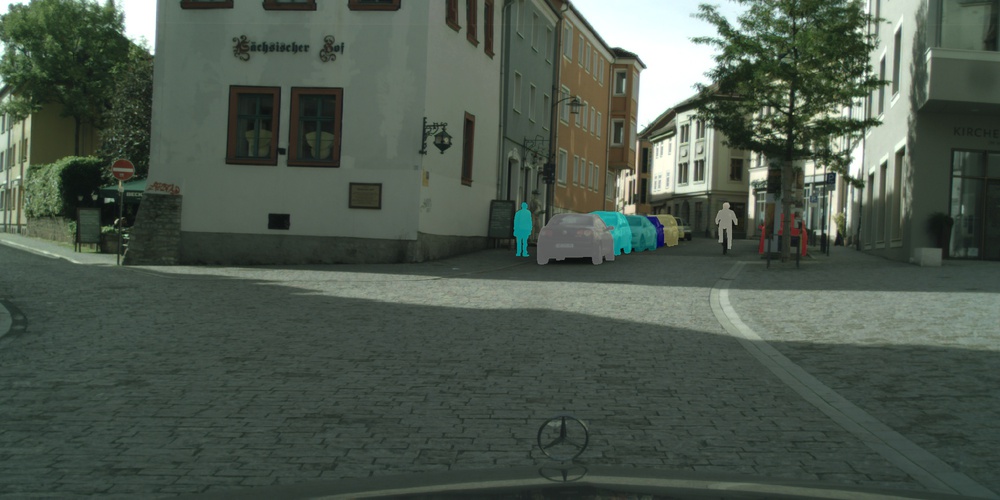} \\
		
	\end{tabular}
	
	\caption{We showcase qualitative segmentation results of our model on the CityscapesVideo validation set and compare it with the ground truth. Red points are the ground truth key point given by the annotator for the new objects. }
	\label{fig:results17}
\end{figure*}

\begin{figure*} 
	\centering
	\setlength\tabcolsep{0.5pt}
	\begin{tabular}{cc}
		

		
		\raisebox{2px}{{Ours}} &	\raisebox{2px}{{Ground Truth}}  \\
		
		\raisebox{40px}{\rotatebox{90}{t =  0 s}}
		\includegraphics[width=.45\textwidth]{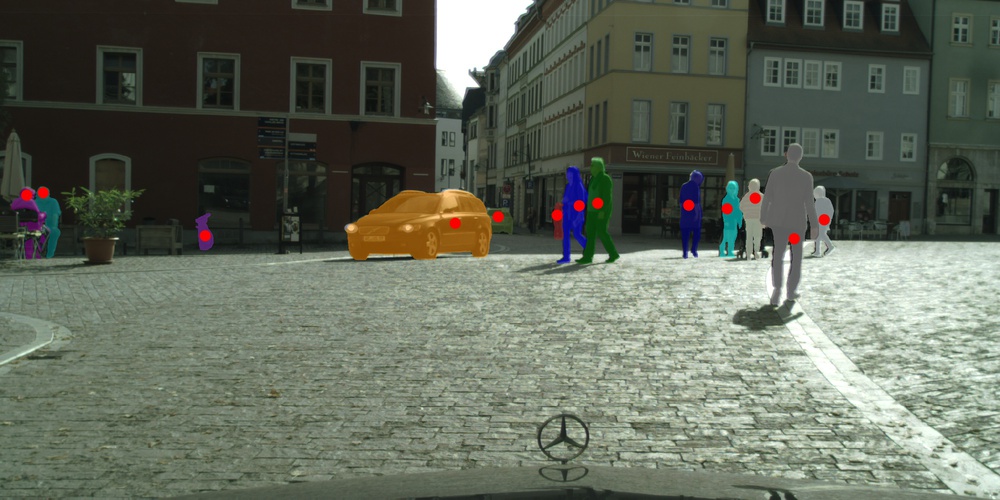} &
		\includegraphics[width=.45\textwidth]{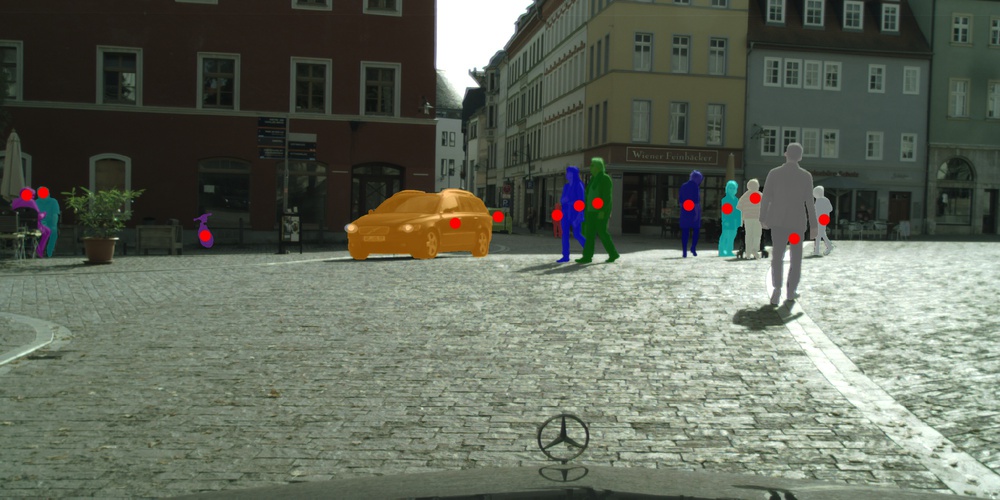} \\
		
		\raisebox{40px}{\rotatebox{90}{t =  0.43 s}}
		\includegraphics[width=.45\textwidth]{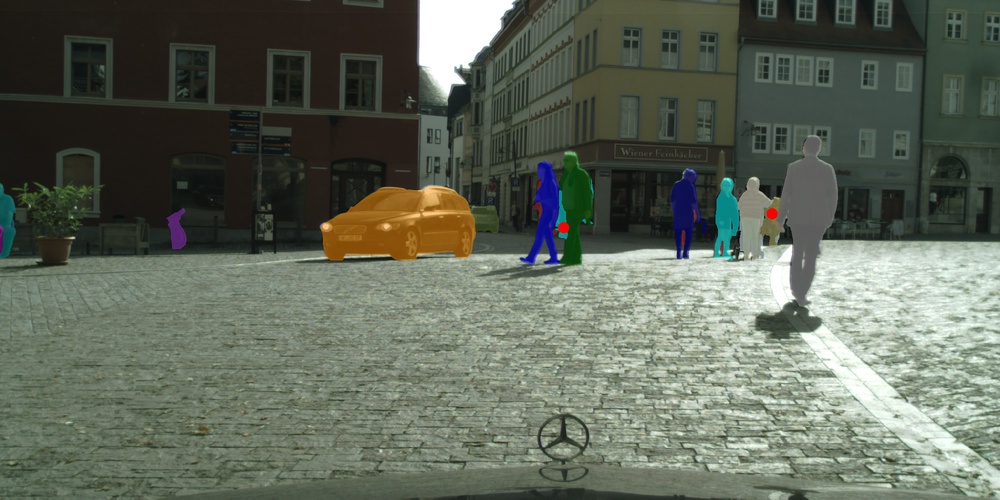} &
		\includegraphics[width=.45\textwidth]{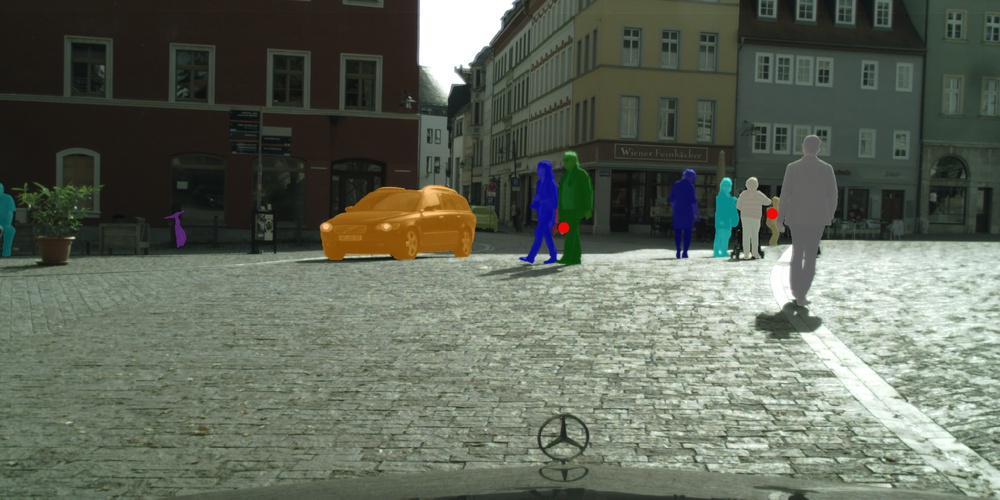} \\
		
		\raisebox{40px}{\rotatebox{90}{t =  0.81 s}}
		\includegraphics[width=.45\textwidth]{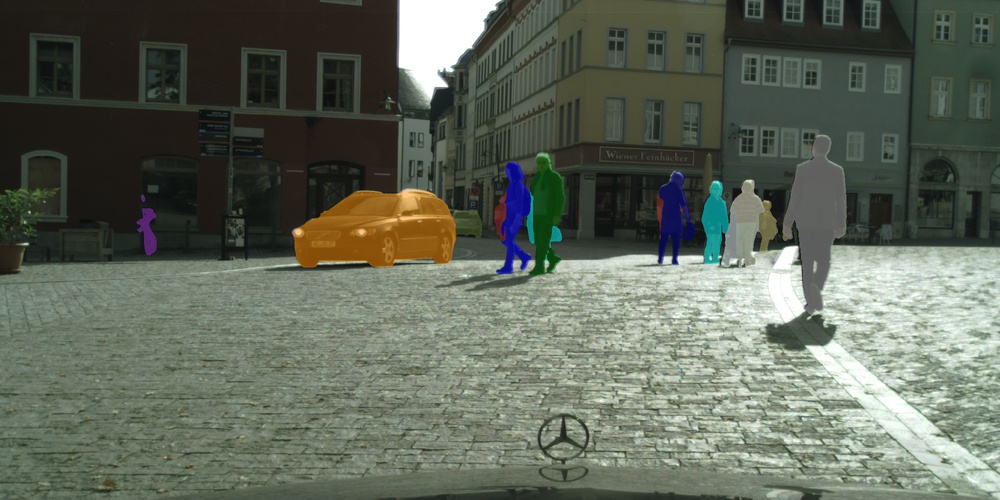} &
		\includegraphics[width=.45\textwidth]{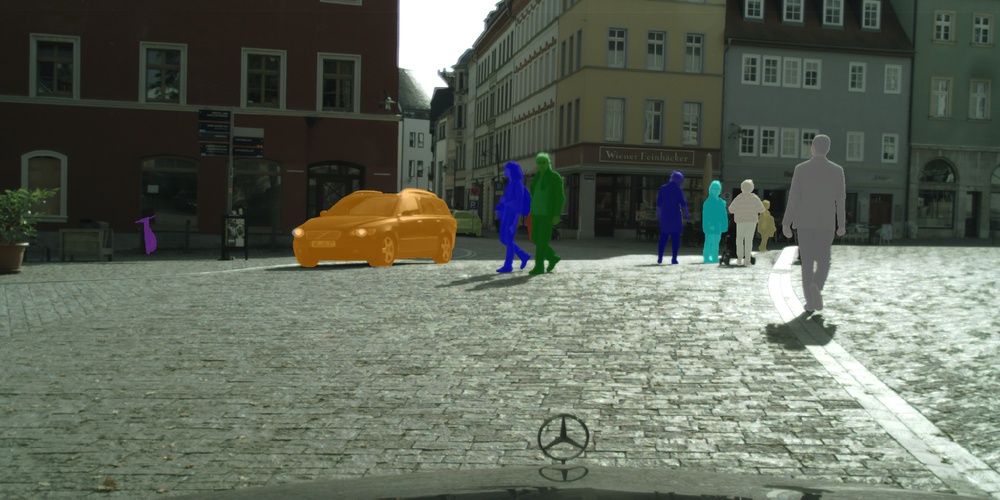} \\
		
		\raisebox{40px}{\rotatebox{90}{t =  1.18 s}}
		\includegraphics[width=.45\textwidth]{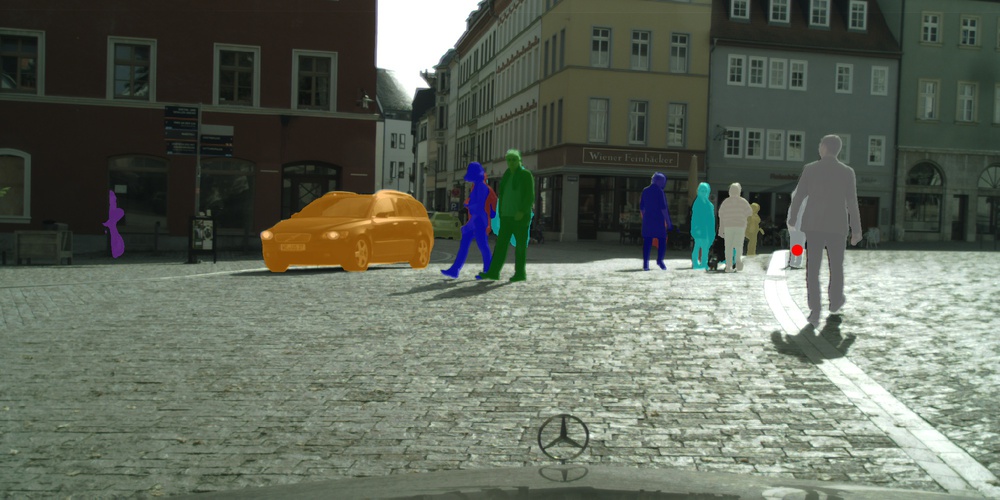} &
		\includegraphics[width=.45\textwidth]{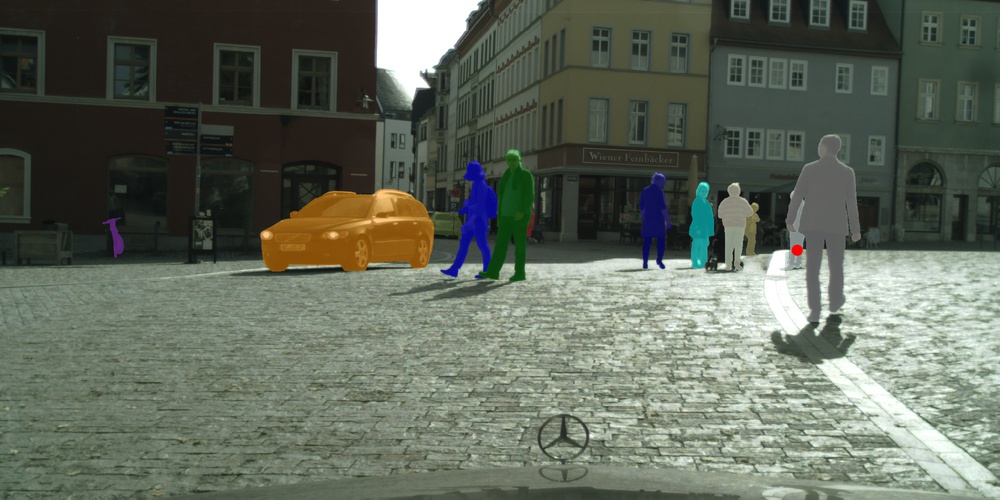} \\
		
		\raisebox{40px}{\rotatebox{90}{t =  1.55 s}}
		\includegraphics[width=.45\textwidth]{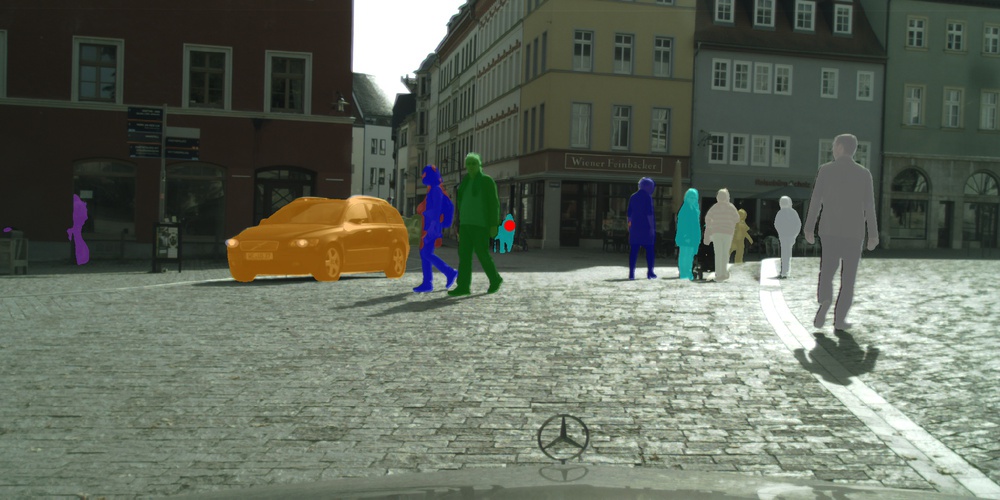} &
		\includegraphics[width=.45\textwidth]{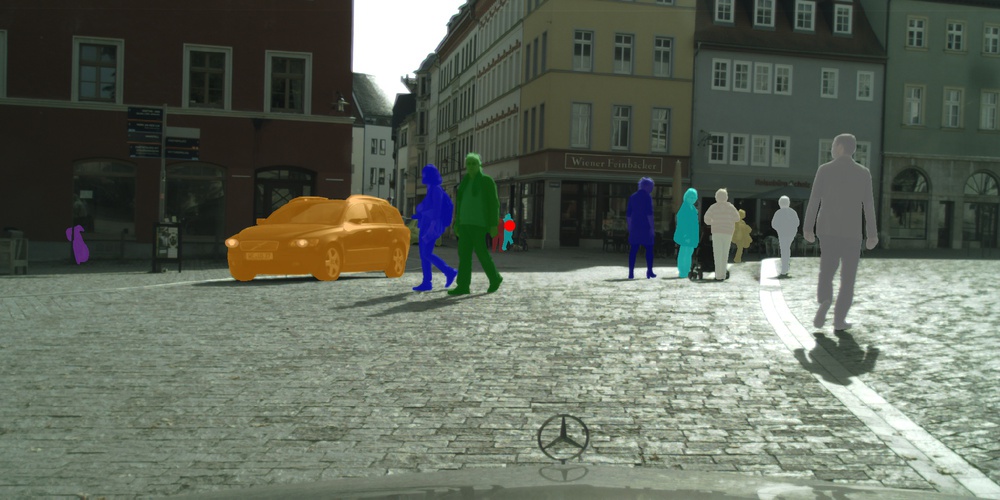} \\
		
	\end{tabular}
	
	\caption{We showcase qualitative segmentation results of our model on the CityscapesVideo validation set and compare it with the ground truth. Red points are the ground truth key point given by the annotator for the new objects. }
	\label{fig:results18}
\end{figure*}

\begin{figure*} 
	\centering
	\setlength\tabcolsep{0.5pt}
	\begin{tabular}{cc}
		

		
		\raisebox{2px}{{Ours}} &	\raisebox{2px}{{Ground Truth}}  \\
		
		\raisebox{40px}{\rotatebox{90}{t =  0 s}}
		\includegraphics[width=.45\textwidth]{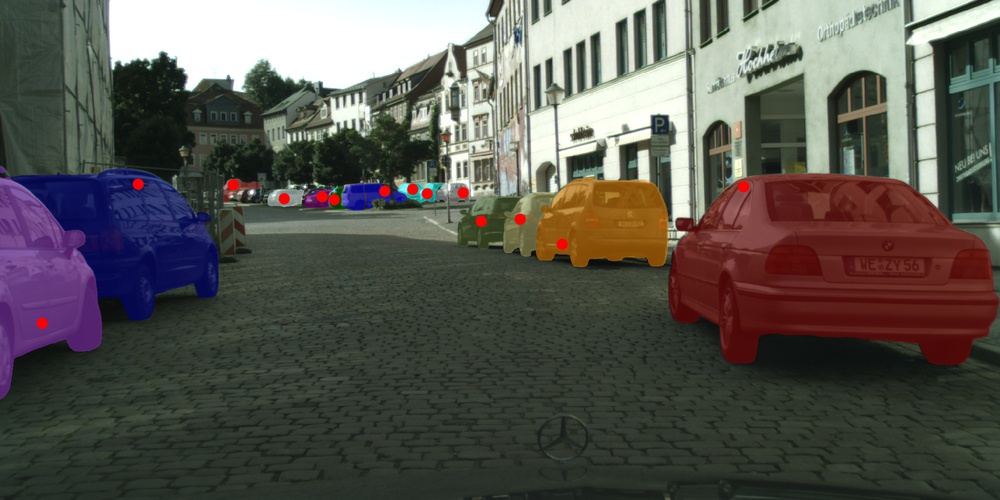} &
		\includegraphics[width=.45\textwidth]{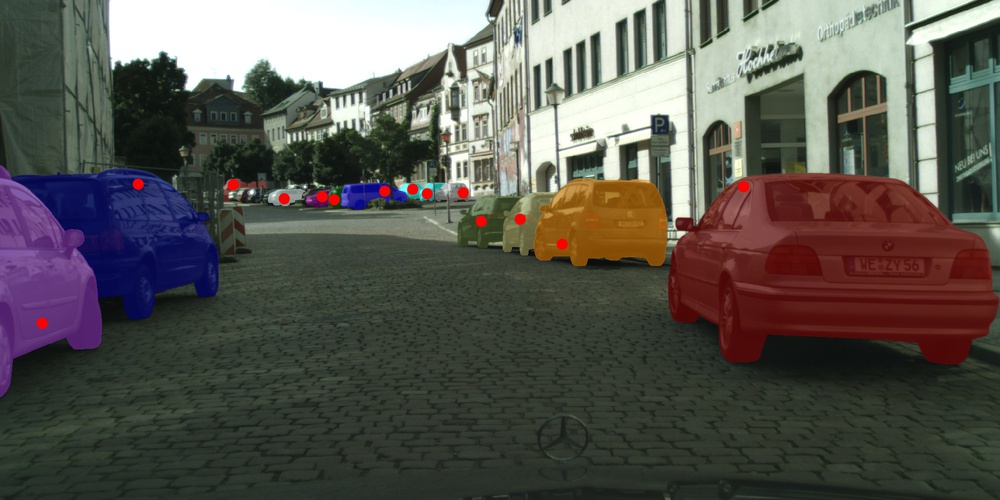} \\
		
		\raisebox{40px}{\rotatebox{90}{t =  0.43 s}}
		\includegraphics[width=.45\textwidth]{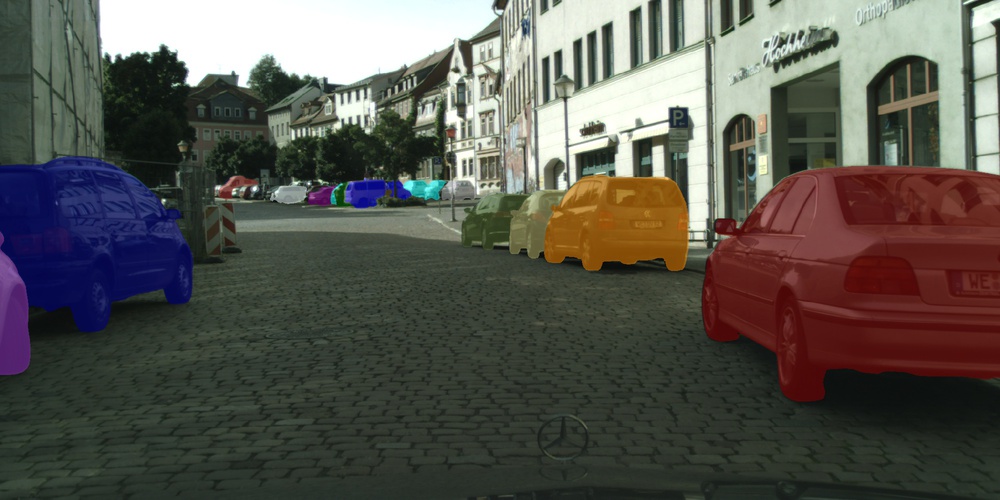} &
		\includegraphics[width=.45\textwidth]{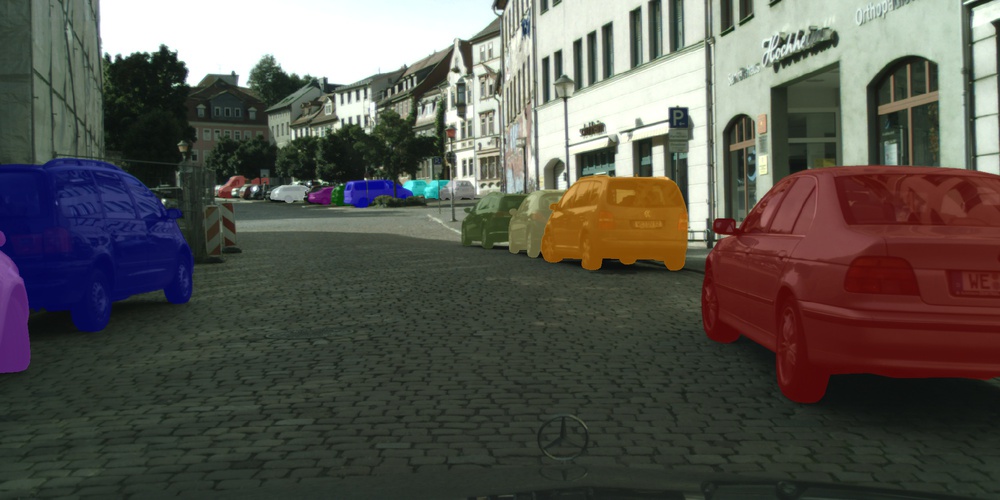} \\
		
		\raisebox{40px}{\rotatebox{90}{t =  0.81 s}}
		\includegraphics[width=.45\textwidth]{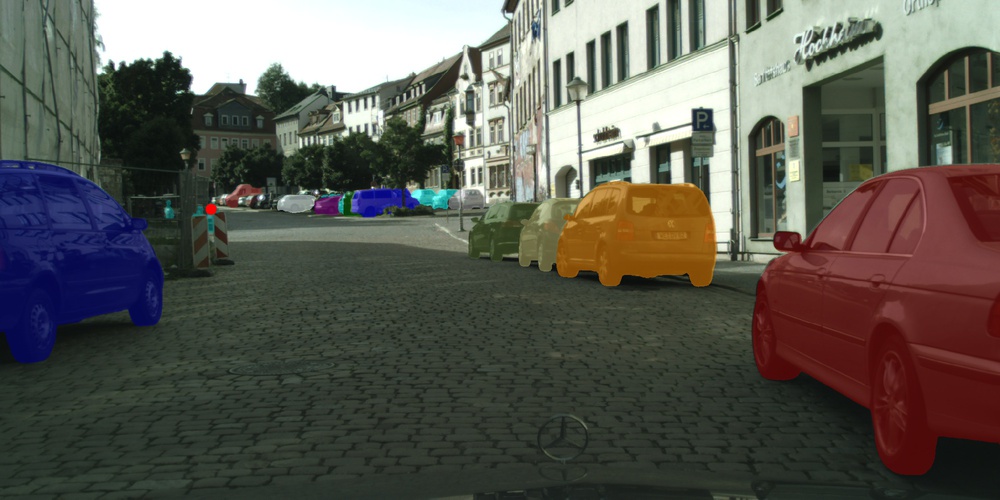} &
		\includegraphics[width=.45\textwidth]{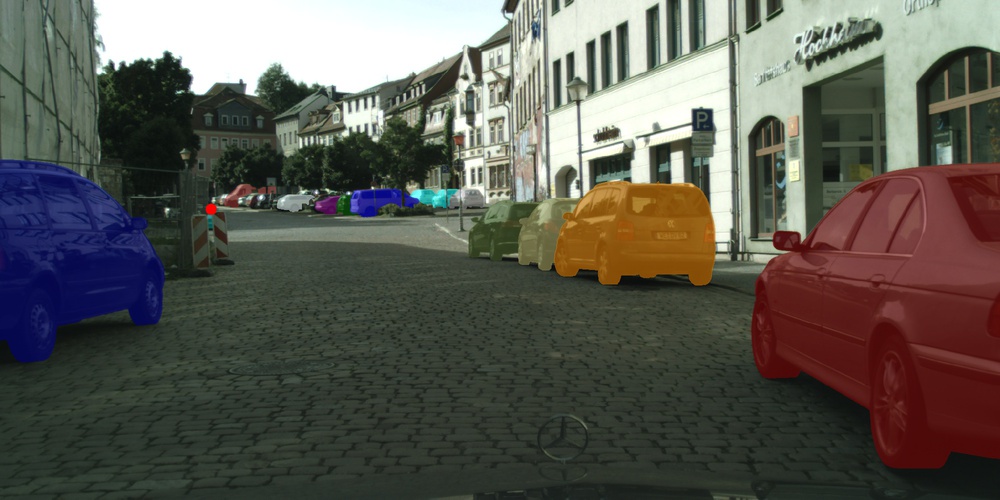} \\
		
		\raisebox{40px}{\rotatebox{90}{t =  1.18 s}}
		\includegraphics[width=.45\textwidth]{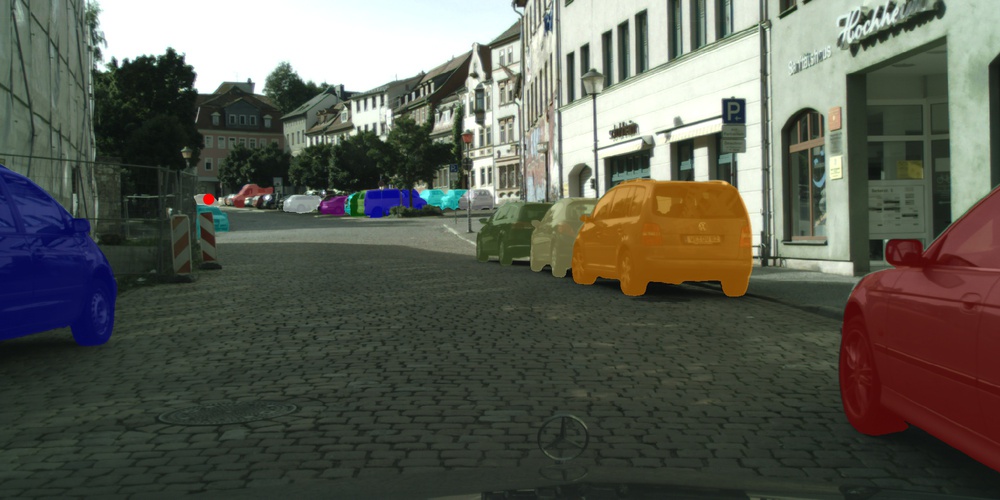} &
		\includegraphics[width=.45\textwidth]{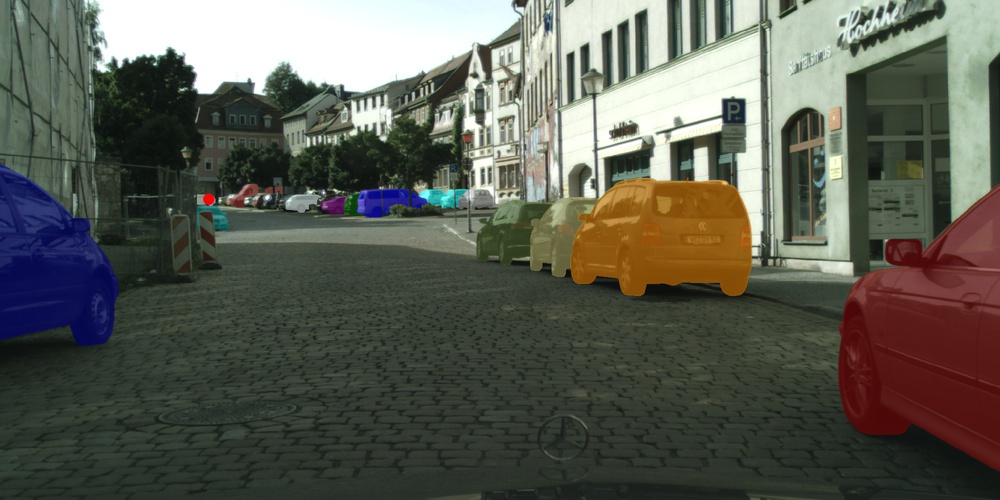} \\
		
		\raisebox{40px}{\rotatebox{90}{t =  1.55 s}}
		\includegraphics[width=.45\textwidth]{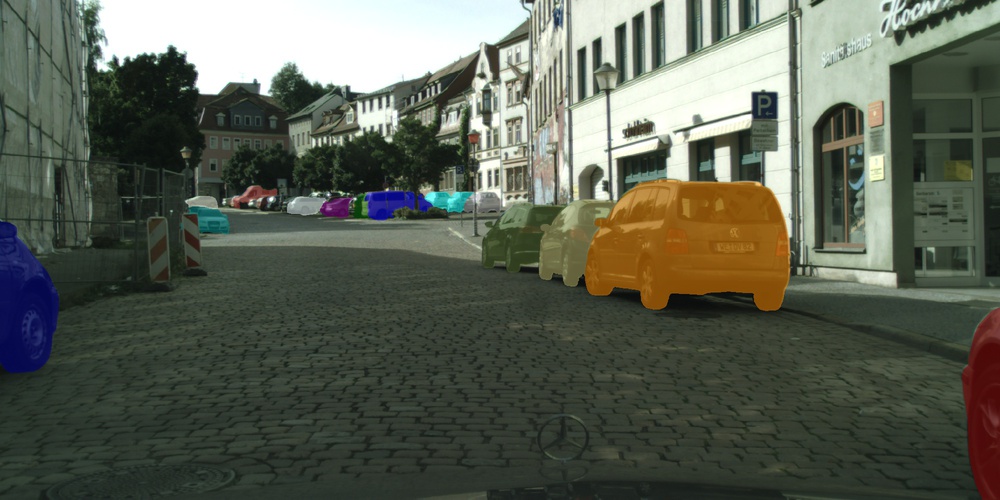} &
		\includegraphics[width=.45\textwidth]{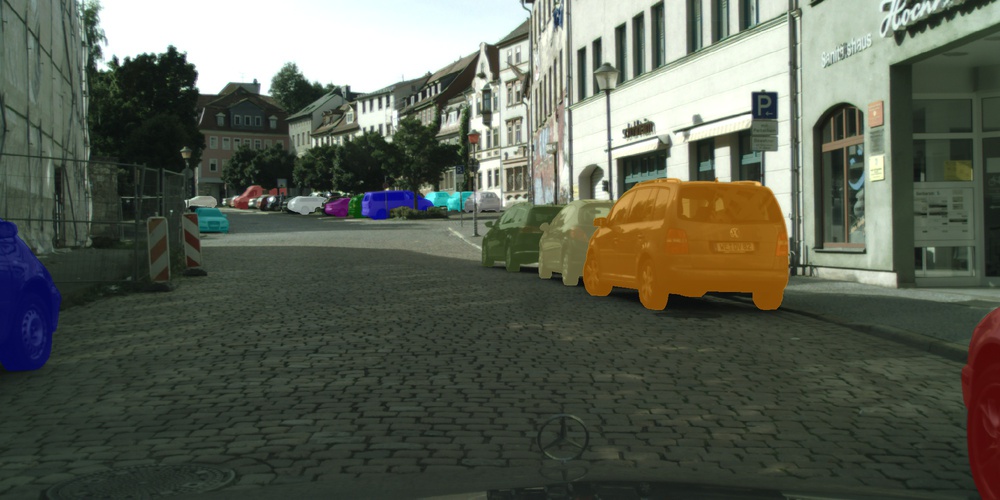} \\
		
	\end{tabular}
	
	\caption{We showcase qualitative segmentation results of our model on the CityscapesVideo validation set and compare it with the ground truth. Red points are the ground truth key point given by the annotator for the new objects. }
	\label{fig:results19}
\end{figure*}

\begin{figure*} 
	\centering
	\setlength\tabcolsep{0.5pt}
	\begin{tabular}{cc}
		

		
		\raisebox{2px}{{Ours}} &	\raisebox{2px}{{Ground Truth}}  \\
		
		\raisebox{40px}{\rotatebox{90}{t =  0 s}}
		\includegraphics[width=.45\textwidth]{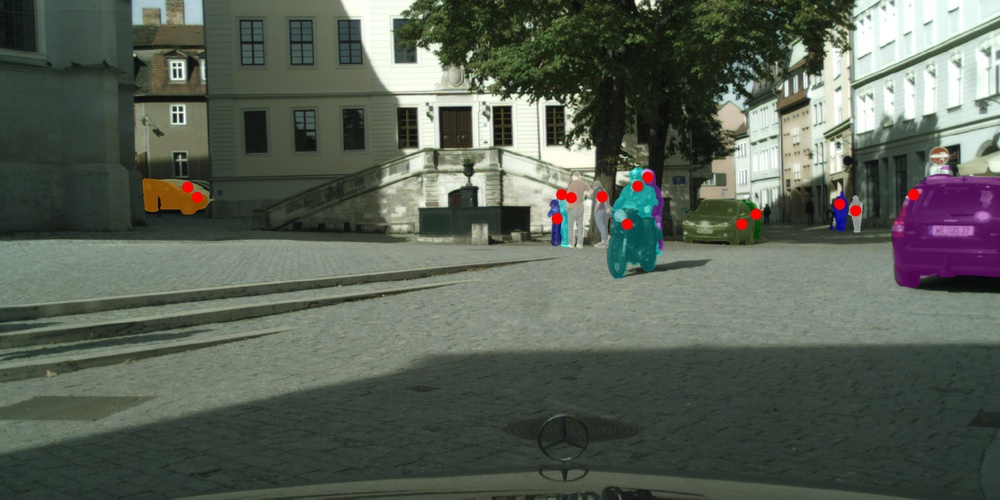} &
		\includegraphics[width=.45\textwidth]{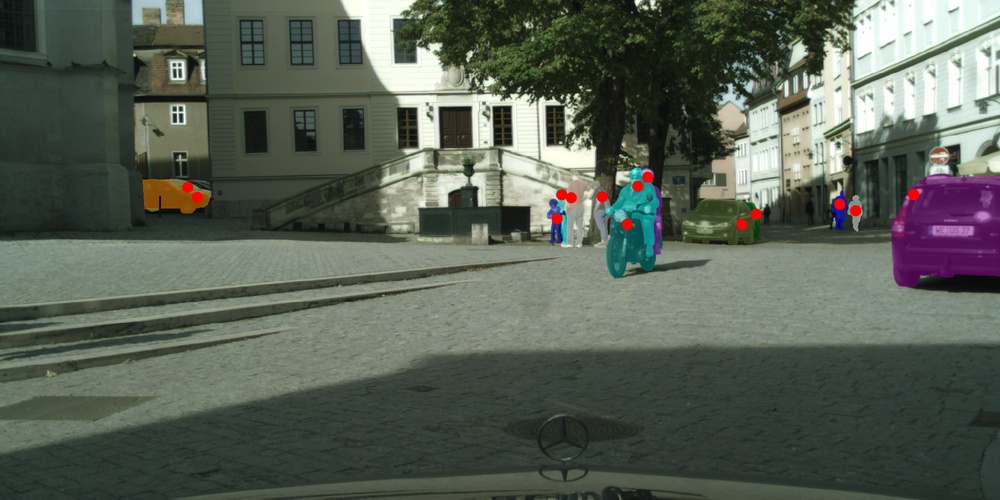} \\
		
		\raisebox{40px}{\rotatebox{90}{t =  0.43 s}}
		\includegraphics[width=.45\textwidth]{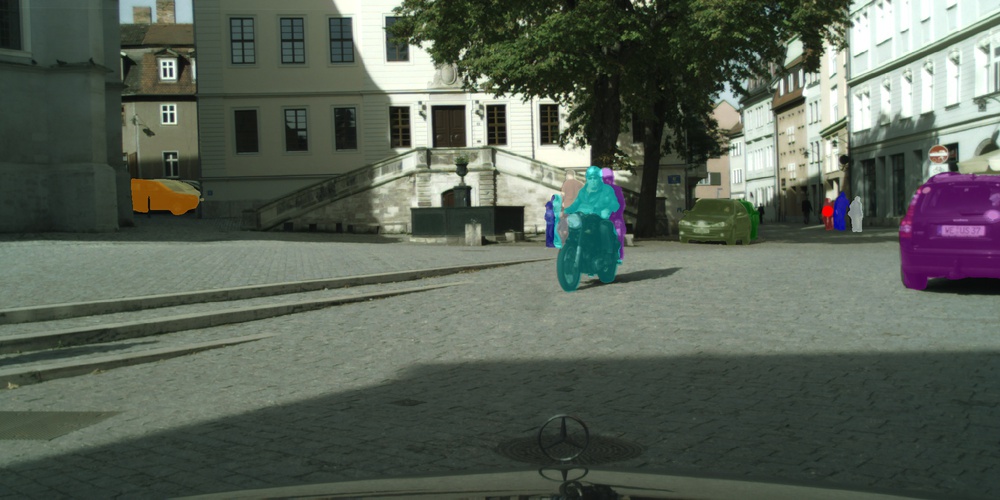} &
		\includegraphics[width=.45\textwidth]{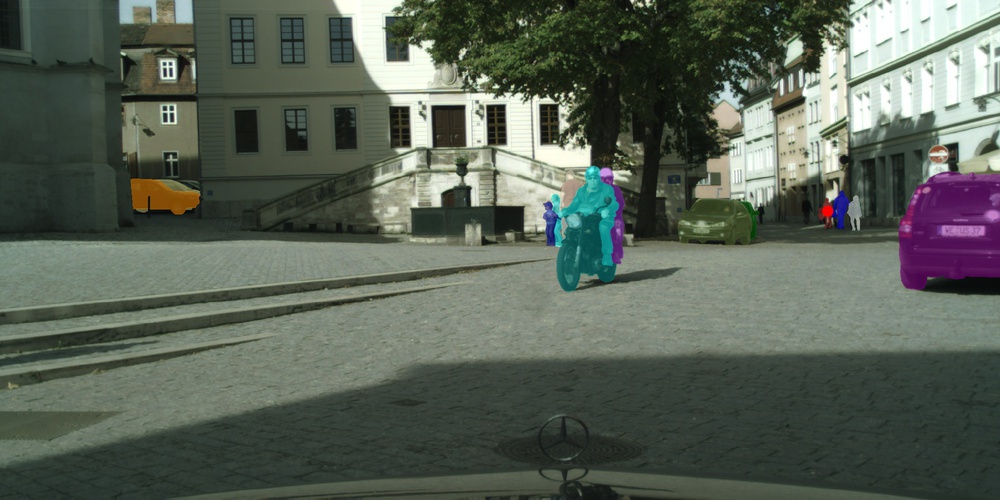} \\
		
		\raisebox{40px}{\rotatebox{90}{t =  0.81 s}}
		\includegraphics[width=.45\textwidth]{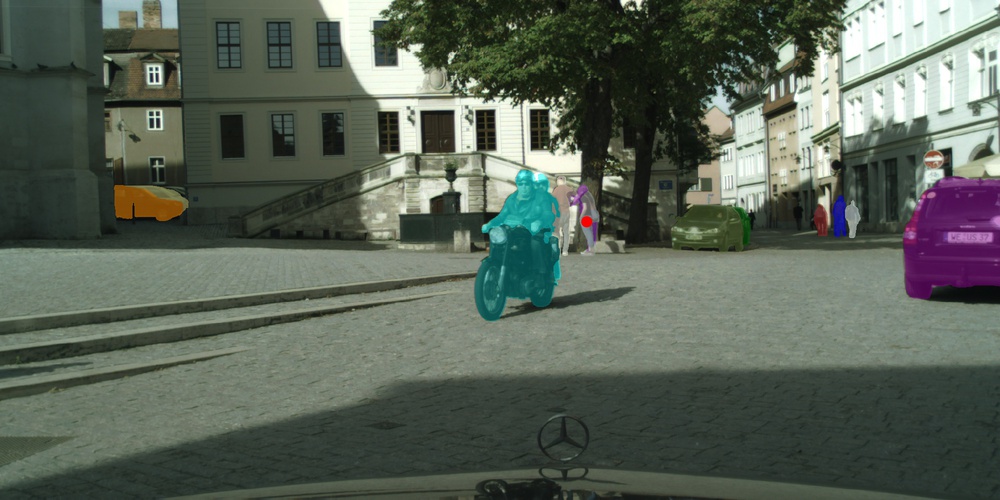} &
		\includegraphics[width=.45\textwidth]{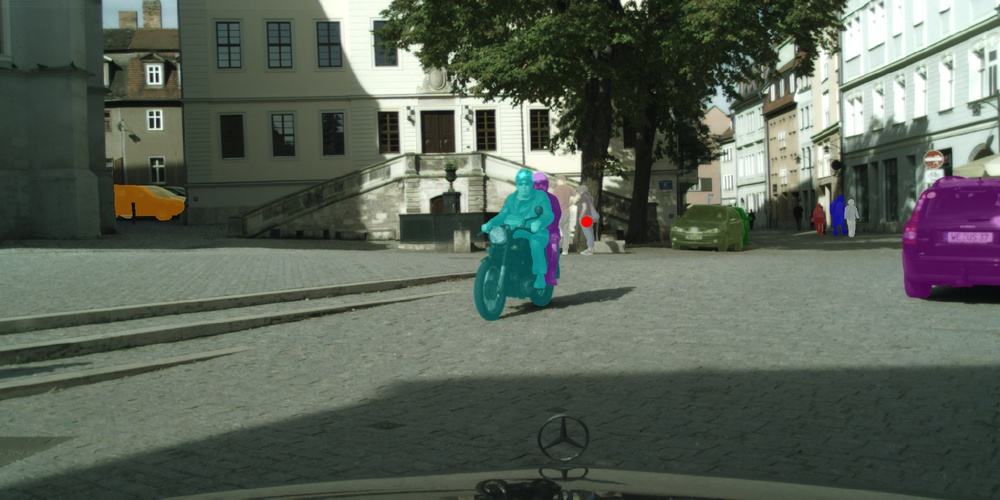} \\
		
		\raisebox{40px}{\rotatebox{90}{t =  1.18 s}}
		\includegraphics[width=.45\textwidth]{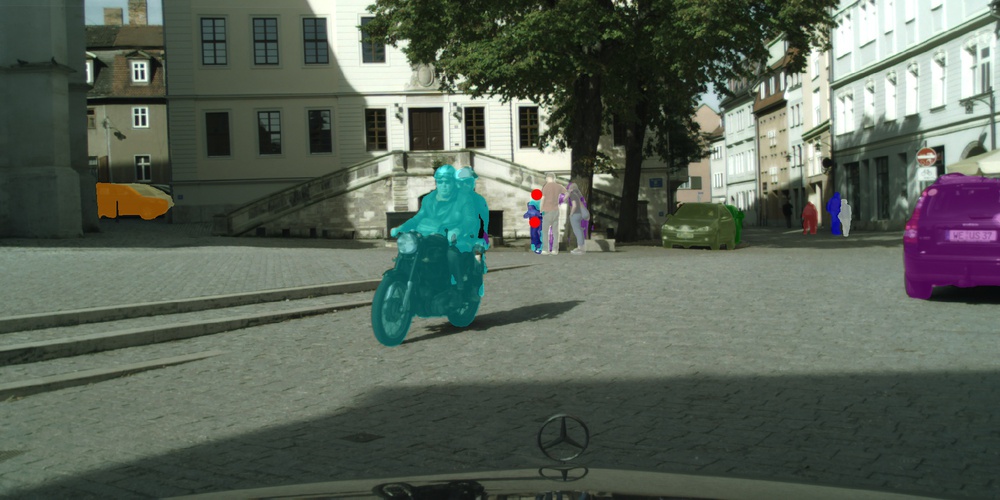} &
		\includegraphics[width=.45\textwidth]{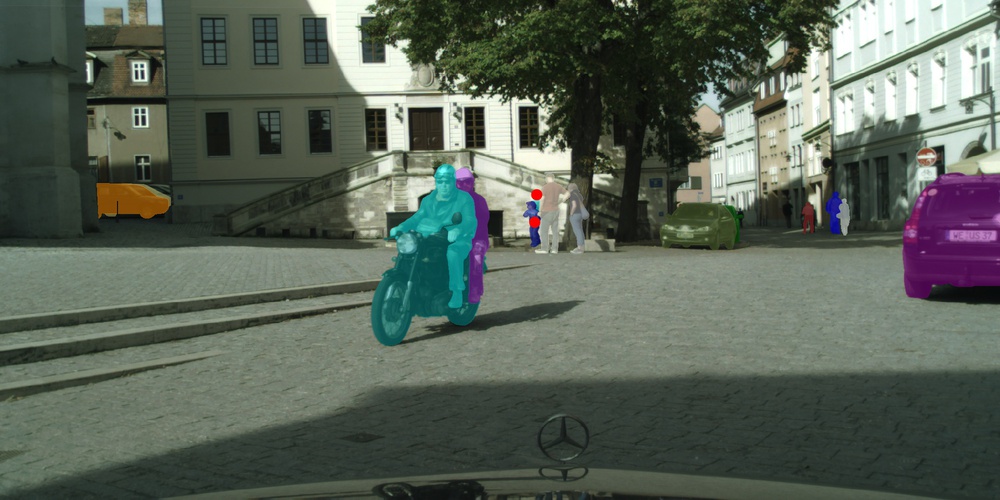} \\
		
		\raisebox{40px}{\rotatebox{90}{t =  1.55 s}}
		\includegraphics[width=.45\textwidth]{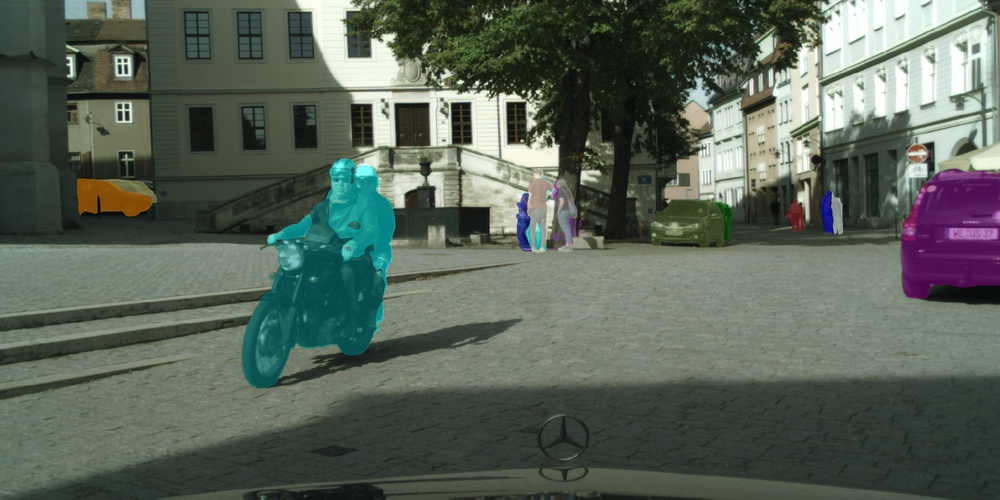} &
		\includegraphics[width=.45\textwidth]{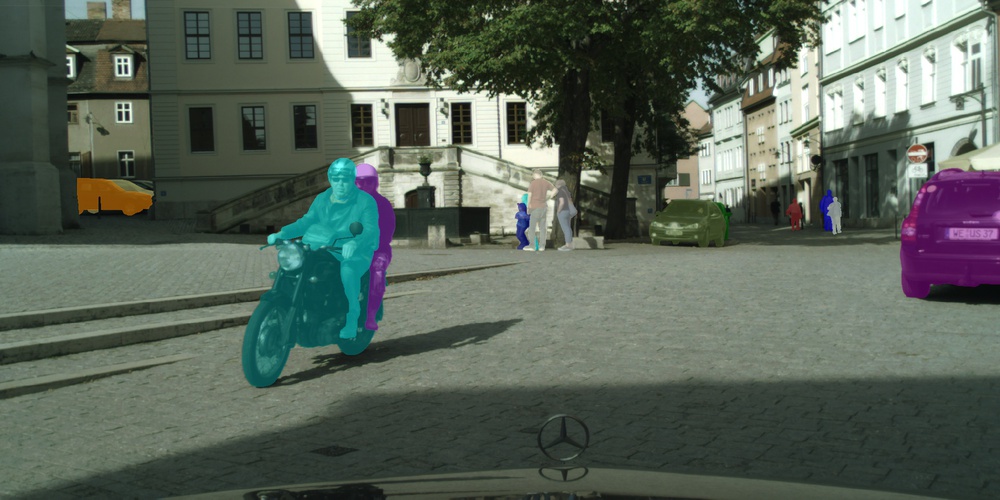} \\
		
	\end{tabular}
	
	\caption{We showcase qualitative segmentation results of our model on the CityscapesVideo validation set and compare it with the ground truth. Red points are the ground truth key point given by the annotator for the new objects. }
	\label{fig:results20}
\end{figure*}
\end{appendices}

\end{document}